\documentclass[10pt,journal,compsoc]{IEEEtran}
\usepackage{hyperref}
\usepackage{enumitem}
\usepackage{bbm}
\usepackage{xcolor}
\usepackage{stackengine}
\usepackage[export]{adjustbox}
\usepackage{algorithm}
\usepackage{algorithmic}
\usepackage{ragged2e}

\ifCLASSOPTIONcompsoc
  \usepackage[nocompress]{cite}
\else
  \usepackage{cite}
\fi

\ifCLASSINFOpdf
\else
\fi

\usepackage{graphicx} 
\usepackage{float} 
\usepackage{subfigure} 
\usepackage{amsmath}
\usepackage{booktabs}
\usepackage{algorithm}
\usepackage{algorithmic}
\newcommand{\daniel}[1]{\textcolor{black}{#1}}
\newcommand{\meng}[1]{\textcolor{black}{#1}}

\begin{document}
%
\title{AnimeDiffusion: Anime Face Line Drawing Colorization via Diffusion Models}
 
%
%
%
%

\author{Yu~Cao$^\dag$,~\IEEEmembership{Student Member,~IEEE},
        Xiangqiao~Meng$^\dag$,
        P.Y.~Mok,~\IEEEmembership{Member,~IEEE},\\
        Xueting~Liu,
        Tong-Yee~Lee,~\IEEEmembership{Senior Member,~IEEE}
        and~Ping~Li,~\IEEEmembership{Member,~IEEE}
\IEEEcompsocitemizethanks{
\IEEEcompsocthanksitem $^\dag$ indicates equal contribution.
\IEEEcompsocthanksitem Y.~Cao, X.~Meng, P.Y.~Mok and P.~Li are with The Hong Kong Polytechnic University, Hong Kong SAR, China. E-mail: \protect\\
\{yu-daniel.cao, xiangqiao.meng\}@connect.polyu.hk,\\ \{tracy.mok, p.li\}@polyu.edu.hk.
\IEEEcompsocthanksitem X.~Liu is with Caritas Institute of Higher Education, Hong Kong SAR, China. \protect E-mail: tliu@cihe.edu.hk.
\IEEEcompsocthanksitem T.-Y. Lee is with National Cheng Kung University, Tainan, Taiwan. \protect \\E-mail: tonylee@mail.ncku.edu.tw.
}
\thanks{Manuscript received April 19, 2005; revised August 26, 2015.}}

%
%

\markboth{Journal of \LaTeX\ Class Files,~Vol.~14, No.~8, August~2015}%
{Shell \MakeLowercase{\textit{et al.}}: Bare Demo of IEEEtran.cls for Computer Society Journals}

\IEEEtitleabstractindextext{%
\begin{abstract}
\justifying  
It is a time-consuming and tedious work for manually colorizing anime line drawing images, which is an essential stage in \meng{cartoon animation} creation pipeline. Reference-based line drawing colorization is a challenging task that relies on the precise cross-domain long-range dependency modelling between the line drawing and reference image. Existing \meng{learning methods} still utilize generative adversarial networks (GANs) as one key module of their model architecture.
In this paper, we propose a novel method called AnimeDiffusion using diffusion models that performs anime face line drawing colorization automatically. To the best of our knowledge, this is the first diffusion model tailored for anime content creation. 
\meng{In order to solve the huge training consumption problem of diffusion models}, we design a hybrid training strategy, first pre-training a diffusion model with classifier-free guidance and then fine-tuning it with image reconstruction guidance. We find that with a few iterations of fine-tuning, the model shows wonderful colorization performance, as illustrated in Fig.~\ref{fig:teaser}. For training AnimeDiffusion, we conduct an anime face line drawing colorization benchmark dataset, which contains 31696 training data and 579 testing data. We hope this dataset can fill the gap of no available high resolution anime face dataset for colorization method evaluation.
Through multiple quantitative metrics evaluated on our dataset and a user study, we demonstrate AnimeDiffusion outperforms state-of-the-art GANs-based models for anime face line drawing colorization. 
We also collaborate with professional artists to test and apply our AnimeDiffusion for their creation work.
We release our code on \href{https://github.com/xq-meng/AnimeDiffusion}{https://github.com/xq-meng/AnimeDiffusion}.
\end{abstract}

}

\maketitle

\IEEEdisplaynontitleabstractindextext

%
\IEEEpeerreviewmaketitle

\IEEEraisesectionheading{\section{Introduction}\label{sec:introduction}}

%
%
%
%
\IEEEPARstart
{L}{ine} drawing colorization is an essential process in the animation industry, however, manually colorizing is time consuming, especially for the line drawings with complex structure content. So, it is necessary and valuable to design a kind of automatic line drawing colorization system.
\meng{Line drawing colorization is challenging, because line drawings, different from grayscale images\cite{9188002, 9186041, 9904484, 8676327}, only contain structure content composing of a series of lines without any luminance or texture information.}
This question has greatly attracted attention of researchers in the field of Computer Graphics, therefore many approaches~\cite{qu2006manga,furusawa2017comicolorization,sykora2009lazybrush,9143503} are proposed for mange and cartoon line drawing colorization during the past time.

Early work~\cite{varga2017automatic} utilized neural network to automatically colorize the cartoon images with random color, which is the first deep learning-based cartoon colorization method. Nevertheless, many interactions are usually needed to refine the colored results to satisfy what the user specified. In order to effectively control the color of the colored result, many user-hint based methods have been proposed successively, such as scribble colors~\cite{ci2018user}, point colors~\cite{zhang2018two}, text-hint~\cite{kim2019tag2pix}, and language-based~\cite{zou2019language}. While these user-hint based methods are still not convenient and intuitive, especially for amateur users without aesthetic judgement. Reference-based colorization methods, such as \cite{lee2020reference,li2022eliminating,zhang2017style,sun2019adversarial,liu2022reference,cao2022attention} provide 
a more convenient way. Users only need to prepare a line drawing and a corresponding reference color image, and the algorithm can automatically complete the colorizing process without other manual intervention.

\begin{figure*}[ht]
\centering  
\includegraphics[width=\linewidth]{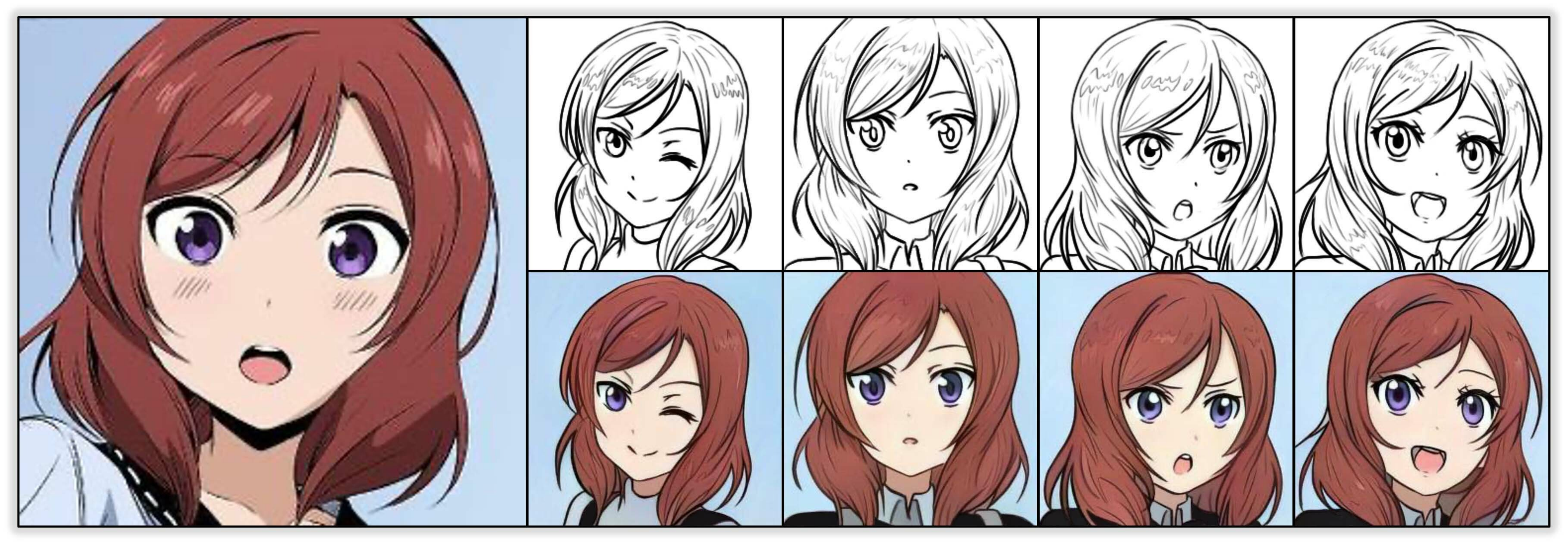}
\caption{We propose AnimeDiffusion that performs reference-based line drawing colorization. Give one reference color image (the left side) and four line drawings of the same character (on the top), AnimeDiffusion generates four colored results (on the bottom) with accurate color and semantic correspondence. In particular, we can generate precise eye color and surprising hair details. The anime character is Nishikino Maki of LoveLive and line drawings are drawn by Ms. Xiao Meng.}
\label{fig:teaser}
\end{figure*}

Reference-base line drawing colorization can be formulated as a conditional image generation task. Since generative adversarial networks (GANs) has become the mainstream model for many generation tasks in the last decade, especially using images as generation conditions. Many previous work for line drawing colorization utilized GANs as one of the most important module of their model architecture design. However, this kind of approach mainly focuses on the improvement of feature aggregation module of two extracted deep features. For example, Lee et al.~\cite{lee2020reference} proposed an attention based Spatial Correspondence Feature Transfer (SCFT) module. Li et al.~\cite{li2022eliminating} eliminated the gradient conflict among attention branches by using Stop-Gradient Attention (SGA) module. Cao et al.~\cite{cao2022attention} designed an attention-aware model for generating high quality colored anime line drawing images. Otherwise, GANs-based models require to deployment of multiple losses, which also increases the instability of the training process.

Along with the diffusion probabilistic models~\cite{sohl2015deep} (“diffusion models” for brevity) has been proved to be an excellent model that is capable of generating high quality images, many algorithms based on diffusion models have been proposed in recent years. As a novel generation algorithm, it has greatly \meng{promoted} the progress of AI-Generated Content (AIGC) technology. Inspired by this, we propose the first diffusion model called AnimeDiffusion tailored for anime face line drawing colorization. Since the diffusion models \meng{usually have} the problem of high computing consumption, we design a hybrid training strategy that consists of classifier-free guidance pre-training stage and image reconstruction guidance fine-tuning stage. 
During the AnimeDiffusion training procedure, we train an U-Net which regards the line drawings and reference images as conditional denoising input. In order to make the model learn semantic correspondence ability, the reference image is a geometry distorted version of original reference image by applying Thin-Plate Splines (TPS) transformation. 
The original reference image is added to a Gaussian noise and concatenated with line drawing and reference image together. The U-Net uses the concatenated images as input and predicts the noise that added onto the original reference feature map. 
This pre-training process mainly makes AnimeDiffusion learn the denoising ability.
In the fine-tuning stage of AnimeDiffusion, we calculate the MSE loss between the reconstructed image and the original reference image and update the parameters of AnimeDiffusion when performing reverse sampling task.
\meng{It is worth to note that our fine-tuning is different from the one applied in other approaches. Existing methods mainly fine-tune the pre-trained image-to-image\cite{meng2021sdedit},\cite{kim2022diffusionclip} or text-to-image\cite{rombach2022high},\cite{zhang2023adding} diffusion models for various kinds of downstream tasks. However, our fine-tuning allows us to train a diffusion model from sketch to perform colorization more efficiently.}
Experimental results demonstrate that AnimeDiffusion generates better results than the state-of-the-art GANs-based line drawing colorization models both qualitatively and quantitatively. In order to train AnimeDiffusion and fill the gap of no available high resolution anime face dataset, we conduct a novel benchmark dataset for academic research purpose. All original images are chosen from 
\cite{danbooru2020}.

Our main contributions can be summarized as follows:
\begin{itemize}[itemsep=0pt, topsep=0pt, parsep=0pt]
    \item
    We propose the first AnimeDiffusion model tailored for anime face line drawing colorization. Experiments demonstrate that AnimeDiffusion notably outperforms the GANs-based counterparts and achieves the state-of-the-art anime face line drawing colorization results.
    \item
    We design a hybrid training strategy for AnimeDiffusion in order to tackle the problem of high computing consumption of diffusion models. The proposed strategy can accelerate the network convergence and improve colorization performance.
    \item
    We conduct a new anime face line drawing colorization benchmark dataset, which contains 31696 training data and 579 testing data. Our dataset aims to fill the gap of no available high resolution ($256\times256$) anime face dataset for training and evaluation.
\end{itemize}

\section{Related Work}
\subsection{Line Drawing Colorization}
Since line drawing contains only structure information with sparse line sets, existing colorization methods for gray-scale images cannot directly be used. Many colorization methods tailored for line drawings have been developed. Traditional line drawing colorization approaches~\cite{qu2006manga}, \cite{sykora2009lazybrush} are commonly optimized-based which allow users to use brushes to inject desired color into specific \meng{regions.} With the advancement of deep learning technology and for the better control of color, many user-hint colorization methods spring up. The color hints are usually concatenated with line drawing and encoded as the input for neural network in many deep learning based methods. Ci et al.~\cite{ci2018user} proposed a conditional GAN model to colorize the anime line drawing using color scribbles, which can generate colored results with accurate shading. Zhang et al.~\cite{zhang2018two} developed a color points hint two-stage colorization method, which divided the complex colorization task into two simpler and goal-clearer subtasks. Kim et al.~\cite{kim2019tag2pix} utilized their SECat module to generate illustrations with quality details using text tags as their hints. Zou et al.~\cite{zou2019language} for the first time presented a language-based system for interactive colorization of scene sketches. However, the complexity of such user-hints methods will become more labor-intensive as the number of line drawings increase, many interactions are usually needed to refine the colored results and these methods are not user-friendly for amateur users without aesthetic judgement, especially for preparing appropriate color hints. Therefore, many reference based colorization methods have been proposed, and they are very suitable for colorizing line drawing sets or videos of anime characters, which need to keep the same characters with consistent colors during each frame. Sato et al.~\cite{sato2014reference} segmented the target and reference image into different regions, then represented regions as nodes of a graph structure and colorized the monochrome target image by matching the graphs of the target and reference images. Furusawa et al.~\cite{furusawa2017comicolorization} proposed the first semi-automatic system to colorize an entire manga with color features extracted from the input reference image. Chen et al.~\cite{9143503} proposed an active learning based framework to match local regions between line arts and reference color image, followed by mixed-integer quadratic programming (MIQP) which considers the spatial contexts to further refine matching results. Shi et al.~\cite{shi2022reference} proposed a new line art video colorization method using 3D convolutional module to refine the temporal consistency of the colored result. Dou et al.~\cite{dou2021dual} is the first work that utilizes the HSV color space for anime sketch colorization. Maejima et al.~\cite{maejima2021anime} proposed colorization method for anime character using few-shot learning. Sun et al.~\cite{sun2019adversarial} trained a dual conditional GAN to colorize contours in different styles which helps designers create icons. Li et al.~\cite{li2022style} presented an icon colorization system that is composed of an encoder-decoder network and a conditional normalizing flow. \meng{Our AnimeDiffusion} is a novel reference-based colorization tailored for anime face line drawing colorization. Compared with previous GANs-based methods, AnimeDiffusion can generate better results both in visual quality and quantitative metrics.

\subsection{Semantic Correspondence}
Semantic correspondence~\cite{xiao2022learning} is one of the fundamental problems in computer vision which goal is to establish dense correspondences across images containing the targets of the same category or with similar semantic information. In computer graphics, it is also very important for exemplar-based image colorization task, and there usually exists the same semantic information between the target image and exemplar image in practical colorization usage. For line art colorization, this can also be viewed as cross-domain correspondence since the texture difference between line drawing and reference color image. Zhang et al.~\cite{zhang2020cross} proposed an exemplar-based image translation system based on cross-domain correspondence learning. Lee et al.~\cite{lee2020reference} proposed an attention-based module to spatially match and aggregate the sketch feature and reference color image feature. Li et al.~\cite{li2021globally} designed a model to colorize grayscale natural image, even if the exemplar image has no similar semantic information of the target grayscale image. He et al.~\cite{he2019progressive} \meng{presented} a novel progressive color transfer model, which jointly optimizes dense semantic correspondences in the deep feature domain and the local color transfer in the image domain. Zhang et al.~\cite{zhang2019deep} proposed the first end-to-end exemplar-based video colorization algorithm, which unified the semantic correspondence and colorization into a single network. Lu et al.~\cite{lu2020gray2colornet} proposed an unified semantic color transfer system from reference image to the target grayscale image. Most of these semantic correspondence learning methods are designed for grayscale image or video colorization. However, we design AnimeDiffusion to perform anime face line drawing colorization, which can generate results with clear and accurate semantic colors.

\subsection{Diffusion Models}
Diffusion models such as denoising diffusion probabilistic models (DDPM)~\cite{ho2020denoising} have achieved great success in image generation tasks. It is shown that image generation models based on diffusion models have better performance in terms of training stability and generation quality~\cite{dhariwal2021diffusion, ho2022cascaded}. 
Denoising diffusion implicit models (DDIM)~\cite{song2020denoising} accelerates the sampling procedure and enables a determined generation process with given Gaussian noise. In addition to generating high quality images based on random noise, diffusion models also show good performance in conditional image-to-image translation tasks.
An image-to-image diffusion models (Palette)~\cite{saharia2022palette} offers a versatile and general framework for image manipulation. Stochastic Differential Editing (SDEdit)~\cite{meng2021sdedit} is a guided image editing and synthesis method, which synthesizes realistic images by iteratively denoising through a stochastic differential equation (SDE). Combined with the contrastive language-image pre-training (CLIP)~\cite{radford2021learning} model, it can also be used for multimodal generation tasks. A text-guided diffusion models (DiffusionCLIP)~\cite{kim2022diffusionclip} shows the flexibility to add text guidance conditions. Latent diffusion
models (LDMs)~\cite{rombach2022high} can be trained on limited computational
resources using powerful pre-trained autoencoders in the latent space. 
Compared with GANs-based models, image generation based on diffusion models can easily add a variety of guidance, such as texts, strokes, and reference images. 
Since existing diffusion models are designed for natural image generation with random noise or text prompt. Some diffusion models can perform natural image colorization based on the prior knowledge of color in real world. They cannot be directly used for our task.
\meng{Very recently, Zhang et al.\cite{zhang2023adding} proposed ControlNet which can generate diversity colored cartoon images according to the sketch input and text prompt. Since it used diffusion models pre-trained on natural image dataset, it sometimes produced some distorted results compared to input sketches.
Our AnimeDiffusion is the first one to perform reference-based anime face line drawing colorization with accurate color information using diffusion models.}

\begin{figure*}[ht]
\centering  
\includegraphics[width=\linewidth]{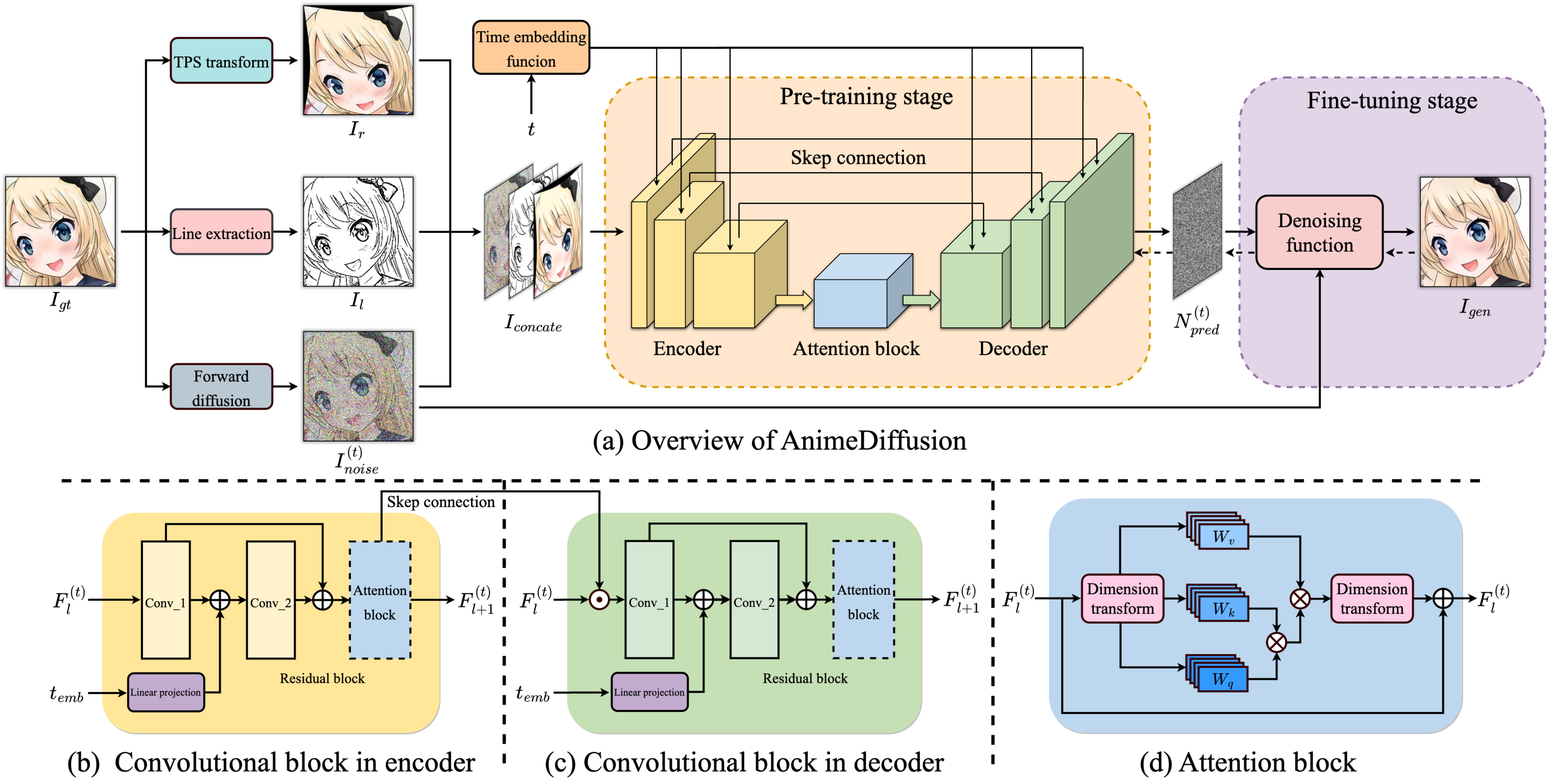}
\caption{The flowchart of AnimeDiffusion.}
\label{fig:framework}
\end{figure*}
\section{Overview}
Given a reference image, we aim to colorize the line drawing with clear geometry structure and accurate semantic color. The core problem to solve is how to inject color from the corresponding position of reference image into line drawing. Previous GANs-based methods usually design two encoders to extract line drawing features and reference image features respectively, and then use feature aggregation block to integrate two cross-domain features in the latent space. This operation can make model learn semantic correspondence ability to generate colored results, and one discriminator is needed to distinguish the generated results from real colored images which makes the colored results more realistic. However, these GANs-based methods \meng{suffer from} two problems, one is that line drawing features and reference images features are integrated in the latent space using complicated feature aggregation module. This part of network design is usually stacked with various novel neural network modules and it is not intuitive and explainable for algorithm designers, \meng{such as \cite{lee2020reference}, \cite{li2022eliminating}, \cite{liu2022reference}.}
The other one is that the weighted sum of multiple loss is employed\meng{\cite{johnson2016perceptual}, \cite{huang2018multimodal}} combined with adversarial loss, this makes it difficult to train the network and the training process is more unstable. Additional normalization techniques \meng{\cite{huang2017arbitrary},\cite{miyato2018spectral}} or improvements of the GAN model itself \meng{\cite{arjovsky2017wasserstein, gulrajani2017improved, mao2017least}} are needed to solve this problem.

\meng{However, diffusion models can fundamentally solve the above two problems.}
We formulate our training procedure as conditional noise prediction task, therefore, multiple condition images including line drawing and reference are directly concatenated in the pixel space. It is easier to extend diffusion models for other conditional generation tasks without the need to design special feature extractors. In addition, the loss function of diffusion models is simple and closely related to the training task. Combined with our designed hybrid training strategy, our model finally has excellent coloring ability. Without introducing additional discriminator network, the quality of the colorized images is extremely close to that of real colored images.

\section{AnimeDiffusion}
\subsection{Model Architecture}
As illustrated in Fig.~\ref{fig:framework}(a), assuming $I_{gt}$ is an original colored image, and an line drawing $I_{l}$ is extracted using XDoG~\cite{winnemoller2012xdog} extractor. The more detailed information about our data preparation will be introduced in section 5.1. Since there are usually large spatial structure discrepancy between the line drawing and reference image, in order to make AnimeDiffusion learn the accurate semantic correspondence ability during the training process, we apply \meng{TPS transformation\cite{24792}} to convert $I_{gt}$ to a geometry distorted version $I_{r}$. The forward diffusion goes from $I_{gt}$ to $I_{noise}^{(t)}$ with random $t$ step in the range of $T$. Then $I_{noise}^{(t)}$, $I_{l}$ and $I_{r}$ are concatenated together to comprise $I_{concate}$ with 7 channels. We build an U-Net to predict the noise with 3 channels that added onto the $I_{gt}$. \meng{We propose a novel conditional noise prediction proxy task for the pre-training stage by introducing 
$I_{l}$ and $I_{r}$ as additional conditional inputs.}
The information of $t$ is embedded using the time embedding function and is transmitted into all convolutional blocks of both encoder and decoder in the U-Net.
As is shown in Fig.~\ref{fig:framework}(b) and (c), the convolutional blocks in encoder and decoder have the same structure containing a residual block followed by an attention block. It is worth to note that we use dotted box to represent attention block is a selective usage. At the shallow layers of the encoder, we do not use attention block due to the large dimension of feature maps.
\meng{The encoder and decoder equipped with multi-head self-attention make our model efficiently capture global and local features in different convolutional layers.}
There is a linear projection module to map the embedded time information $t_{emb}$ to the one with the same size as the feature map after the first convolution operation. We use the common add operation to encode the time information into the convolutional block. The attention block is not only used as a sub-block in the encoder and decoder of U-Net, but also as an independent block in the bottleneck of U-Net. The detailed structure of attention block is illustrated in Fig.~\ref{fig:framework}(d). The use of attention block can make model learn long-range features and multi-scale features which is essential for our colorization task. Then we use denoising function to transfer the predicted noised $N_{pred}^{(t)}$ combined with $I_{noise}^{(t)}$ to the generated colored image $I_{gen}$.

\subsection{Line Extraction}
Due to the lack of a large amount of hand-drawn line data, it is quite time-consuming and laborious to expand the data volume by hand-drawn method. We use \meng{XDoG\cite{winnemoller2012xdog}} line style as the intermediate representation of line drawings during the training and inference stage. Any input line drawings created by artist will be automatically converted to XDoG style to fit the model. For this reason, we build a line extraction module integrated in AnimeDiffusion. During the training stage, reference color images are used as input to extract the line drawings, and during the inference stage, hand-drawn line drawings are used as input to transform the line draft style into XDoG style.

For given image $x$, the line extractor is described in the form\cite{winnemoller2012xdog}

\begin{equation}
\label{equ:xdog1}
    S_{\sigma, k, p}(x) = (1 + p) \cdot G_{\sigma}(x) - p \cdot G_{k\sigma}(x)
\end{equation}
where $G_{\sigma}$ and $G_{k\sigma}$ is Gaussian convolution operation,  $\sigma$ is the variance of Gaussian convolution kernel, $k$ is scaling ratio of the variance between two convolution, and $p$ is used to control the edge emphasis lines. 

We need a line extractor \meng{in which} line extraction results are as close as possible to the effect of a painter's hand-drawn line.
The variance $\sigma$ of the Gaussian convolution has a significant effect on the line thickness of the line drawing, and we choose $\sigma$ to be 0.3 to get a reasonable line width.
We hired a professional artist to draw line art for some of the images in our dataset. 
For a color image $I_c$, its line drawing extraction \meng{result} is $S_{k, p}(I_c)$, and the hand-drawn image by the painter is represented as $H(I_c)$.
The objective of parameters for the line extractor is

\begin{equation}
\label{equ:xdog2}
    \arg\mathop{\min}_{k, p}\ \sum_i \lVert S_{k, p}(I_i) - H(I_i)  \rVert_2^2
\end{equation}

\subsection{Training Strategy}
We design a hybrid training strategy to train AnimeDiffusion, which consists of classifier-free guidance pre-training stage and an image reconstruction guidance fine-tuning stage. This training strategy separates the denoising task from the image reconstruction task, makes the network learning a specific task at each stage, and is more beneficial to network training and weight updating. Each training step will be introduced in detail in \meng{section 4.3.1 and 4.3.2.}

\subsubsection{Pre-training Stage}

During the classifier-free guidance pre-training stage, AnimeDiffusion mainly learns denoising ability. 
As shown in Fig. \ref{fig:framework}, the original image $I_{gt}$ goes through a forward diffusion process which is a Markov chain since it adds Gaussian noise to $I_{gt}$ and obtains noisy image $I_{noise}^{t}$ for time step $t$ iteratively.
Each step of the forward process is a Gaussian transition.

\begin{equation}
\label{equ:1}
     q({I}_{noise}^{(t)} | {I}_{noise}^{(t-1)}) = \mathcal{N}({I}_{noise}^{(t)}; \sqrt{1 - \beta_t}{I}_{noise}^{(t-1)}, \beta_t \mathbf{I})
\end{equation}
where $\beta_t$ is variance schedule at time step $t$. 
The forward process of the diffusion model represents the addition of noise from step $0$ to $t$.
For the cumulative $t$ steps of noise addition $I_{noise}^{(1:T)}$, the marginal distribution is

\begin{equation}
\label{equ:2}
    q({I}_{noise}^{(1:T)} | {I}_{gt}) = \Pi_{t = 1}^T q({I}_{noise}^{(t)} | {I}_{noise}^{(t-1)}) 
\end{equation}
Under the condition of equation \ref{equ:1}, the marginal distribution of each forward step is a standard Gaussian distribution

\begin{equation}
\label{equ:3}
    q({I}_{noise}^{(t)} | {I}_{gt}) = \mathcal{N}({I}_{noise}^{(t)}; \sqrt{\bar{\alpha}_t} {I}_{gt}, (1 - \bar{\alpha}_t)\mathbf{I})
\end{equation}
where $\bar{\alpha}_t = \Pi_{i = 1}^t (1 - \beta_i)$. After $t$ times iteration, the result latent variable $I_{noise}^{(t)}$ can be simplified as

\begin{equation}
\label{equ:4}
    {I}_{noise}^{(t)} = \sqrt{\bar{\alpha}_t} {I}_{gt} + \sqrt{1 - \bar{\alpha}_t} \epsilon, \epsilon \sim \mathcal{N}(\mathbf{0}, \mathbf{I})
\end{equation}
The training objective of the model is to predict the noise $N_{pred}$=$\epsilon_\theta({I}_{l},{I}_{r},{I}_{noise}^{(t)}, t)$ with given noised data point ${I}_{noise}^{(t)}$, time step $t$ and condition $I_{l}$ and $I_{r}$, and optimizing the objective

\begin{equation}
    \label{equ:5}
    \mathbbm{E}_{{I}_{gt} \sim q({I}_{gt}), \epsilon \sim \mathcal{N}(\mathbf{0}, \mathbf{I}), {I}_{l},{I}_{r}, t} \Vert \epsilon - \epsilon_\theta({I}_{l},{I}_{r}, {I}_{noise}^{(t)}, t) \Vert_p^p
\end{equation}

For a simplified description, we subsequently use $\epsilon_\theta$ represent $\epsilon_\theta({I}_{l},{I}_{r}, {I}_{noise}^{(t)}, t)$.
Palette\cite{saharia2022palette} indicates that the L2 norm can capture the output distribution more faithfully, and we adopt $p = 2$ in our pre-training stage.

We used a model of U-Net with attention blocks to predict the noise $\epsilon$ added in equation \ref{equ:4}.
The U-Net needs to accept line drawing $I_l$, reference image $I_r$, and noisy image $I_{noise}^{t}$ as inputs. 
Considering the high spatial consistency between line drawing and color images, we concatenate the above three in the channel dimension.
And the subsequent experimental in section \ref{sec:exp} results demonstrate that without a complex feature fusion mechanism, the model we proposed can achieve the semantic correspondence between the line drawings and the reference maps. 
\meng{Our previous work\cite{cao2022attention} for the first time emphasized the importance of semantic correspondence for reference-based line drawing colorization. The method in this paper uses clever algorithm design to achieve more amazing results in accurate semantic correspondence, especially in the region of anime character face.}

\subsubsection{Fine-tuning Stage}
After training $\epsilon_\theta({I}_{l},{I}_{r},{I}_{noise}^{(t)}, t)$, diffusion models inference through the learned reverse process. Since the result distribution of forward process $p({I}_{noise}^{(T)})$ approximates a standard Gaussian distribution $\mathcal{N}(\mathbf{0}, \mathbf{I})$, the sampling process starts from pure Gaussian noise, followed by $T$ rounds of denoising.
On the one hand, training diffusion models often requires a large batch size and long iteration rounds with large computing consumption.
On the other hand, we want to strike a balance between the diversity and accuracy of generated results. 
Based on the pre-trained model already having some denoising ability, we introduce the image reconstruction guidance fine-tuning stage to improve the generation ability of AnimeDiffusion. 
Since the great diversity of the generated results of the diffusion model, as the training iterations grow, the quality of generated images is affected less by the input noise and more by the guidance condition.
Therefore we input the noise generated according to the reference image instead of random noise in the fine-tuning and inference process of the model.
According to equation \ref{equ:4}, $I_{gt}$ is estimated as

\begin{equation}
\label{equ:inf_x0}
    \widetilde{I}_{gt} = \frac{1}{\sqrt{\bar{\alpha}_t}}({I}_{noise}^{(t)} - \sqrt{1 - \bar{\alpha}_t} \epsilon_\theta({I}_{l},{I}_{r},{I}_{noise}^{(t)}, t))
\end{equation}
The mean value of reverse process $p_\theta({I}_{noise}^{(t-1)} | {I}_{noise}^{(t)}, {I}_{l}, {I}_{r})$ is parameterize as

\begin{equation}
\label{equ:inf_mu}
    \widetilde{\mu}_\theta({I}_{noise}^{(t)}, t) = \frac{\sqrt{\bar{\alpha}_{t - 1}} \beta_t}{1 - \bar{\alpha}_t} \cdot \widetilde{I}_{gt} + \frac{(1 - \bar{\alpha}_{t - 1}) \sqrt{\alpha_t}}{1 - \bar{\alpha}_t} \cdot {I}_{noise}^{(t)}
\end{equation}
With the estimation of $\mu_\theta({I}_{noise}^{(t-1)}, t)$, each iteration of reverse process is

\begin{equation}
\label{equ:inf_ddpm}
    {I}_{noise}^{(t-1)} = \widetilde{\mu}_\theta({I}_{noise}^{(t)}, t) + \sigma_t^2 \epsilon, \epsilon \sim \mathcal{N}(\mathbf{0}, \mathbf{I})
\end{equation}
where $\sigma_t$ is the sampling variance with $\sigma_t^2 = \frac{1 - \bar{\alpha}_{t - 1}}{1 - \bar{\alpha}_t}\beta_t$.

The time consumption and space consumption for fine-tuning by directly adopting the reverse process of DDPM is vast, so we use DDIM\cite{song2020denoising} as our denoising function, which is an alternative non-Markov chain denoising process with different sampling or reverse process

\begin{equation}
\label{equ:inf_ddim}
    {I}_{noise}^{(t' - 1)} = \sqrt{\bar{\alpha}_{t' - 1}}\widetilde{I}_{gt} + \sqrt{1 - \bar{\alpha}_{t' - 1} - \eta \sigma_{t'}^2}\epsilon_\theta + \eta \sigma_{t'} \epsilon
\end{equation}
DDIM obtains a sub-sequence of time $[0, T)$, where $t'$ is sampled time sequence\cite{song2020denoising}. 
$\eta$ is a hyper-parameter that controls whether noise is added during the reverse process. If $\eta$ is set to $0$, the process of image generation is deterministic.

Since both the forward and reverse processes of DDPM are random, the colorization results are different for the same sample. 
DDIM provides a deterministic reverse sampling strategy, but due to the different initial noise, there is no guarantee that the image reconstruction can be completed with the original image as the reference image.
To fully utilize the image synthesis performance of the diffusion model for image processing purposes, we borrow the deterministic forward process from DiffusionCLIP\cite{kim2022diffusionclip} in our fine-tuning stage.
According to equation \ref{equ:inf_x0} and \ref{equ:inf_ddim}, DDIM is condidered as an Euler method to solve an \meng{ordinary differential equation (ODE)}

\begin{equation}
    \label{equ:ode}
    \mathrm{d}\frac{{I}_{noise}^{(t)}}{\sqrt{\bar{\alpha}_t}} = \epsilon_\theta \cdot \mathrm{d}\sqrt{\frac{1 - \bar{\alpha}_t}{\bar{\alpha}_t}}
\end{equation}

The above ODE holds in a finite number of steps, and to obtain an accelerated forward process, we use the same sampling time series $t'$ as the reverse process of DDIM.
The recursive relation from $I_{noise}^{(t')}$ to $I_{noise}^{(t' + 1)}$ is simplified as

\begin{equation}
    I_{noise}^{(t' + 1)} = \sqrt{\bar{\alpha}_{t' + 1}}\widetilde{I}_{gt}(I_{noise}^{(t')}) + \sqrt{1 - \bar{\alpha}_{t' + 1}}\epsilon_\theta
\end{equation}
$\widetilde{I}_{gt}(I_{noise}^{(t')})$ represents the ground truth estimation function of equation \ref{equ:inf_x0}.
Accordingly, to obtain the deterministic reverse process, we apply $\eta$ as 0 in the reverse process of DDIM.

After several iterations, the denoising function finally generates the colored image $I_{gen}$. We calculate the mean square error (MSE) between $I_{gen}$ and $I_{gt}$, in order to constrain the generated image as close to the original image as possible in pixel level. It is worth noted that the dotted arrow in Fig. \ref{fig:framework} represents reverse gradient propagation to update the parameters of U-Net during the fine-tuning stage.
\begin{eqnarray}
    L_{rec} = \mathbbm{E}_{{I}_{gt} \sim q({I}_{gt}), \epsilon \sim \mathcal{N}(\mathbf{0}, \mathbf{I})}\parallel I_{gen} - I_{gt} \parallel_p^p
\end{eqnarray}

\renewcommand{\algorithmicrequire}{\textbf{Input:}}
\renewcommand{\algorithmicensure}{\textbf{Output:}}

\begin{algorithm}
\caption{Anime Diffusion Fine-tuning}
\begin{algorithmic}[1]
\REQUIRE{\, $\epsilon_\theta$ pretrained noise prediction model, \\
\qquad $\mathcal{I}$ training anime face images, \\
\qquad $S$ number of sampling steps 
}
\ENSURE{$\hat{\epsilon}_\theta$ fine-tuned noise prediction model}
\STATE Initialize list of forward noisy images $\hat{\mathcal{I}}$;
\STATE Sampling a increasing sub-sequence $\mathcal{T}' = \{t'_{0}, \cdots T'\}$ of length $S$;
\FOR{$I_{gt}$ \textbf{in} $\mathcal{I}$}
    \FOR{$t'$ \textbf{in} $\mathcal{T}'$}
        \STATE Calculate $\widetilde{\epsilon} \leftarrow \epsilon_\theta(I_l, I_r, I_{noise}^{t'}, t')$;
        \STATE Predict the ground truth $\widetilde{I}_{gt}(I_{noise}^{t'}, t')$;
        \STATE Forward step $I_{noise}^{(t' + 1)} \leftarrow \sqrt{\bar{\alpha}_{t' + 1}} \widetilde{I}_{gt} + \sqrt{1 - \bar{\alpha}_{t' + 1}}\widetilde{\epsilon}$;
    \ENDFOR
    \STATE Update $\hat{\mathcal{I}}$;
\ENDFOR
\FOR{$I_{noise}^{T'}$ \textbf{in} $\mathcal{\hat{I}}$}
    \FOR{$t'$ \textbf{in} \texttt{reverse}($\mathcal{T}'$)}
        \STATE Calculate $\widetilde{\epsilon} \leftarrow \epsilon_\theta(I_l, I_r, I_{noise}^{t'}, t')$;
        \STATE Predict the ground truth $\widetilde{I}_{gt}(I_{noise}^{t'}, t')$;
        \STATE Reverse step $I_{noise}^{(t' - 1)} \leftarrow \sqrt{\bar{\alpha}_{t' - 1}}\widetilde{I}_{gt} + \sqrt{1 - \bar{\alpha}_{t' - 1}}\widetilde{\epsilon}$;
    \ENDFOR
    \STATE Update reconstruction loss $L_{rec}$;
    \STATE Gradient step $\nabla_{\hat{\epsilon}_\theta} L_{rec}$
\ENDFOR
\end{algorithmic}
\end{algorithm}

We perform a small number of steps of fine-tuning based on the pre-trained model, as shown in algorithm 1. 
\meng{In summary, we perform our proposed hybrid training strategy to reduce computation consumption of training and inference significantly in comparison with the traditional manner of diffusion method.}
\daniel{The training objective of the pre-training stage is to obtain the solution of equation \ref{equ:ode}, the derivative of the path from the initial distribution to the target distribution at time step $t$.
When applying equation \ref{equ:inf_ddim} for inference, the distribution of the sub-sequences of time step $\{t'\}$ will affect the colorization results due to the deviation between the predicted derivatives $\epsilon_\theta$ and the true path from time step $t'$ to $t'-1$.
Therefore, we fix the sub-sequence $t'$ in the fine-tuning stage and select a smaller number of sampling steps to make our model save inference time.
Fine-tuning the pre-trained model so that the noise prediction at time step $t'$ is closer to the direction pointing to the next time step $t' - 1$, rather than accurately predicting the derivative.}

\subsection{User Interface}

\begin{figure}[htb]
\centering  
\includegraphics[width=0.95\linewidth]{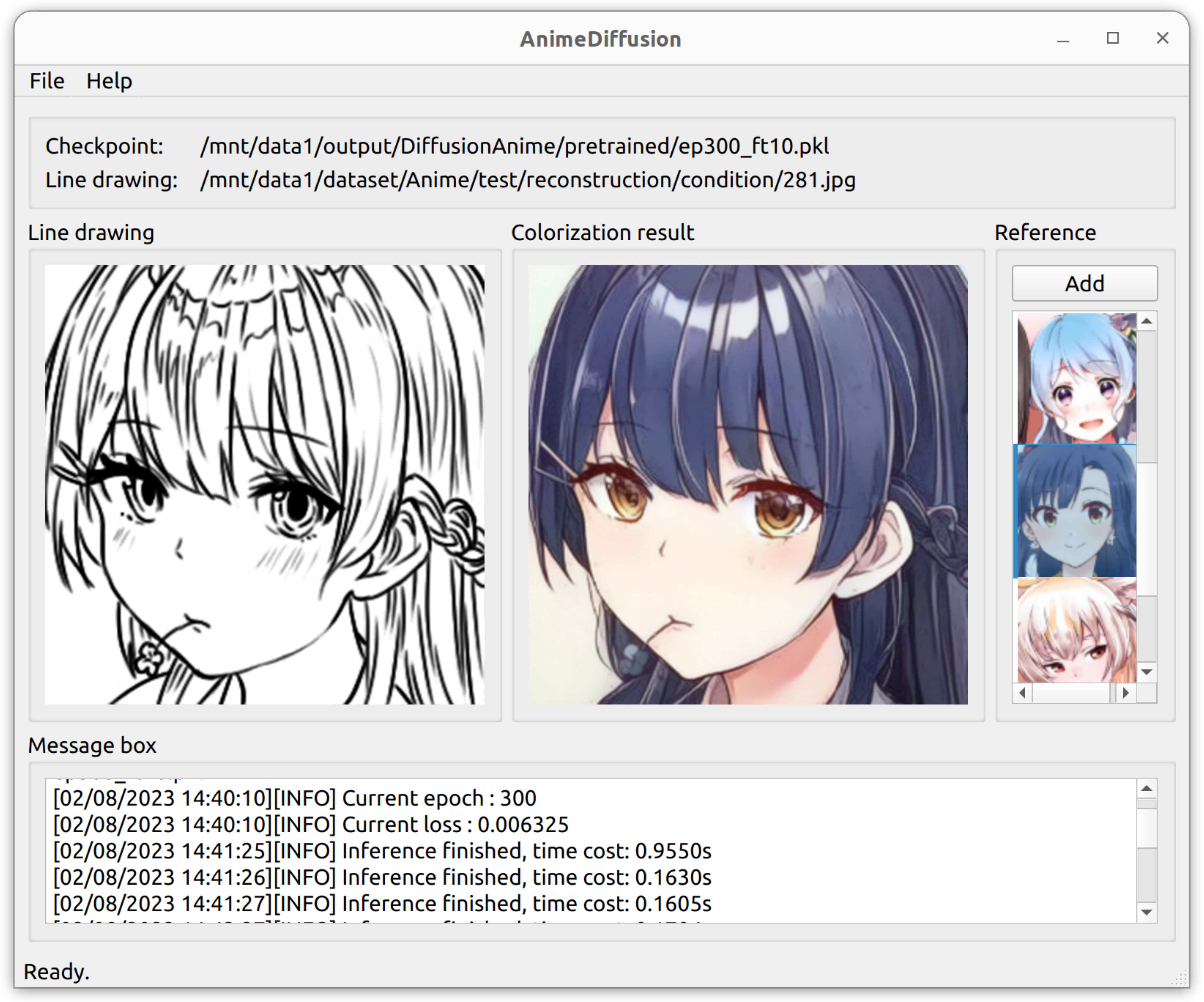}
\caption{User interface of AnimeDiffusion for colorizing anime face line drawings.}
\label{fig:ui}
\end{figure}

As is shown in Fig. \ref{fig:ui}, a user interface is developed for users to perform line drawing colorization by our AnimeDiffusion. Users only need to provide the line drawing and the reference image as two inputs of AnimeDiffusion, then it can one-key automatically complete the coloring process to generate colored results without additional human intervention. The elapsed time of the colorizing operation is printed at the bottom of the interface.
\daniel{Owing to the adoption of the DDIM acceleration sampling strategy and the fine-tuning stage, our method generated high-quality colorization results with only a small number of sampling steps.
The colorization time consumption for a line drawing can be controled within 0.2s on a machine equipped with RTX 4090, excluding the initialization period.}
Our end-to-end AnimeDiffusion model can be directly integrated into the practical colorization pipeline in the animation creation industry. 
\meng{In contrast to other diffusion methods\cite{zhang2023adding},\cite{saharia2022palette}, our method can accurately edit face line drawings according to reference images. Especially as shown in Fig.\ref{fig:teaser}, the artist added a lot of detail lines to the hair, our AnimeDiffusion can also add detailed textures instead of just flat color to the hair after coloring.}

\section{Experiments}
\label{sec:exp}

\subsection{Dataset}
In this paper, we focus on the anime face line drawing colorization task. In order to train AnimeDiffusion, a large dataset of anime face images is necessary, and the image resolution should not be too low in order to adequately express the color and detail information of the face. However, there is no dataset that meets our needs and can be directly used for training. So we build a benchmark dataset for anime face line drawing colorization. All anime character images are collected from Danbooru2020\cite{danbooru2020}, which is a large-scale anime image database with 4.2m+ images.  According to our task requirements, we only cut out the face part. After simple manual alignment and denoising operation, a total of 31696 training data and 579 testing data are produced. Due to limitations in GPU memory and model computing efficiency, all images are resized to $256 \times 256$ resolution. To simulate the manual line drawing style by artists and generate paired line drawing images, we use XDoG\cite{winnemoller2012xdog} to extract line drawings from colored anime images and set the parameters of XDoG algorithm with $\phi = 1 \times 10^9$ to keep a step transition at the border of lines in line drawings. We randomly set $\sigma$ to be $0.3/0.4/0.5$ to get different levels of line thickness, which generalizes AnimeDiffusion on various line widths to avoid overfitting. And we set $p=9, k=4.5, \epsilon=0.01$ in XDoG. 
\meng{As mentioned before, in the practical colorization scenario, there are large space discrepancies between the target line drawing and reference image.}
In order to make AnimeDiffusion learn accurate semantic correspondence ability during training and \meng{avoid learning trivial solution by directly using pixel aligned training data, we randomly set the parameters of TPS transformation on colored reference images when loading training data, i.e. each image of one batch data will have different geometry distortions. To some extent, this is a data augmentation trick.}

\subsection{Implementation Details}
We implement our AnimeDiffusion model based on the PyTorch framework, and it is trained on 1 NVIDIA A100 GPU. All input image size is fixed at $256 \times 256$. 
For the diffusion hyper-parameters setting, we use a linear noise schedule of $(1e^{-6}, 1e^{-2})$ with 1000 time steps. 
We pre-train the model with a batch size of 32 for 300 epochs, and we don't find overfitting, and we fine-tune the model with a batch size of 4 for 1 epoch. 
\daniel{On our devices, the pre-training stage \daniel{takes} 40 hours and the fine-tuning stage \daniel{takes} 110 minutes.}
We apply the Adam optimizer with a learning rate of $1e^{-5}$ for both of the above stages. Besides, we have no other hyper-parameters to adjust, like learning rate decay or warm-up schedule.

\subsection{Qualitative Evaluation}
We compare our AnimeDiffusion with another three state-of-the-art GANs-based methods. 
Lee et al.~\cite{lee2020reference} proposed the self-augmented supervised training strategy and designed a model with an attention based Spatial Correspondence Feature Transfer (SCFT) module. We regard it as the baseline for line drawing colorization task. 
Li et al.~\cite{li2022eliminating} designed a Stop-Gradient Attention (SGA) module to  eliminate the gradient conflict among attention branches. 
Cao et al.~\cite{cao2022attention} proposed an attention-aware improved method based on \cite{lee2020reference}, which focus on the anime line drawing colorization task. 
Since the variety of anime characters' faces and combine with the actual needs of anime character creation, we set two anime cases separately including anime face with homochromatic pupils and anime face with heterochromatic pupils. The latter is a very challenging case, which needs model require a high precision extraction of local features and semantic correspondence. To the best of our knowledge, we are the first learning based work can generate results with accurate color in pupils according to the reference image with no extra eyes segmentation label~\cite{9143503} or pupil position estimation network~\cite{akita2020colorization}.

\begin{figure}[ht]
    \centering
    \subfigure[]{
    \begin{minipage}[b]{0.31\linewidth}
        \includegraphics[width=0.95\linewidth, cframe={black 0.75pt}]{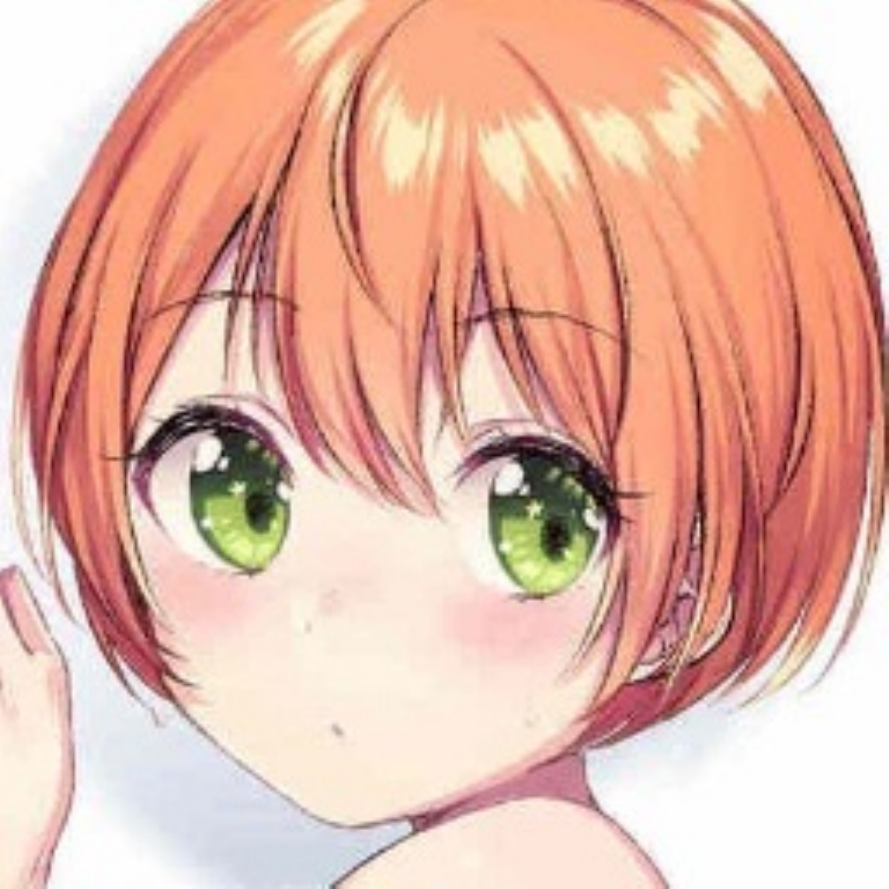}
    \end{minipage}
    }\hspace{-3pt}
    \subfigure[]{
    \begin{minipage}[b]{0.31\linewidth}
        \includegraphics[width=0.95\linewidth, cframe={black 0.75pt}]{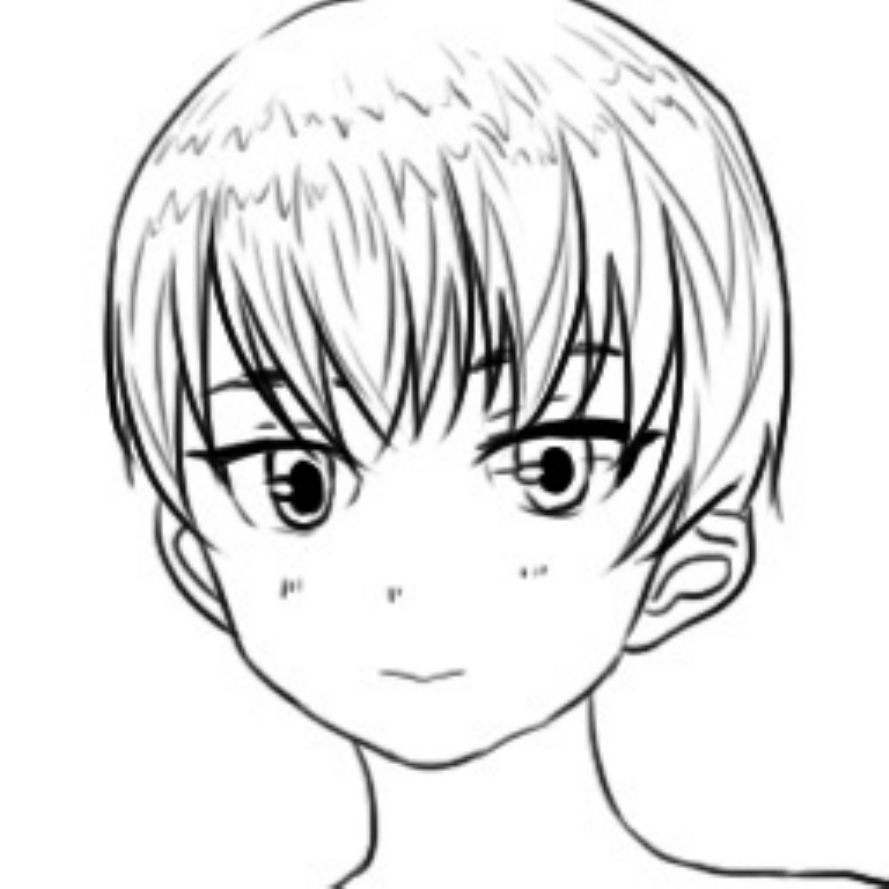}
    \end{minipage}
    }\hspace{-3pt}
    \subfigure[]{
    \begin{minipage}[b]{0.31\linewidth}
        \includegraphics[width=0.95\linewidth, cframe={black 0.75pt}]{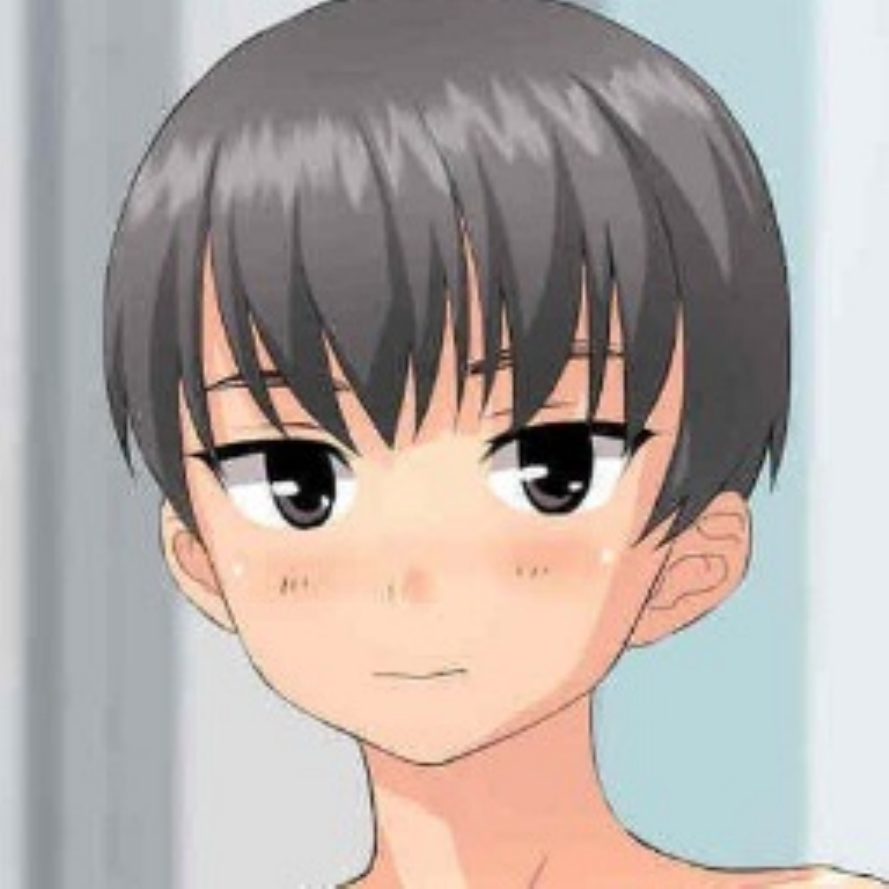}
    \end{minipage}
    } \\
    \subfigure[]{
    \begin{minipage}[b]{0.24\linewidth}
        \includegraphics[width=0.95\linewidth, cframe={black 0.75pt}]{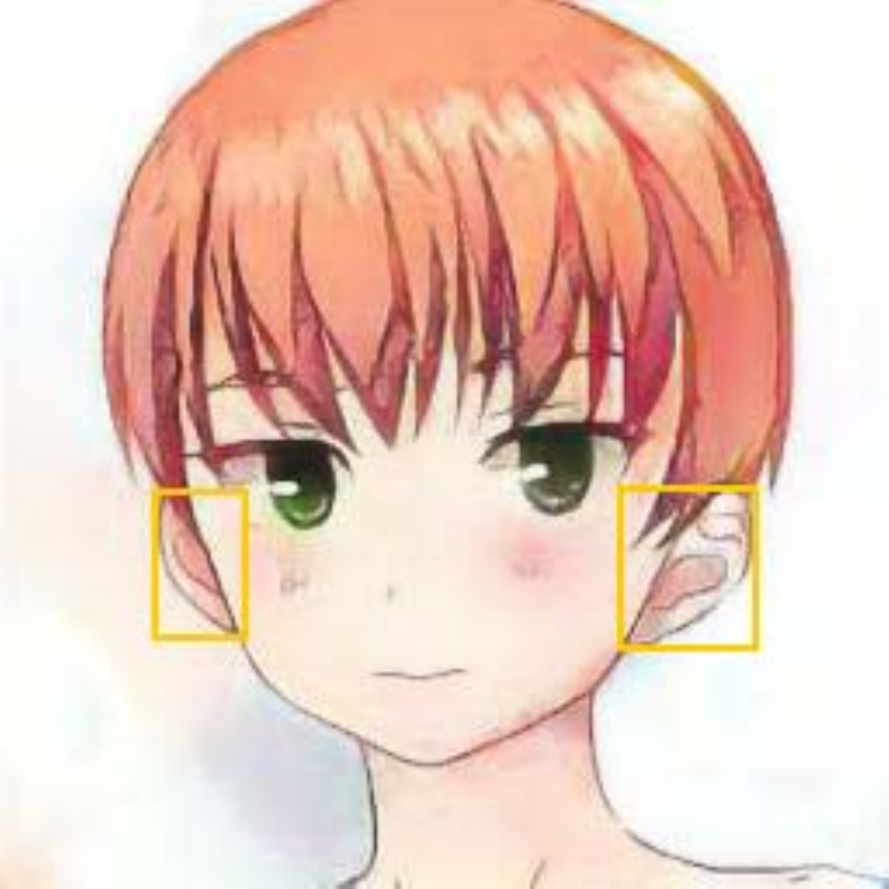}\\
        \includegraphics[width=0.95\linewidth, cframe={green 0.75pt}]{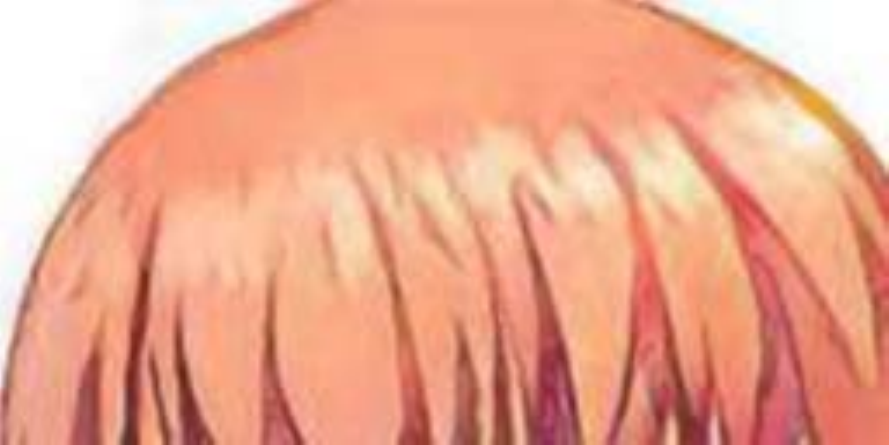}\\
        \includegraphics[width=0.95\linewidth, cframe={blue 0.75pt}]{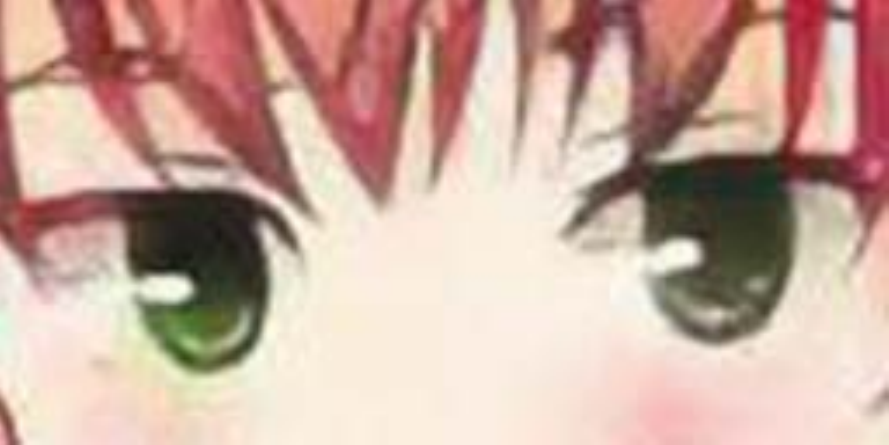}
    \end{minipage}
    }\hspace{-8pt}
    \subfigure[]{
    \begin{minipage}[b]{0.24\linewidth}
        \includegraphics[width=0.95\linewidth, cframe={black 0.75pt}]{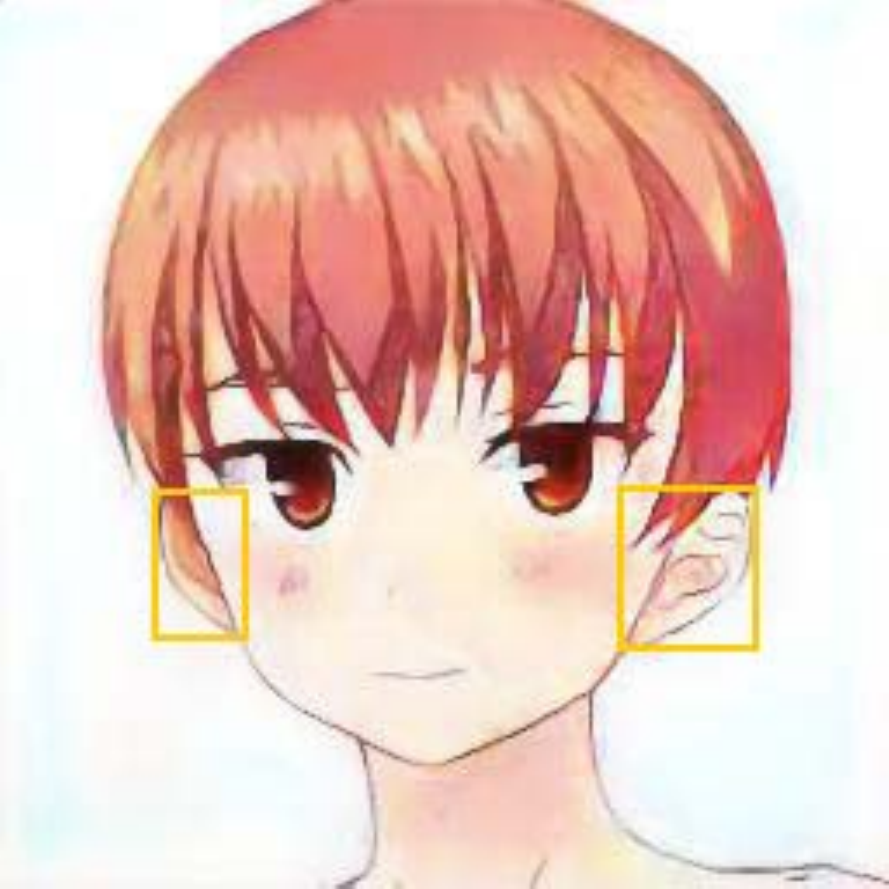}\\
        \includegraphics[width=0.95\linewidth, cframe={green 0.75pt}]{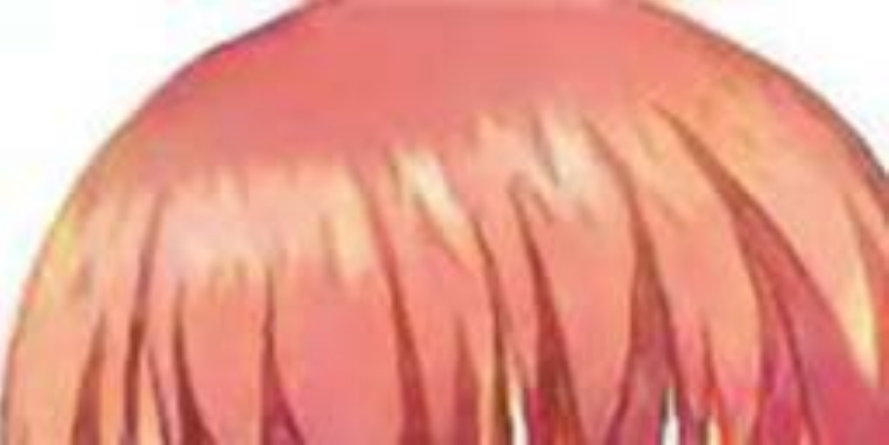}\\
        \includegraphics[width=0.95\linewidth, cframe={blue 0.75pt}]{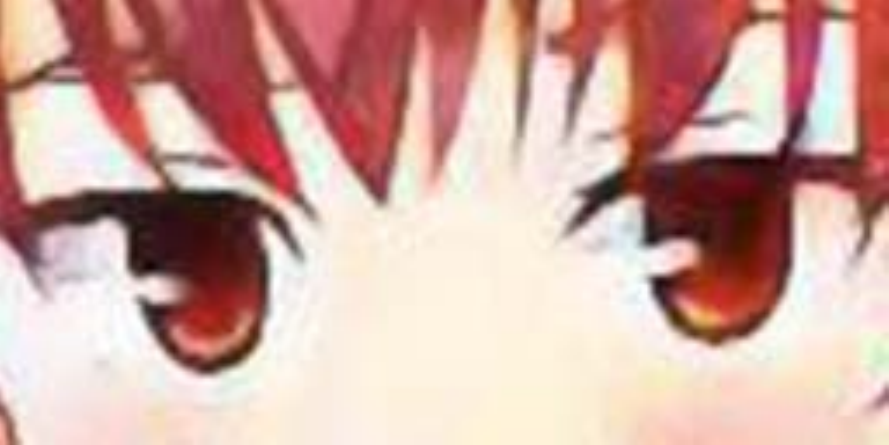}
    \end{minipage}
    }\hspace{-8pt}
    \subfigure[]{
    \begin{minipage}[b]{0.24\linewidth}
        \includegraphics[width=0.95\linewidth, cframe={black 0.75pt}]{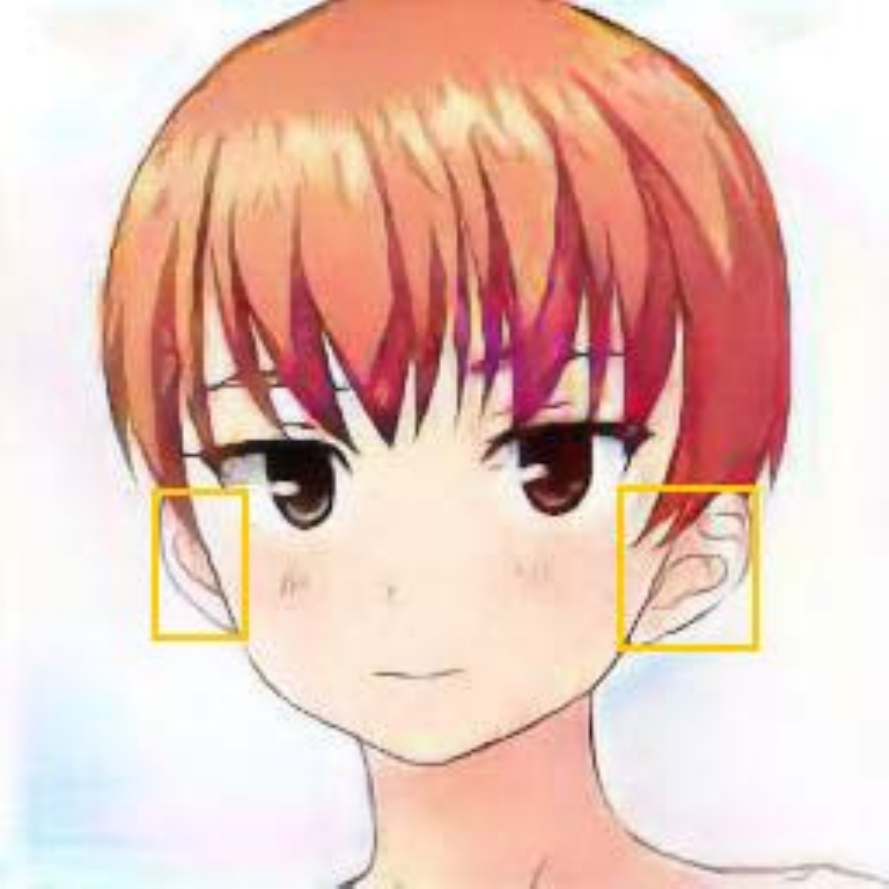}\\
        \includegraphics[width=0.95\linewidth, cframe={green 0.75pt}]{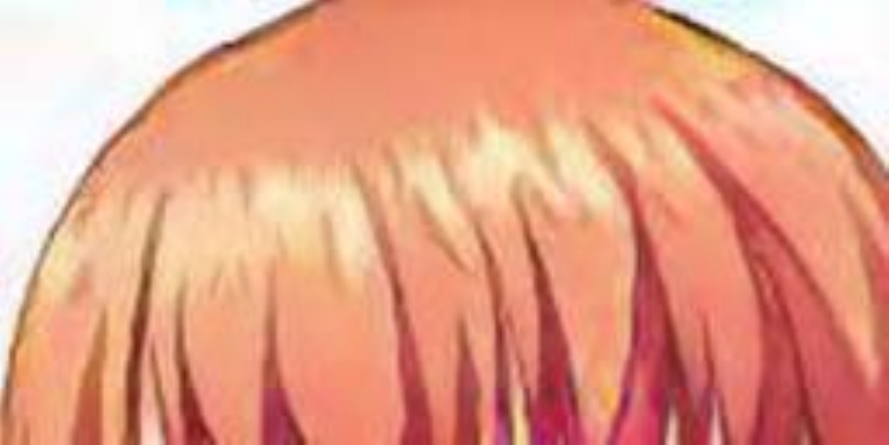}\\
        \includegraphics[width=0.95\linewidth, cframe={blue 0.75pt}]{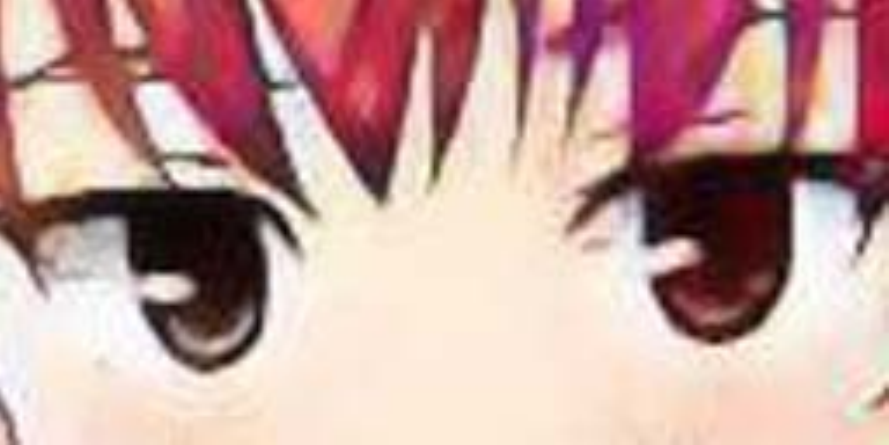}
    \end{minipage}
    }\hspace{-8pt}
    \subfigure[]{
    \begin{minipage}[b]{0.24\linewidth}
        \includegraphics[width=0.95\linewidth, cframe={black 0.75pt}]{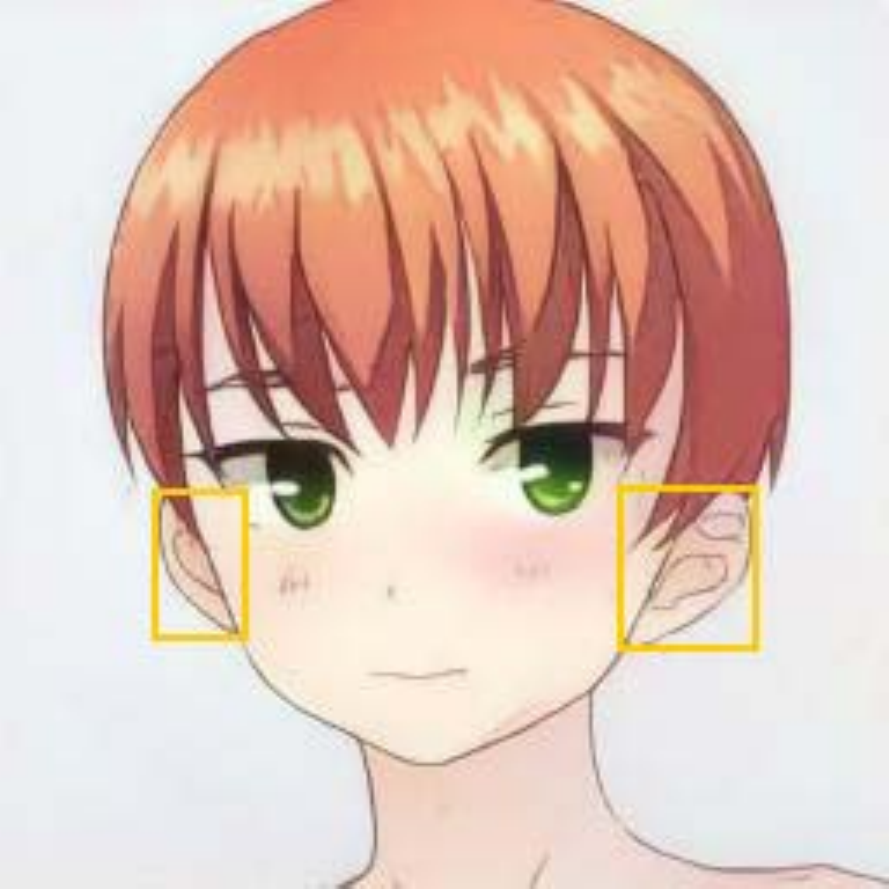}\\
        \includegraphics[width=0.95\linewidth, cframe={green 0.75pt}]{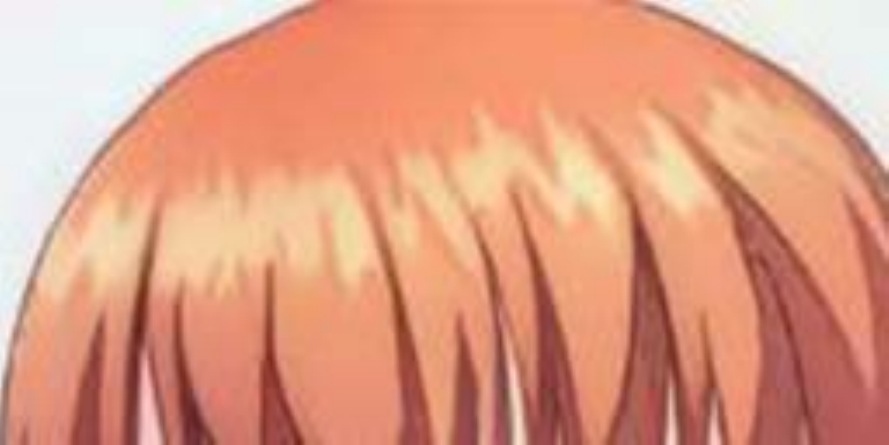}\\
        \includegraphics[width=0.95\linewidth, cframe={blue 0.75pt}]{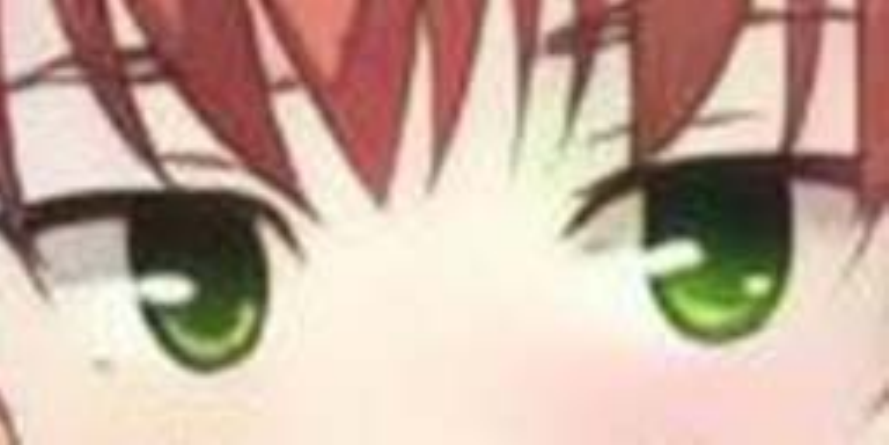}
    \end{minipage}
    }
    \caption{Detailed comparison of colorization results. (a) reference image, (b) line drawing, (c) original color image, (d) Lee et al.\cite{lee2020reference}, (e) Li et al.\cite{li2022eliminating}, (f) Cao et al.\cite{cao2022attention}, (g) AnimeDiffusion }
    \label{fig:detail_single_color}
\end{figure}

For homochromatic pupils case, we show detailed comparison results in Fig.~\ref{fig:detail_single_color}. Yellow region shows that AnimeDiffusion can recognize the ear semantic information from line drawing and inject the right color the same as face. Green region shows that AnimeDiffusion can maintain the light-reflecting effect compared with the original color image (Fig.~\ref{fig:detail_single_color}(c)). Blue region indicates that AnimeDiffusion can accurately transfer the color information from reference image (Fig.~\ref{fig:detail_single_color}(a)) into the line drawing (Fig.~\ref{fig:detail_single_color}(b)) in the eyes part. However, the other three methods have flaws in the areas we have marked in three colors.
\meng{Since the diffusion models are based on maximum likelihood estimation method that can estimate the probability density more accurately than GANs-based method. So our results are much clearer than the other three \daniel{methods}. The performance of feature aggregation modules in other three models is not the best, so the coloring effect is defective.
Combined with our proposed training strategy, denoising and reconstruction tasks are separated during the training procedure, making the network training more stable. Therefore, our model acquires better detailed features capture ability.}

\begin{figure*}[p]
    \centering
    \subfigure[]{
    \begin{minipage}[b]{0.125\linewidth}
        \includegraphics[width=\linewidth]{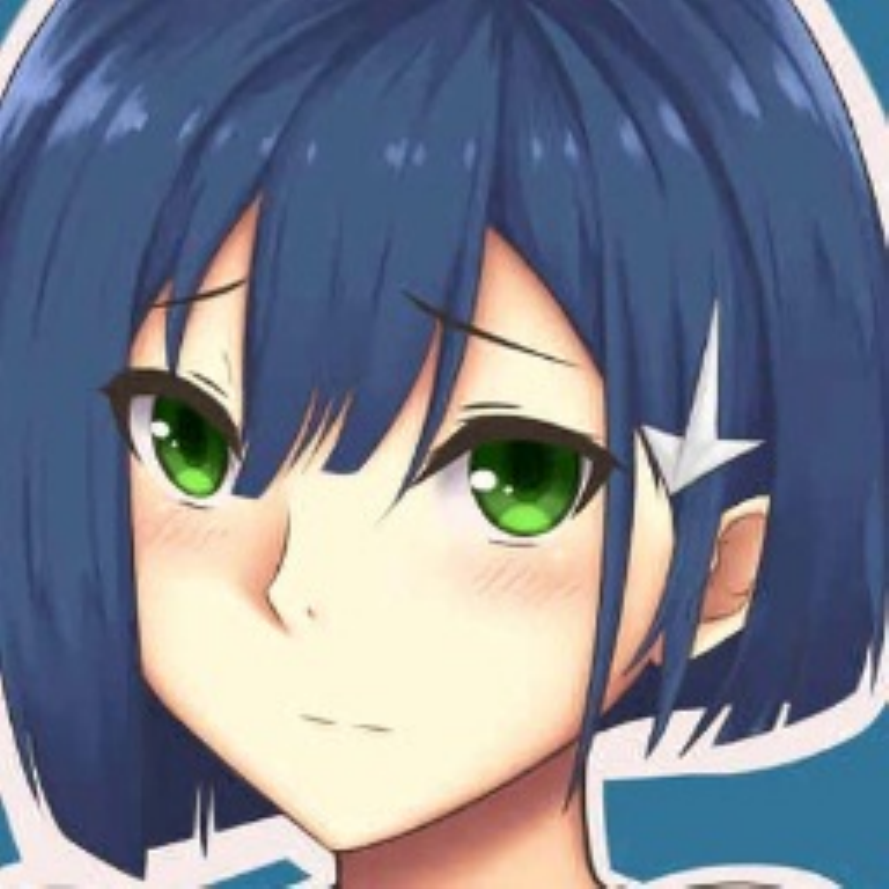}\vspace{4pt}
        \includegraphics[width=\linewidth]{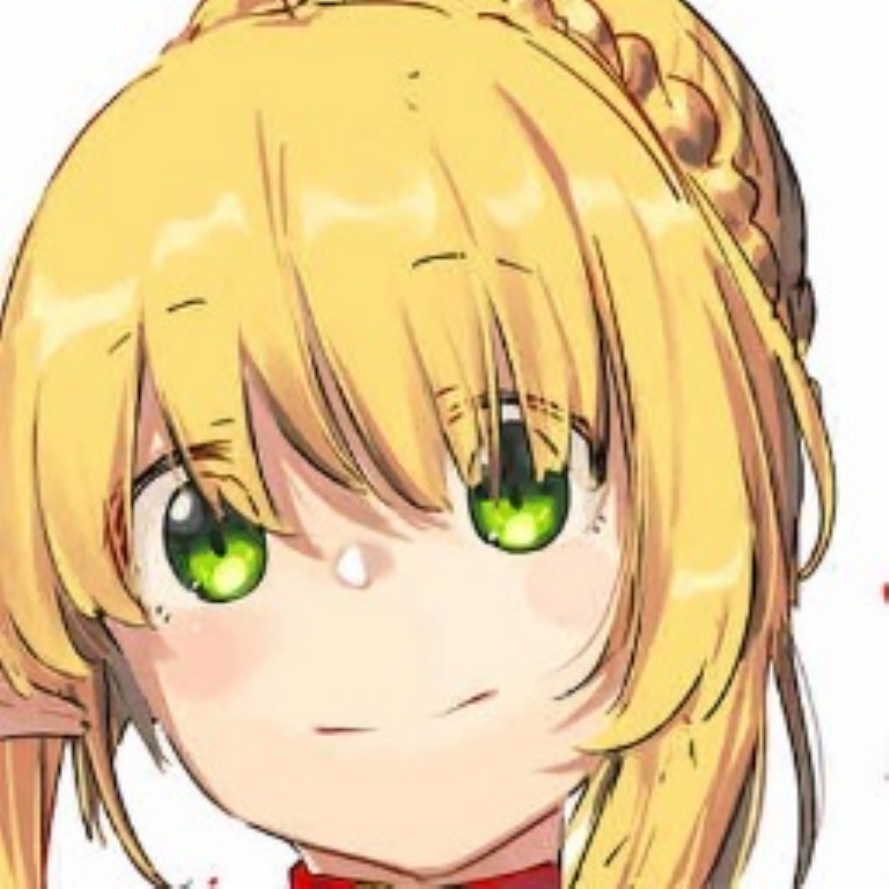}\vspace{4pt}
        \includegraphics[width=\linewidth]{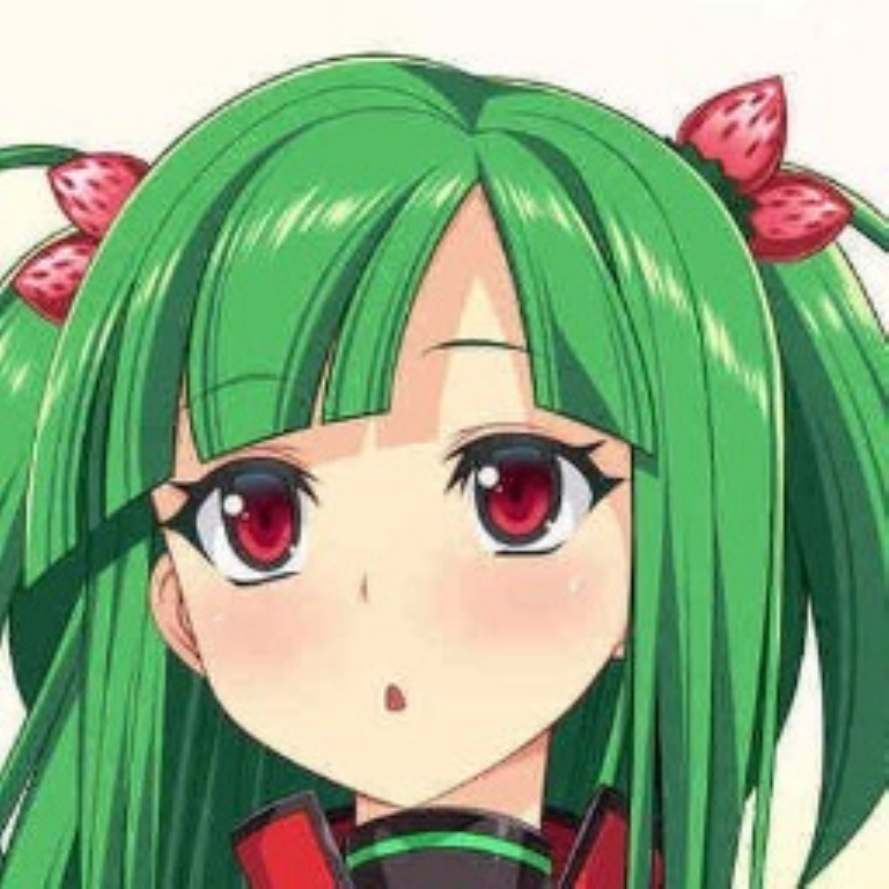}\vspace{4pt}
        \includegraphics[width=\linewidth]{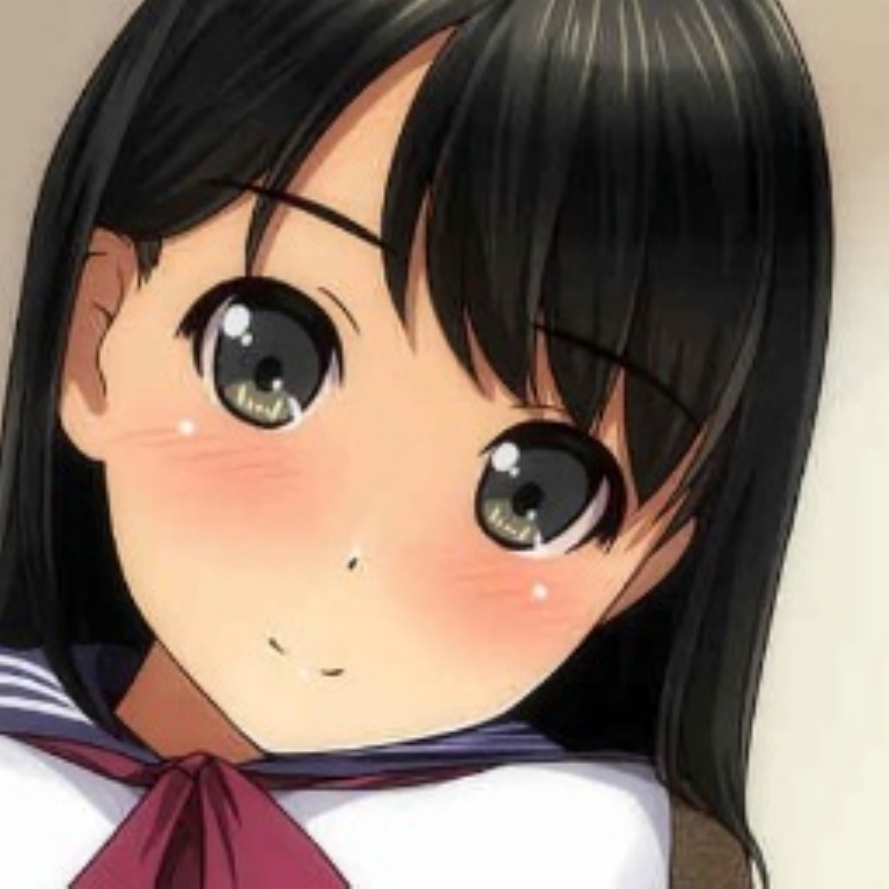}\vspace{4pt}
        \includegraphics[width=\linewidth]{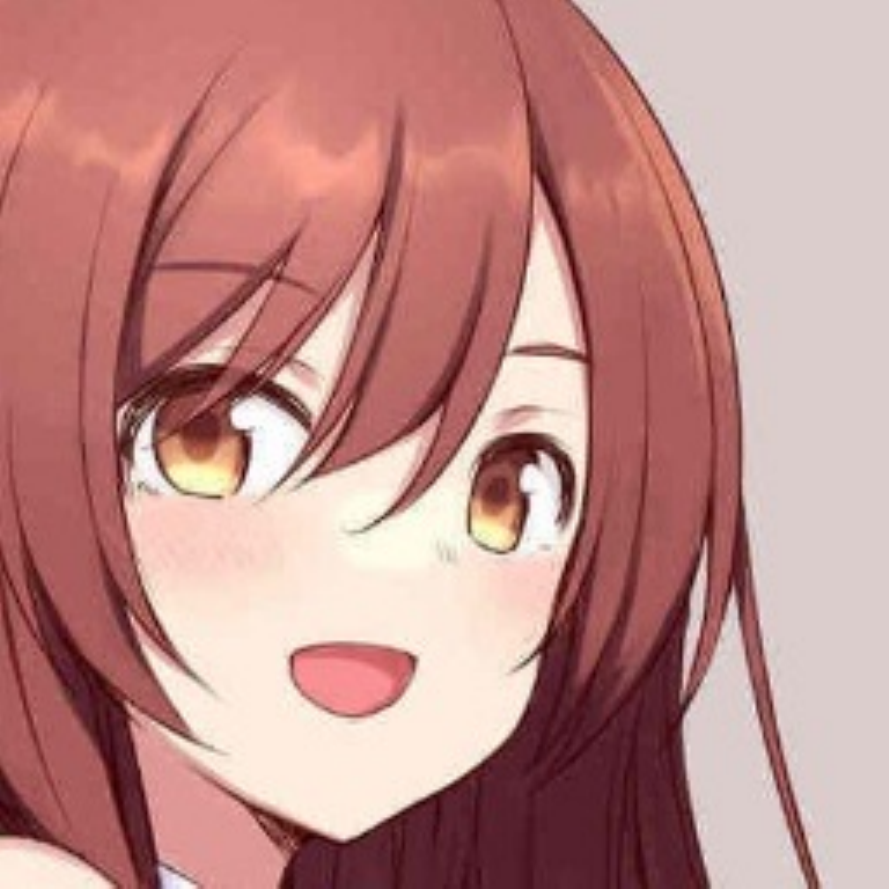}\vspace{4pt}
        \includegraphics[width=\linewidth]{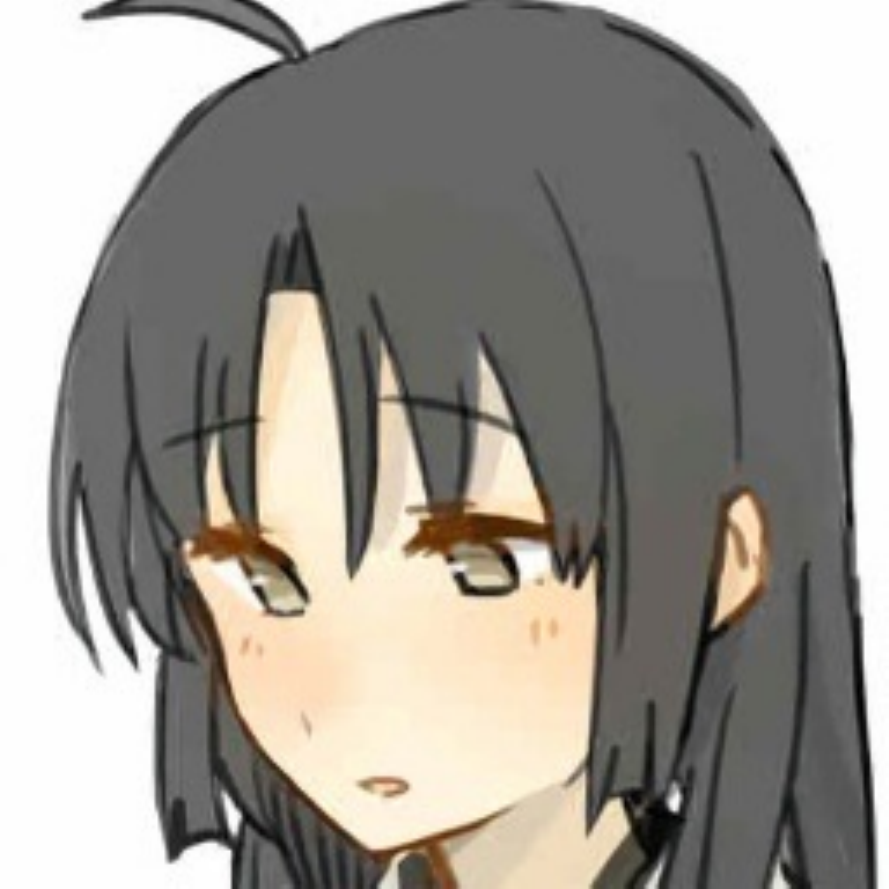}\vspace{4pt}
        \includegraphics[width=\linewidth]{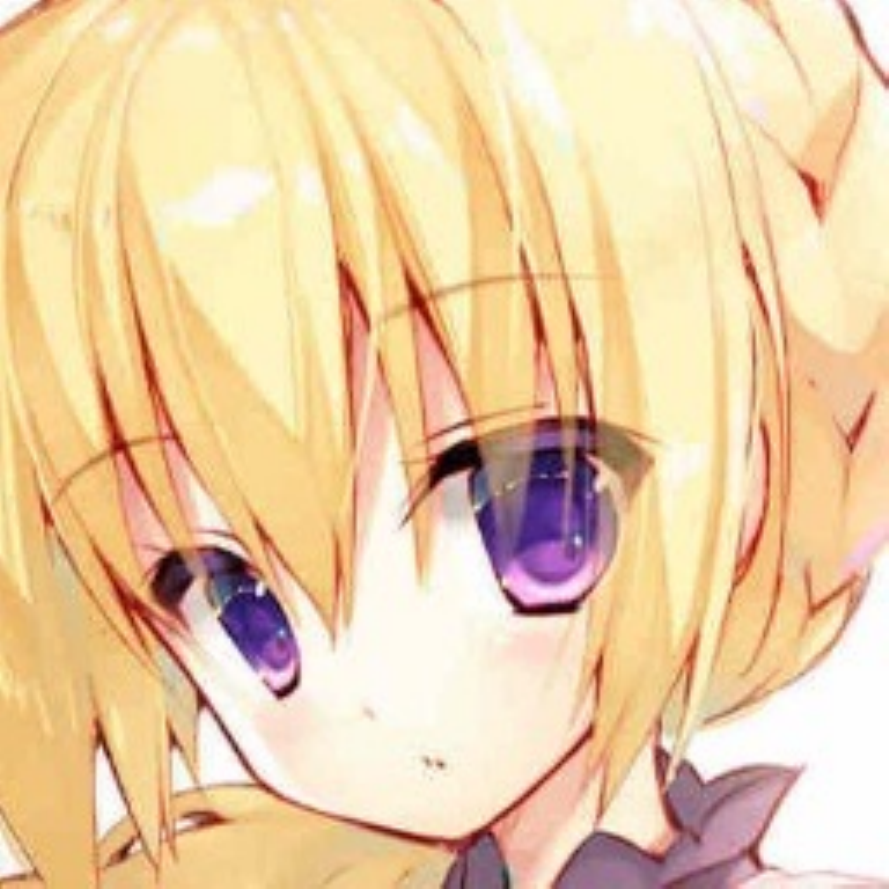}\vspace{4pt}
        \includegraphics[width=\linewidth]{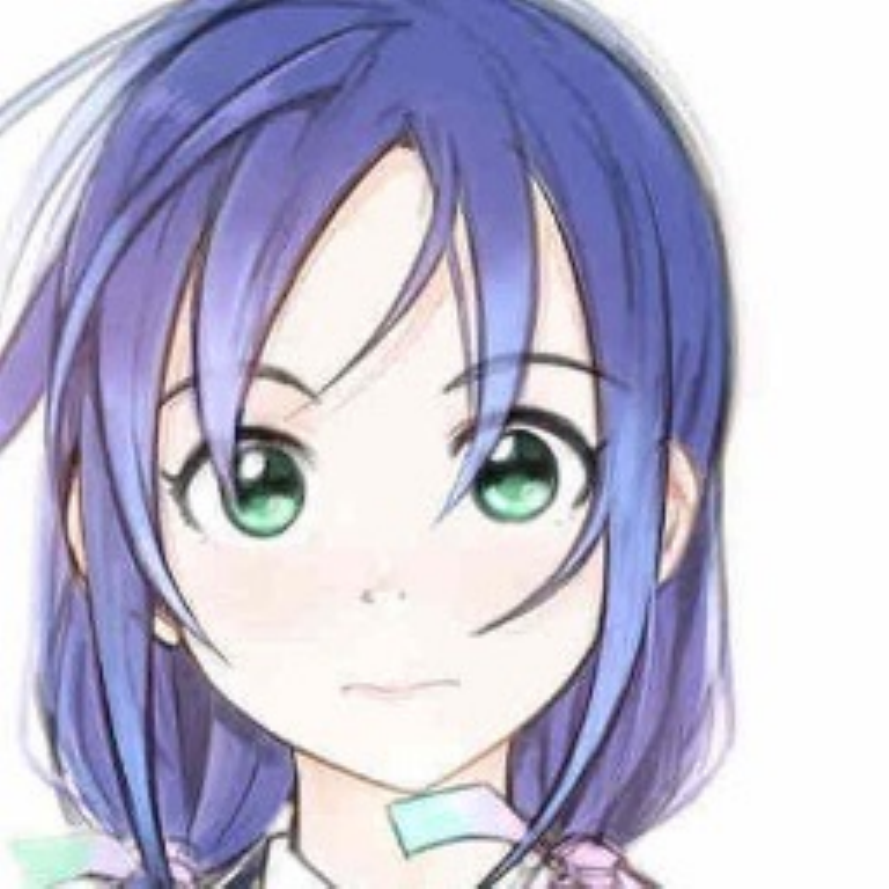}\vspace{4pt}
        \includegraphics[width=\linewidth]{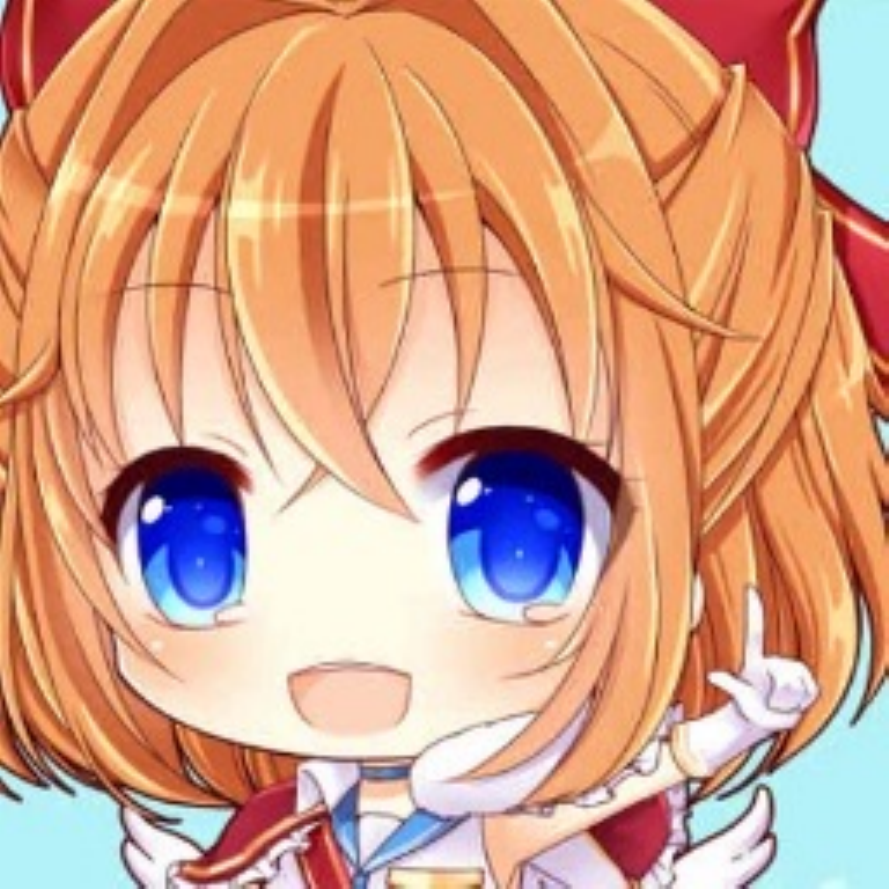}\vspace{4pt}
    \end{minipage}
    }
    \subfigure[]{
    \begin{minipage}[b]{0.125\linewidth}
        \includegraphics[width=\linewidth]{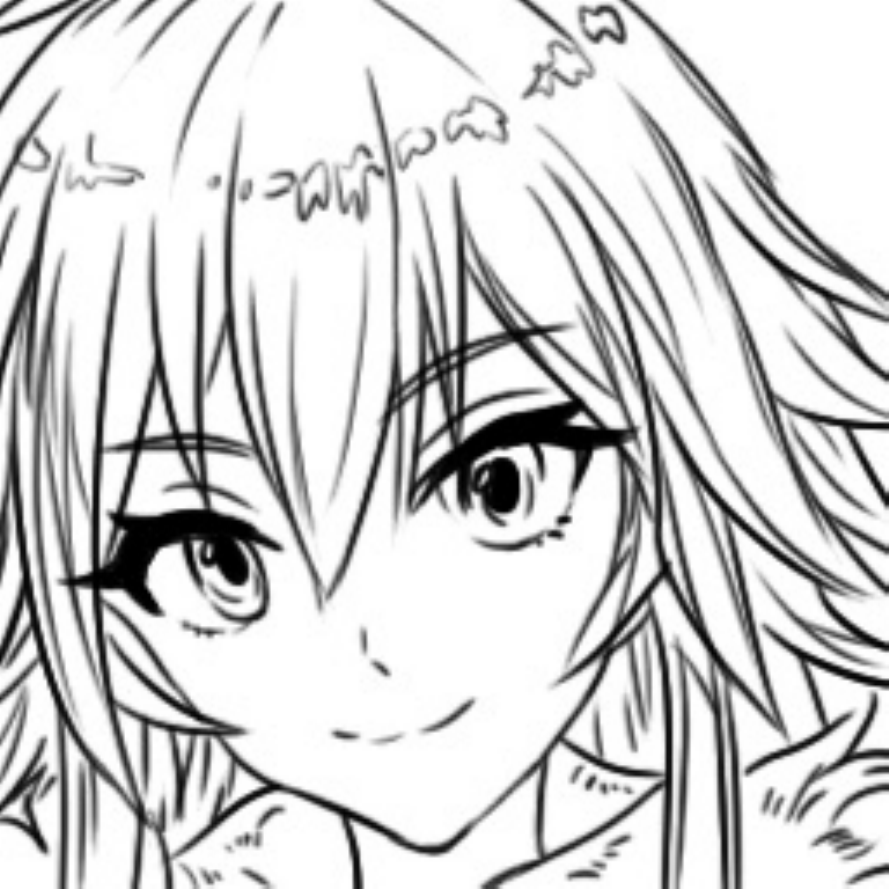}\vspace{4pt}
        \includegraphics[width=\linewidth]{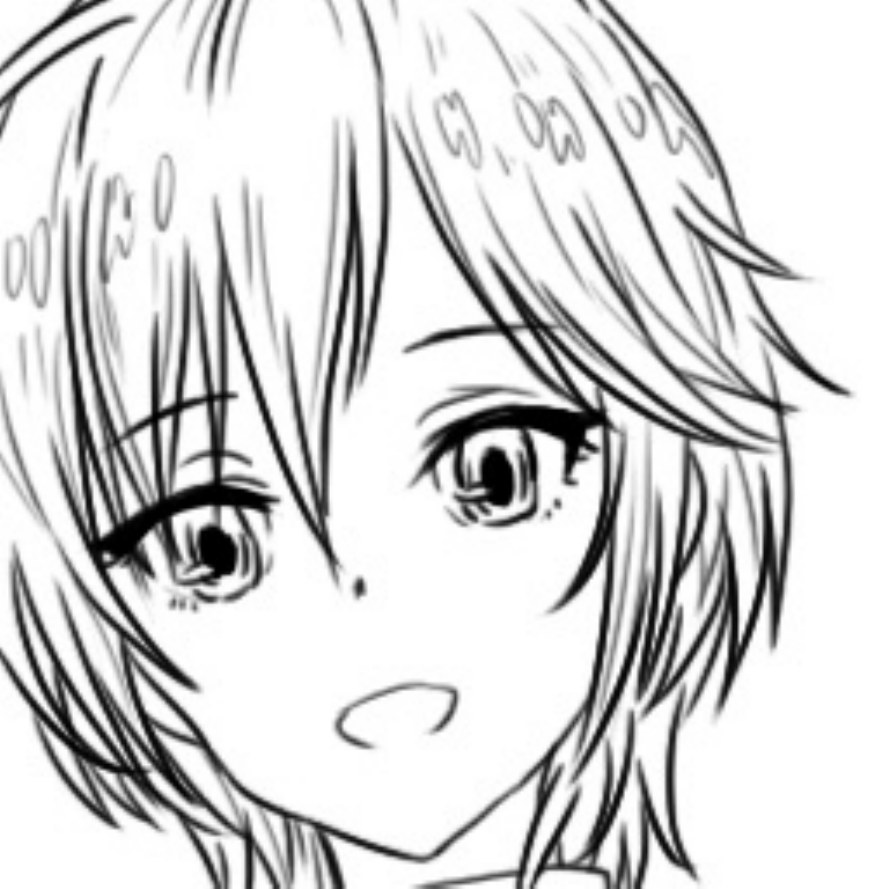}\vspace{4pt}
        \includegraphics[width=\linewidth]{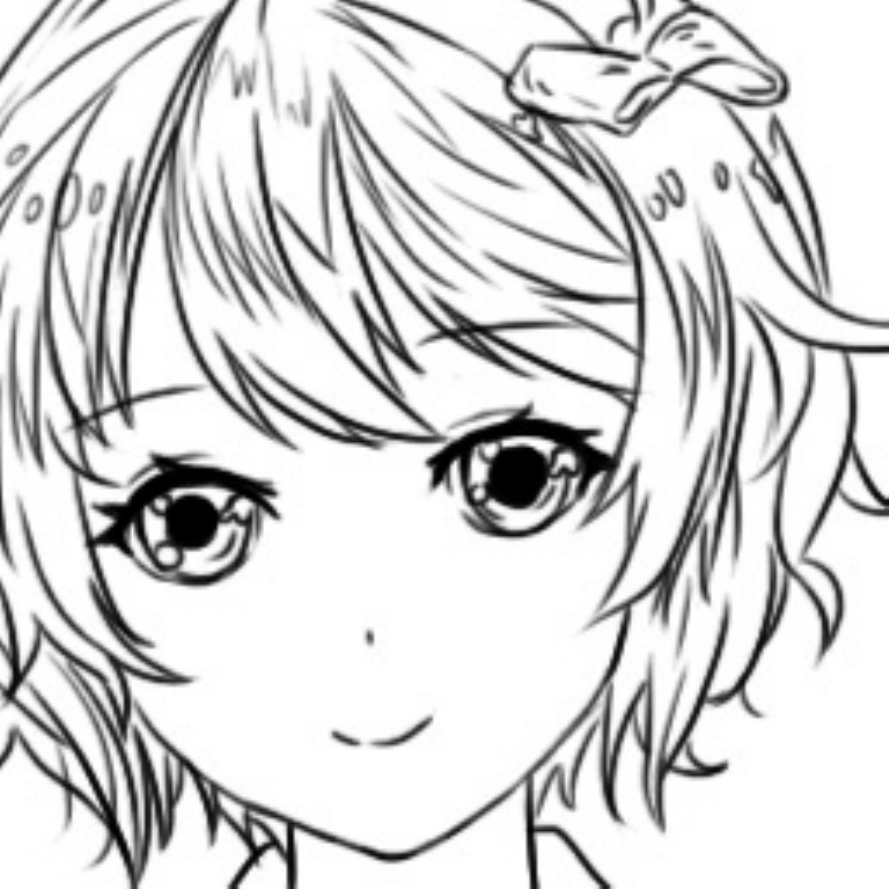}\vspace{4pt}
        \includegraphics[width=\linewidth]{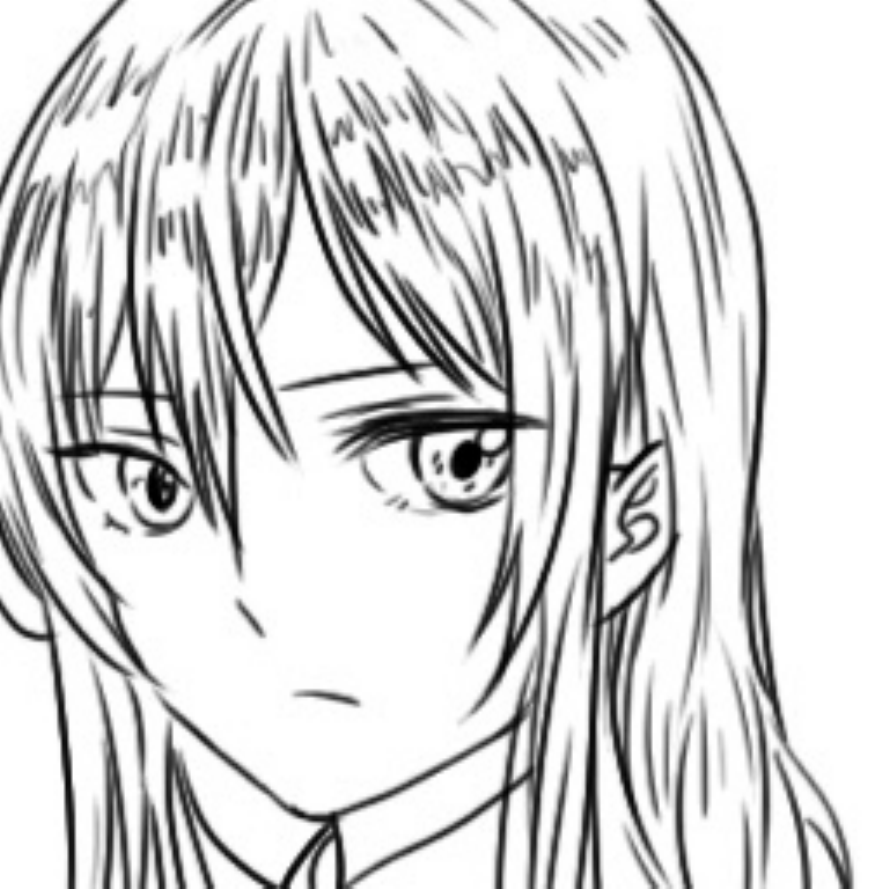}\vspace{4pt}
        \includegraphics[width=\linewidth]{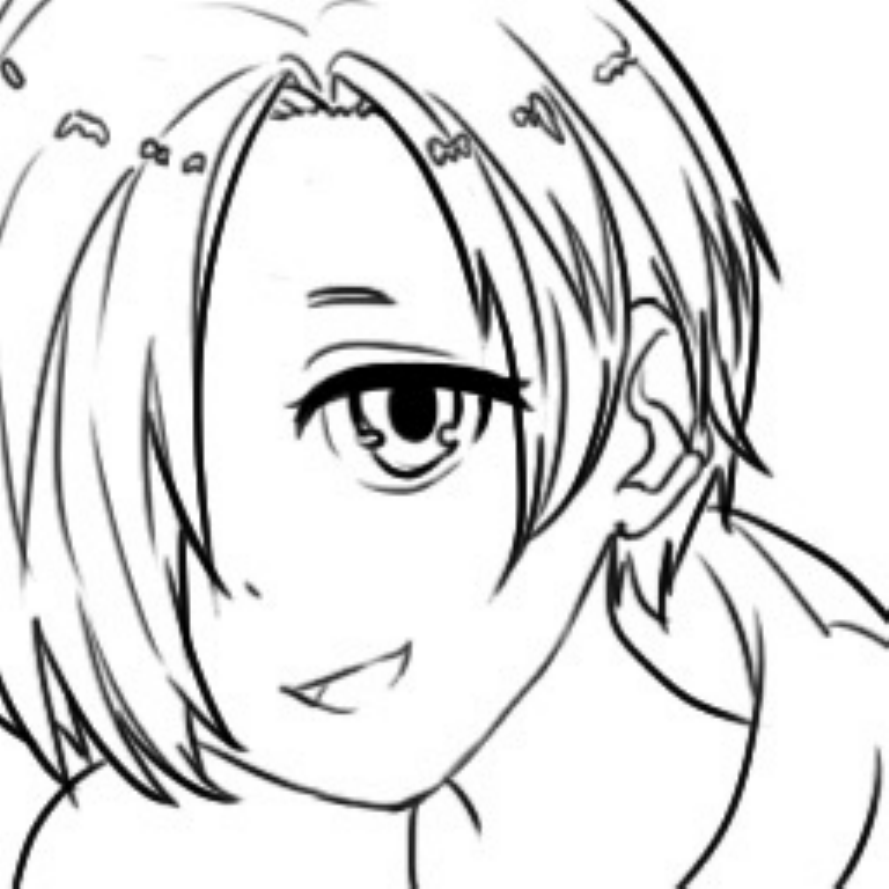}\vspace{4pt}
        \includegraphics[width=\linewidth]{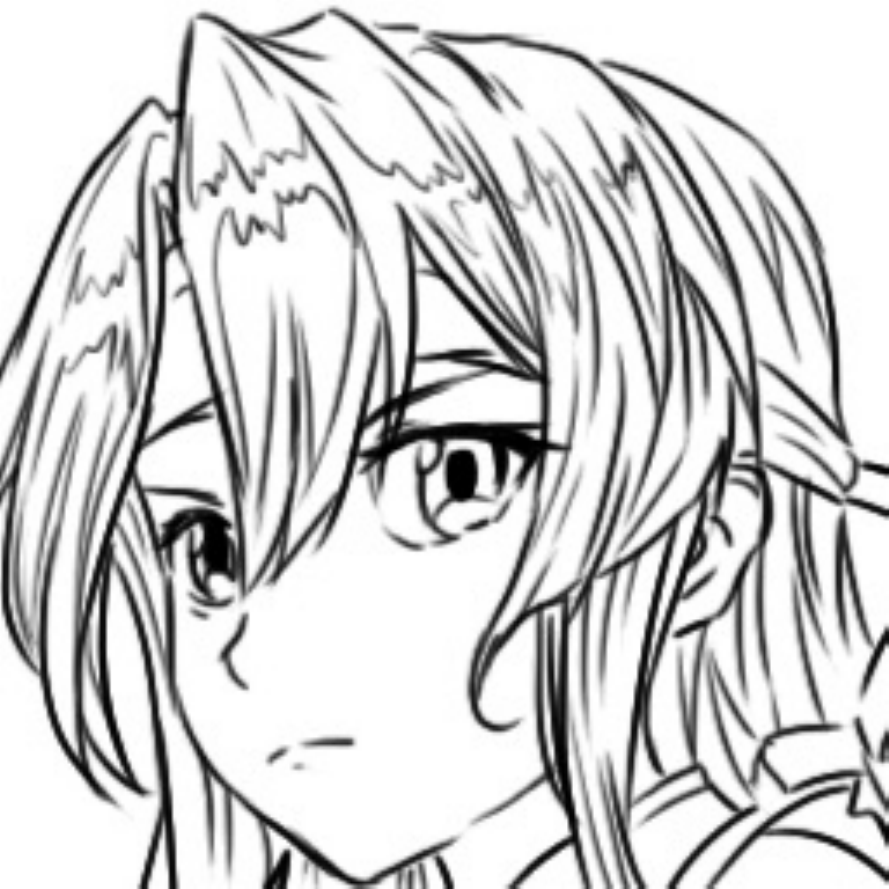}\vspace{4pt}
        \includegraphics[width=\linewidth]{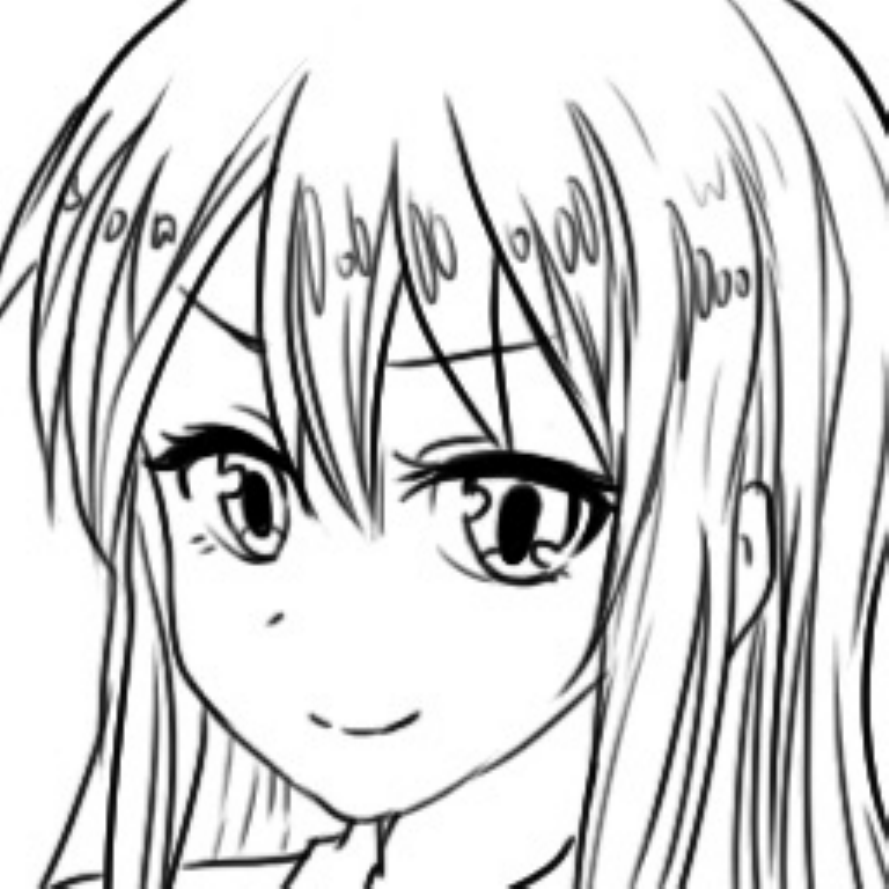}\vspace{4pt}
        \includegraphics[width=\linewidth]{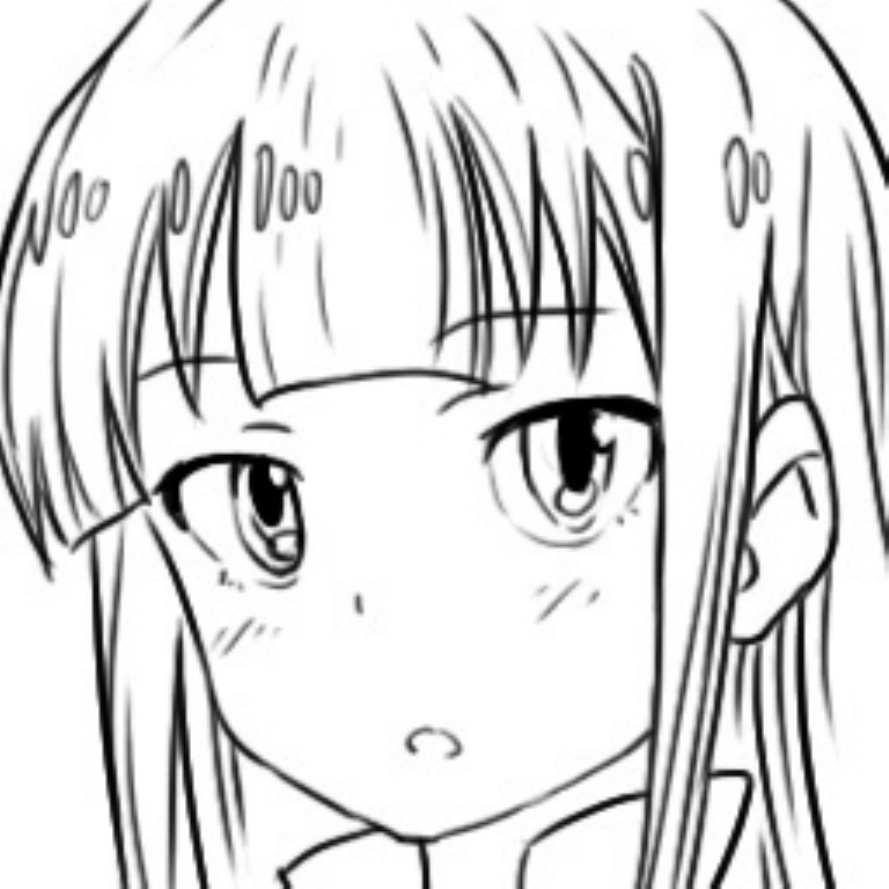}\vspace{4pt}
        \includegraphics[width=\linewidth]{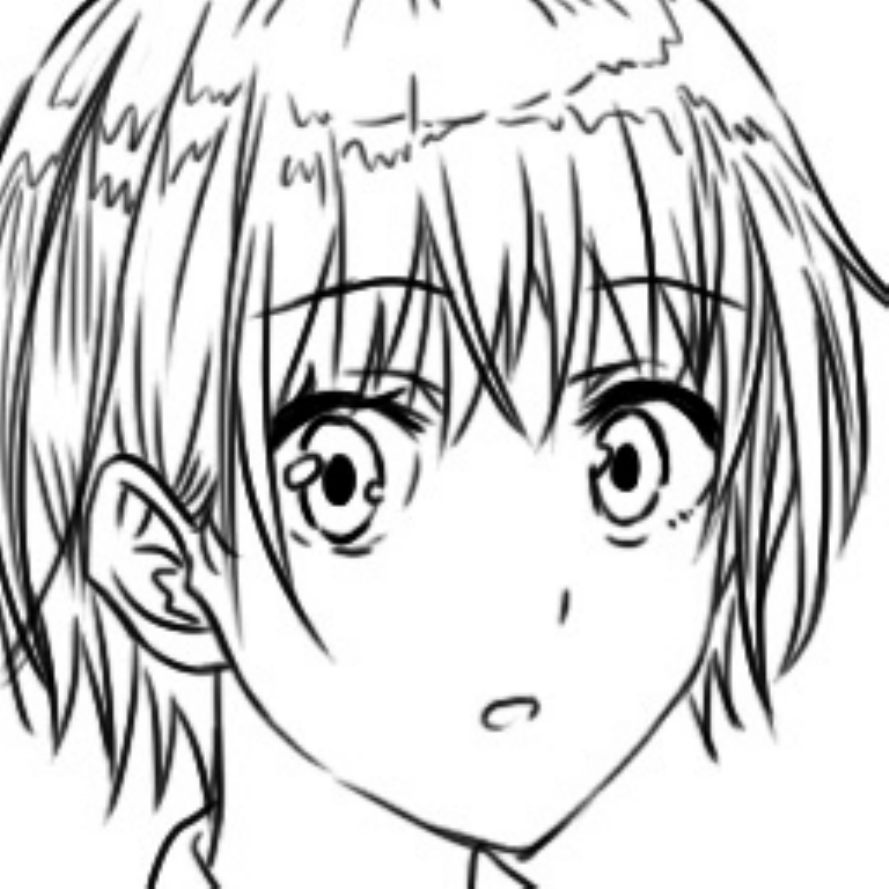}\vspace{4pt}
    \end{minipage}
    }
    \subfigure[]{
    \begin{minipage}[b]{0.125\linewidth}
        \includegraphics[width=\linewidth]{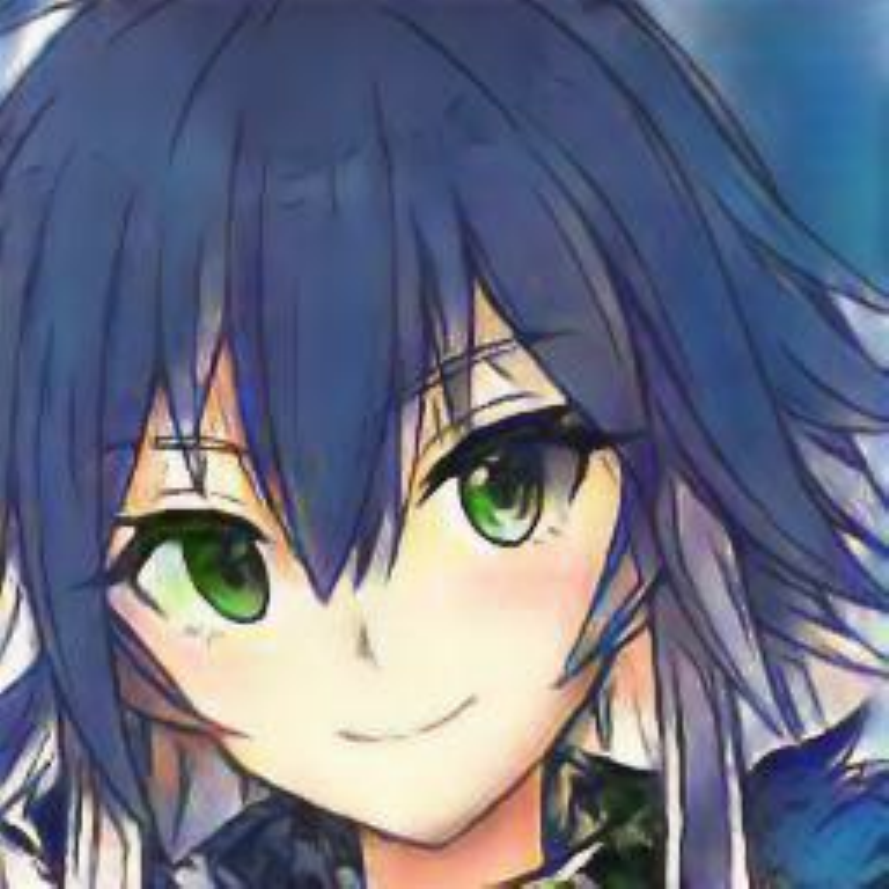}\vspace{4pt}
        \includegraphics[width=\linewidth]{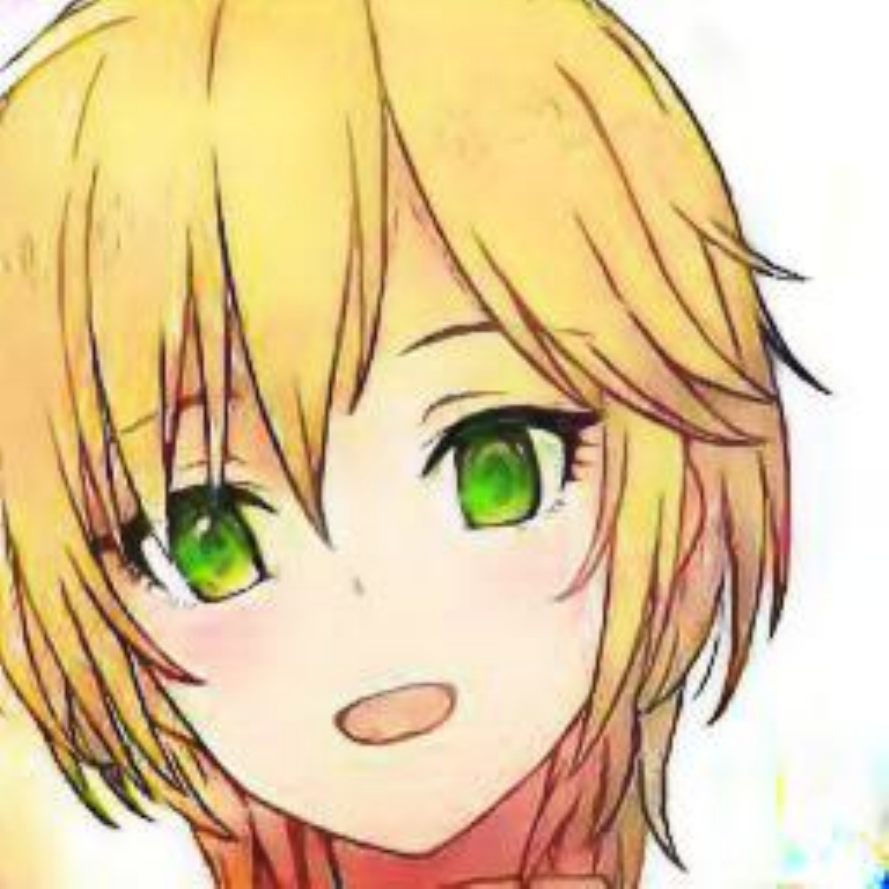}\vspace{4pt}
        \includegraphics[width=\linewidth]{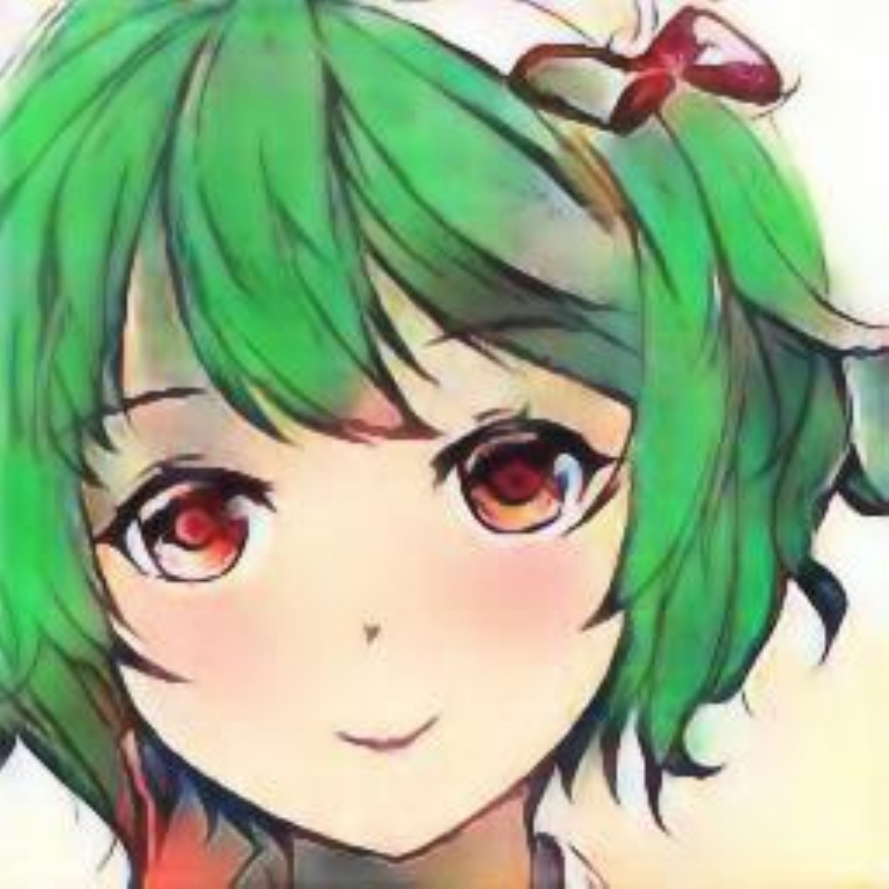}\vspace{4pt}
        \includegraphics[width=\linewidth]{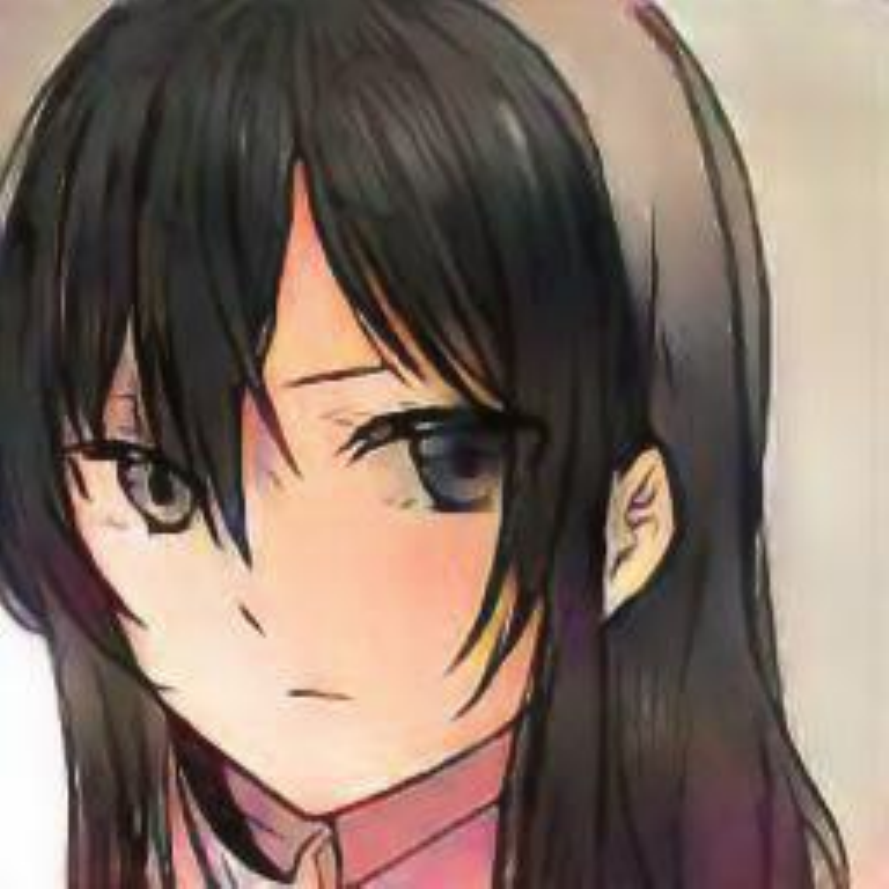}\vspace{4pt}
        \includegraphics[width=\linewidth]{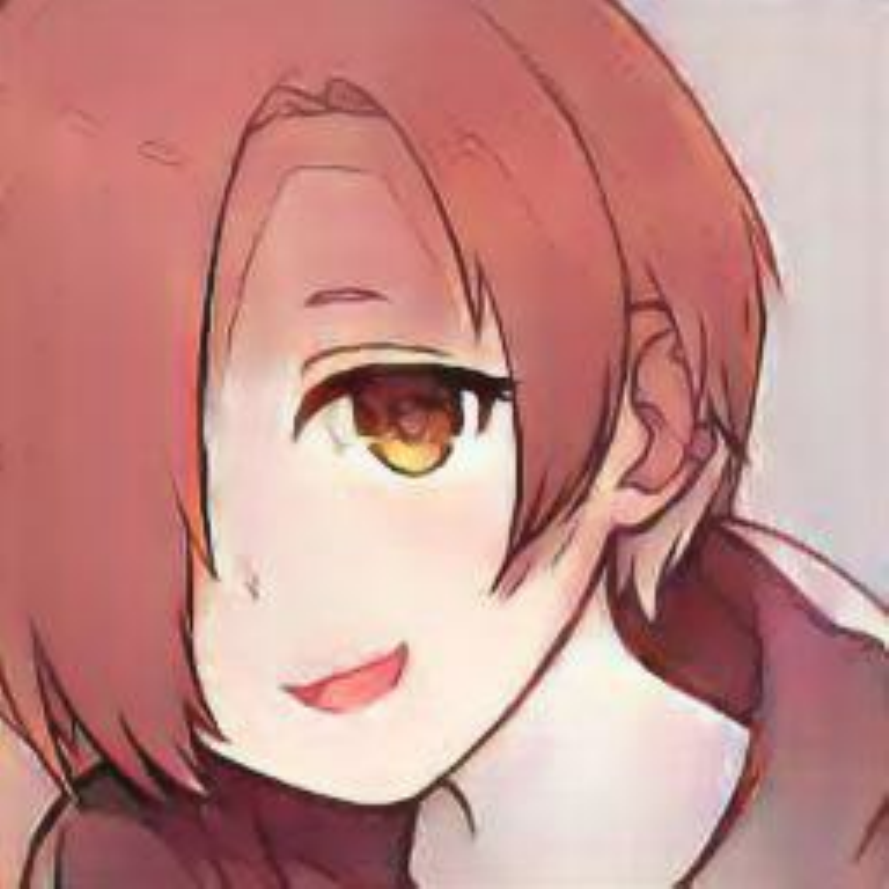}\vspace{4pt}
        \includegraphics[width=\linewidth]{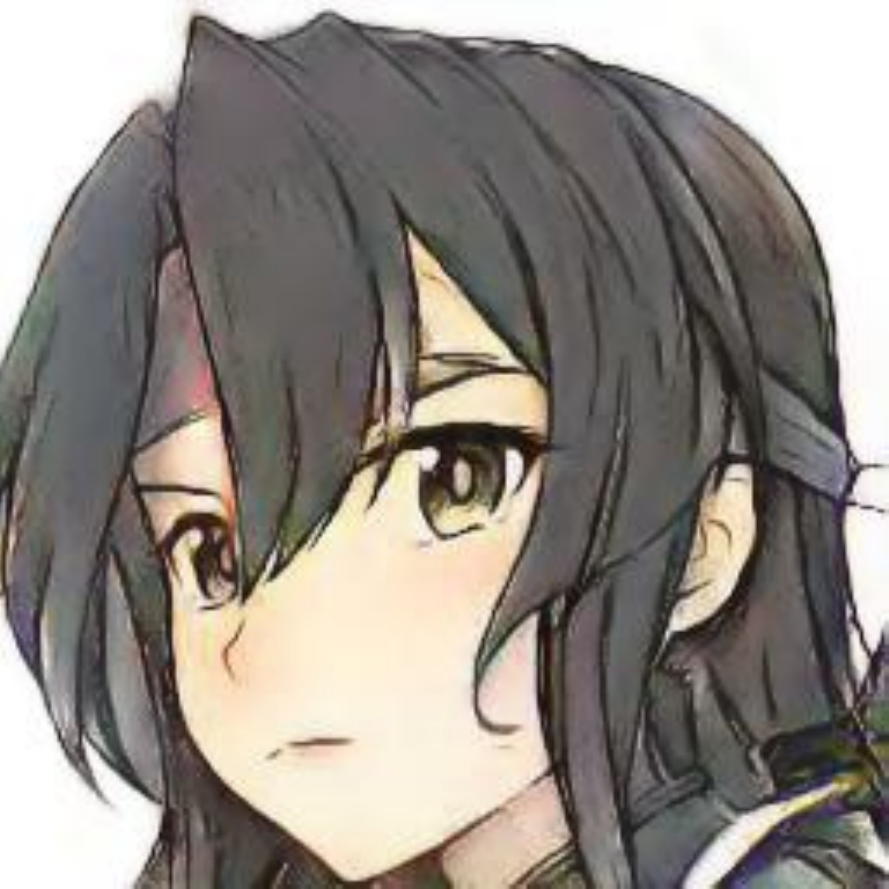}\vspace{4pt}
        \includegraphics[width=\linewidth]{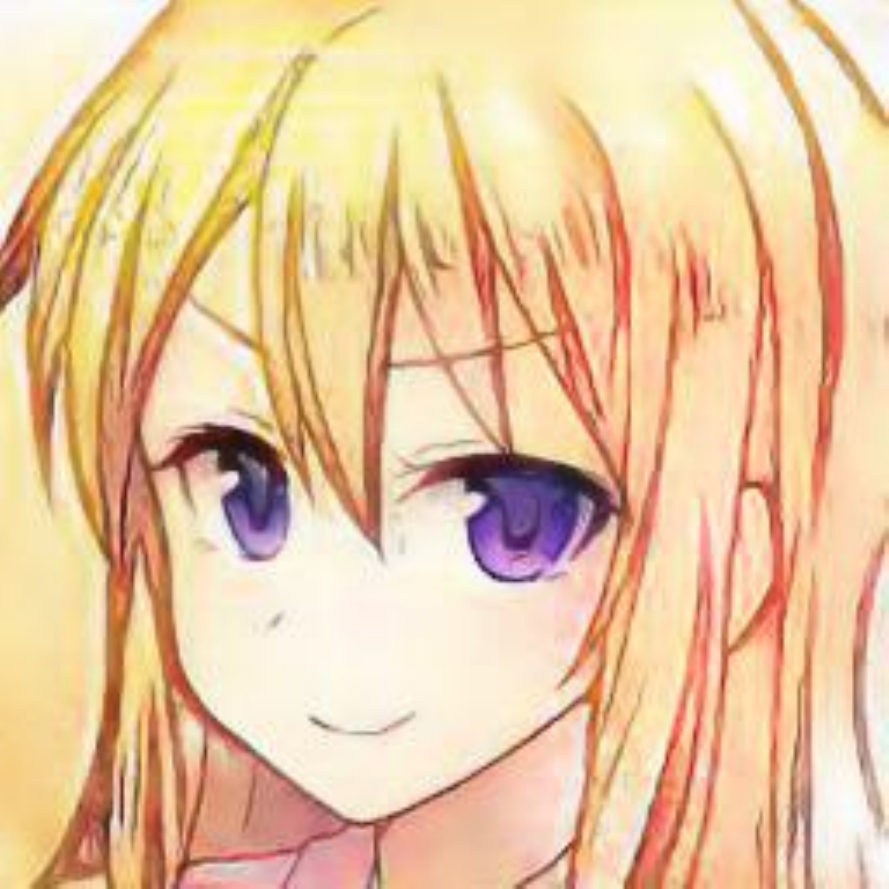}\vspace{4pt}
        \includegraphics[width=\linewidth]{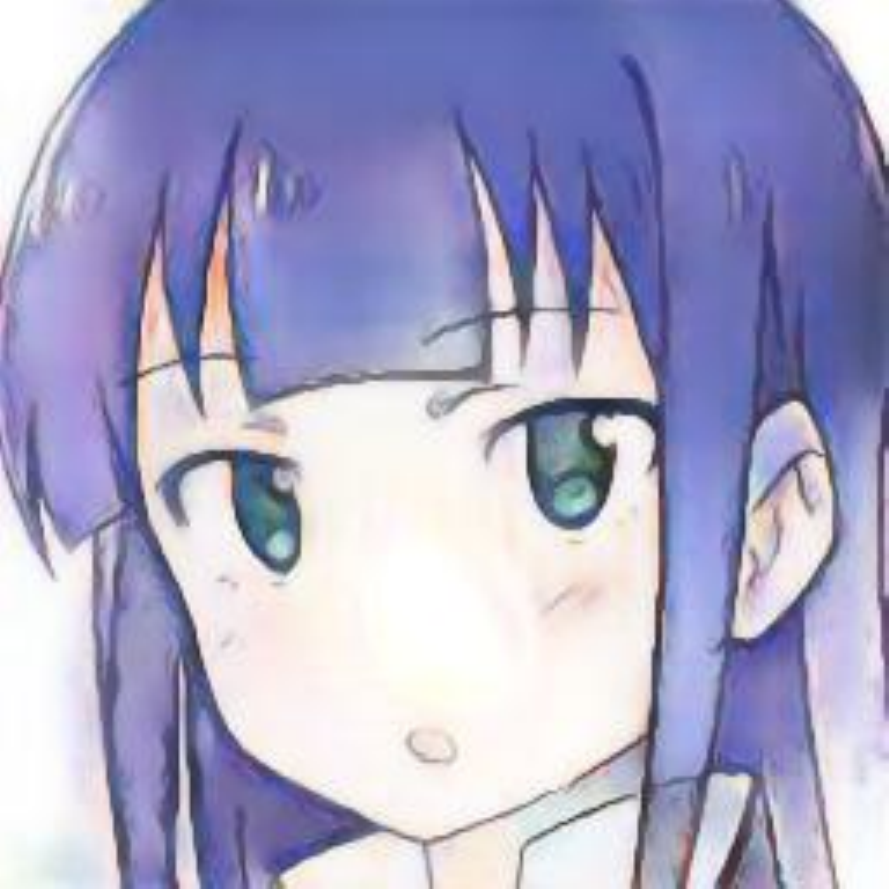}\vspace{4pt}
        \includegraphics[width=\linewidth]{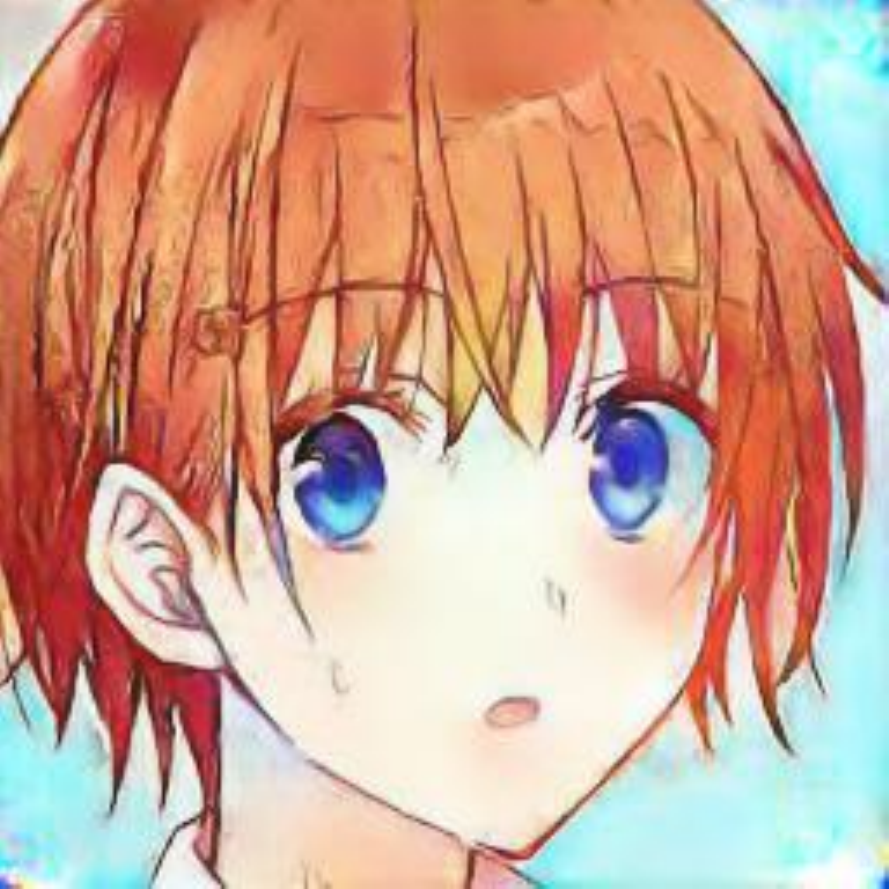}\vspace{4pt}
    \end{minipage}
    }
    \subfigure[]{
    \begin{minipage}[b]{0.125\linewidth}
        \includegraphics[width=\linewidth]{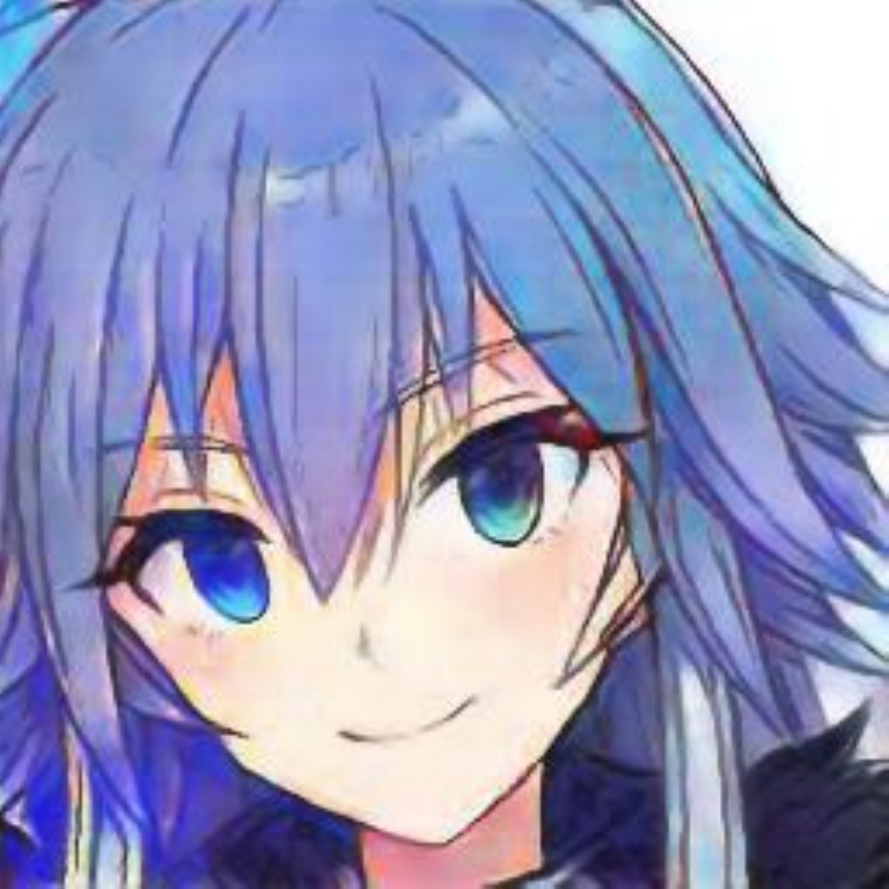}\vspace{4pt}
        \includegraphics[width=\linewidth]{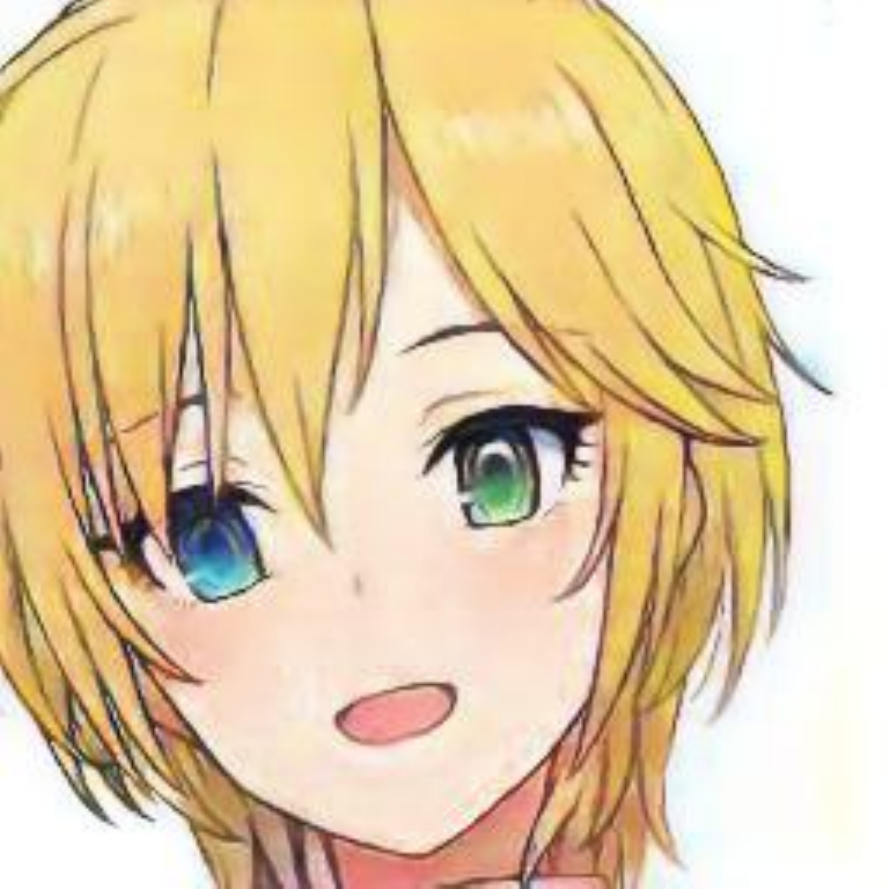}\vspace{4pt}
        \includegraphics[width=\linewidth]{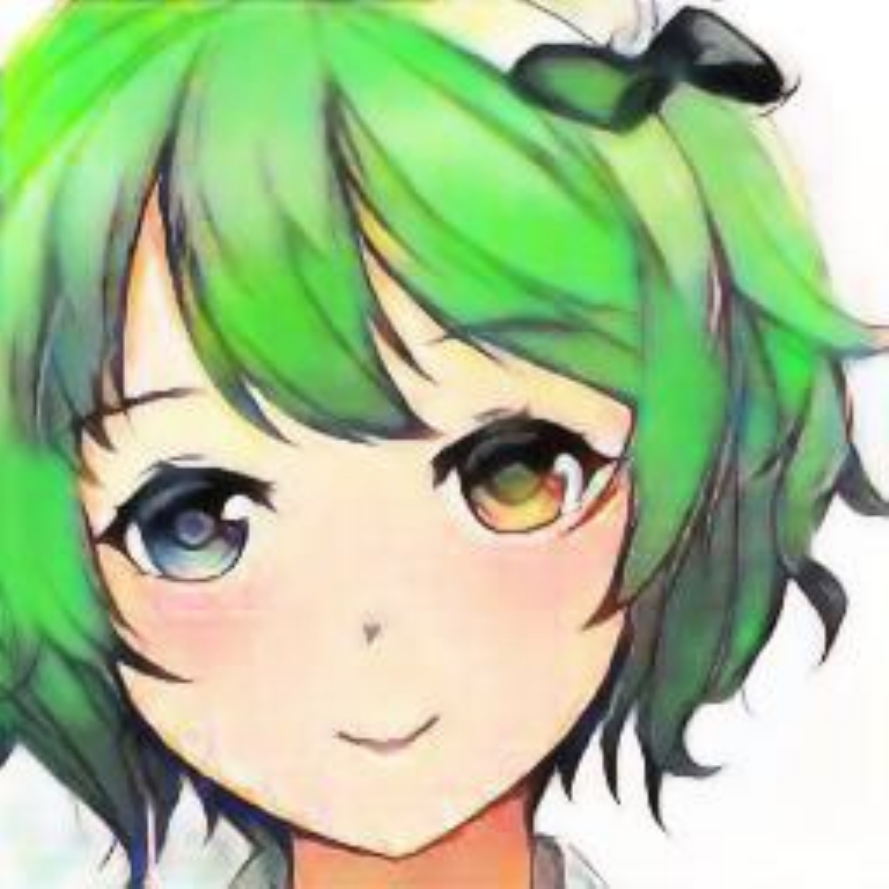}\vspace{4pt}
        \includegraphics[width=\linewidth]{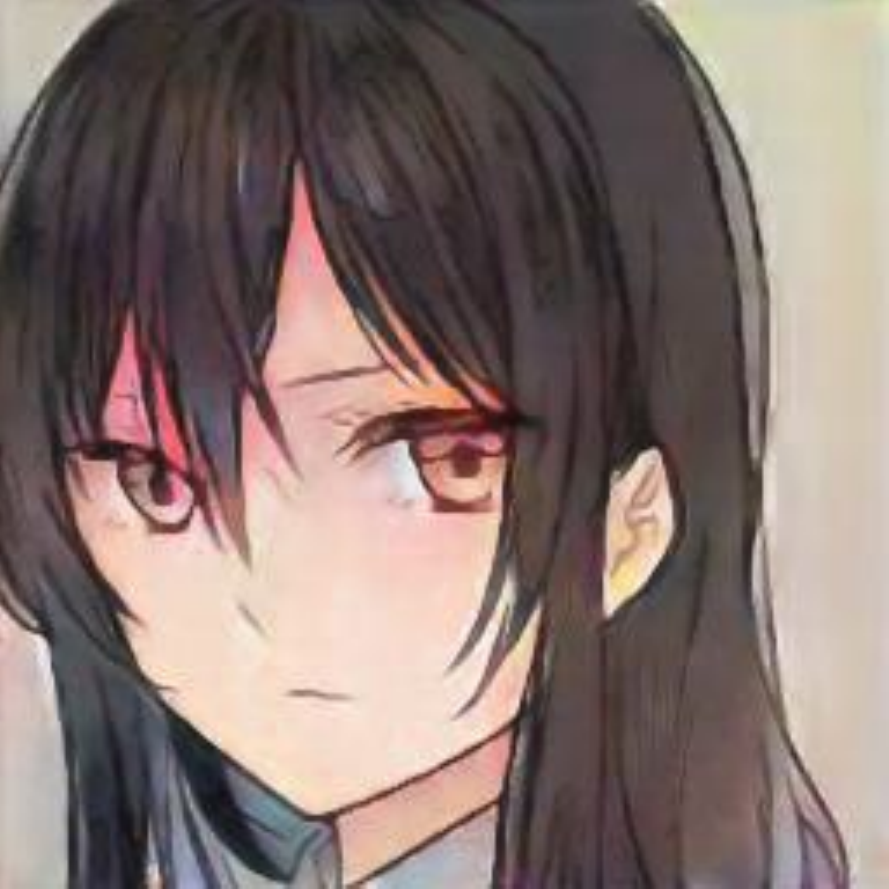}\vspace{4pt}
        \includegraphics[width=\linewidth]{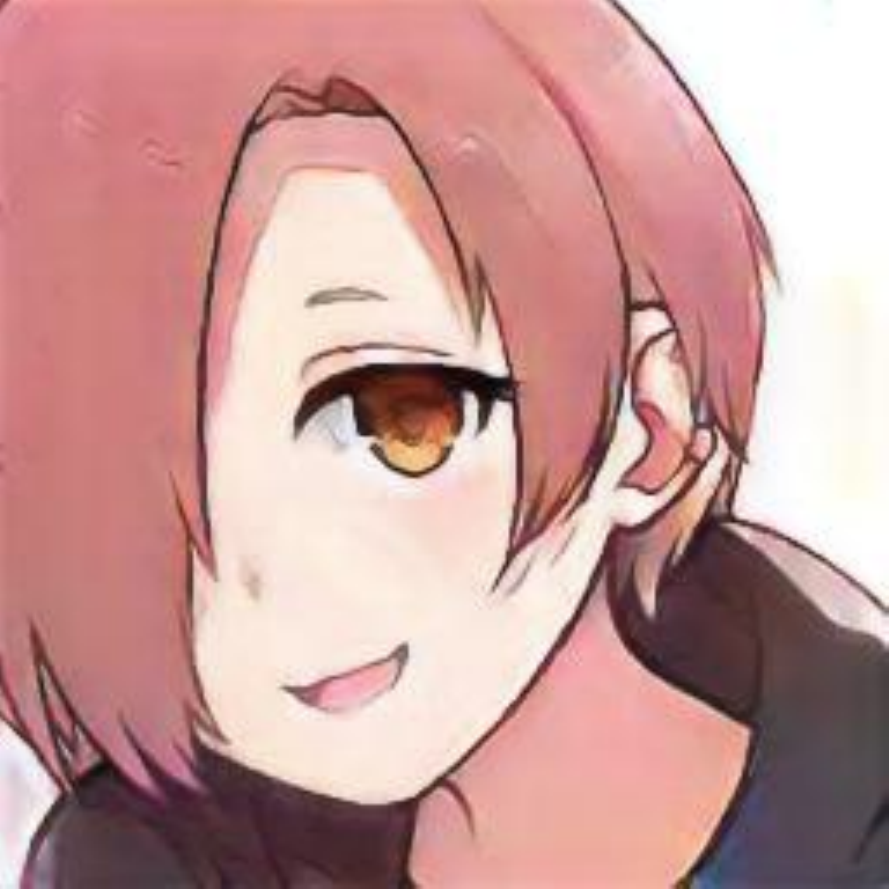}\vspace{4pt}
        \includegraphics[width=\linewidth]{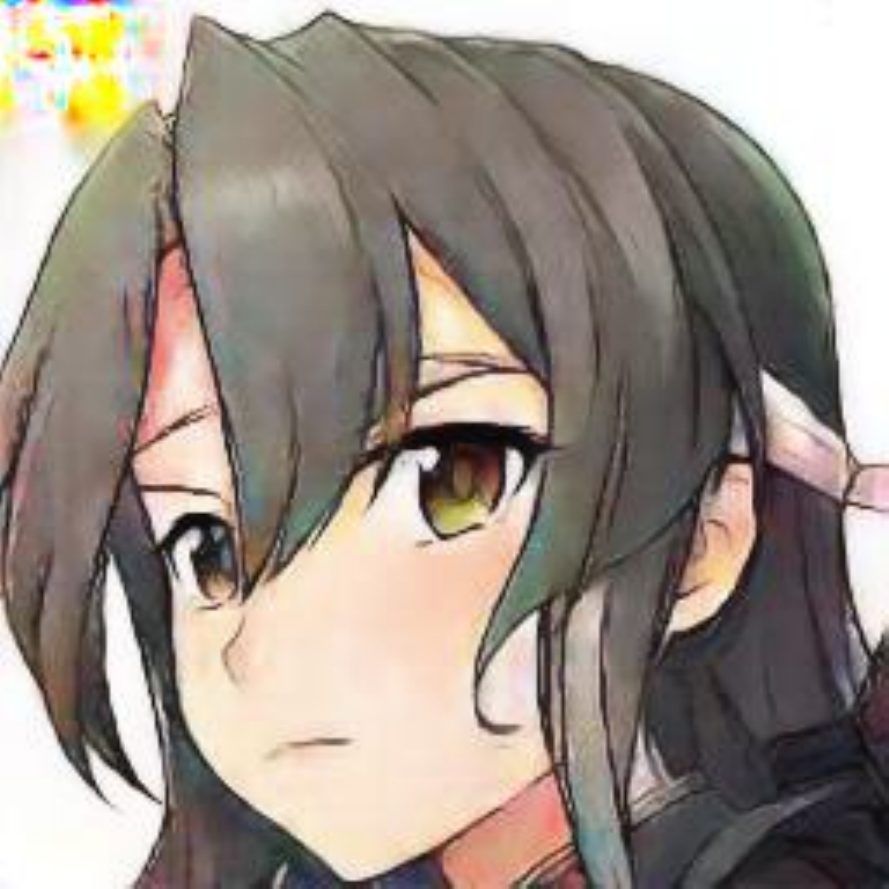}\vspace{4pt}
        \includegraphics[width=\linewidth]{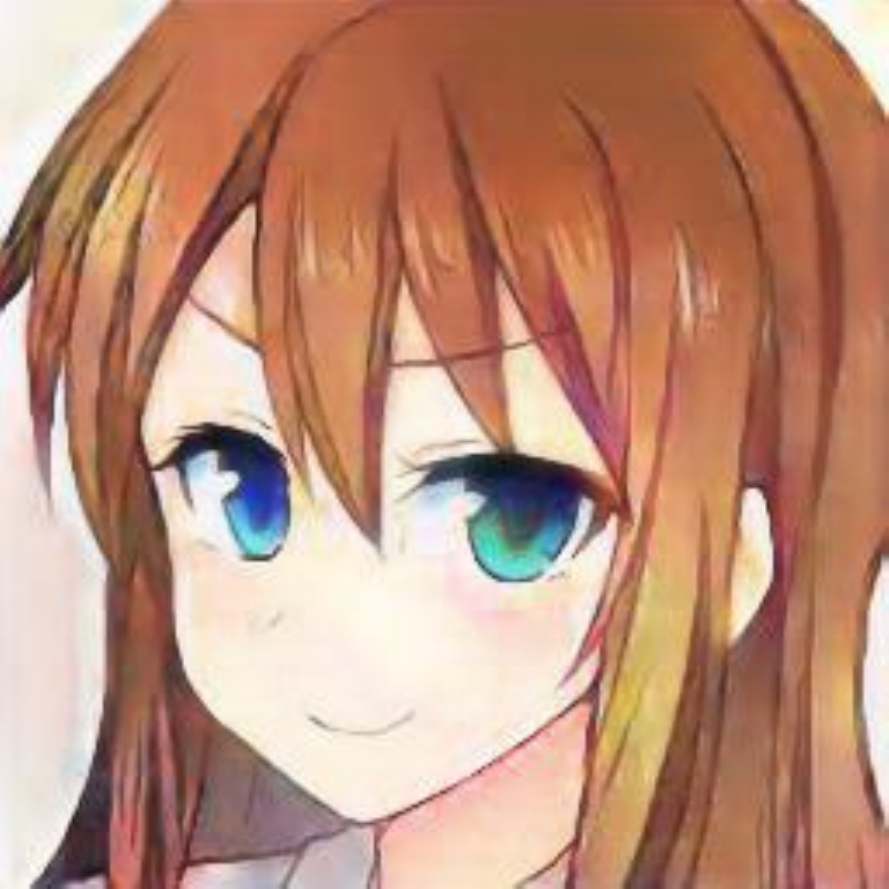}\vspace{4pt}
        \includegraphics[width=\linewidth]{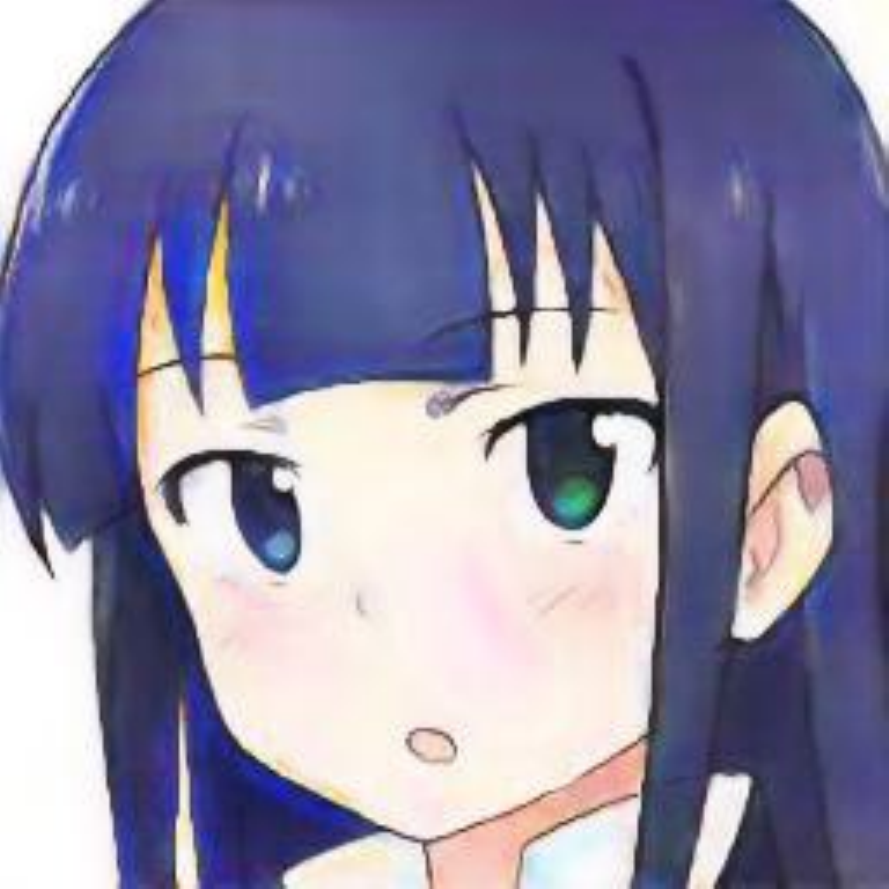}\vspace{4pt}
        \includegraphics[width=\linewidth]{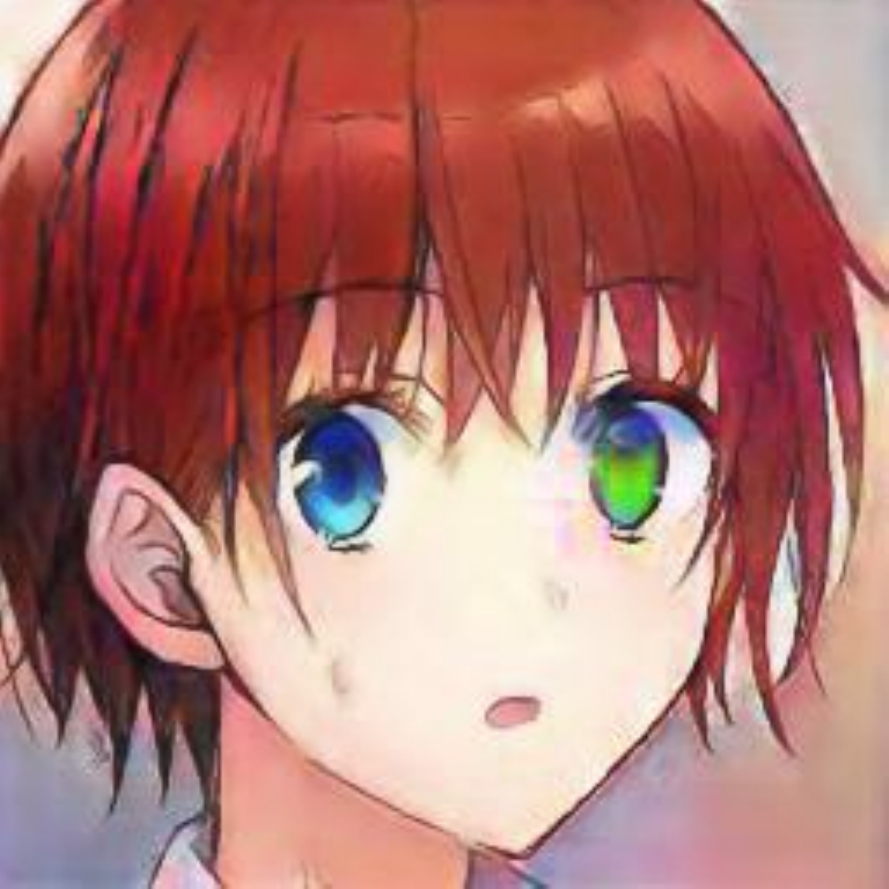}\vspace{4pt}
    \end{minipage}
    }
    \subfigure[]{
    \begin{minipage}[b]{0.125\linewidth}
        \includegraphics[width=\linewidth]{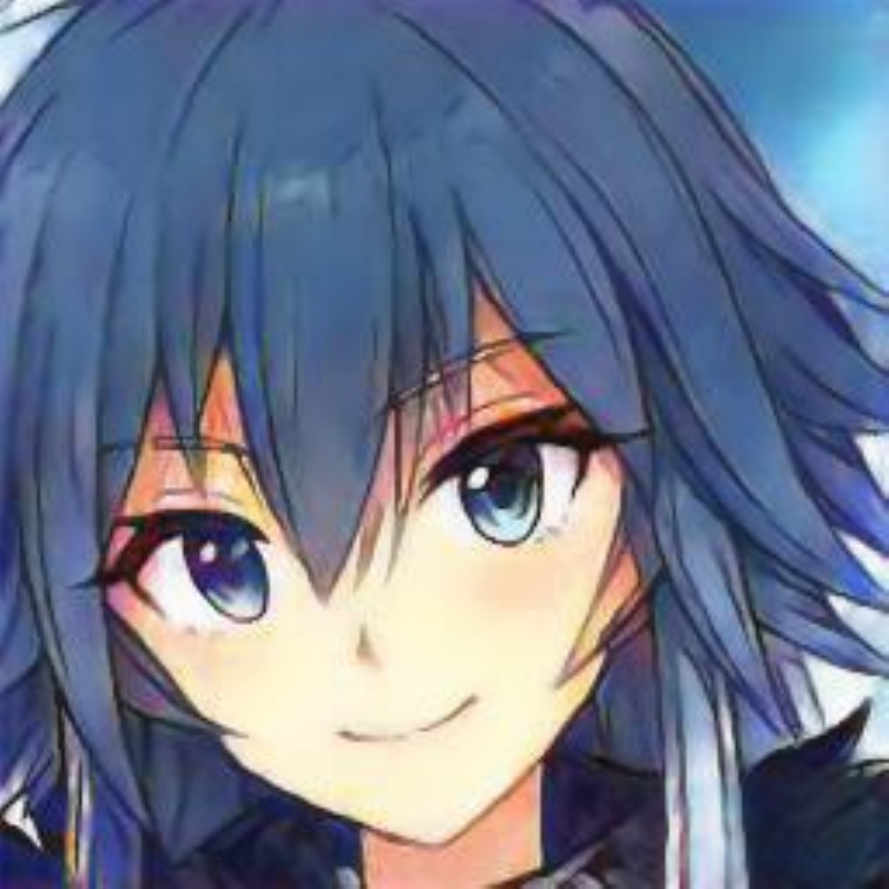}\vspace{4pt}
        \includegraphics[width=\linewidth]{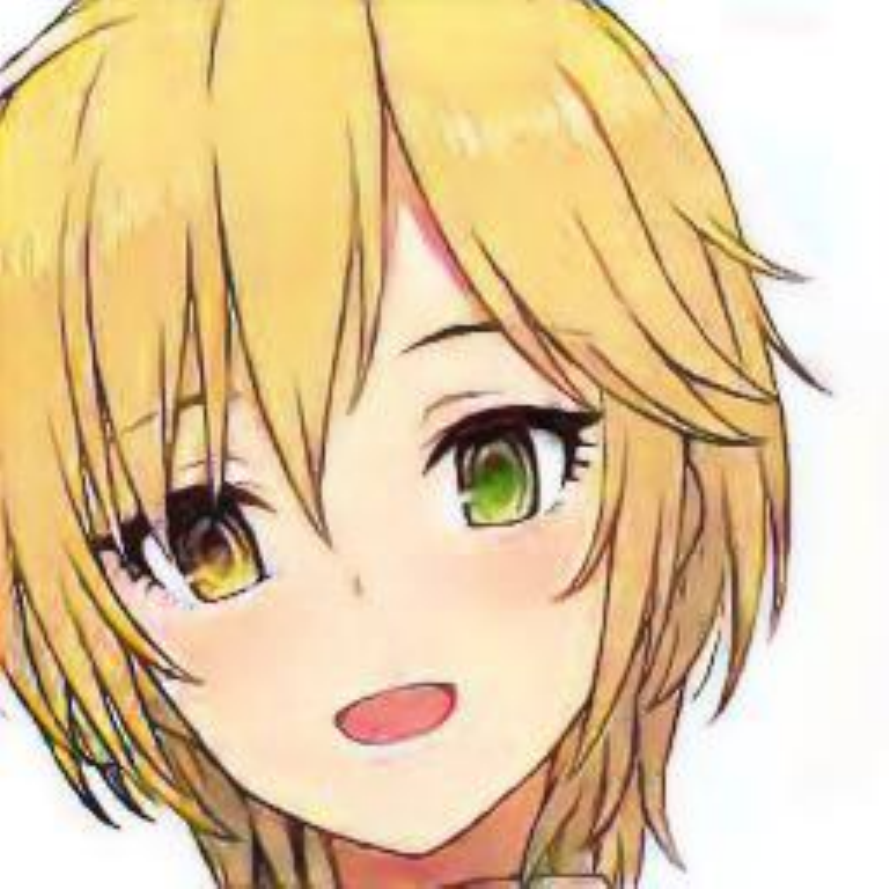}\vspace{4pt}
        \includegraphics[width=\linewidth]{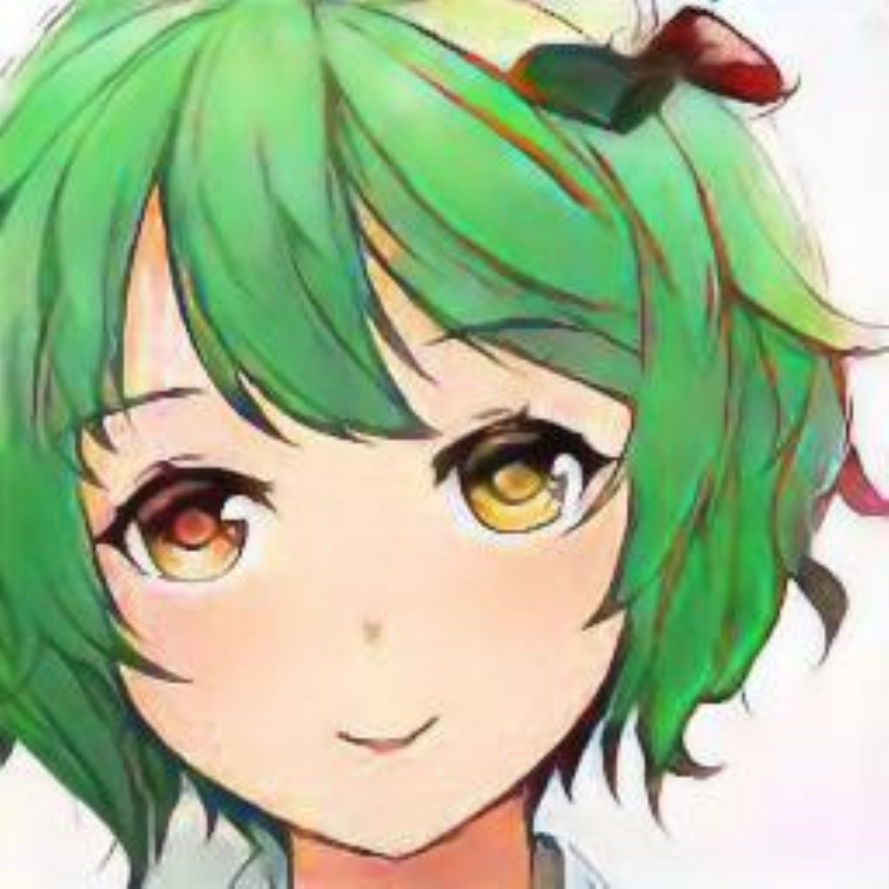}\vspace{4pt}
        \includegraphics[width=\linewidth]{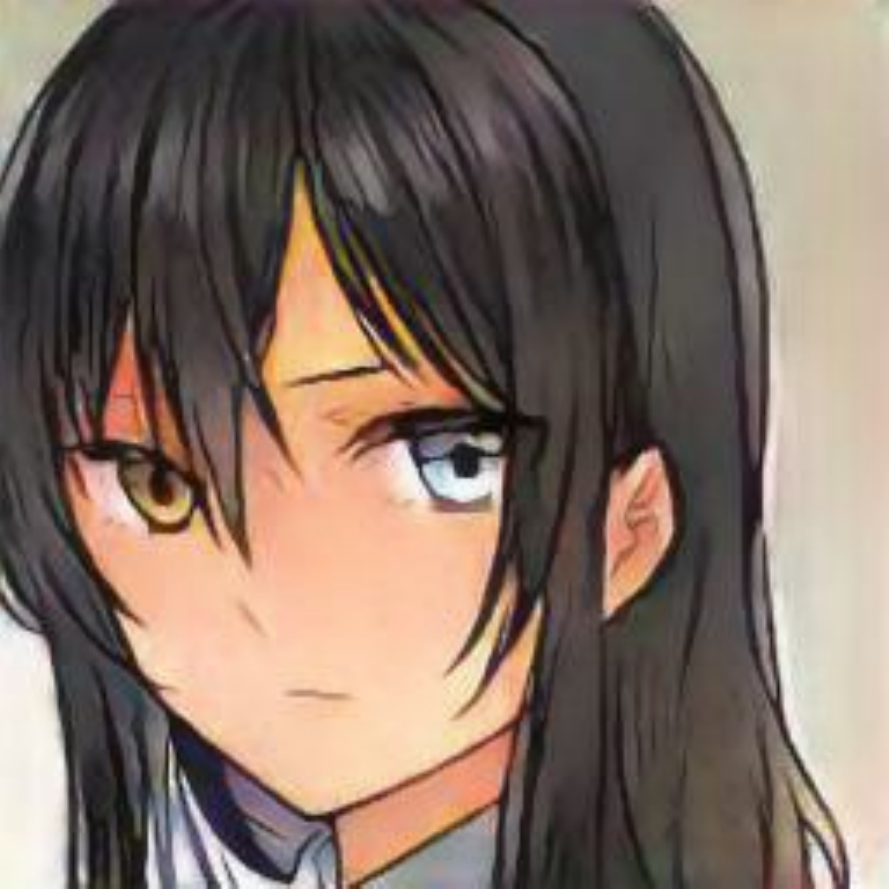}\vspace{4pt}
        \includegraphics[width=\linewidth]{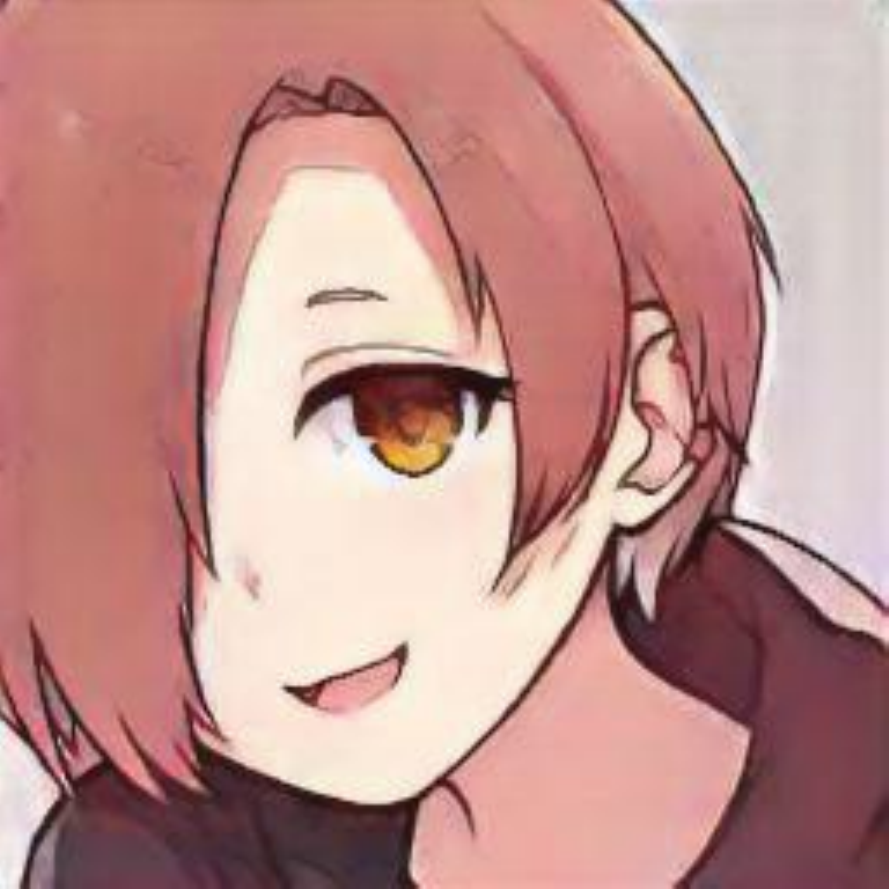}\vspace{4pt}
        \includegraphics[width=\linewidth]{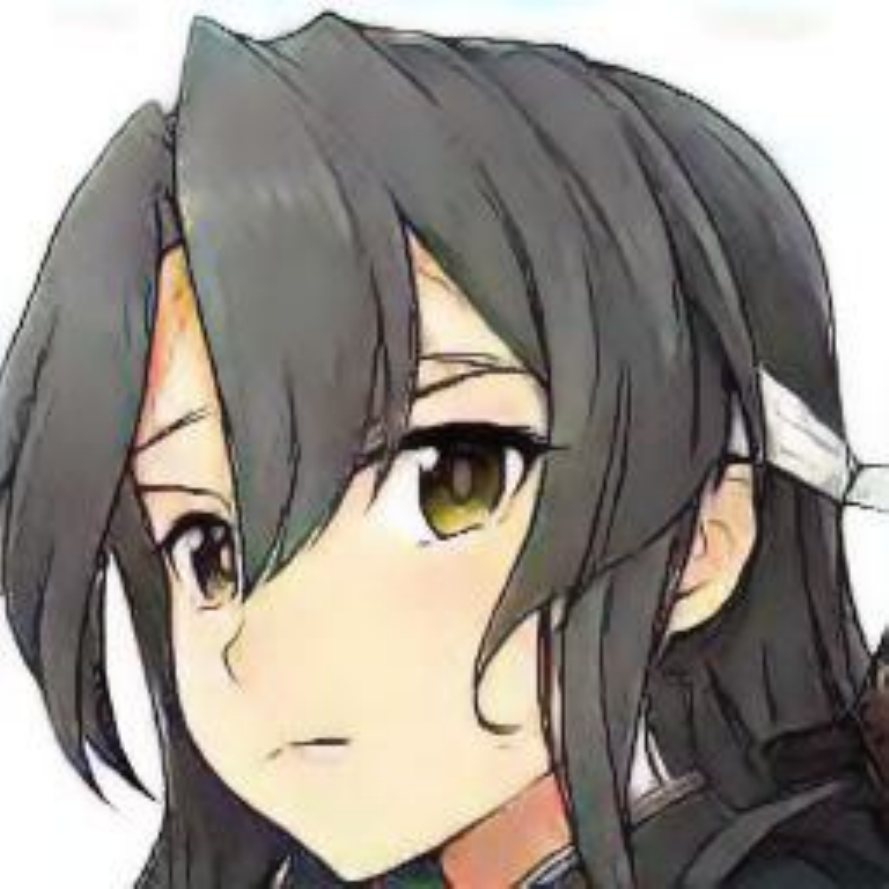}\vspace{4pt}
        \includegraphics[width=\linewidth]{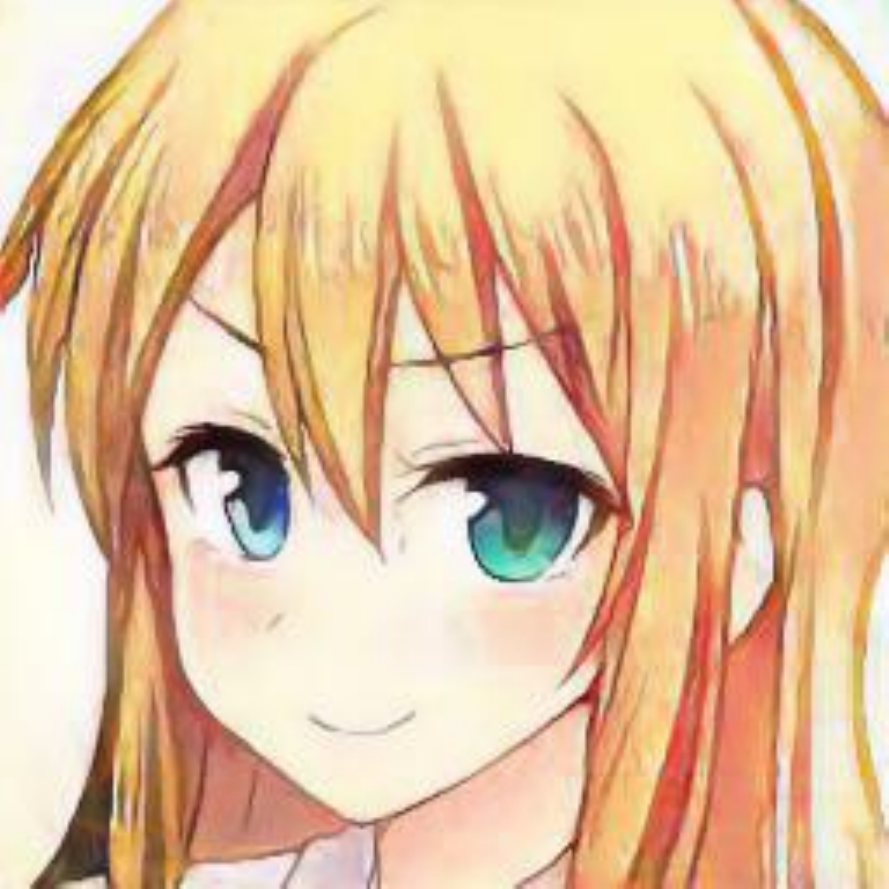}\vspace{4pt}
        \includegraphics[width=\linewidth]{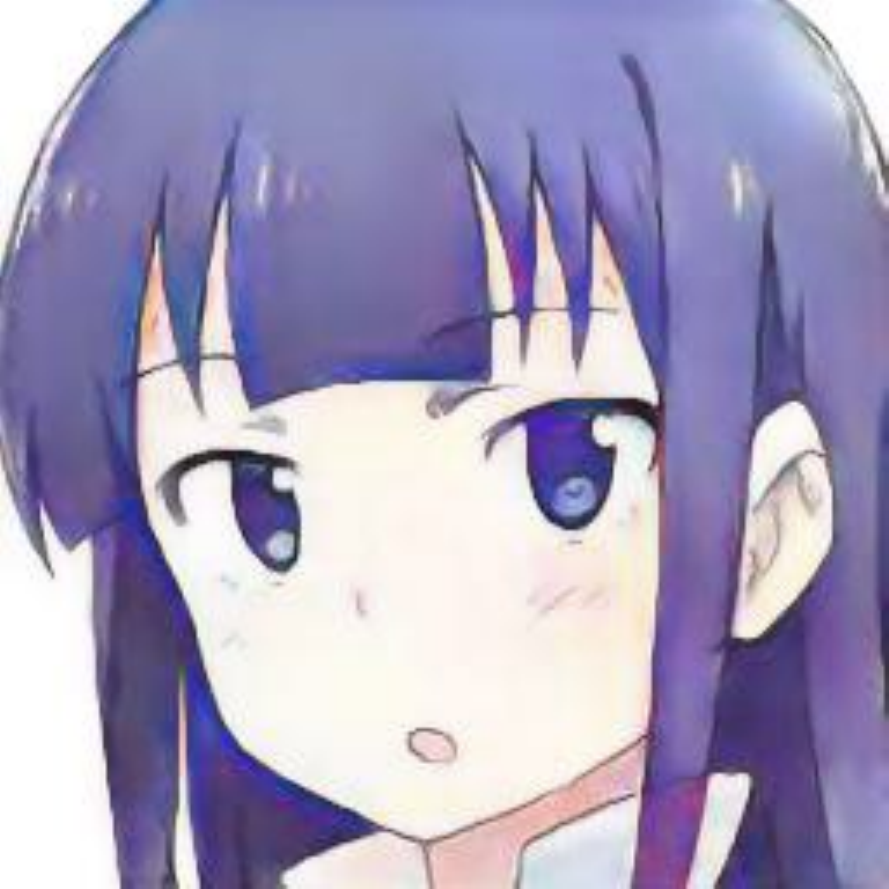}\vspace{4pt}
        \includegraphics[width=\linewidth]{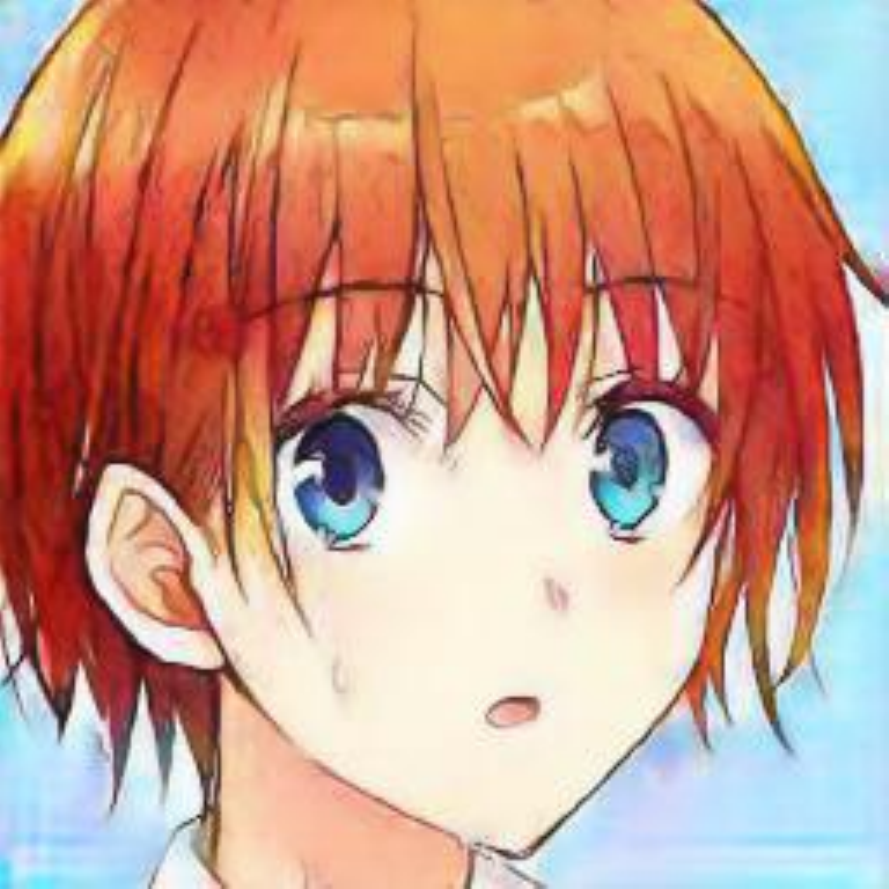}\vspace{4pt}
    \end{minipage}
    }
    \subfigure[]{
    \begin{minipage}[b]{0.125\linewidth}
        \includegraphics[width=\linewidth]{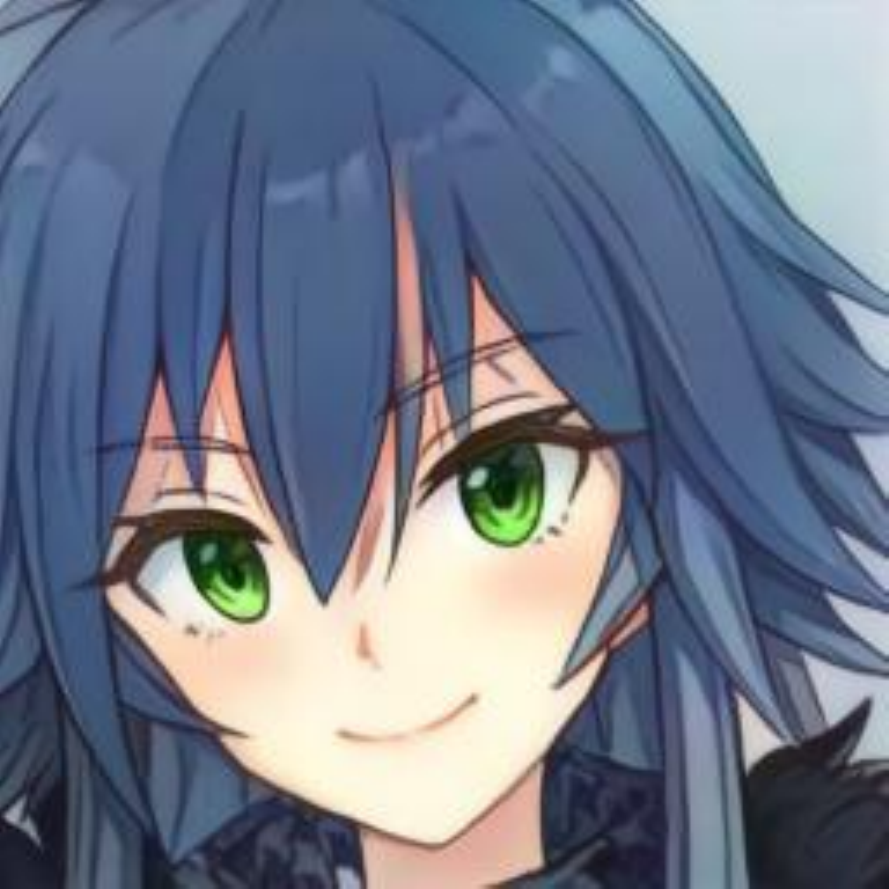}\vspace{4pt}
        \includegraphics[width=\linewidth]{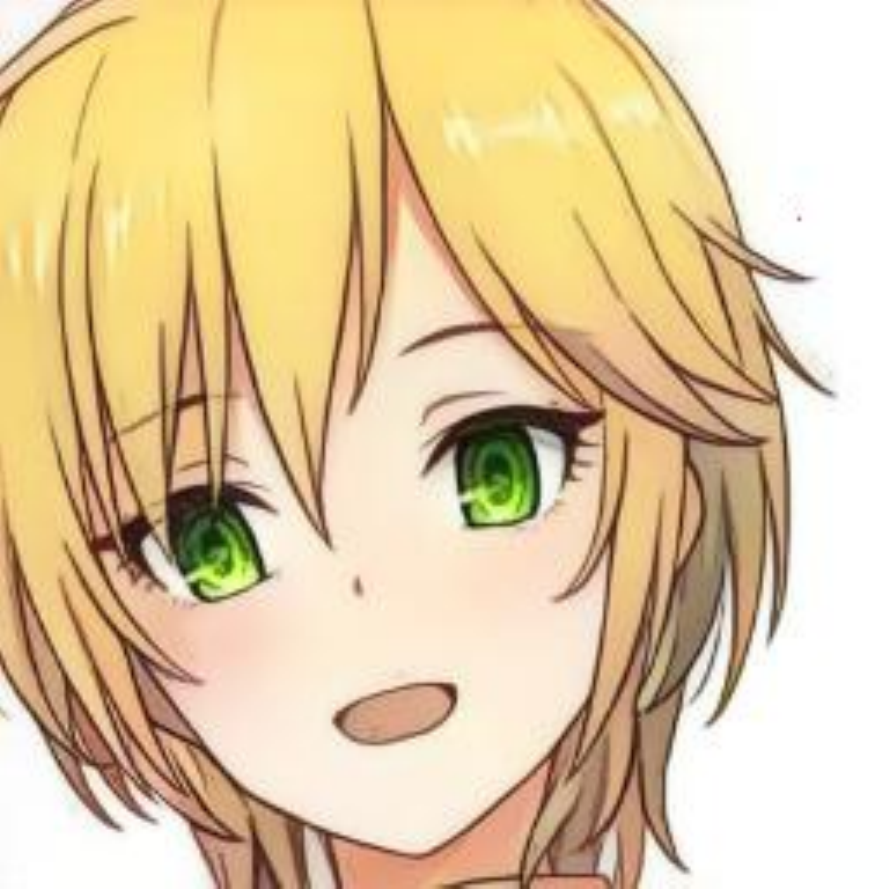}\vspace{4pt}
        \includegraphics[width=\linewidth]{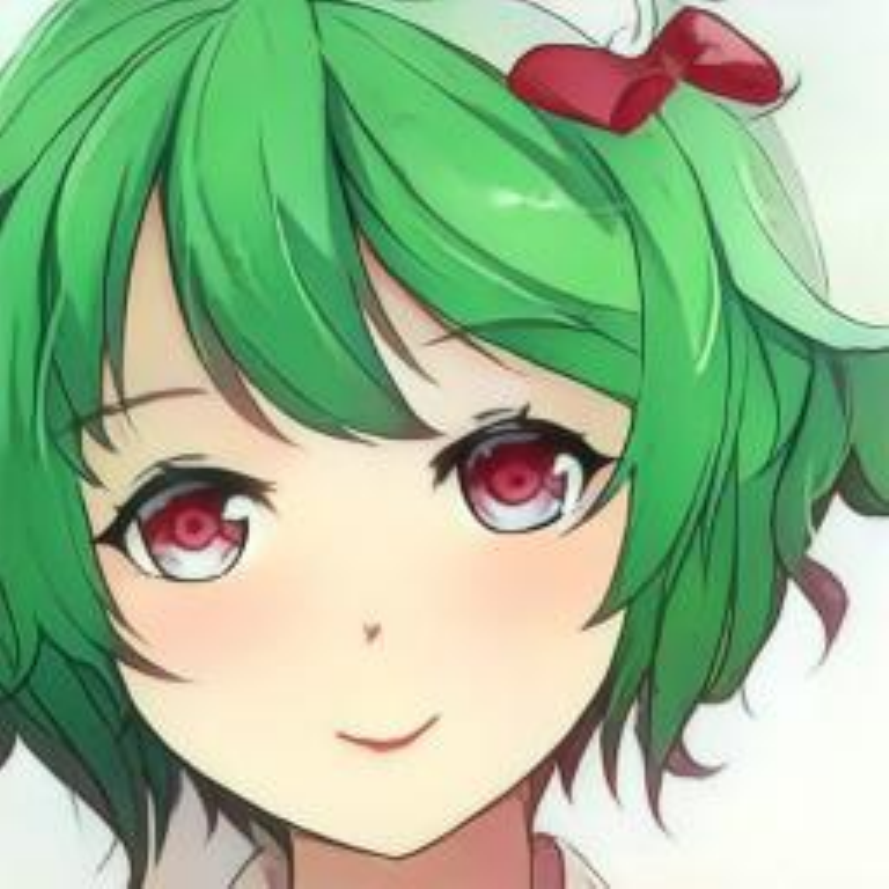}\vspace{4pt}
        \includegraphics[width=\linewidth]{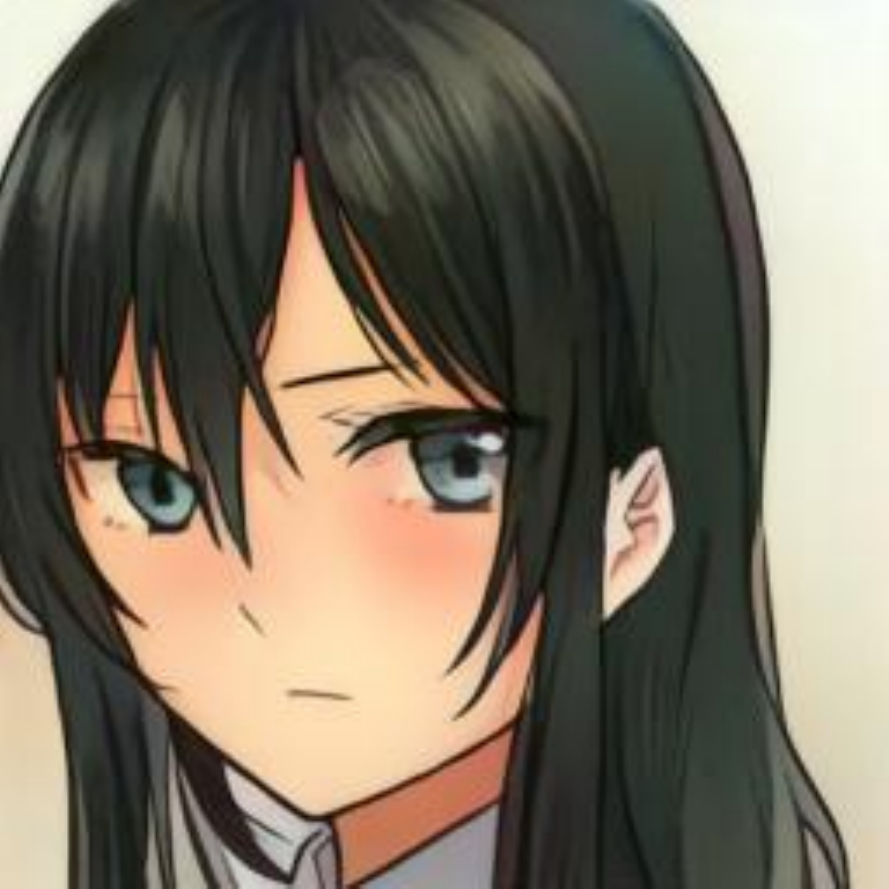}\vspace{4pt}
        \includegraphics[width=\linewidth]{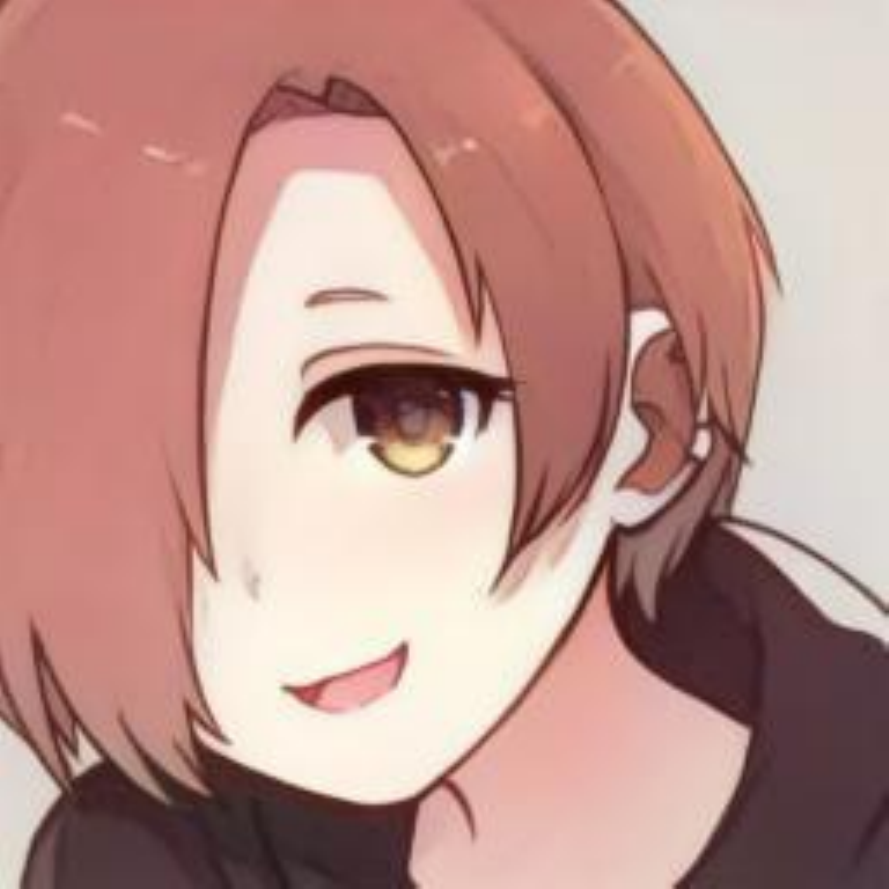}\vspace{4pt}
        \includegraphics[width=\linewidth]{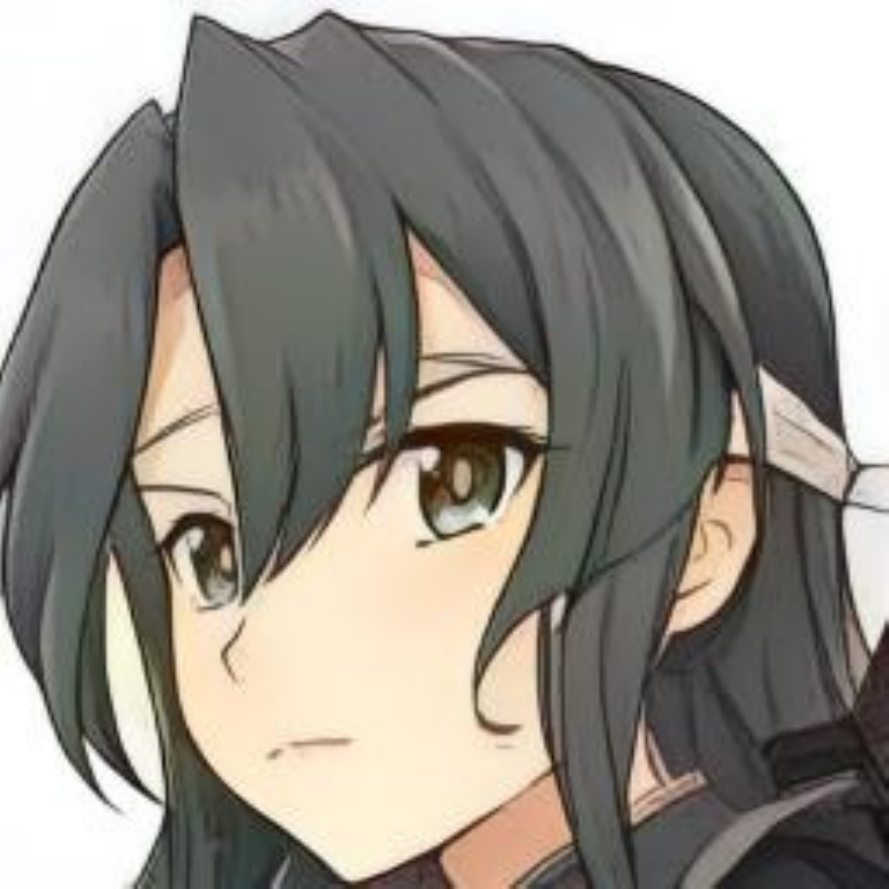}\vspace{4pt}
        \includegraphics[width=\linewidth]{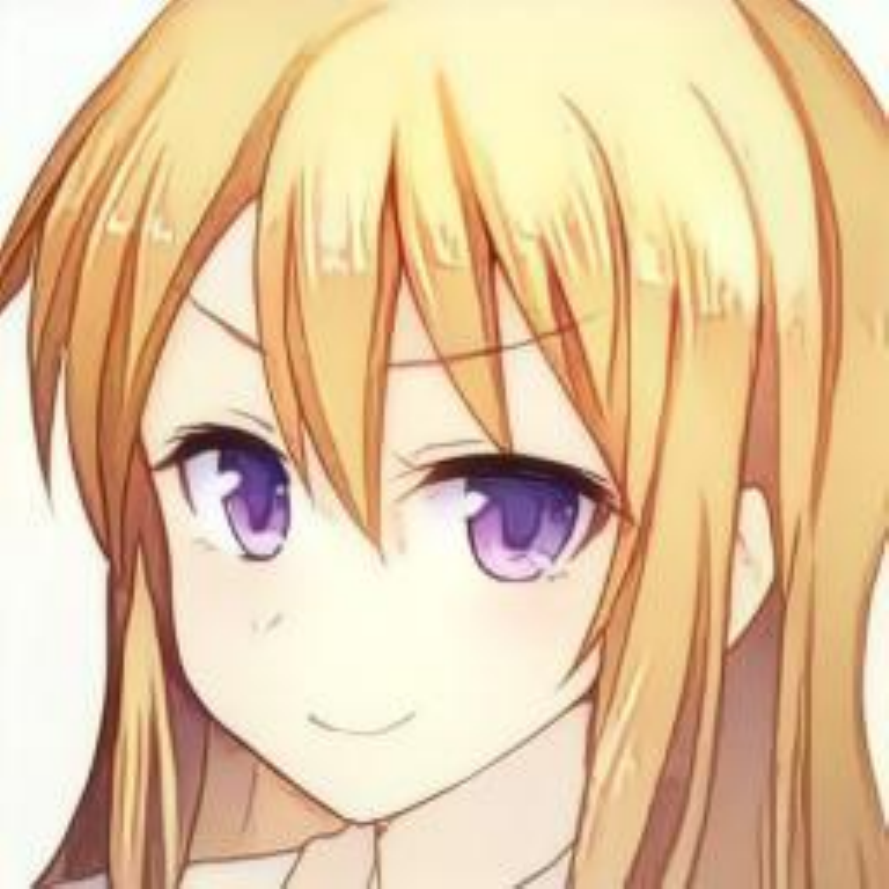}\vspace{4pt}
        \includegraphics[width=\linewidth]{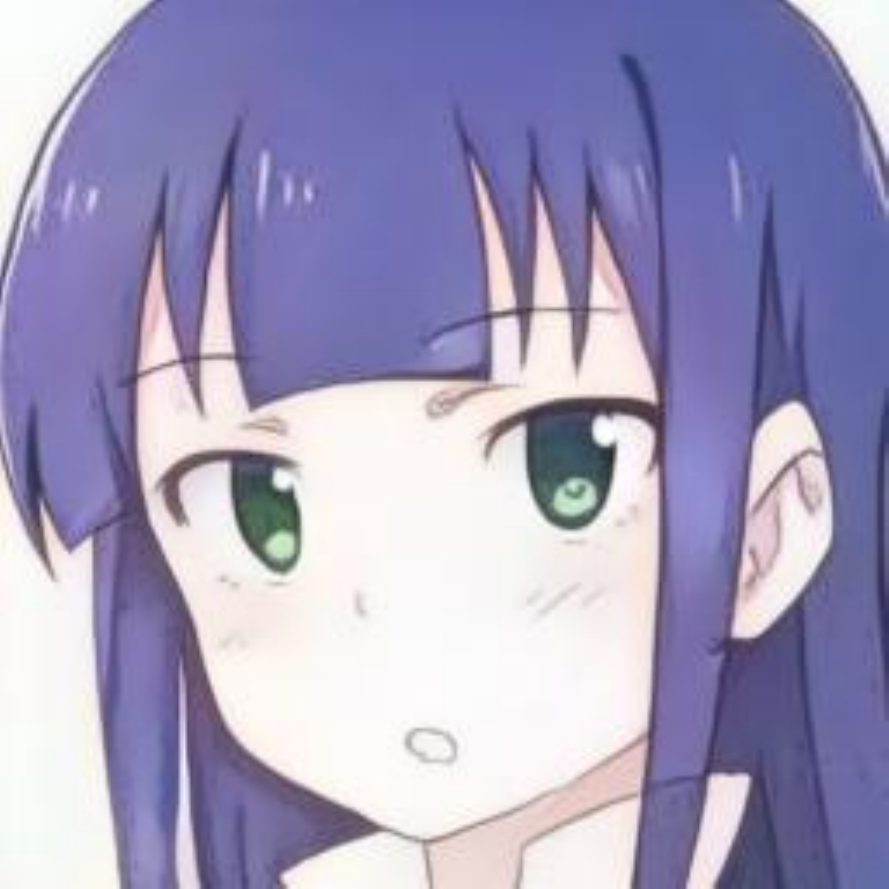}\vspace{4pt}
        \includegraphics[width=\linewidth]{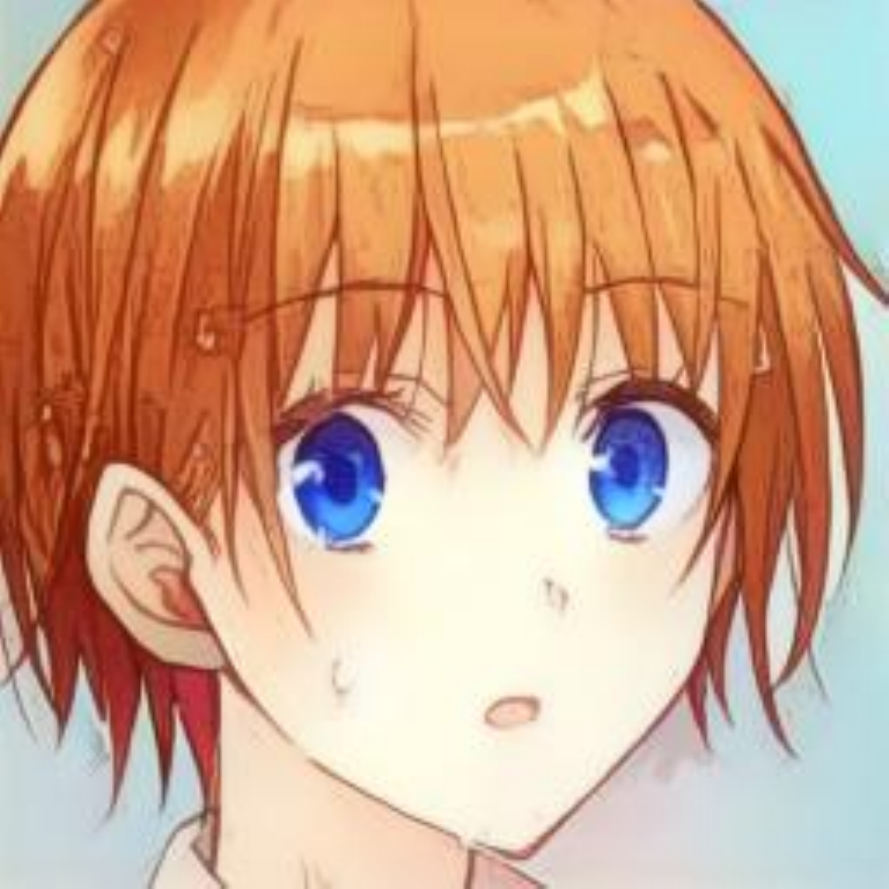}\vspace{4pt}
    \end{minipage}
    }
    \subfigure[]{
    \begin{minipage}[b]{0.125\linewidth}
        \includegraphics[width=\linewidth]{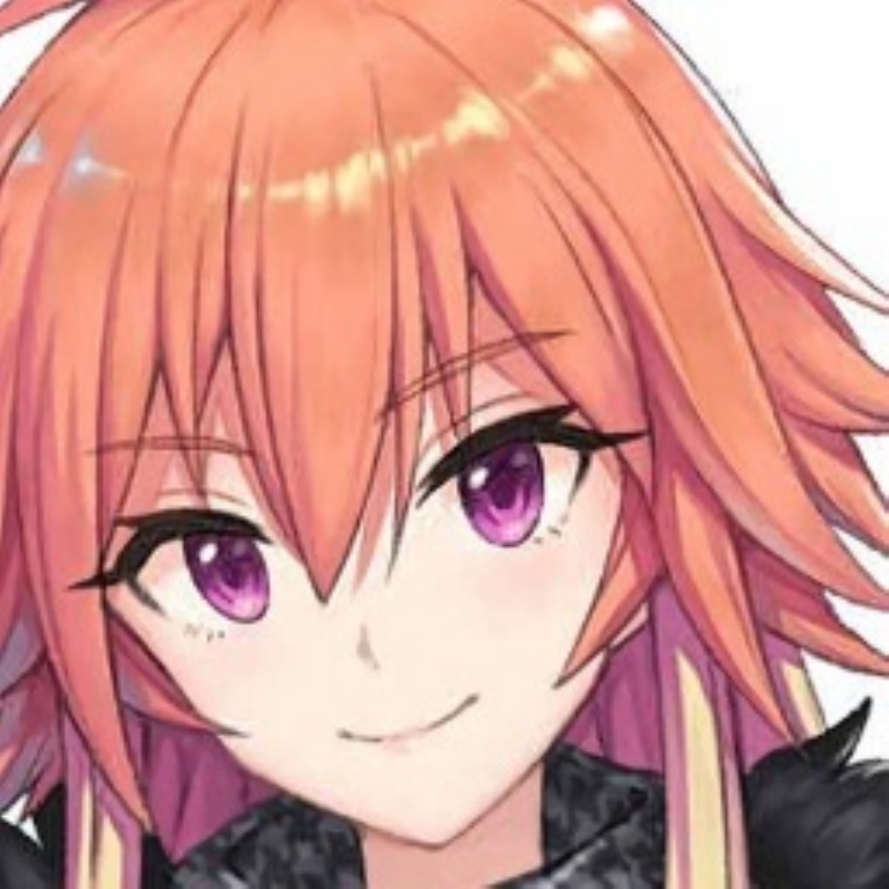}\vspace{4pt}
        \includegraphics[width=\linewidth]{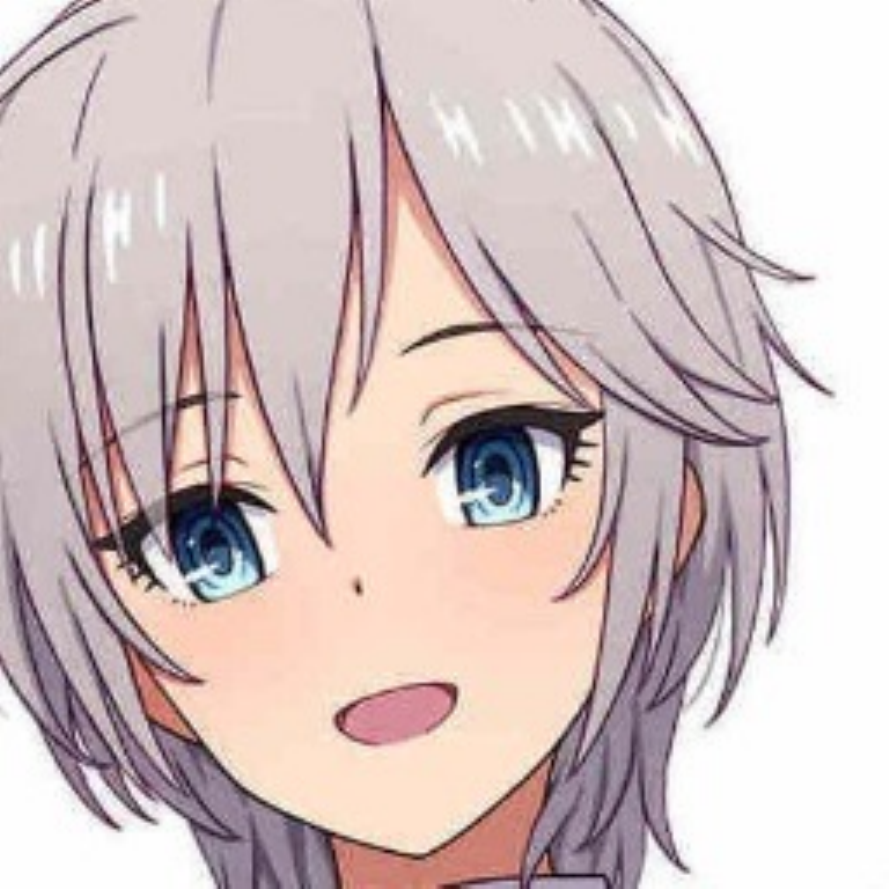}\vspace{4pt}
        \includegraphics[width=\linewidth]{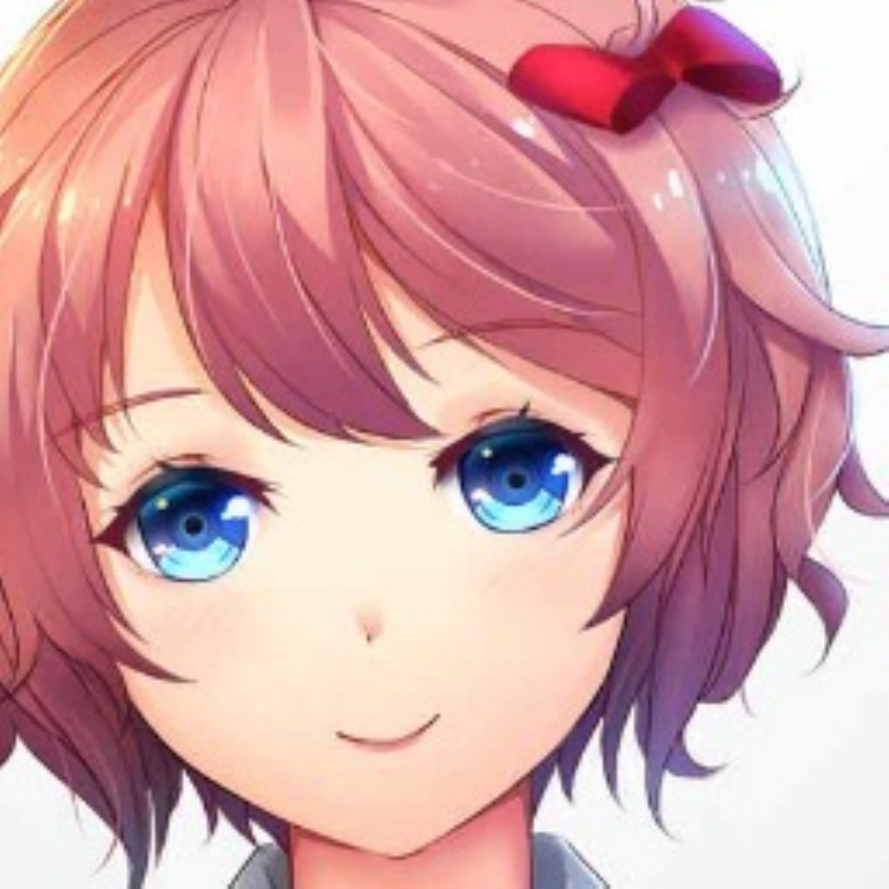}\vspace{4pt}
        \includegraphics[width=\linewidth]{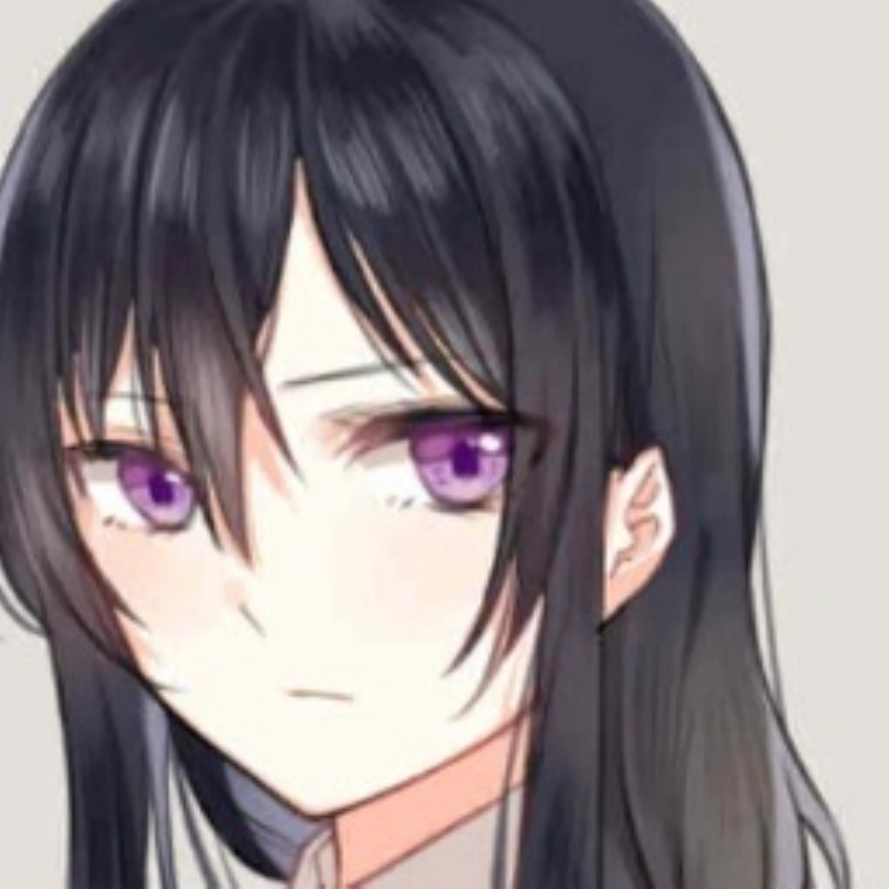}\vspace{4pt}
        \includegraphics[width=\linewidth]{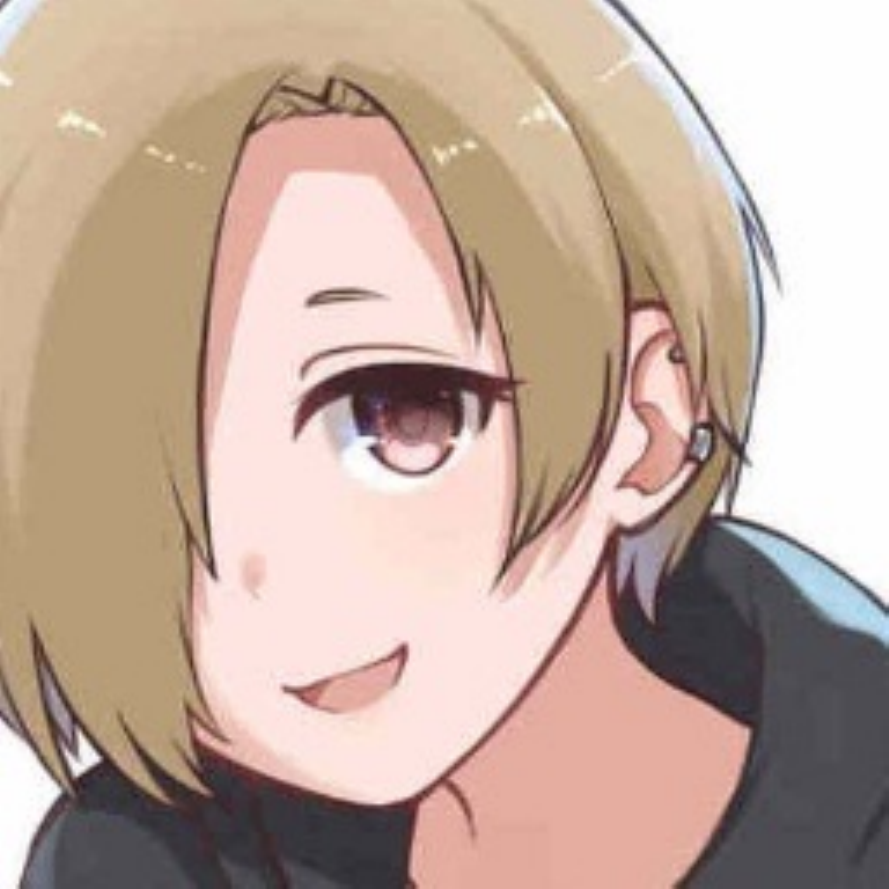}\vspace{4pt}
        \includegraphics[width=\linewidth]{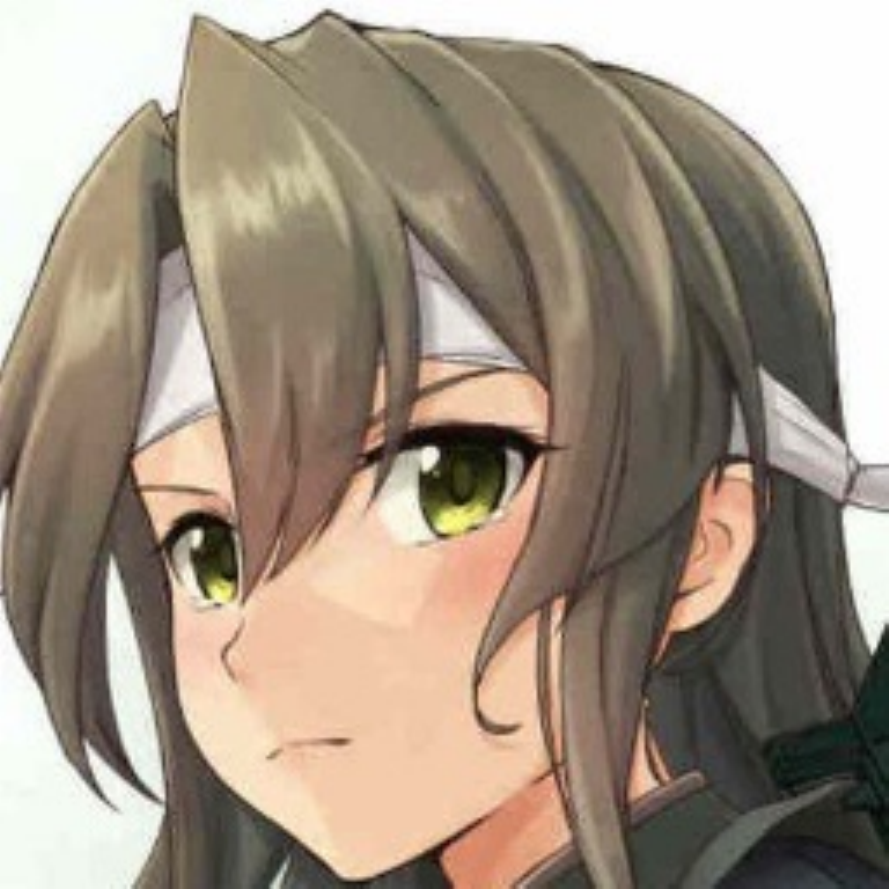}\vspace{4pt}
        \includegraphics[width=\linewidth]{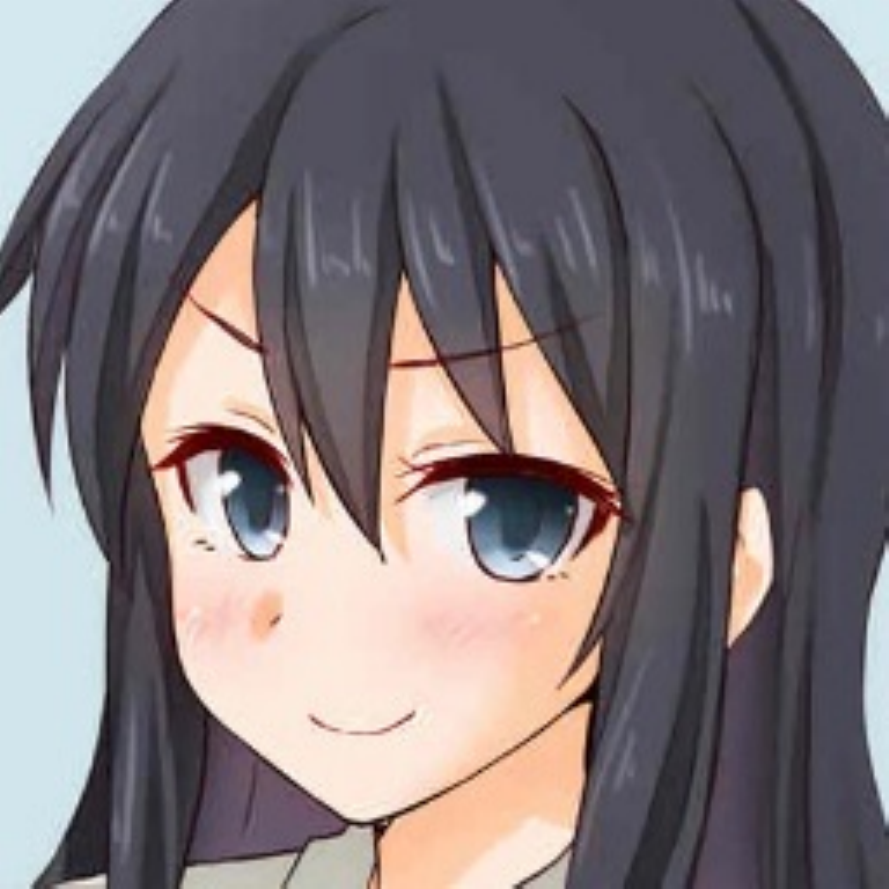}\vspace{4pt}
        \includegraphics[width=\linewidth]{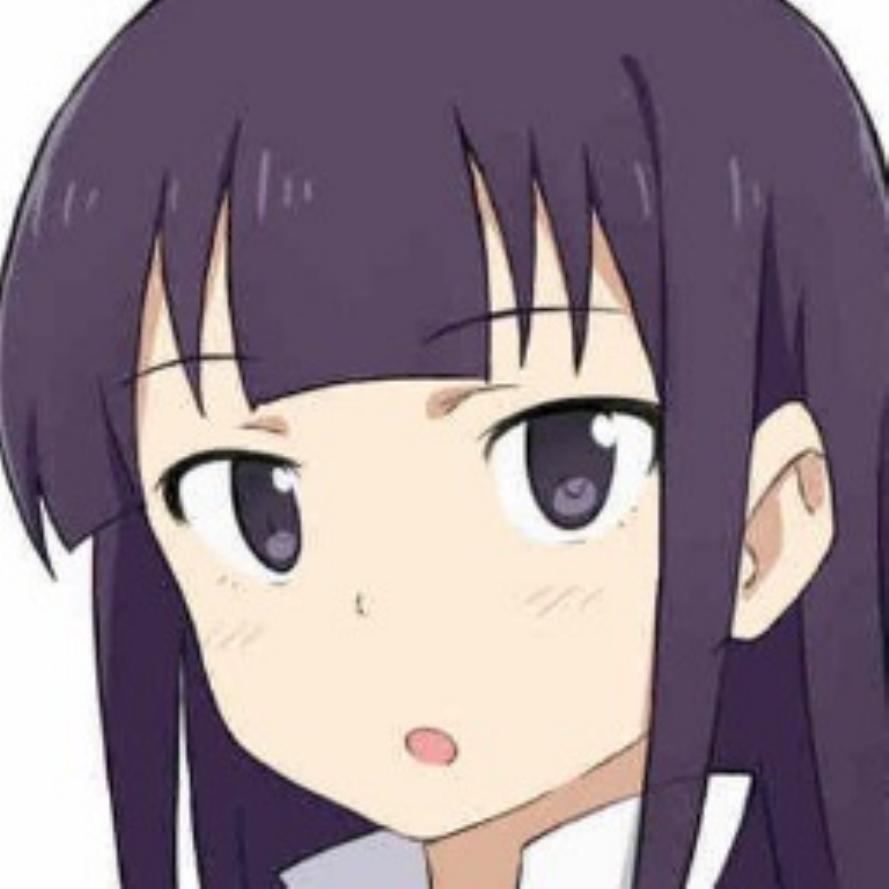}\vspace{4pt}
        \includegraphics[width=\linewidth]{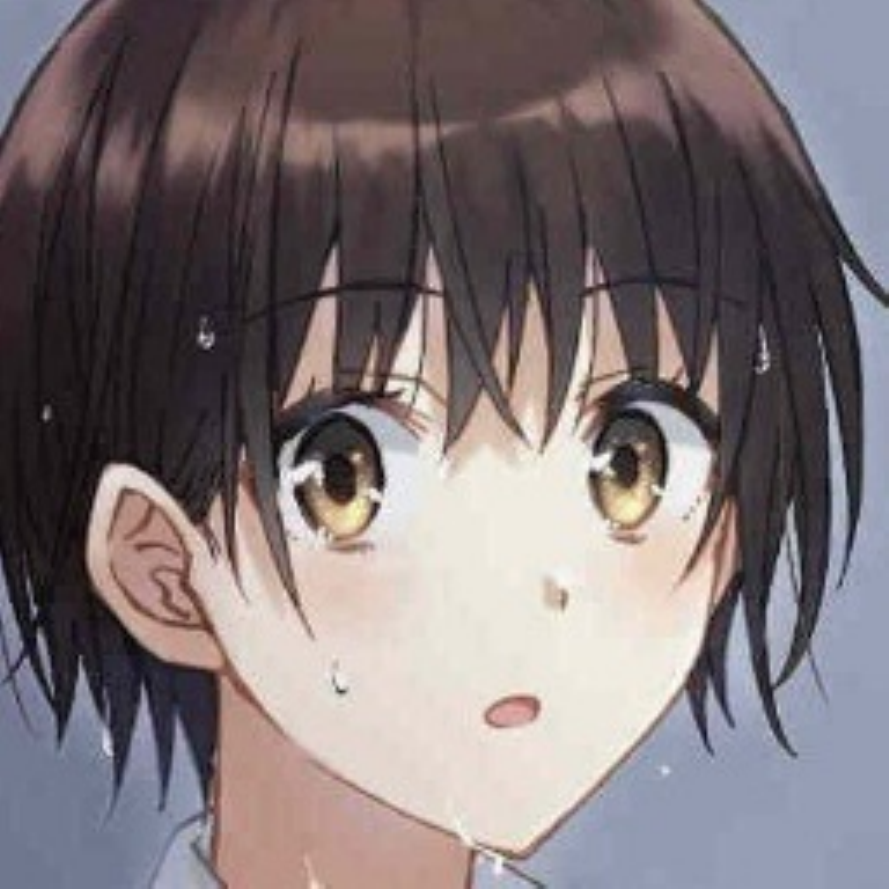}\vspace{4pt}
    \end{minipage}
    }
    \caption{Qualitative comparison for anime face with homochromatic pupils. (a) reference images, (b) line drawings, (c) Lee et al.\cite{lee2020reference}, (d) Li et al.\cite{li2022eliminating}, (e) Cao et al.\cite{cao2022attention}, (f) AnimeDiffusion, and (g) original color images.}
    \label{fig:single_qualitative}
\end{figure*}
\begin{figure*}[hbt]
    \centering
    \subfigure[]{
    \begin{minipage}[b]{0.125\linewidth}
        \includegraphics[width=\linewidth]{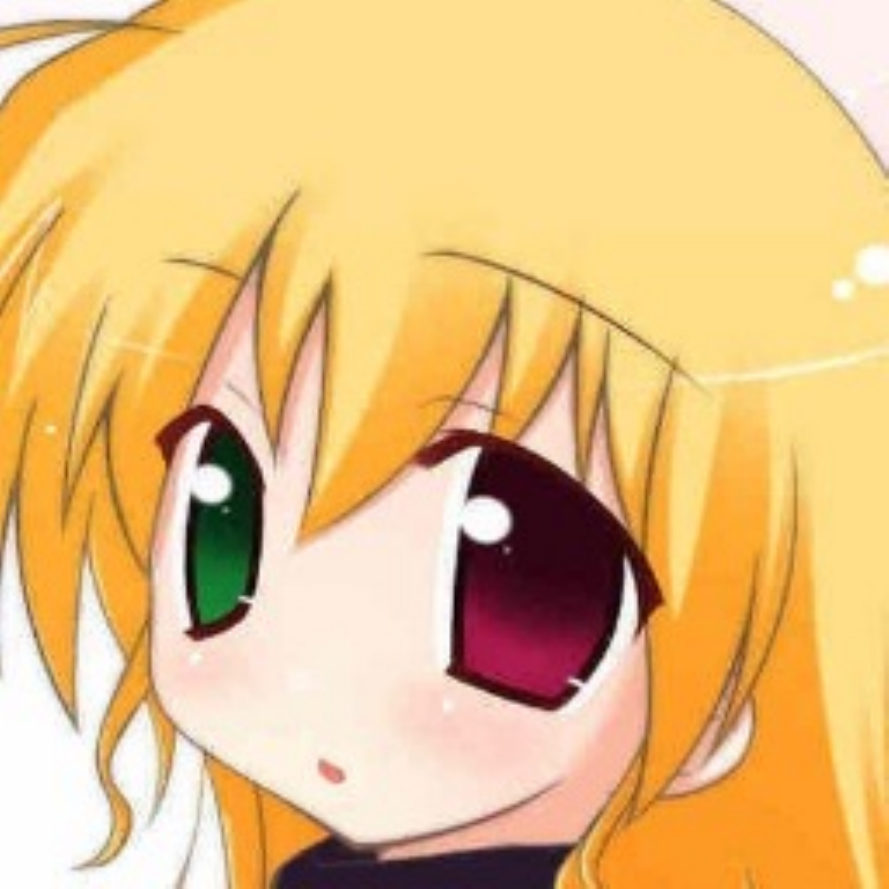}\vspace{4pt}
        \includegraphics[width=\linewidth]{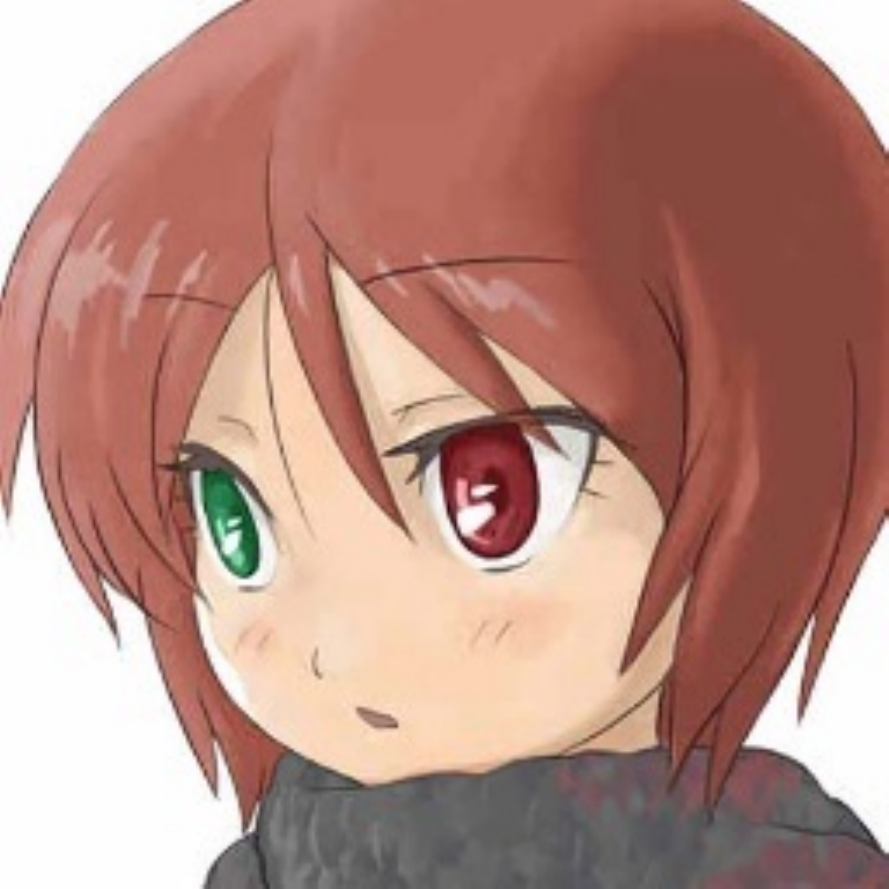}\vspace{4pt}
        \includegraphics[width=\linewidth]{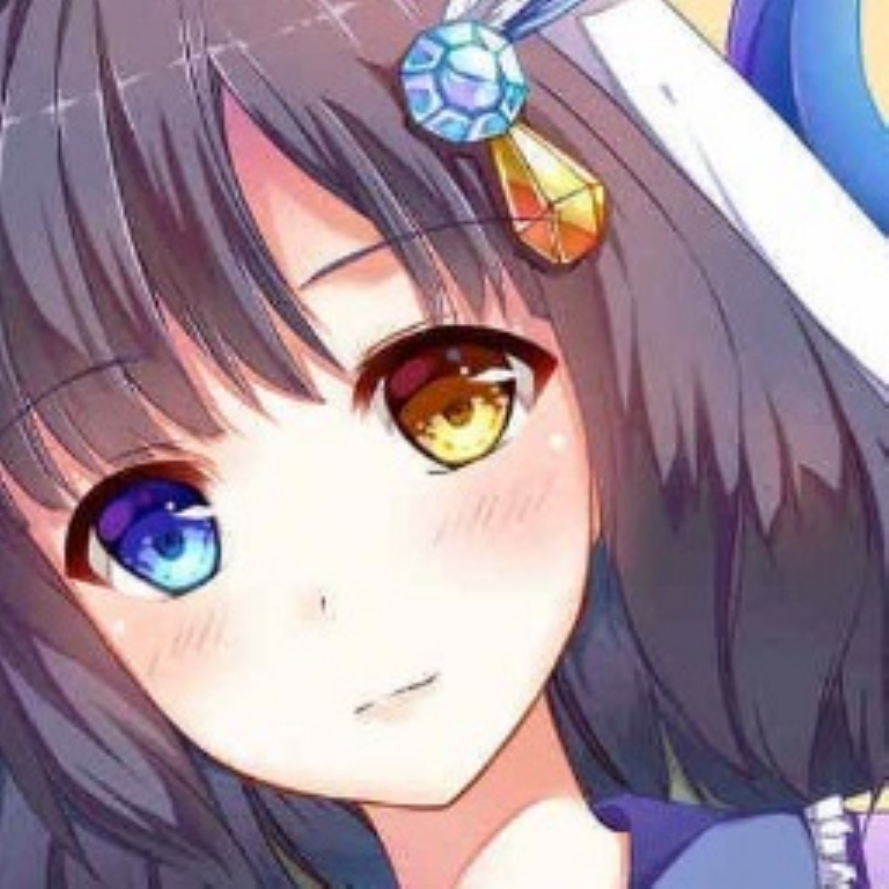}\vspace{4pt}
        \includegraphics[width=\linewidth]{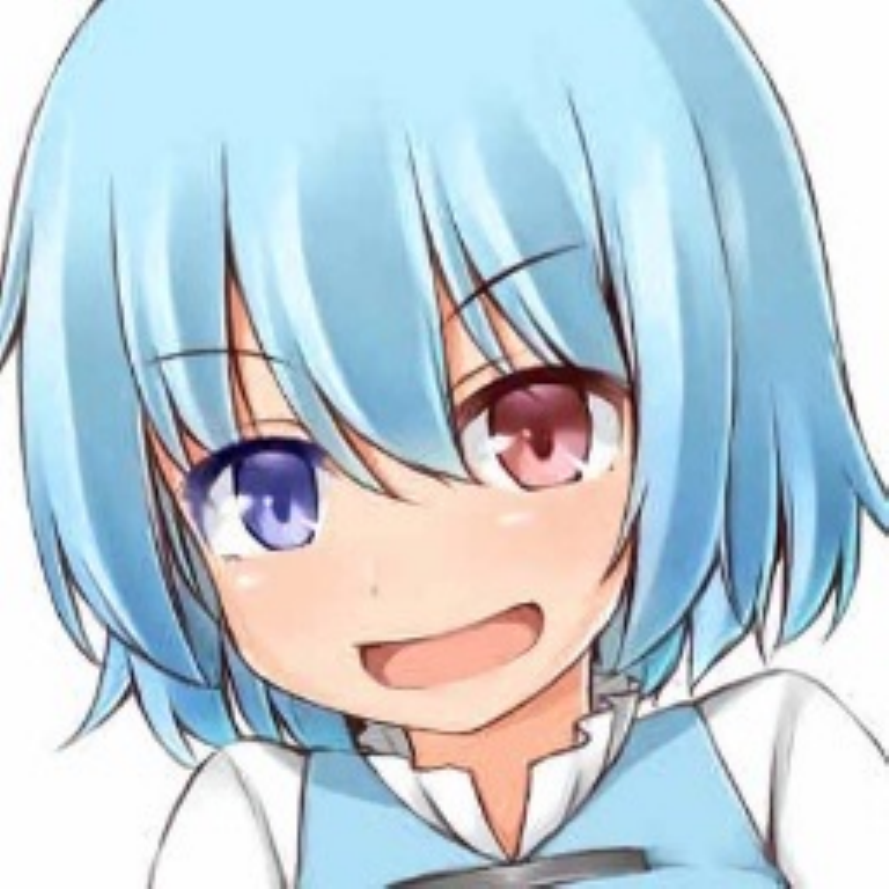}\vspace{4pt}
        \includegraphics[width=\linewidth]{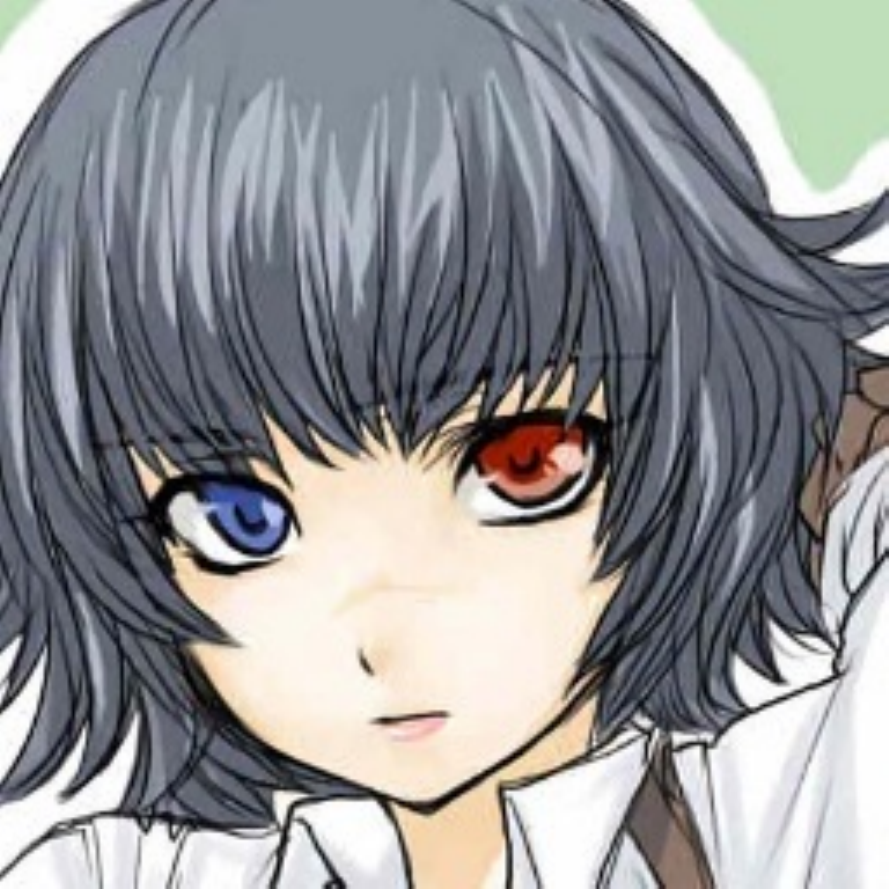}\vspace{4pt}
    \end{minipage}
    }
    \subfigure[]{
    \begin{minipage}[b]{0.125\linewidth}
        \includegraphics[width=\linewidth]{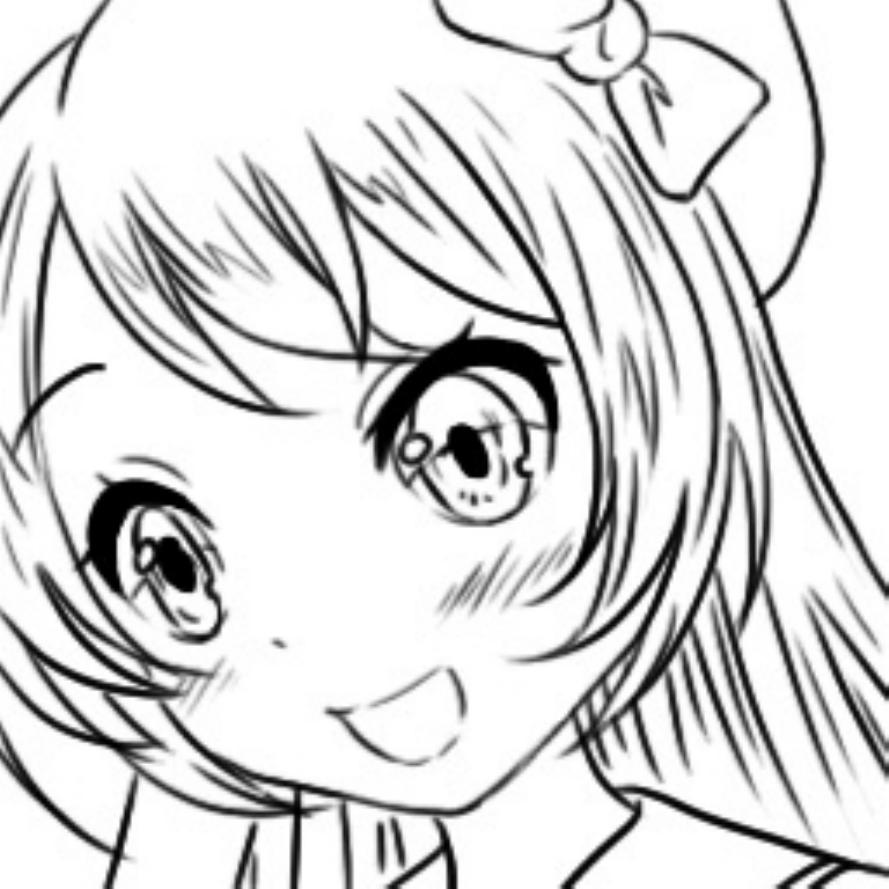}\vspace{4pt}
        \includegraphics[width=\linewidth]{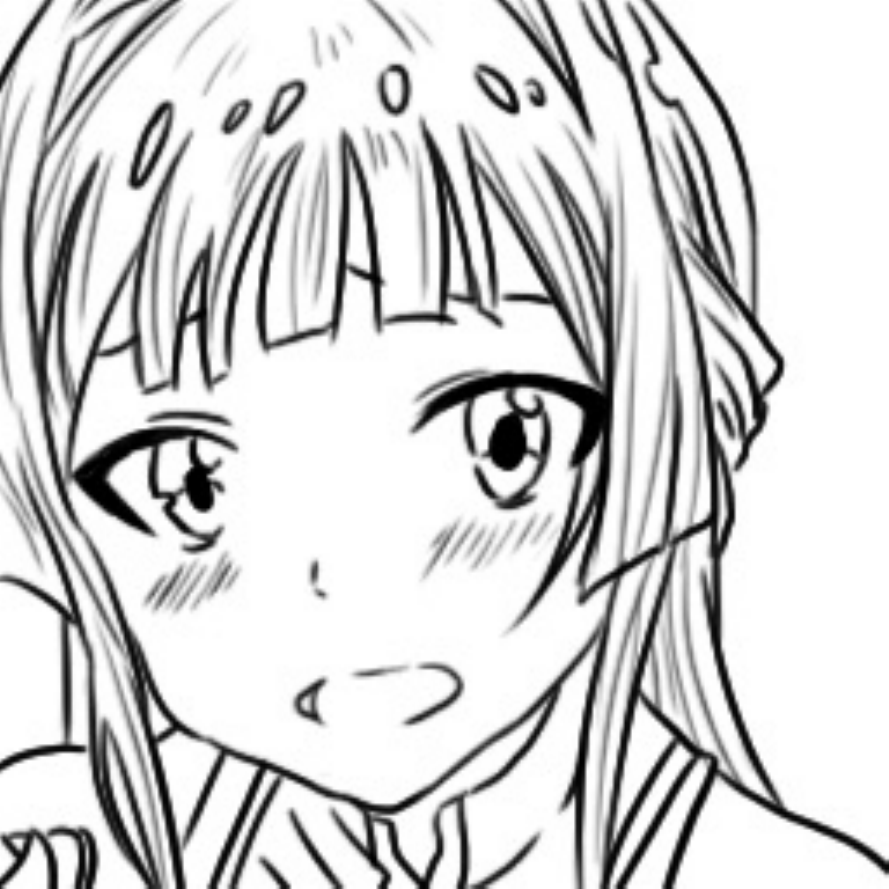}\vspace{4pt}
        \includegraphics[width=\linewidth]{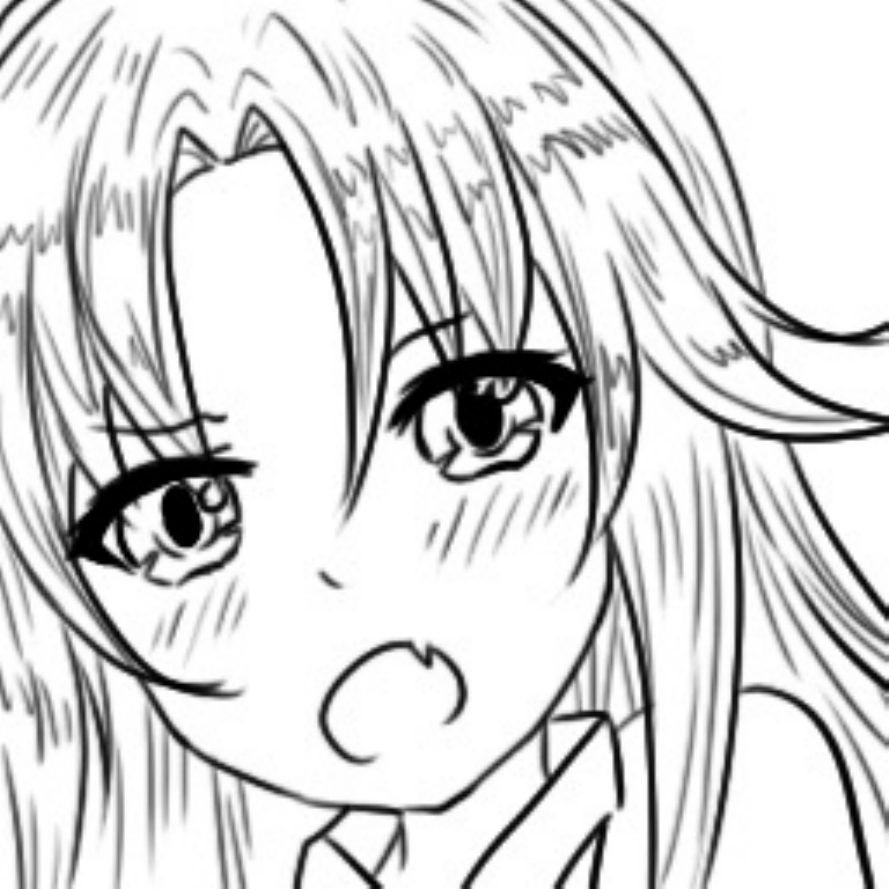}\vspace{4pt}
        \includegraphics[width=\linewidth]{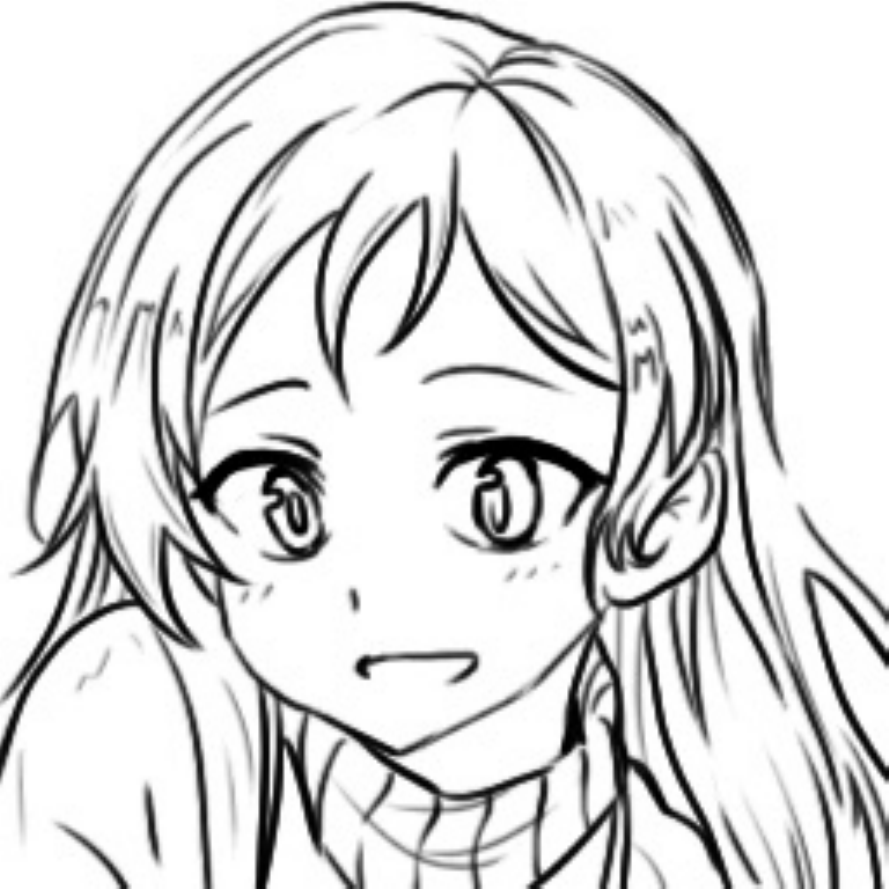}\vspace{4pt}
        \includegraphics[width=\linewidth]{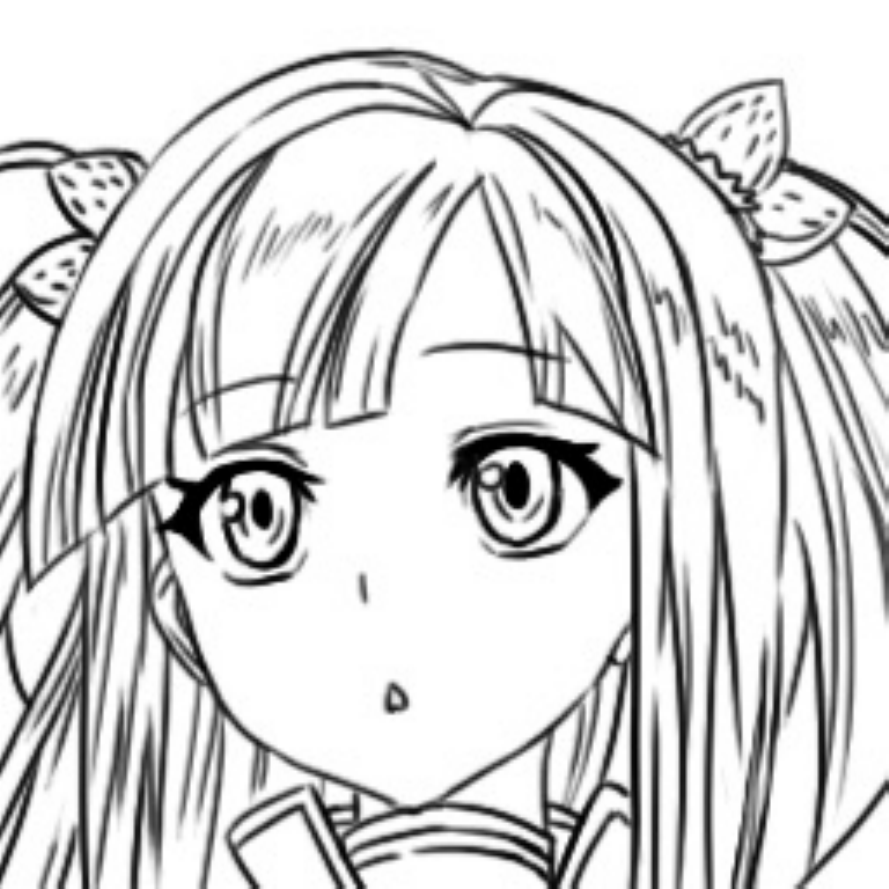}\vspace{4pt}
    \end{minipage}
    }
    \subfigure[]{
    \begin{minipage}[b]{0.125\linewidth}
        \includegraphics[width=\linewidth]{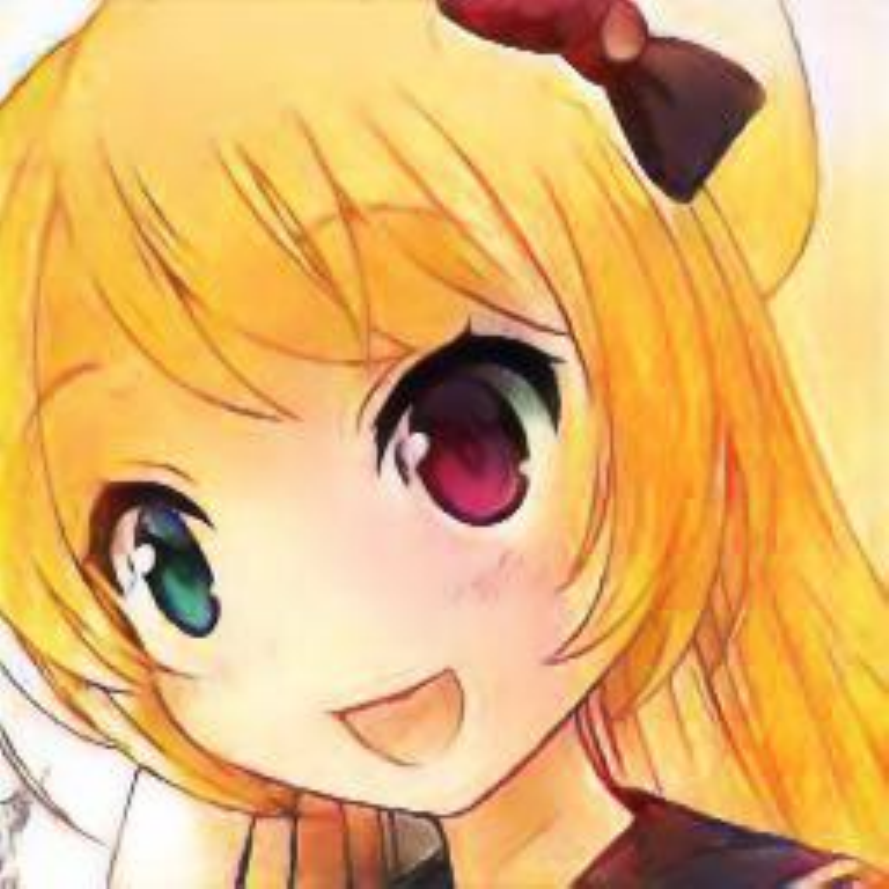}\vspace{4pt}
        \includegraphics[width=\linewidth]{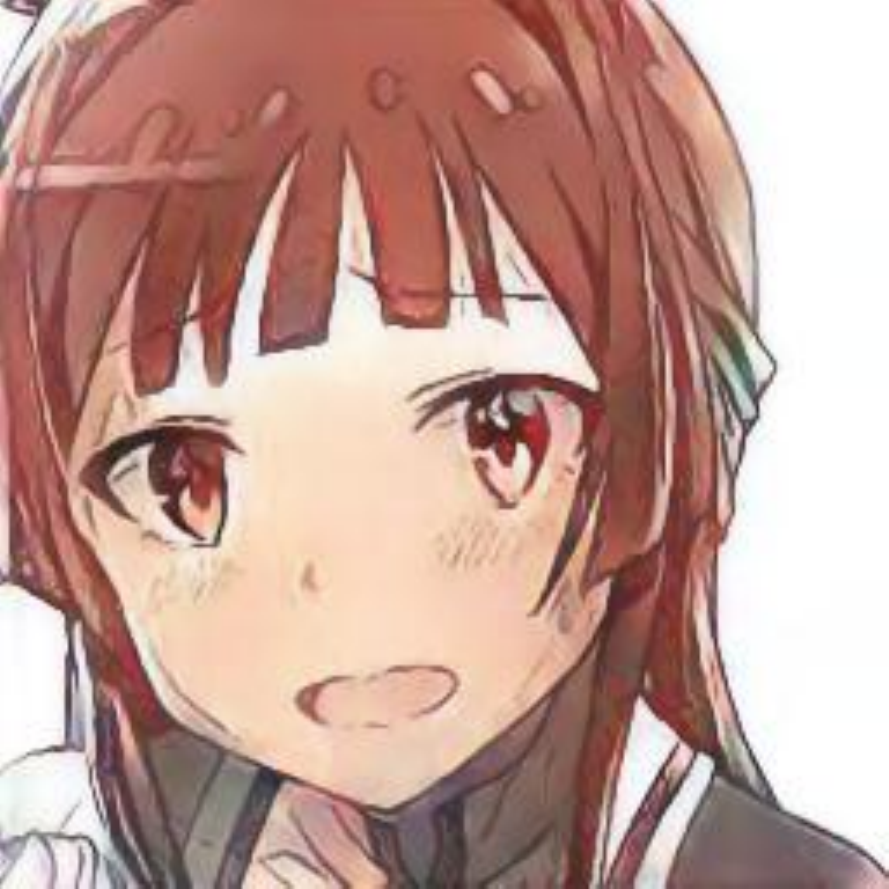}\vspace{4pt}
        \includegraphics[width=\linewidth]{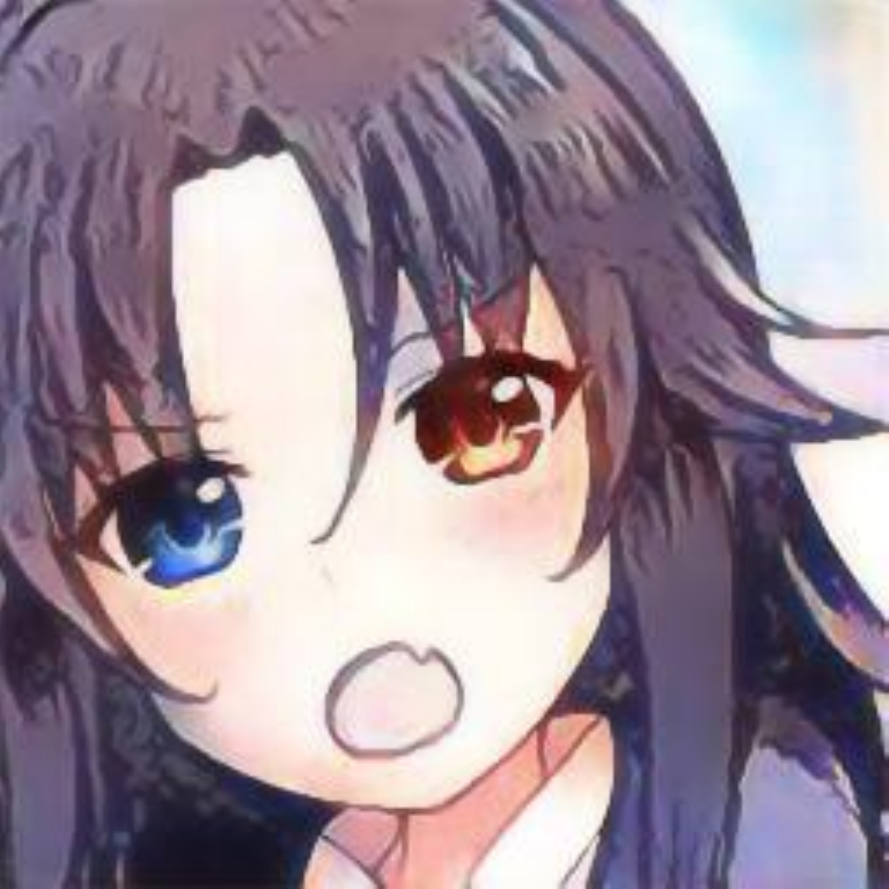}\vspace{4pt}
        \includegraphics[width=\linewidth]{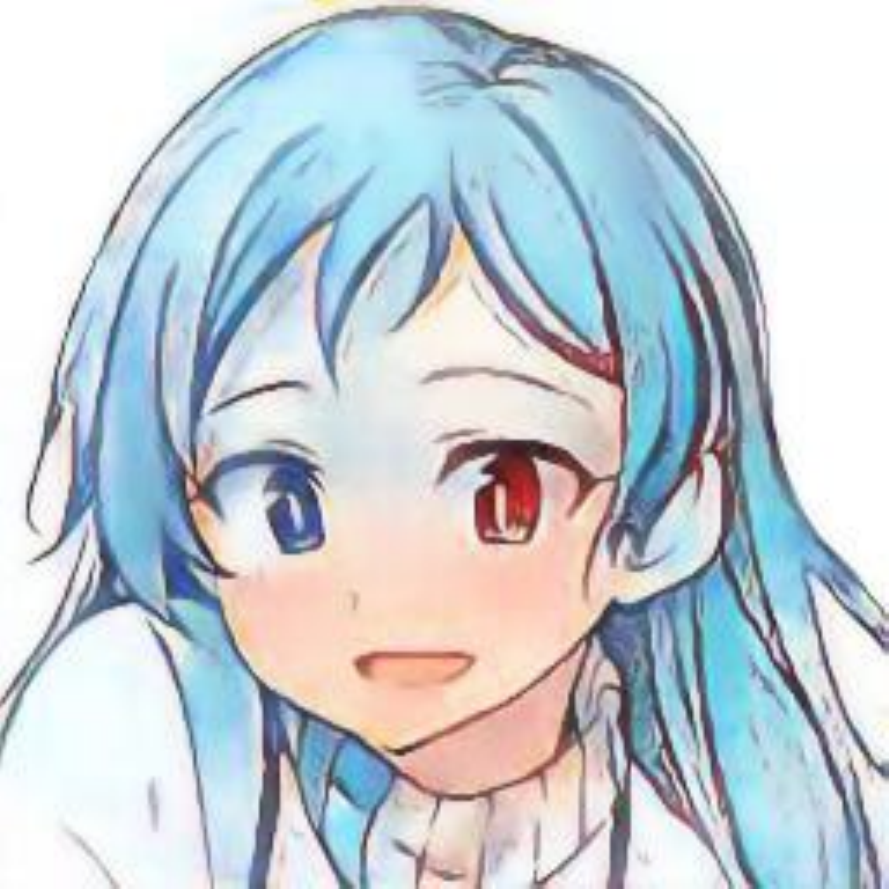}\vspace{4pt}
        \includegraphics[width=\linewidth]{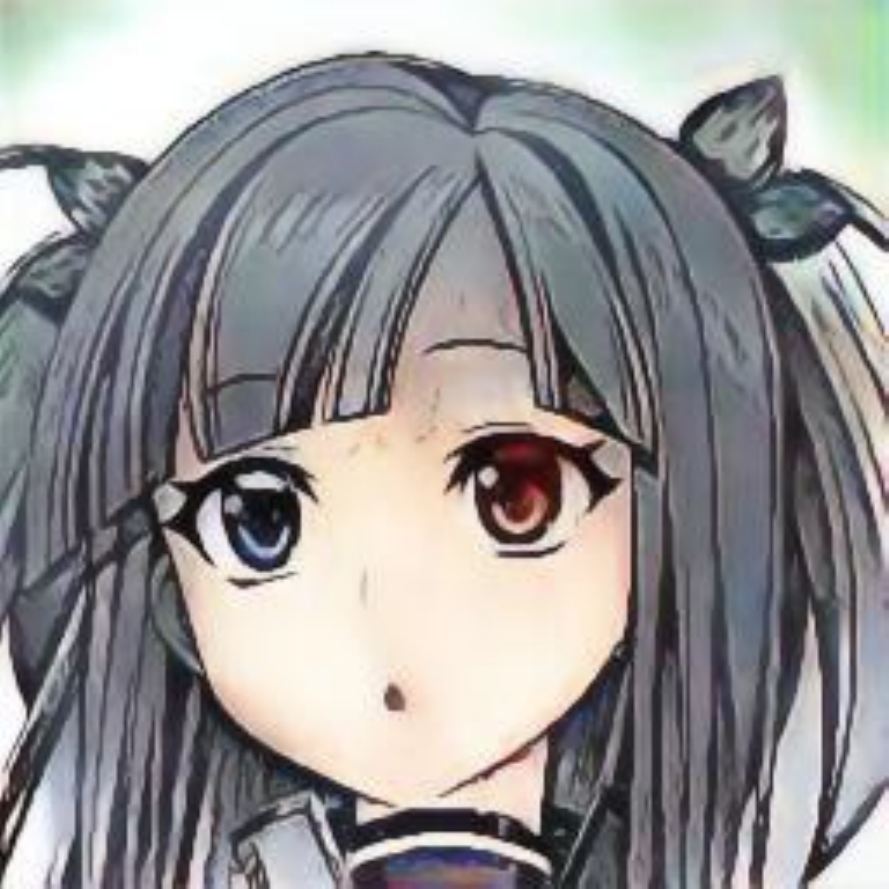}\vspace{4pt}
    \end{minipage}
    }
    \subfigure[]{
    \begin{minipage}[b]{0.125\linewidth}
        \includegraphics[width=\linewidth]{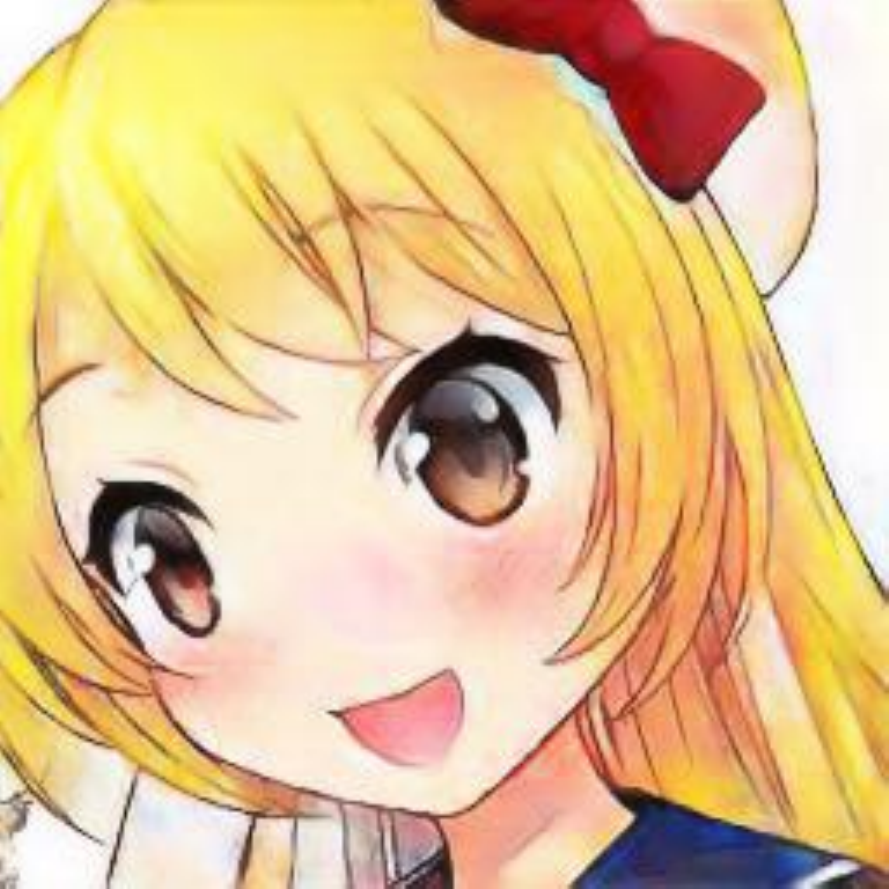}\vspace{4pt}
        \includegraphics[width=\linewidth]{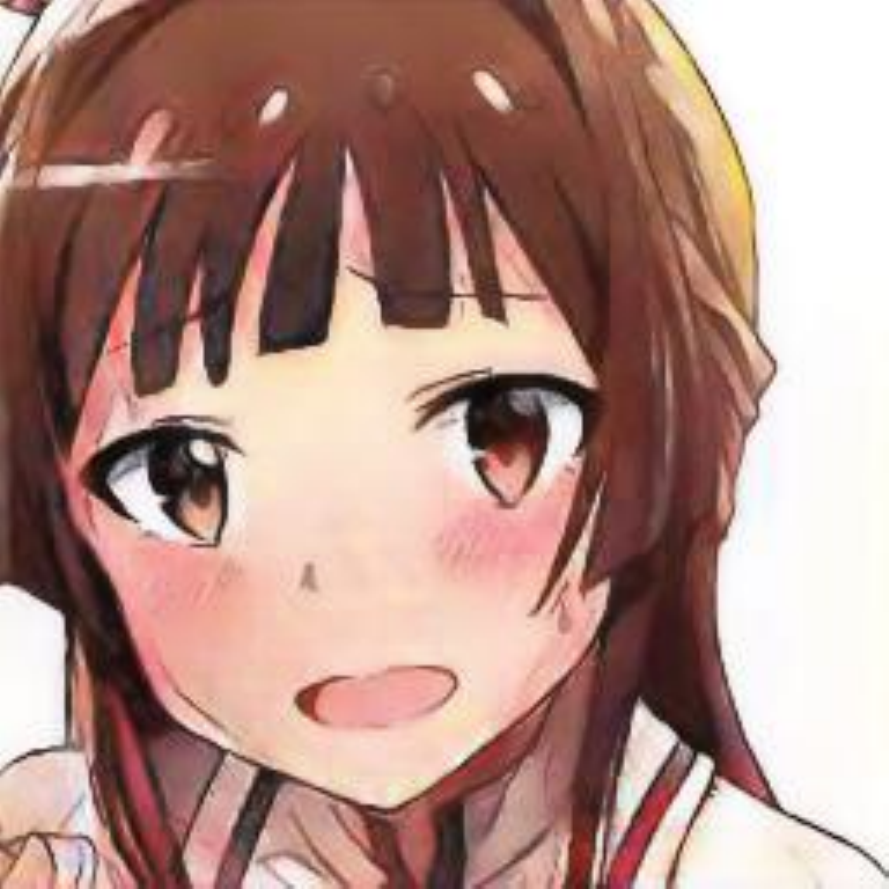}\vspace{4pt}
        \includegraphics[width=\linewidth]{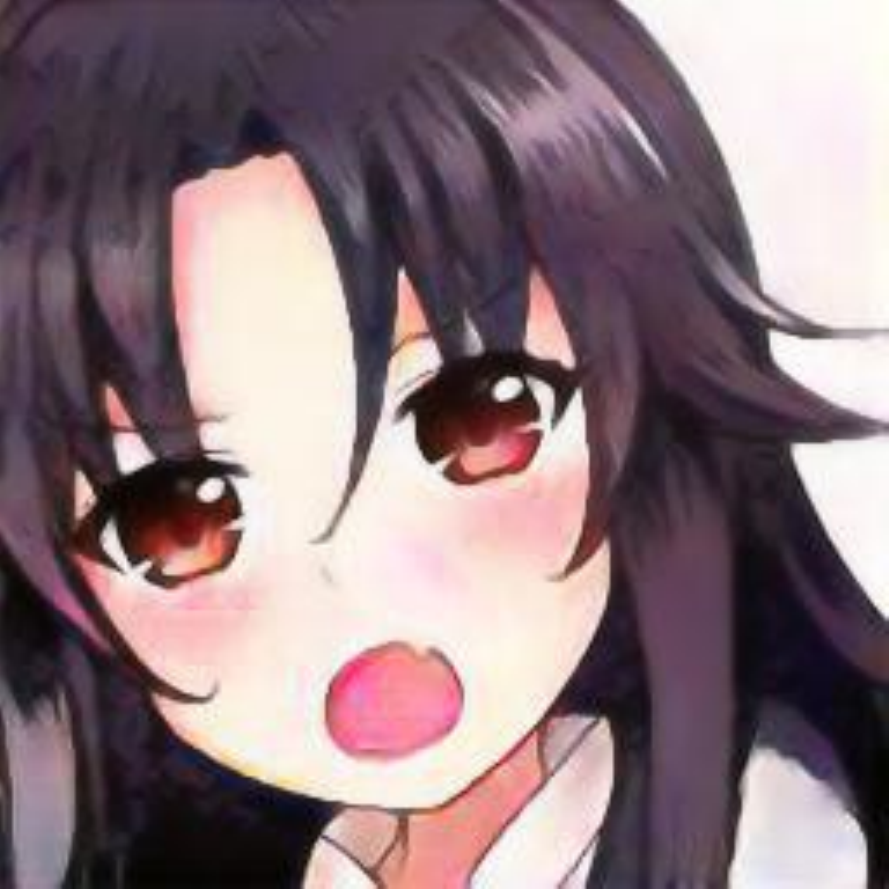}\vspace{4pt}
        \includegraphics[width=\linewidth]{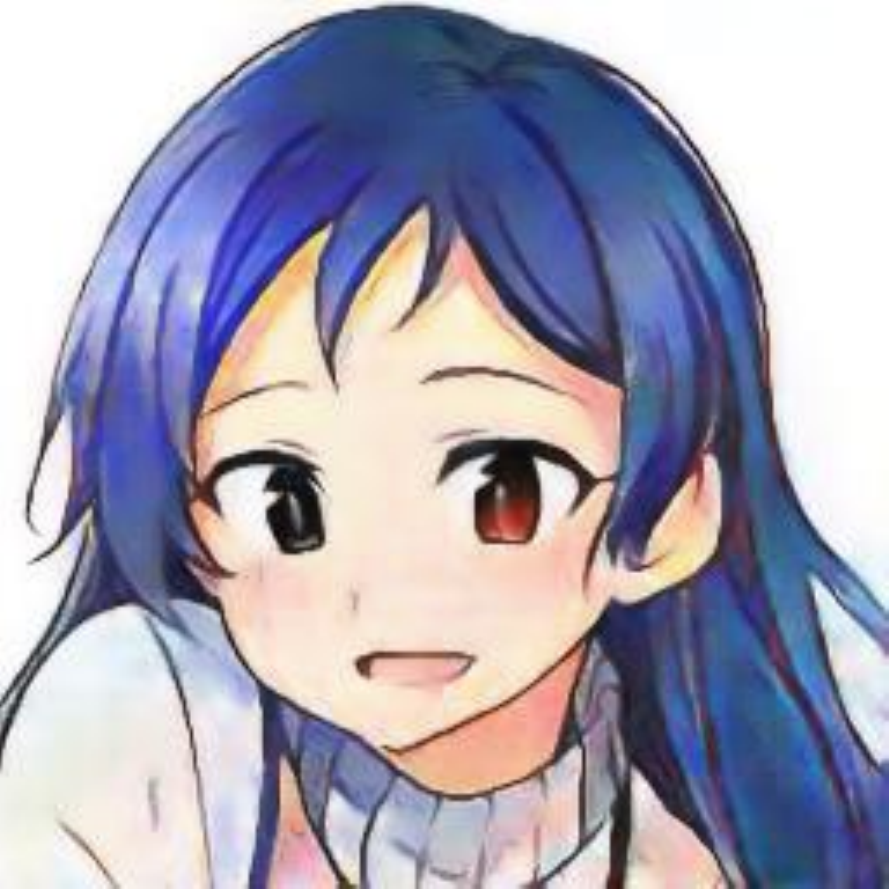}\vspace{4pt}
        \includegraphics[width=\linewidth]{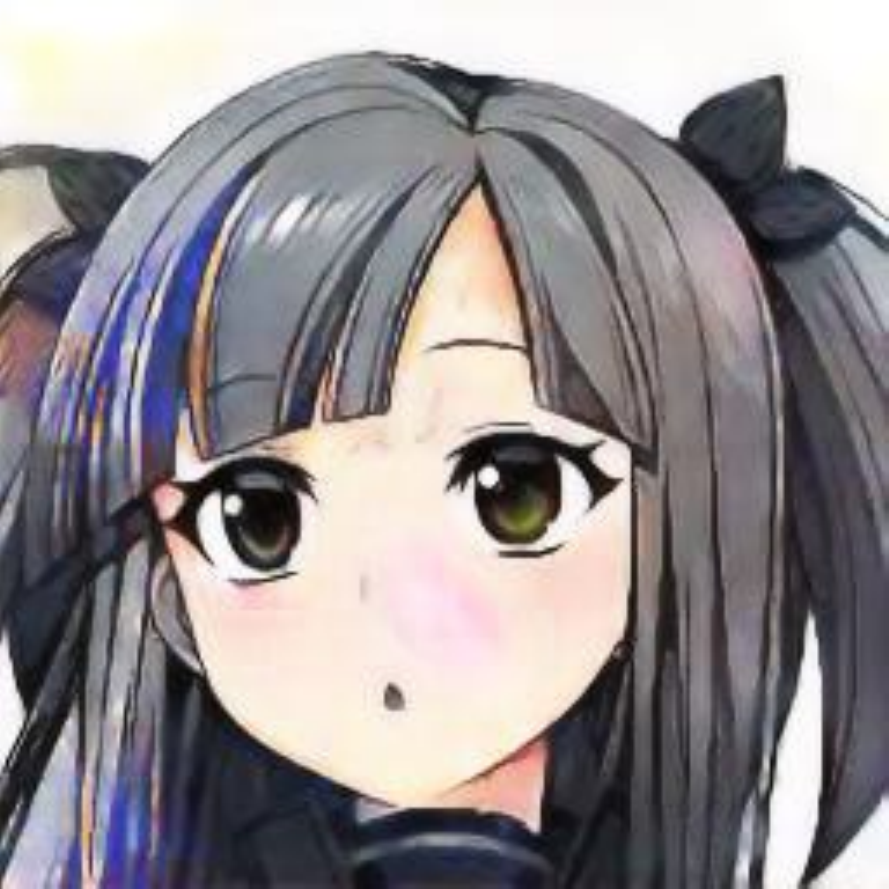}\vspace{4pt}
    \end{minipage}
    }
    \subfigure[]{
    \begin{minipage}[b]{0.125\linewidth}
        \includegraphics[width=\linewidth]{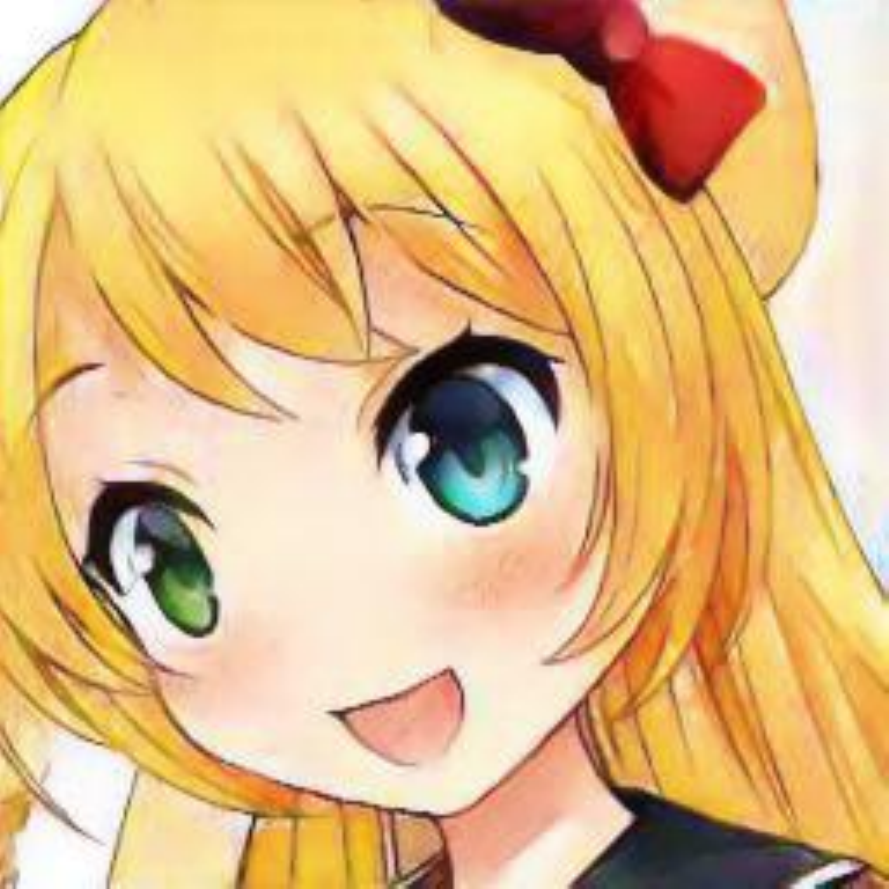}\vspace{4pt}
        \includegraphics[width=\linewidth]{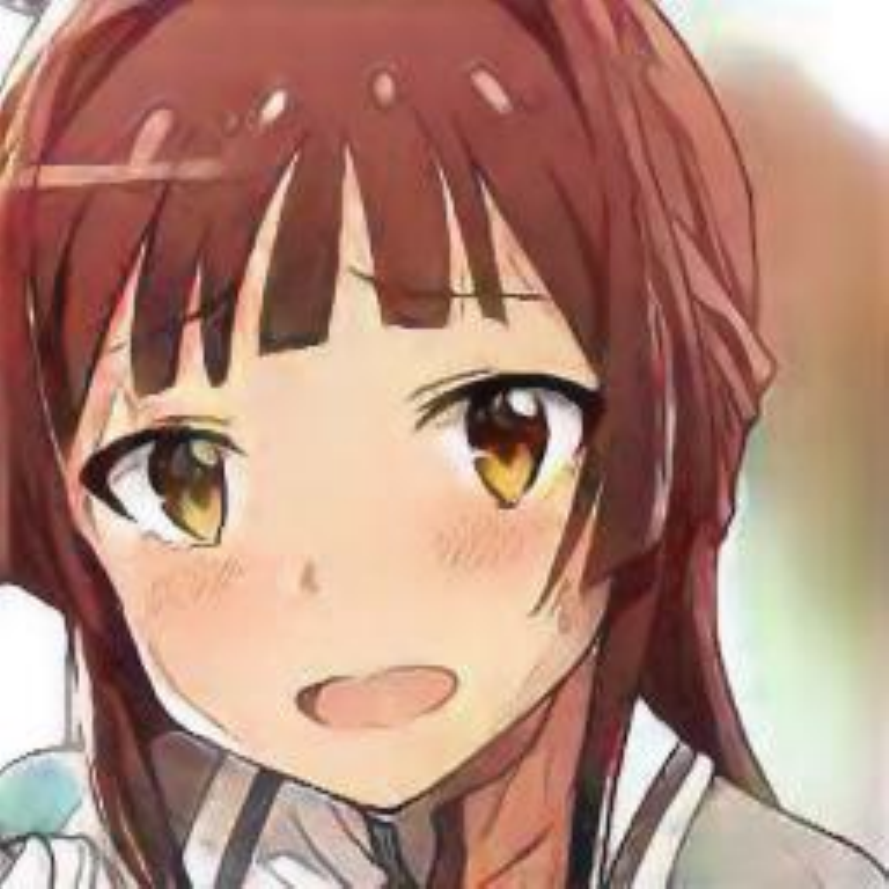}\vspace{4pt}
        \includegraphics[width=\linewidth]{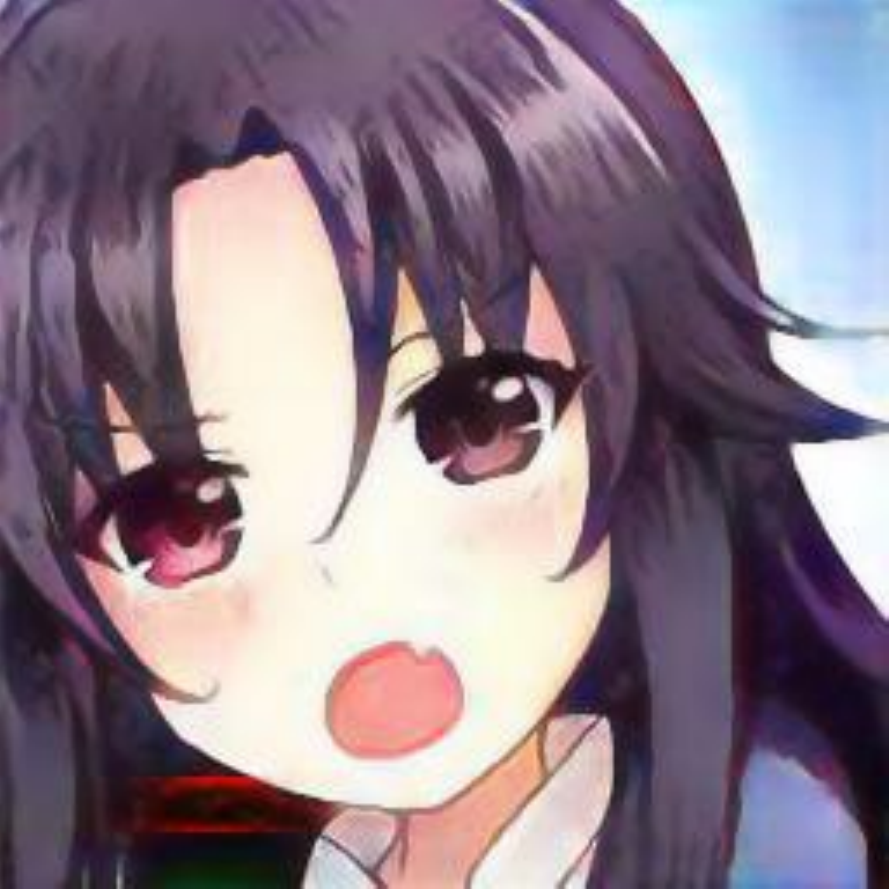}\vspace{4pt}
        \includegraphics[width=\linewidth]{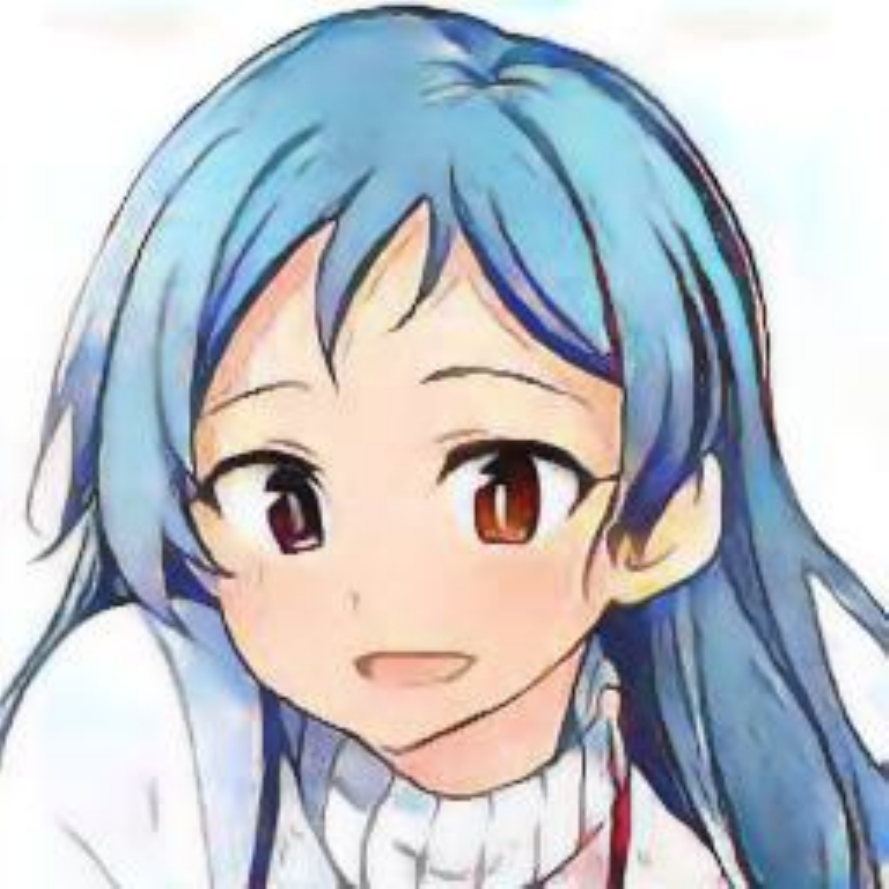}\vspace{4pt}
        \includegraphics[width=\linewidth]{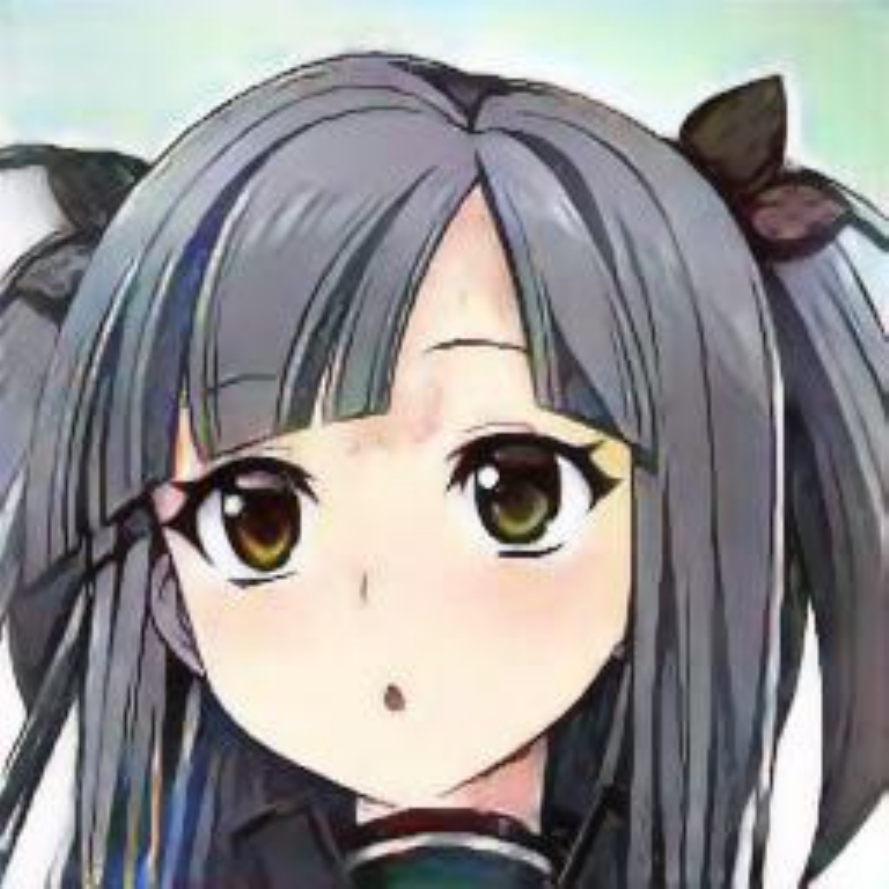}\vspace{4pt}
    \end{minipage}
    }
    \subfigure[]{
    \begin{minipage}[b]{0.125\linewidth}
        \includegraphics[width=\linewidth]{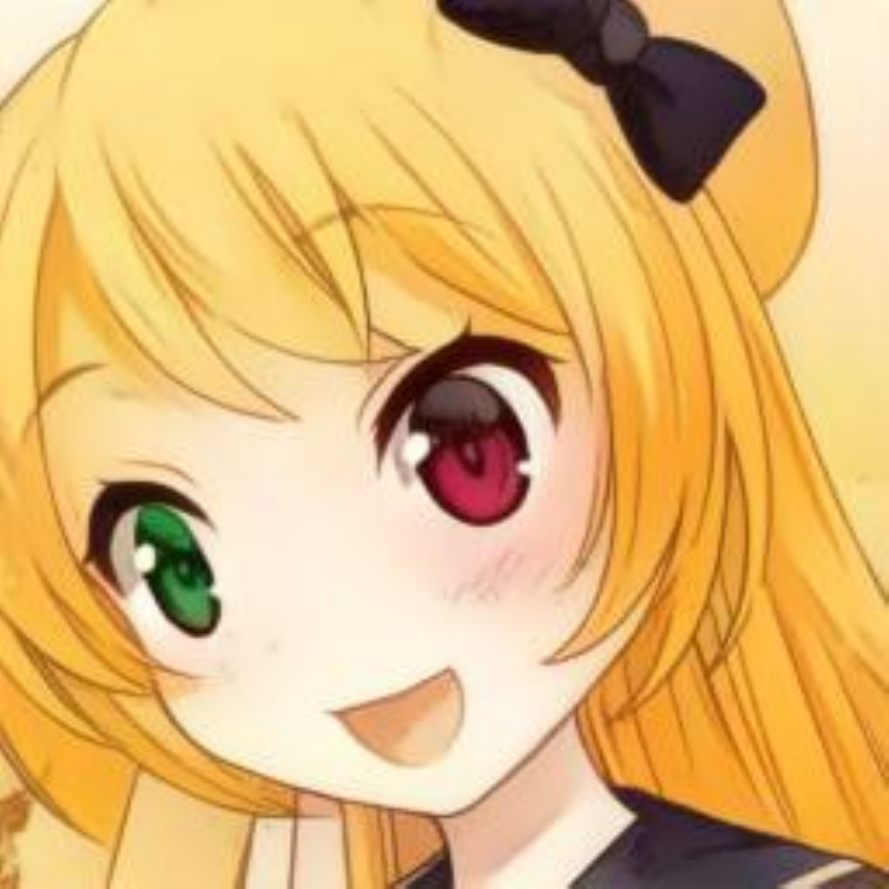}\vspace{4pt}
        \includegraphics[width=\linewidth]{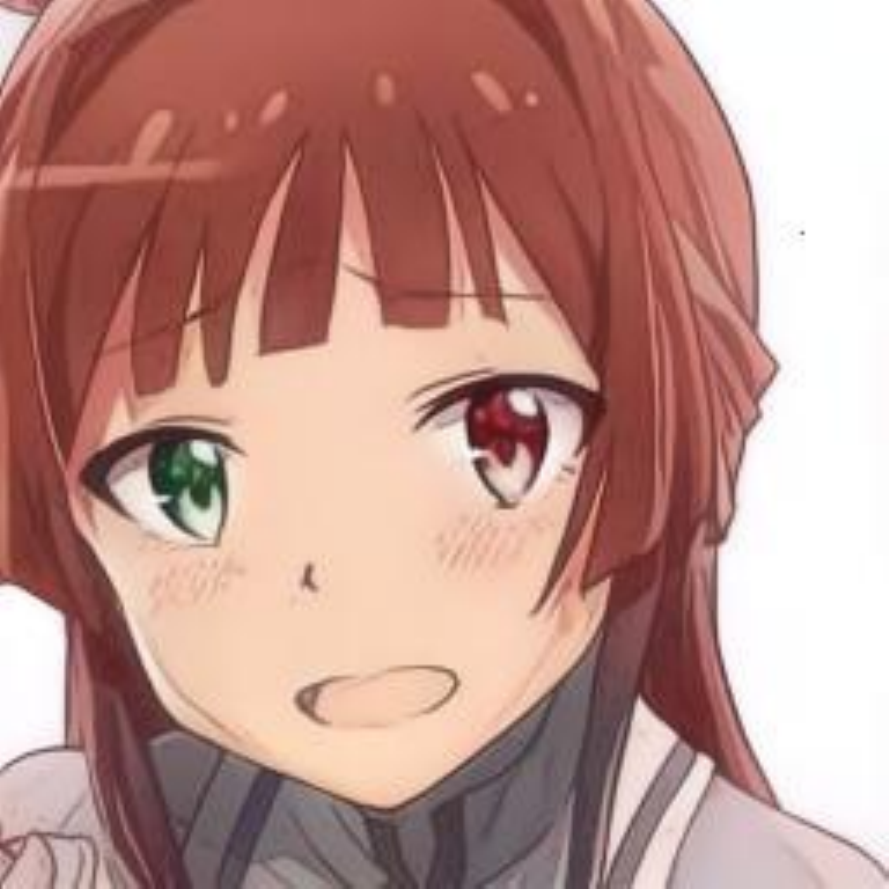}\vspace{4pt}
        \includegraphics[width=\linewidth]{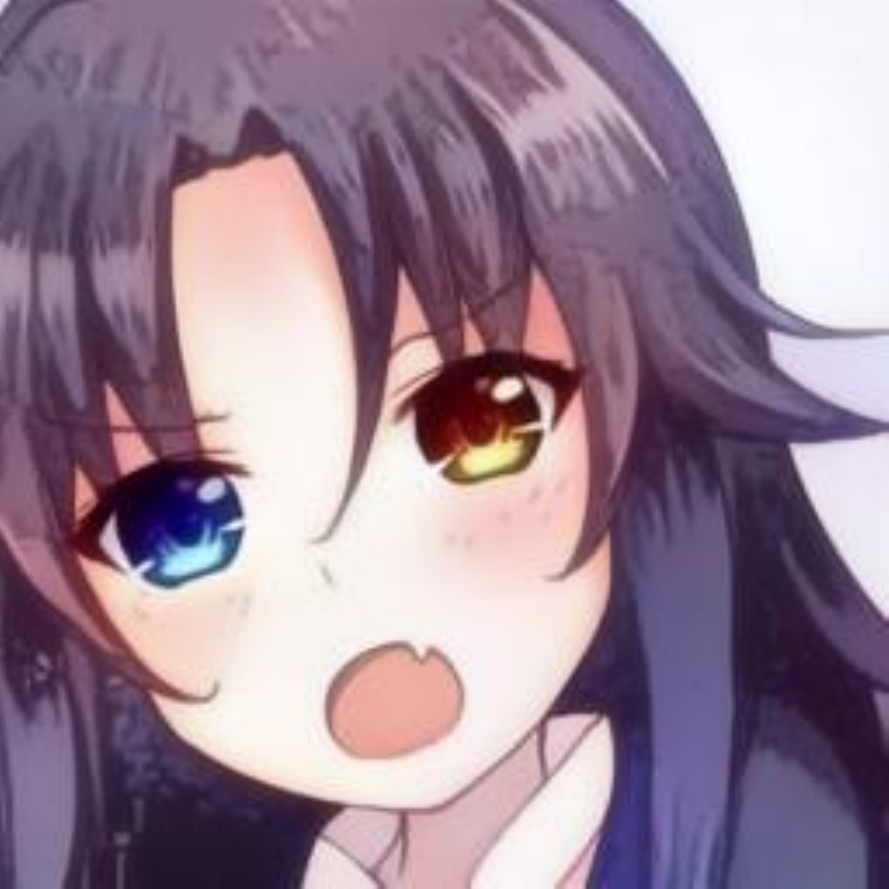}\vspace{4pt}
        \includegraphics[width=\linewidth]{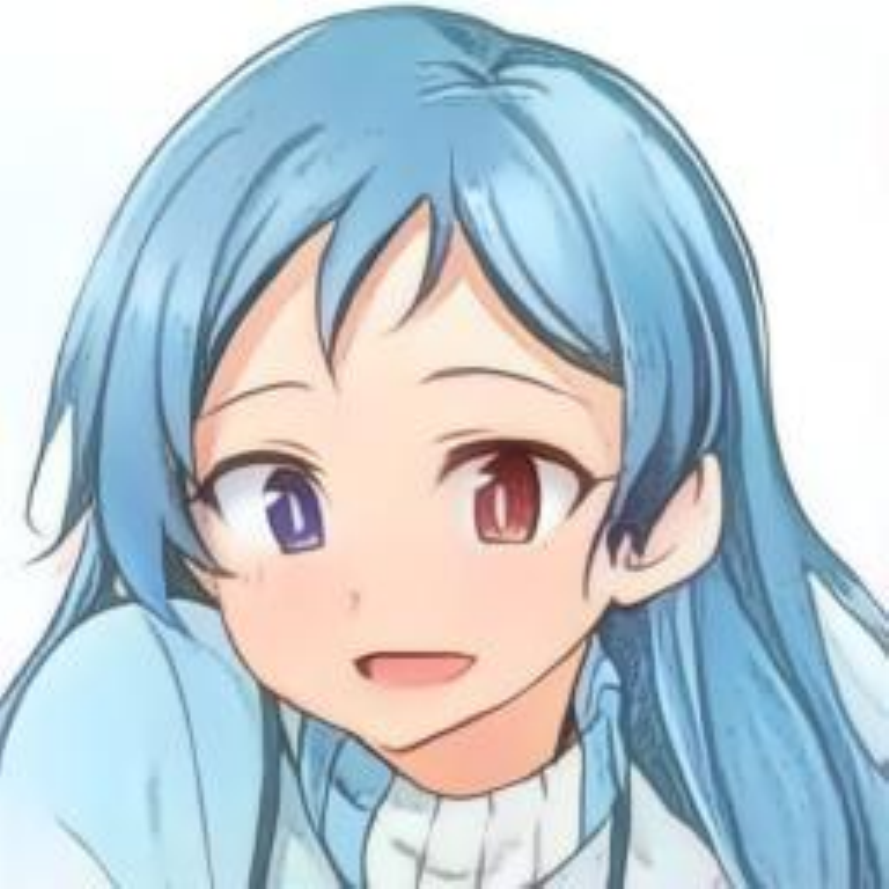}\vspace{4pt}
        \includegraphics[width=\linewidth]{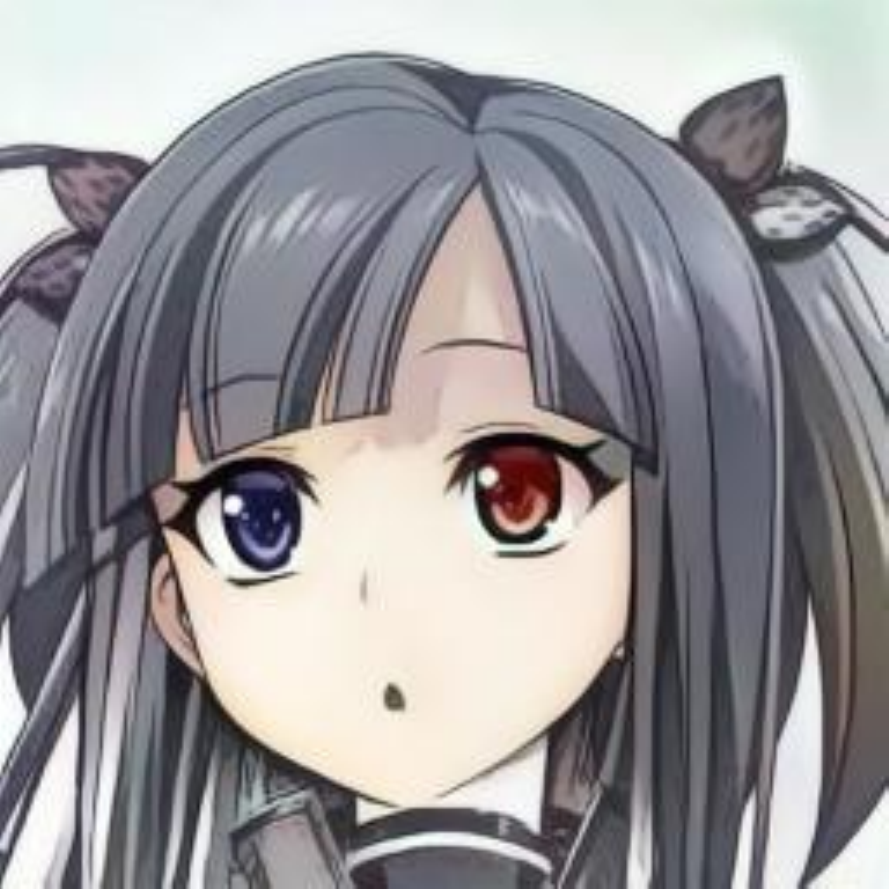}\vspace{4pt}
    \end{minipage}
    }
    \subfigure[]{
    \begin{minipage}[b]{0.125\linewidth}
        \includegraphics[width=\linewidth]{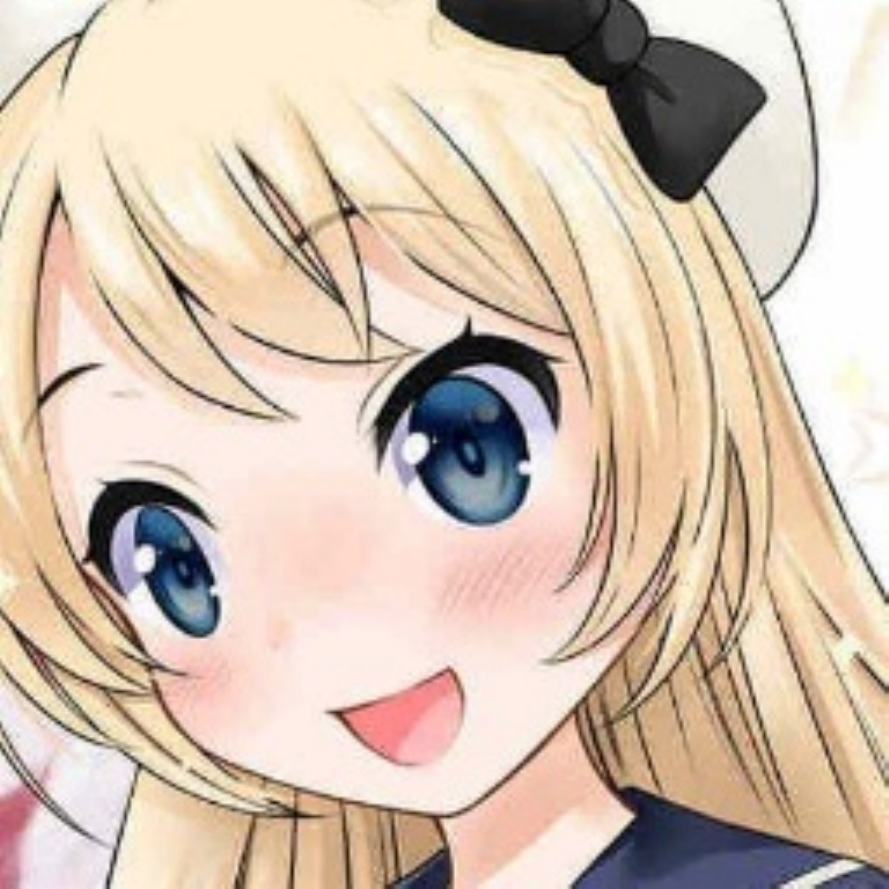}\vspace{4pt}
        \includegraphics[width=\linewidth]{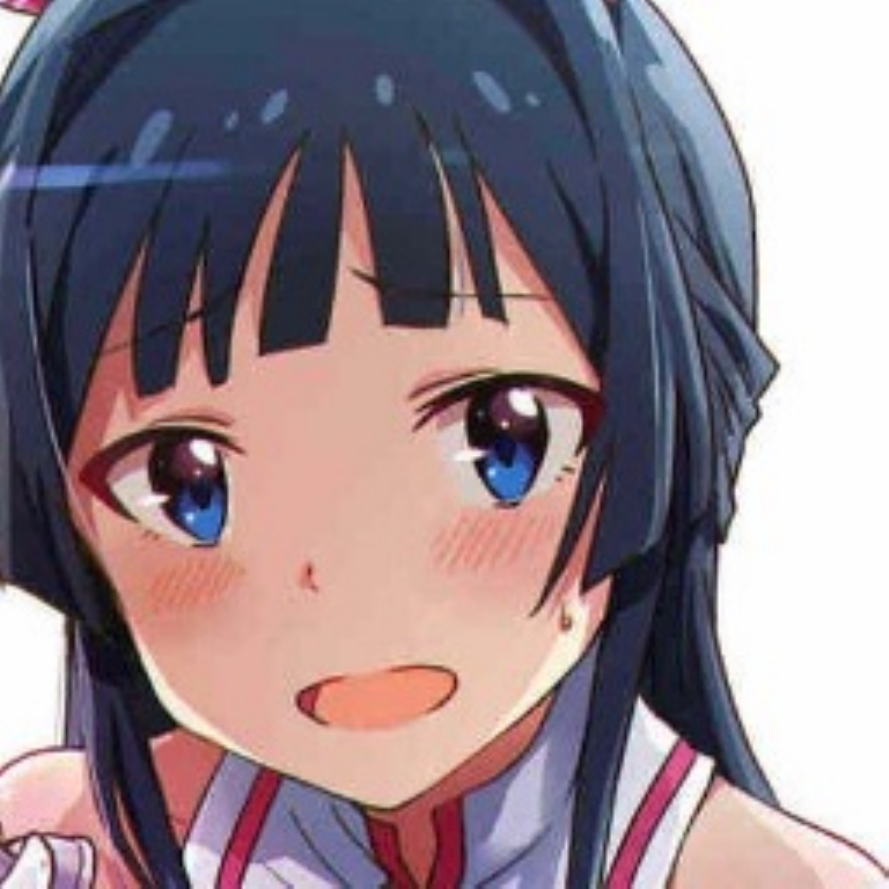}\vspace{4pt}
        \includegraphics[width=\linewidth]{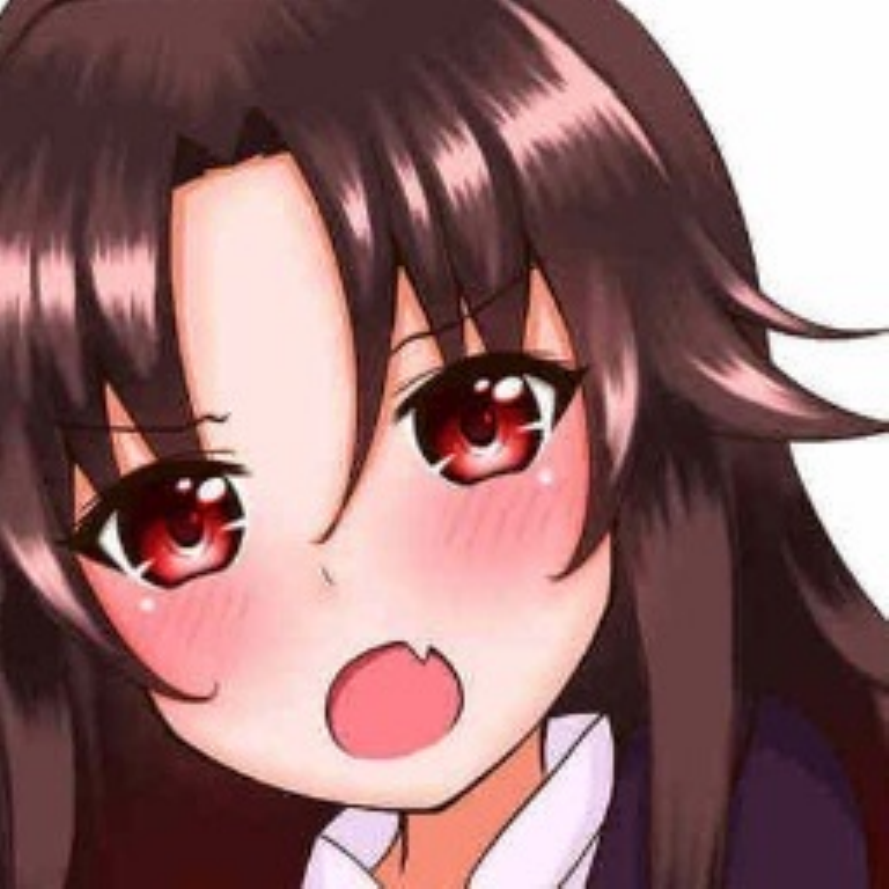}\vspace{4pt}
        \includegraphics[width=\linewidth]{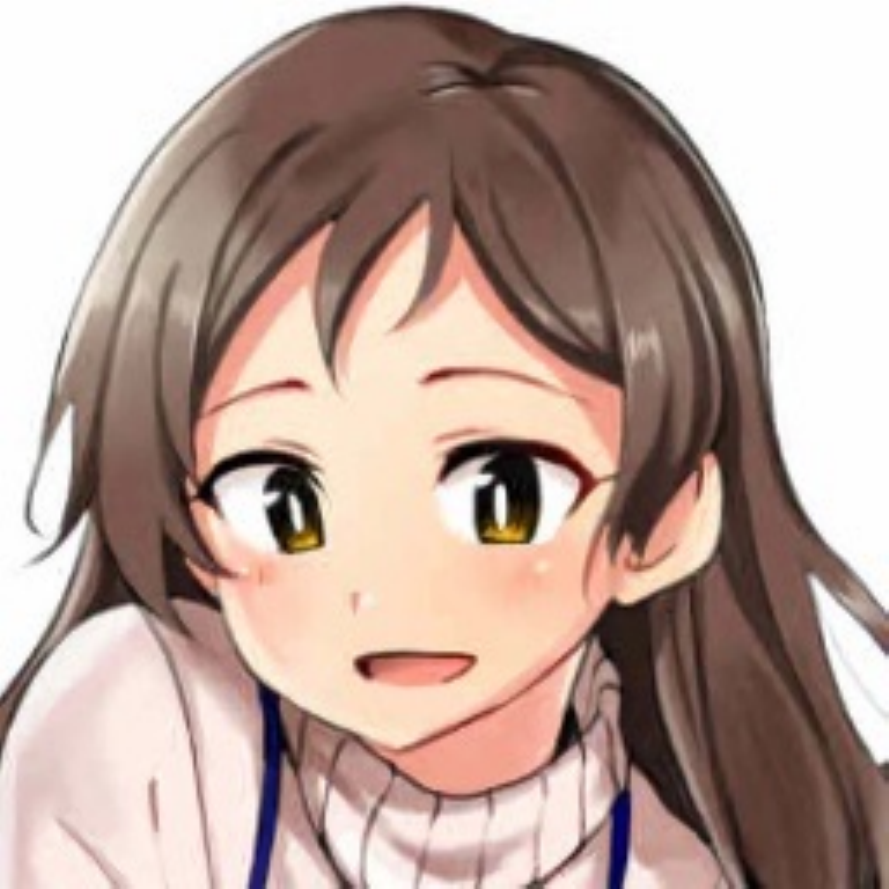}\vspace{4pt}
        \includegraphics[width=\linewidth]{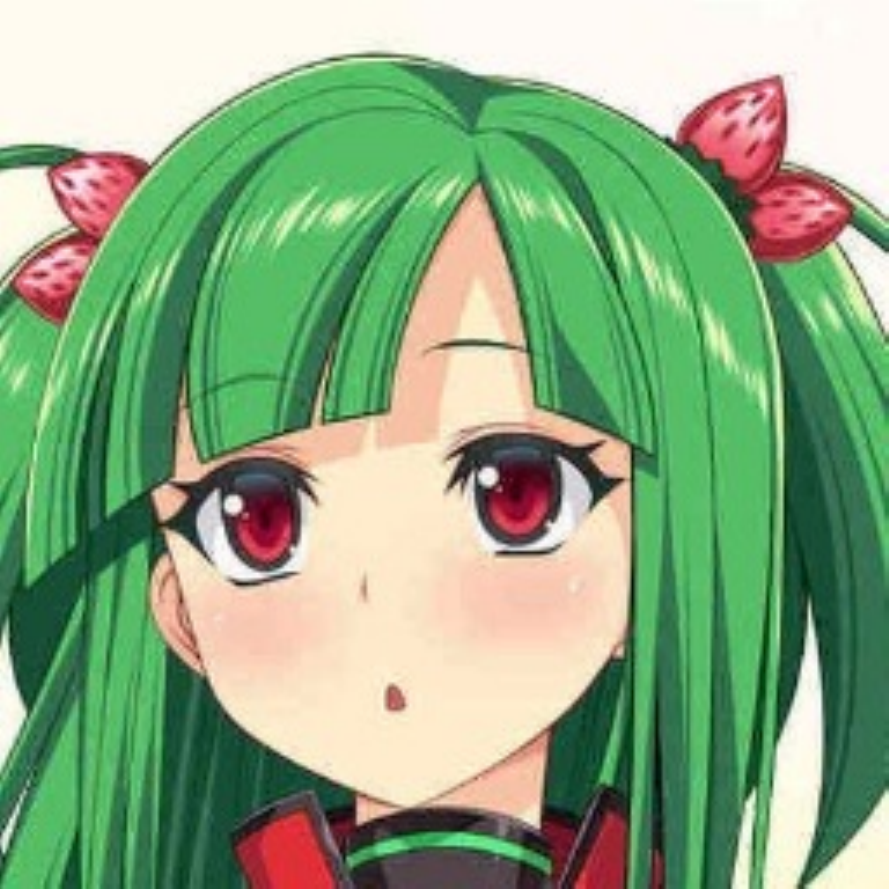}\vspace{4pt}
    \end{minipage}
    }
    \caption{Qualitative comparison for anime face with heterchromatic pupils. (a) reference images, (b) line drawings, (c) Lee et al.\cite{lee2020reference}, (d) Li et al.\cite{li2022eliminating}, (e) Cao et al.\cite{cao2022attention}, (f) AnimeDiffusion, and (g) original color images.}
    \label{fig:double_qualitative}
\end{figure*}

More comparison results are illustrated in Fig.~\ref{fig:single_qualitative}, given line drawings and reference images, AnimeDiffusion generates colored results with accurate color and good semantic correspondence. The image is clear without noise, and the color texture is smooth and soft. Especially in the eyes, the color is very precise, and the sense of light in the eyes is kept very good and full of charm. By contrast, Lee et al.~\cite{lee2020reference} generates results with color bleeding and wrong semantic correspondence, the color texture is rough and detail information is unclear. Li et al.~\cite{li2022eliminating} generates results with inaccurate color. Cao et al.~\cite{cao2022attention} generates results with sharpen image quality and wrong color in eyes. In general, the quality of images produced by GANs-based methods is not very stable and random flaw sometimes occurs. Our results are of high quality with beautiful color and rich details. Compared with the other three GANs-based methods, the image texture quality has been significantly improved.

For heterochromatic pupils case which is a very challenging task of anime face line drawing colorization. As is illustrated in Fig.~\ref{fig:double_qualitative}, AnimeDiffusion can generate results with accurate color in different pupils according to the reference image. Although Lee et al.~\cite{lee2020reference} produces some results with heterochromatic pupils, but the overall quality of results is not high, there are still color bleeding and inaccurate semantic correspondence. Li et al.~\cite{li2022eliminating} generates results with distorted color globally and inaccurate color in pupils.
Cao et al.~\cite{cao2022attention} fail to handle the heterochromatic pupils case, but the image sharpness and semantic information are still accurate.
\meng{As heterochromatic pupil is a fine-grained feature in the image space, and GAN is not accurate in data distribution modeling, the three SOTA GANs-based methods\daniel{\cite{lee2020reference, li2022eliminating, cao2022attention}} cannot handle it well. In contrast, our diffusion-based solution takes advantage of its precise data modeling property, combined with the use of multi-scale feature self-attention modules. Therefore, pupils can be colorized accurately according to the reference images without introducing additional processing modules.}

\subsection{Quantitative Evaluation}
\subsubsection{Evaluation Metrics}
We mainly use three evaluation metrics for quantitative comparison AnimeDiffusion with other methods. The popular Fréchet Inception Distance (FID) is used to assess the generation ability of algorithms in perceptual level. Besides measuring the perceptual credibility, we also adopt Peak Signal-to-Noise Ratio (PSNR) and Multi-Scale Structural Similarity Index
Measure (MS-SSIM) to evaluate the image reconstruction ability of algorithms in pixel level.
We design two kinds of colorization tasks respectively including self-reference reconstruction and random-reference colorization to analysize the colorization performance of AnimeDiffusion.  

\subsubsection{Self-reference Reconstruction}
For self-reference colorization, the line drawing and reference image are paired, ideally the colorized output should be exactly the same as the reference image. We directly use our paired testing data for conducting self-reference colorization. In fact, during the training phase, the main task of AnimeDiffusion is to do the image self-reconstruction, through this proxy task, the network can learn the colorization ability. For fairness, we train AnimeDiffusion and other three GANs-based methods sufficiently to compute PSNR and MS-SSIM. As is shown in Table~\ref{tab:quantitative_compar}, AnimeDiffusion acquires the best image reconstruction performance.

\subsubsection{Random-reference Colorization}
For the random-reference colorization, it is more like the common practical usage when using reference-based colorization method. We shuffle all the reference images in our testing data, then use unpaired line drawings and reference images to perform random-reference colorization. Using the total 579 generated images and 579 reference images to compute FID score. A smaller FID indicates that the distribution of the colored images is closer to the reference images and indicates that the model with wonderful generation ability. As is shown in Table~\ref{tab:quantitative_compar}, AnimeDiffusion shows better generation ability than other three GANs-based methods. One thing needs to note is that although Cao et al.~\cite{cao2022attention} shows little poor image reconstruction performance than Lee et al.~\cite{lee2020reference}, but shows better generation ability than Lee et al.\cite{lee2020reference} and Li et al.~\cite{li2022eliminating}, that is because Lee et al.~\cite{lee2020reference} just learns a trivial solution. This point is also discovered and discussed in Li et al.~\cite{li2022eliminating}.

\begin{table}[ht]
    \centering
    \caption{Quantitative Comparison between AnimeDiffusion and Other Three SOTA GANs-based Methods}
    \label{tab:quantitative_compar}
    \begin{tabular}{lccc}
    \toprule
        Method & PSNR$\uparrow$  &  MS-SSIM$\uparrow$ & FID$\downarrow$  \\
    \midrule
        Lee et al.\cite{lee2020reference} & 23.8901 & 0.9224 & 57.19\\
        Li et al.\cite{li2022eliminating} & 18.6347 & 0.8209 & 49.33 \\
        Cao et al.\cite{cao2022attention}   & 19.7746 & 0.8388 & 46.39 \\
        AnimeDiffusion     & \textbf{25.4658} & \textbf{0.9596} & \textbf{44.19} \\
    \bottomrule
    \end{tabular}
\end{table}

\subsection{Ablation Study}
We perform extensive ablation experiments to verify the effectiveness of our designed fine-tuning strategy when training AnimeDiffusion. 
We find that the denoising model obtained by classifier-free guidance pre-training can generate images with high diversity, but this diversity also means that the colorization results are unstable since randomness is introduced by Gaussian noise. 

We believe that the pre-trained model already acquires the ability to capture the line structure and can inject different colors in the corresponding area according to the reference image. 
Fine-tuning is only to eliminate color gaps due to insufficient pre-training.
To validate our idea, we perform image reconstruction test and reference-based line drawing colorization test respectively. We fine-tune AnimeDiffusion with 1 epoch and 10 epochs with a batch size of 4 for comparison.

\begin{figure}[tb]
    \centering
    \subfigure[]{
    \begin{minipage}[b]{0.23\linewidth}
        \includegraphics[width=\linewidth]{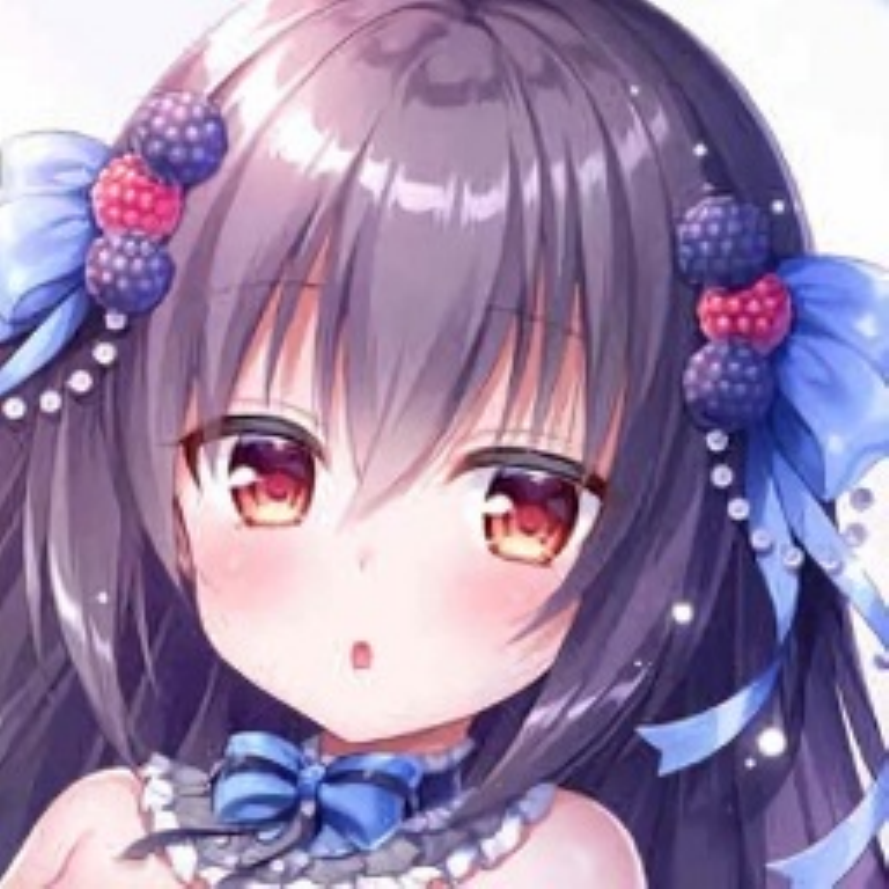}\vspace{4pt}
        \includegraphics[width=\linewidth]{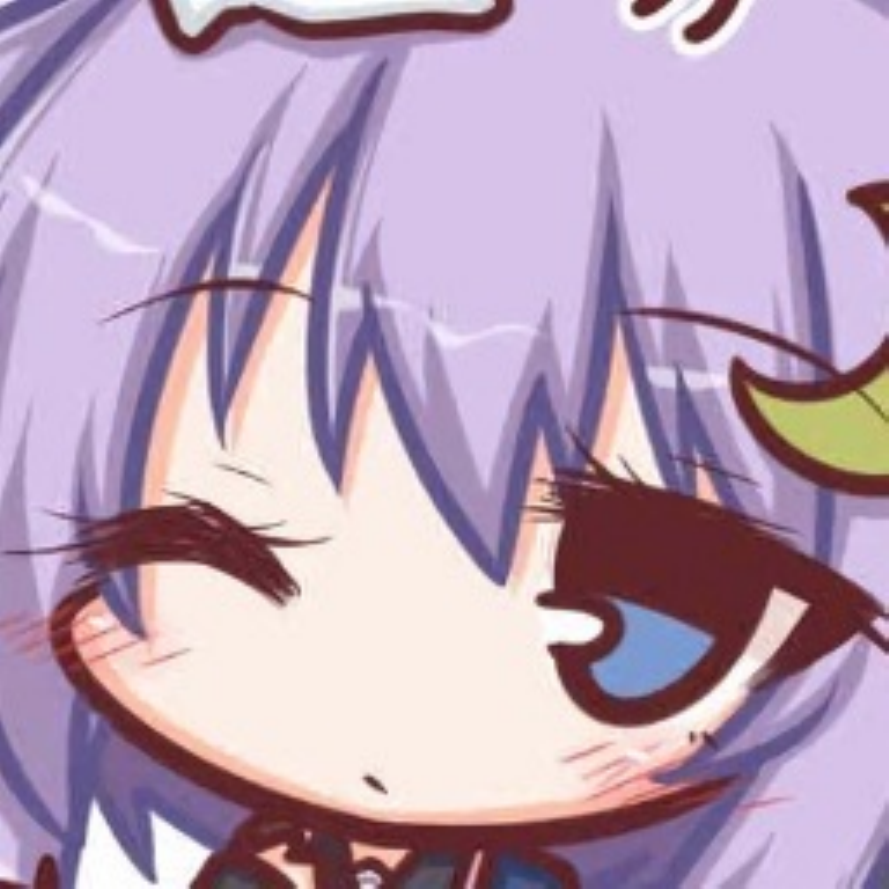}\vspace{4pt}
        \includegraphics[width=\linewidth]{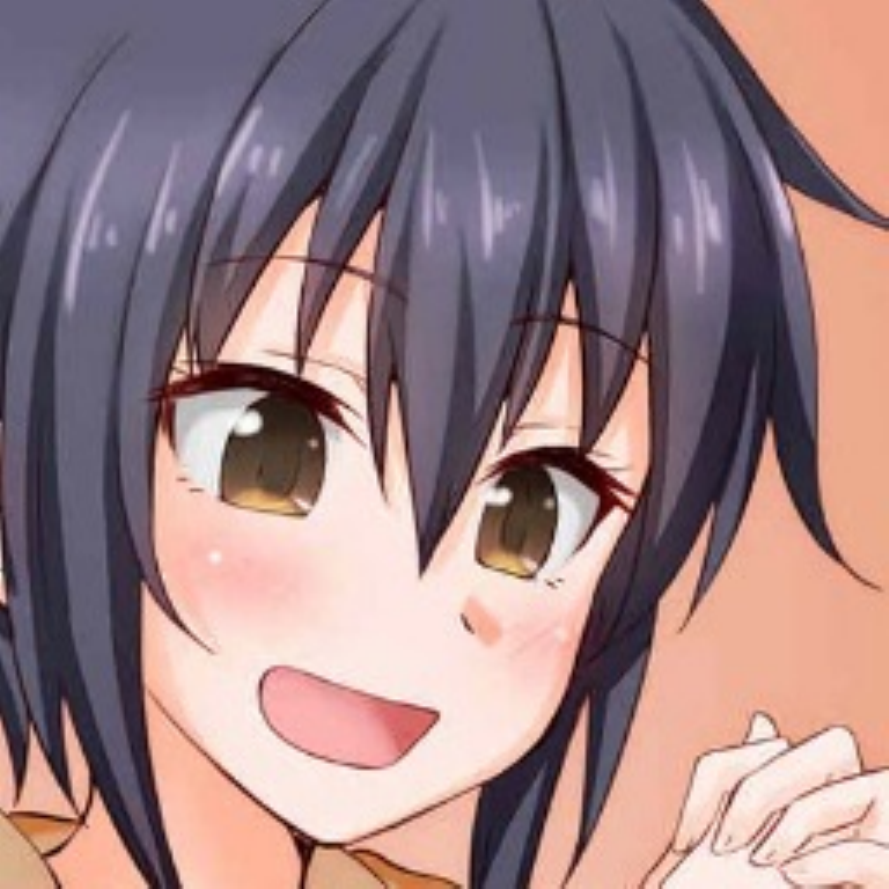}
    \end{minipage}
    }\hspace{-3pt}
    \subfigure[]{
    \begin{minipage}[b]{0.23\linewidth}
        \includegraphics[width=\linewidth]{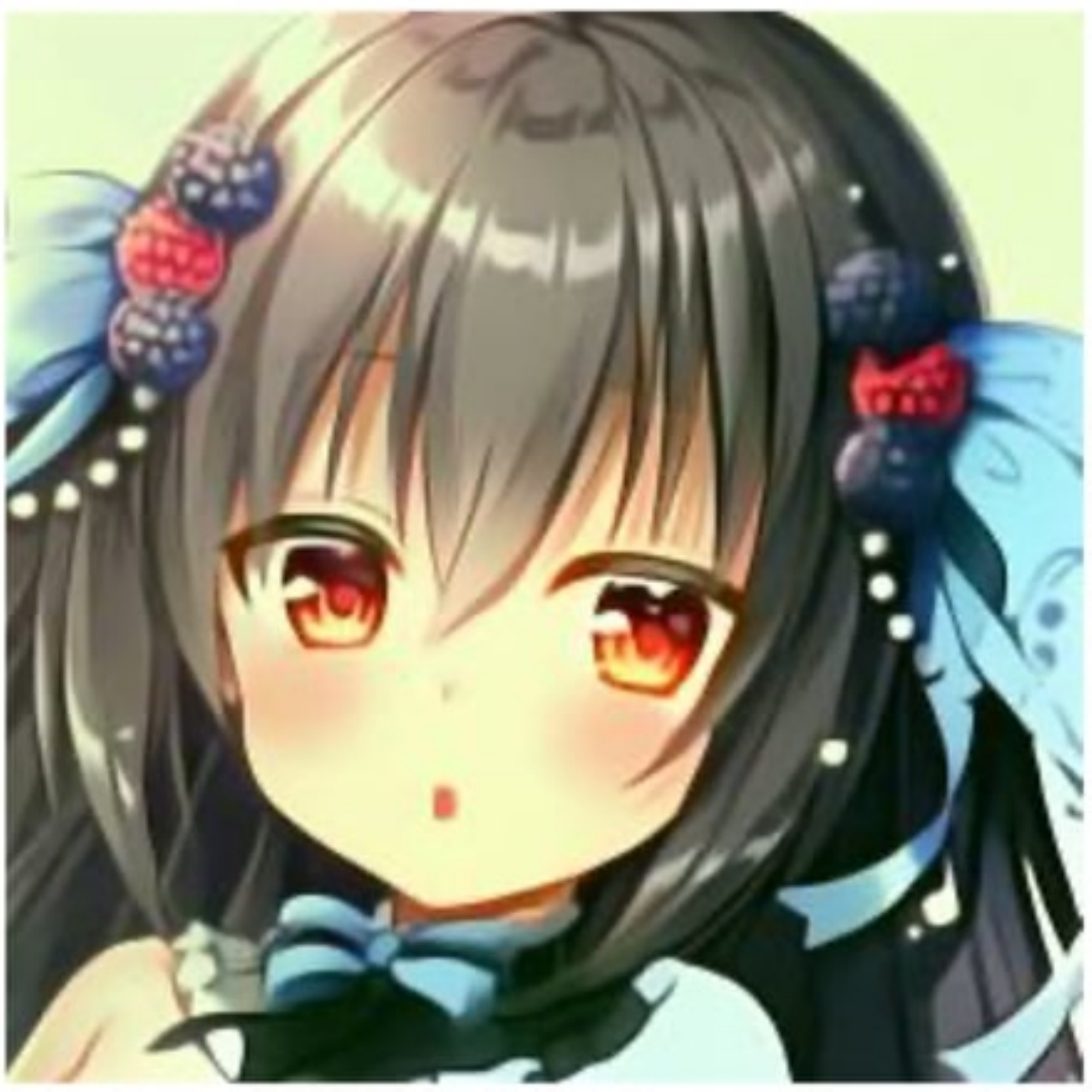}\vspace{4pt}
        \includegraphics[width=\linewidth]{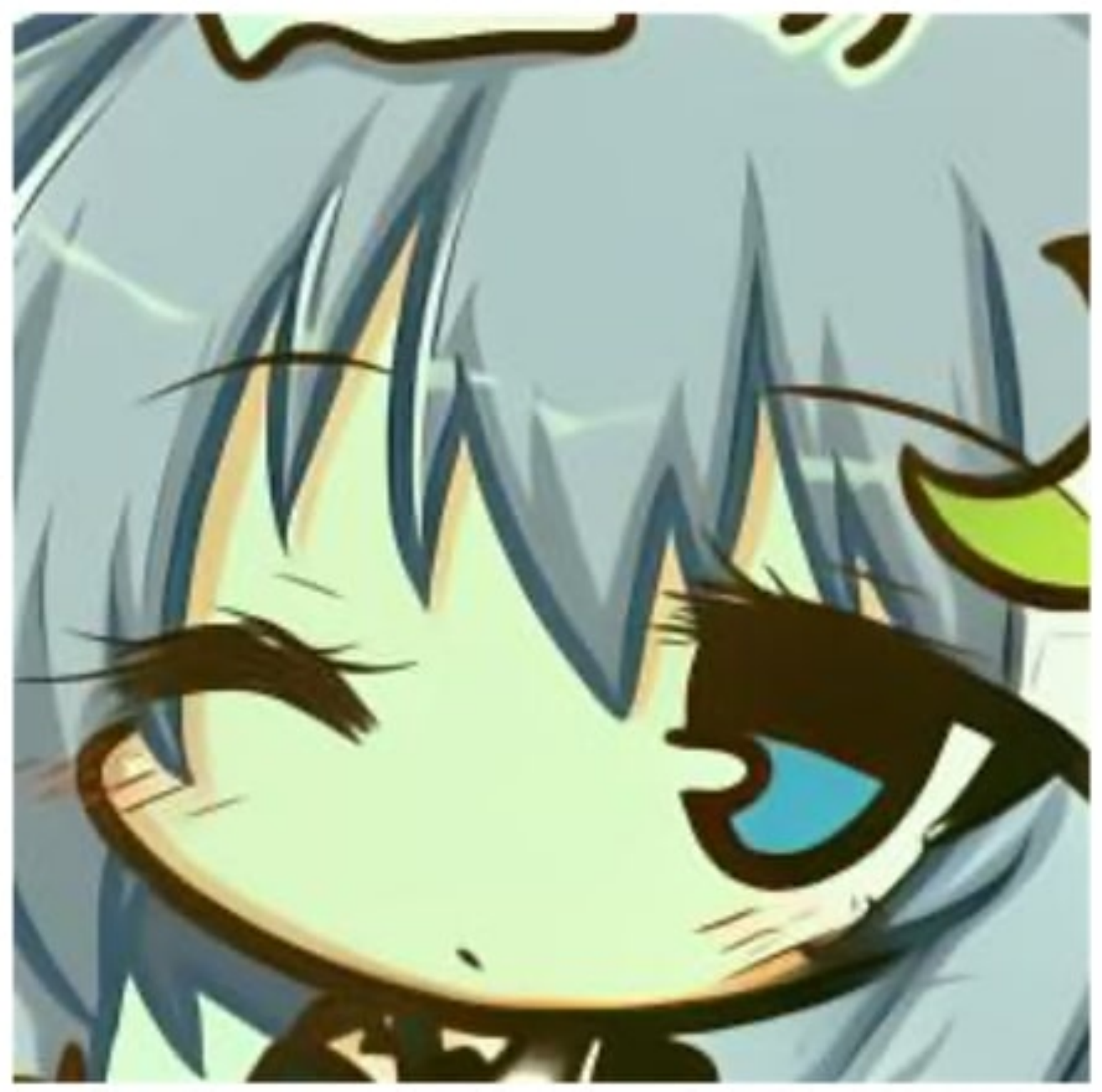}\vspace{4pt}
        \includegraphics[width=\linewidth]{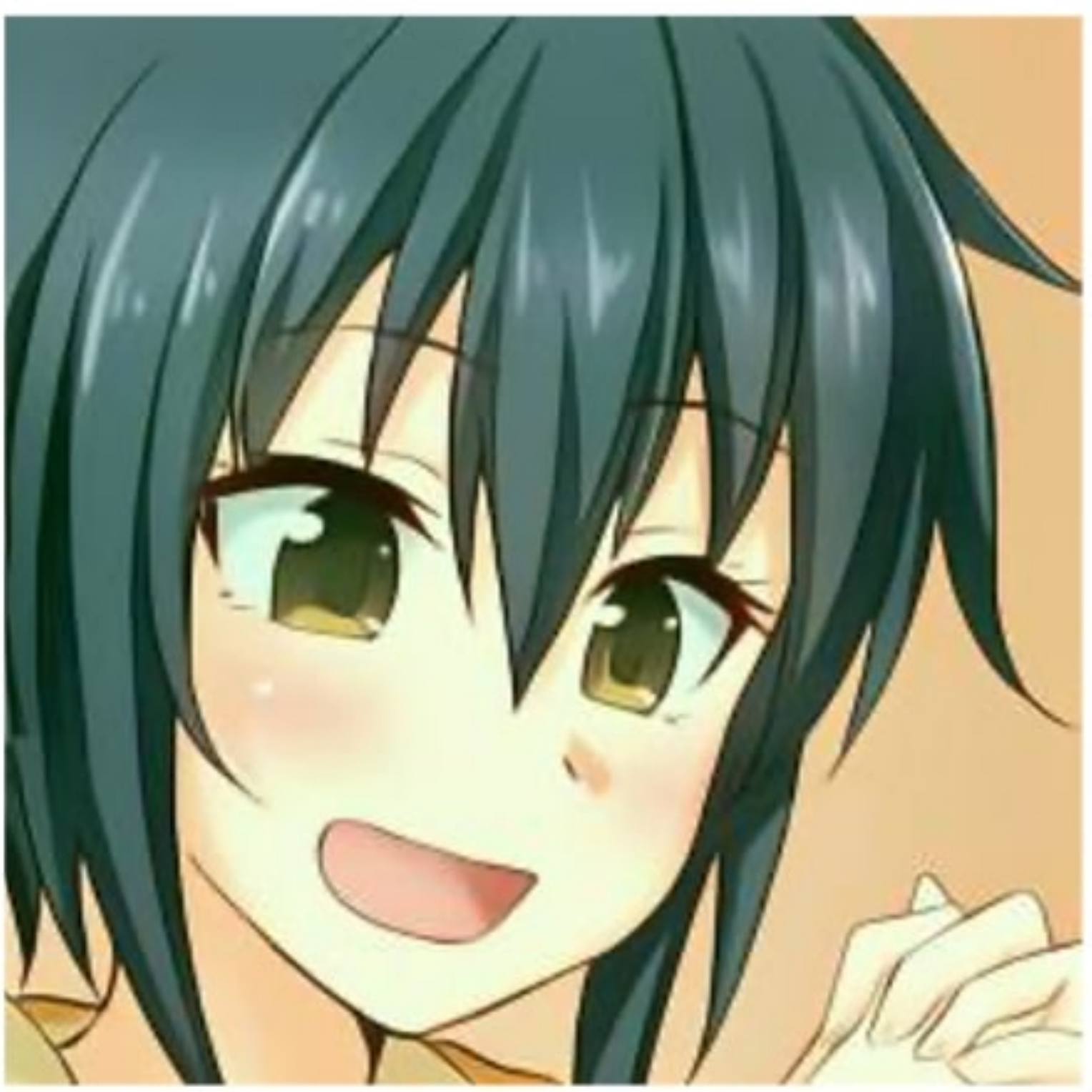}
    \end{minipage}
    }\hspace{-3pt}
    \subfigure[]{
    \begin{minipage}[b]{0.23\linewidth}
        \includegraphics[width=\linewidth]{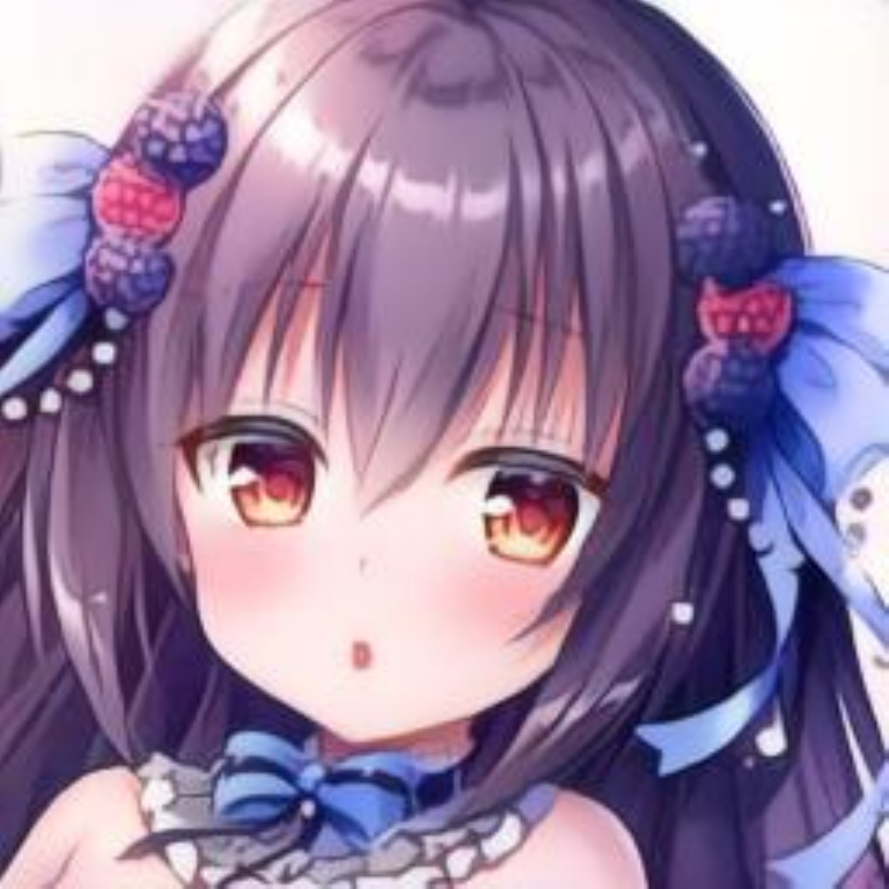}\vspace{4pt}
        \includegraphics[width=\linewidth]{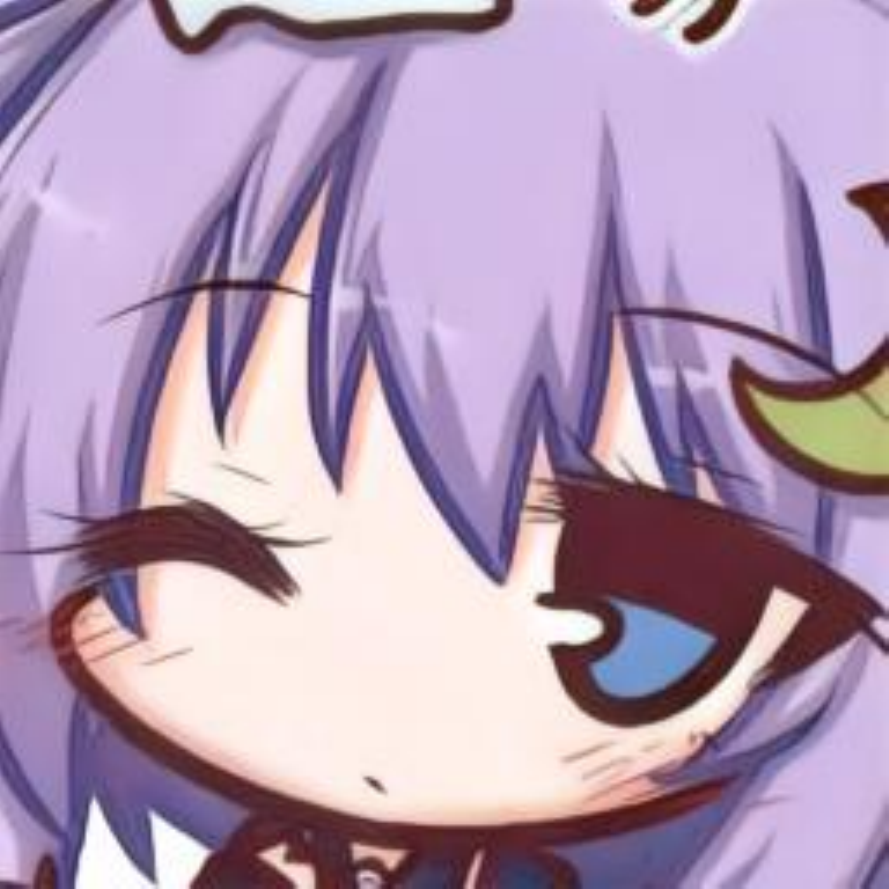}\vspace{4pt}
        \includegraphics[width=\linewidth]{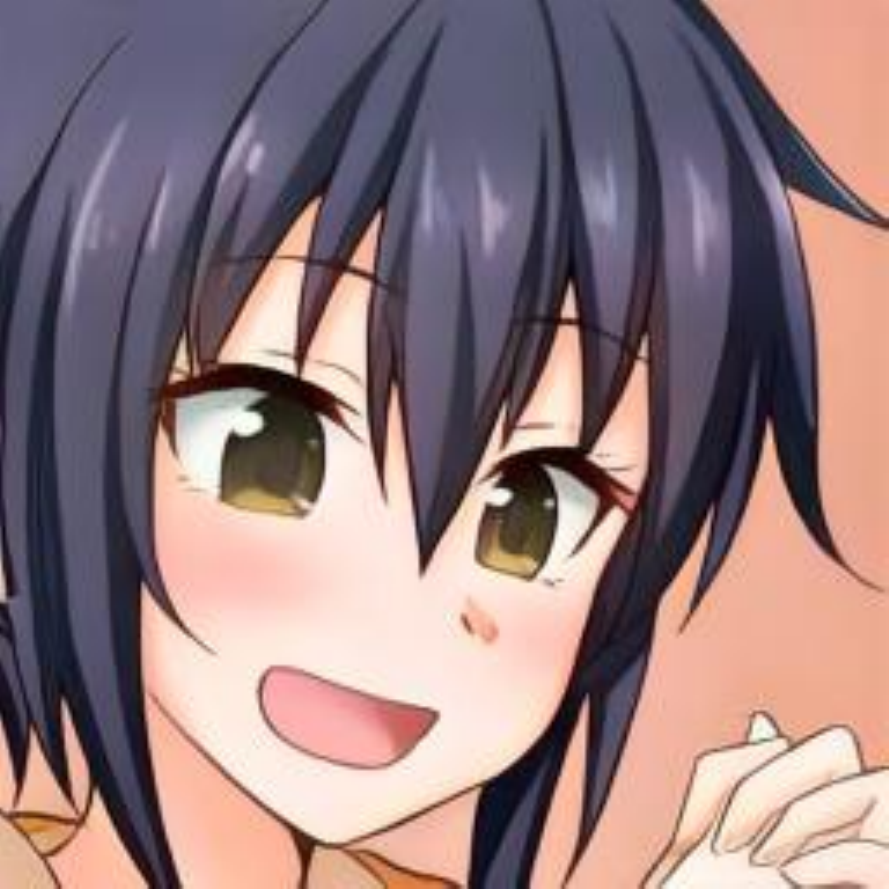}
    \end{minipage}
    }\hspace{-3pt}
    \subfigure[]{
    \begin{minipage}[b]{0.23\linewidth}
        \includegraphics[width=\linewidth]{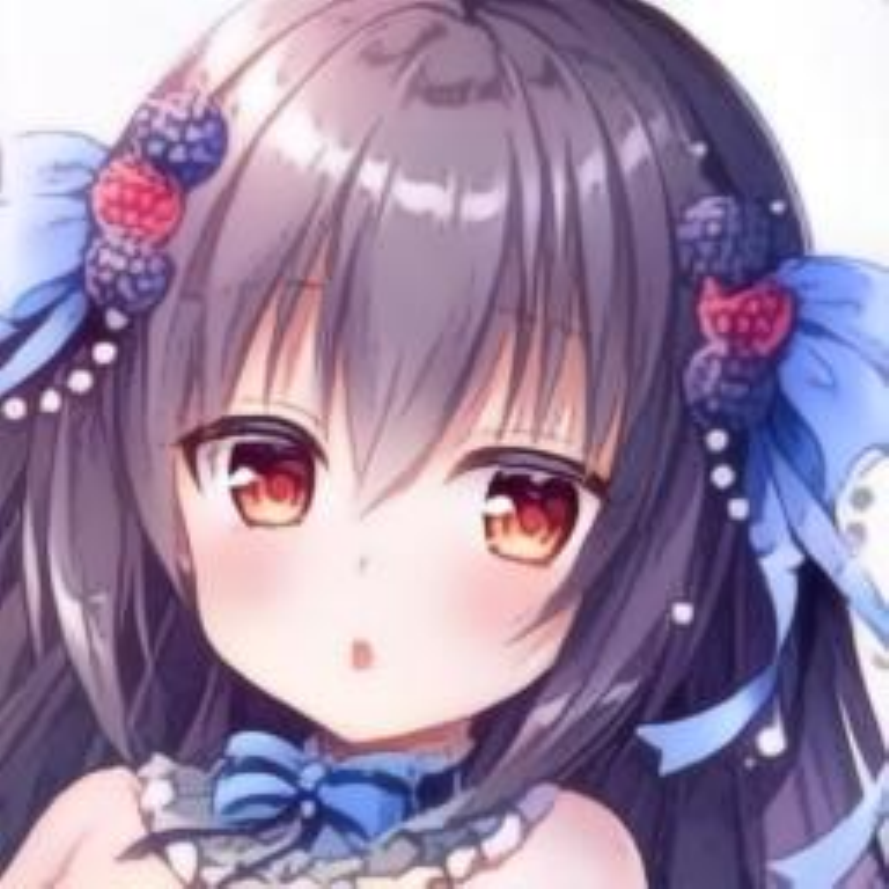}\vspace{4pt}
        \includegraphics[width=\linewidth]{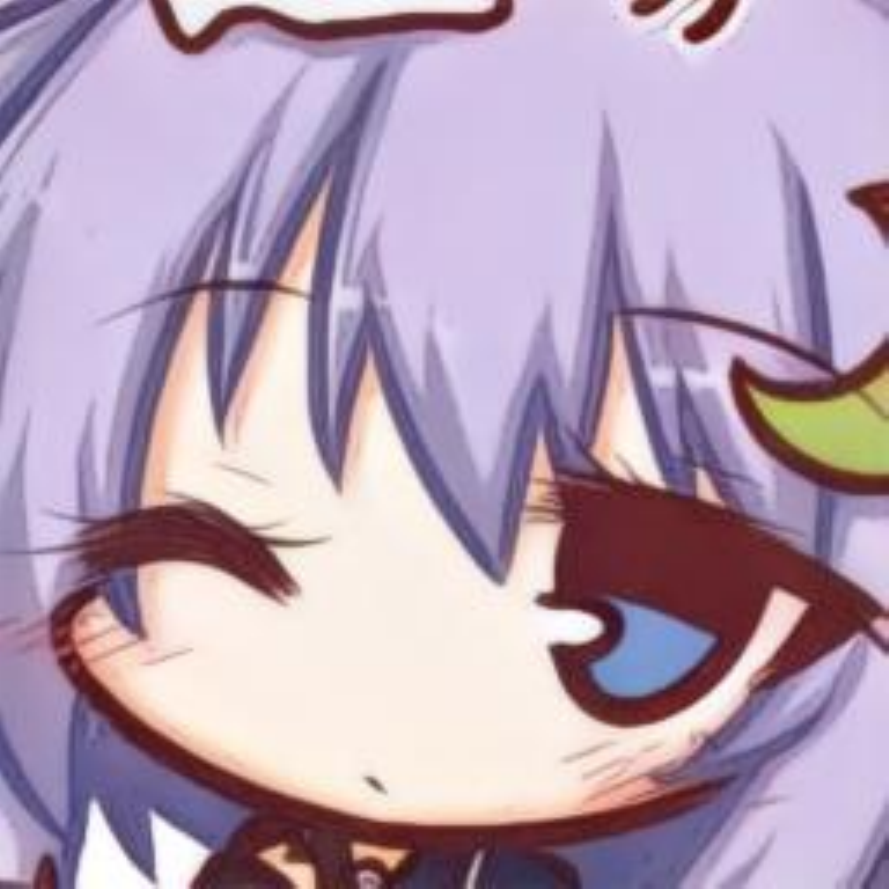}\vspace{4pt}
        \includegraphics[width=\linewidth]{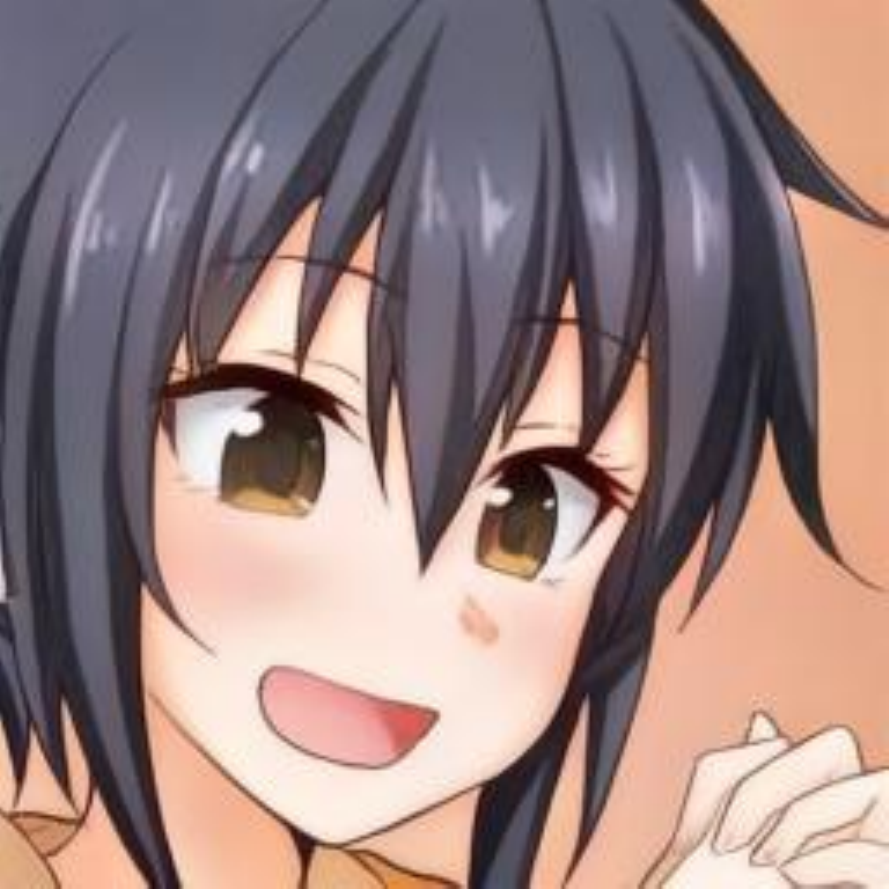}
    \end{minipage}
    }
    \caption{Image reconstruction test for ablation study. (a) Ground truth images, (b) Results without fine-tuning, (c) Results with fine-tuning for 1 epoch, (d) Results with fine-tuning for 10 epochs.}
    \label{fig:ablation_reconstruction}
\end{figure}
\begin{figure}[htb]
    \centering
    \subfigure[]{
    \begin{minipage}[b]{0.23\linewidth}
        \includegraphics[width=\linewidth]{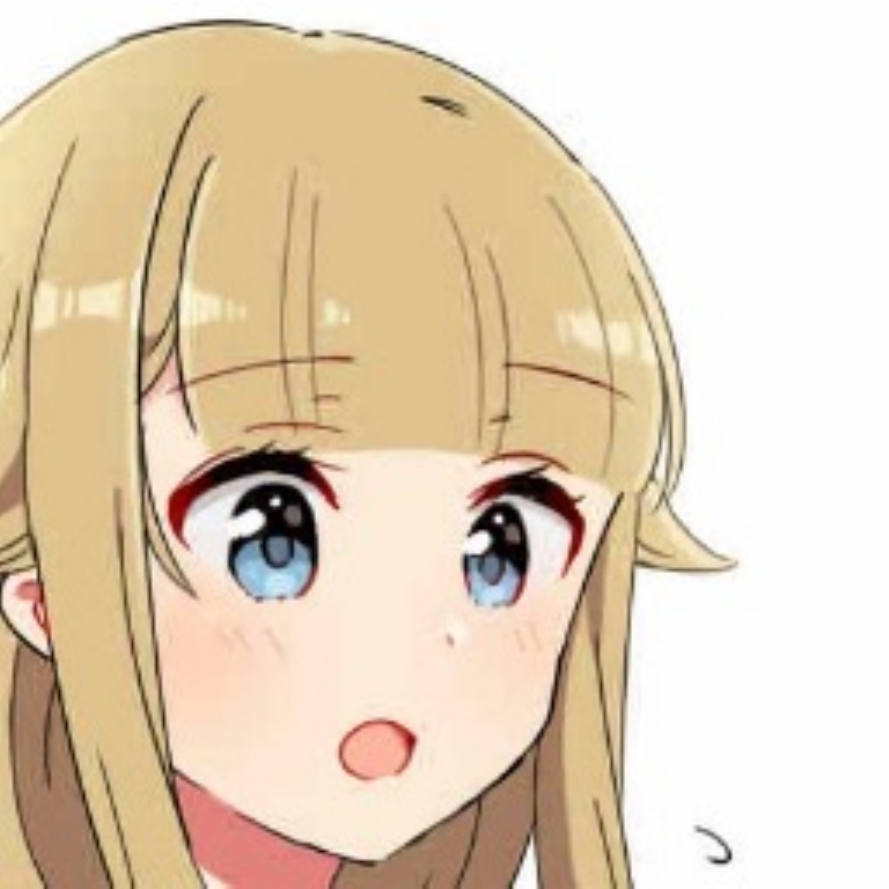}\vspace{4pt}
        \includegraphics[width=\linewidth]{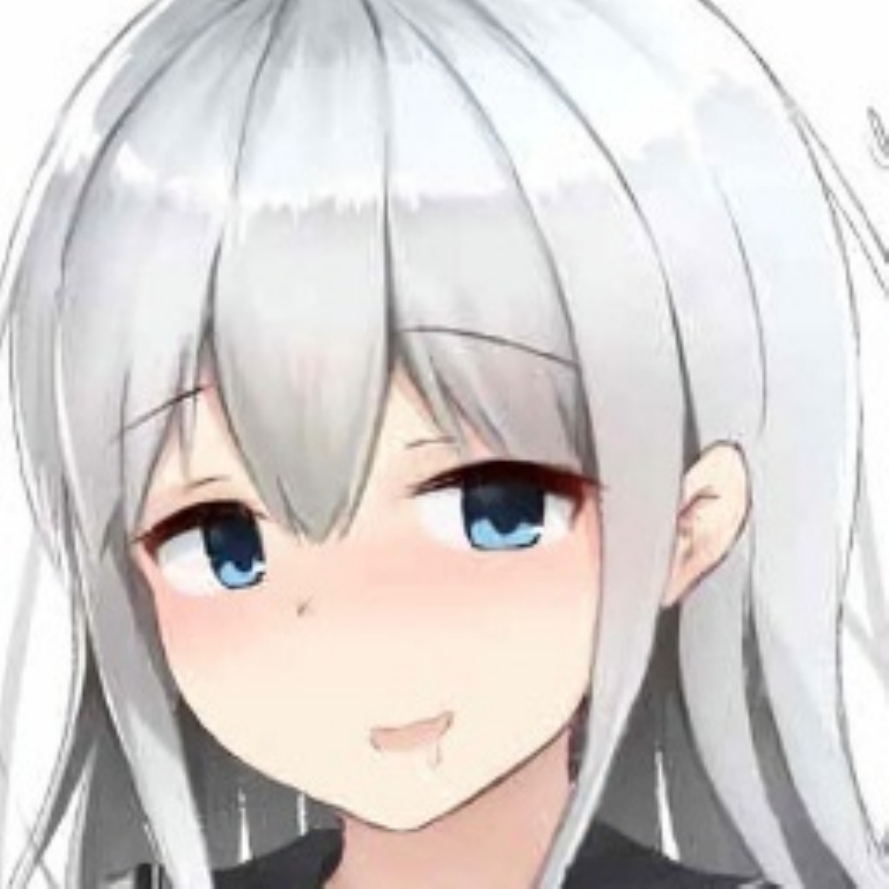}\vspace{4pt}
        \includegraphics[width=\linewidth]{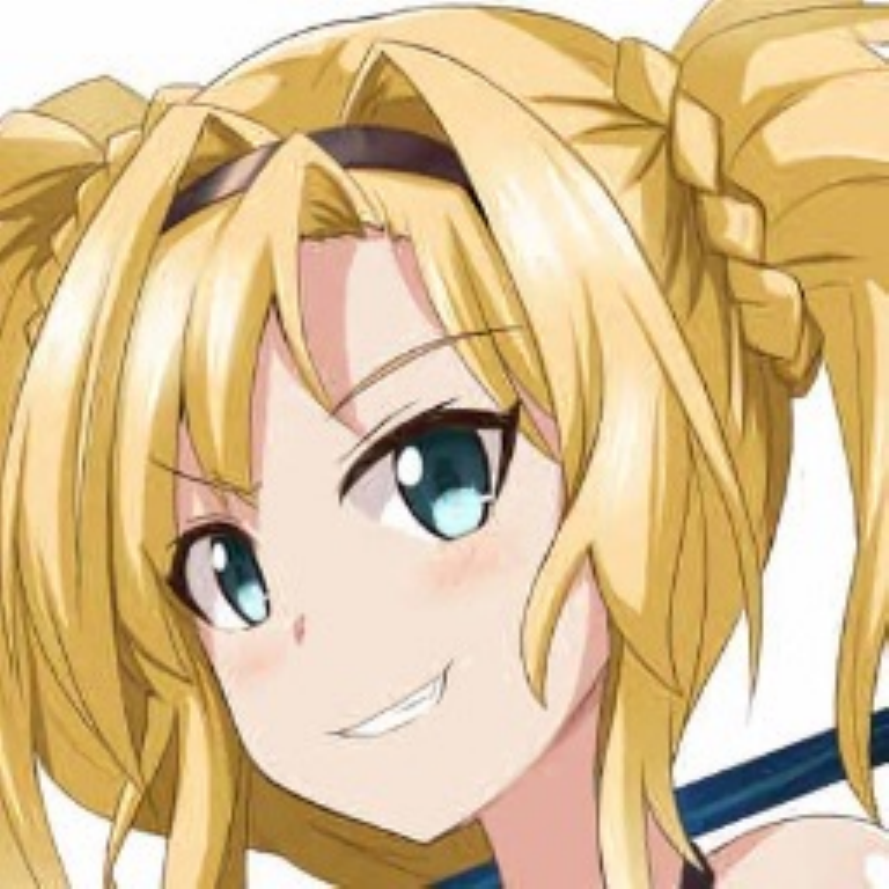}
    \end{minipage}
    }\hspace{-4pt}
    \subfigure[]{
    \begin{minipage}[b]{0.23\linewidth}
        \includegraphics[width=\linewidth]{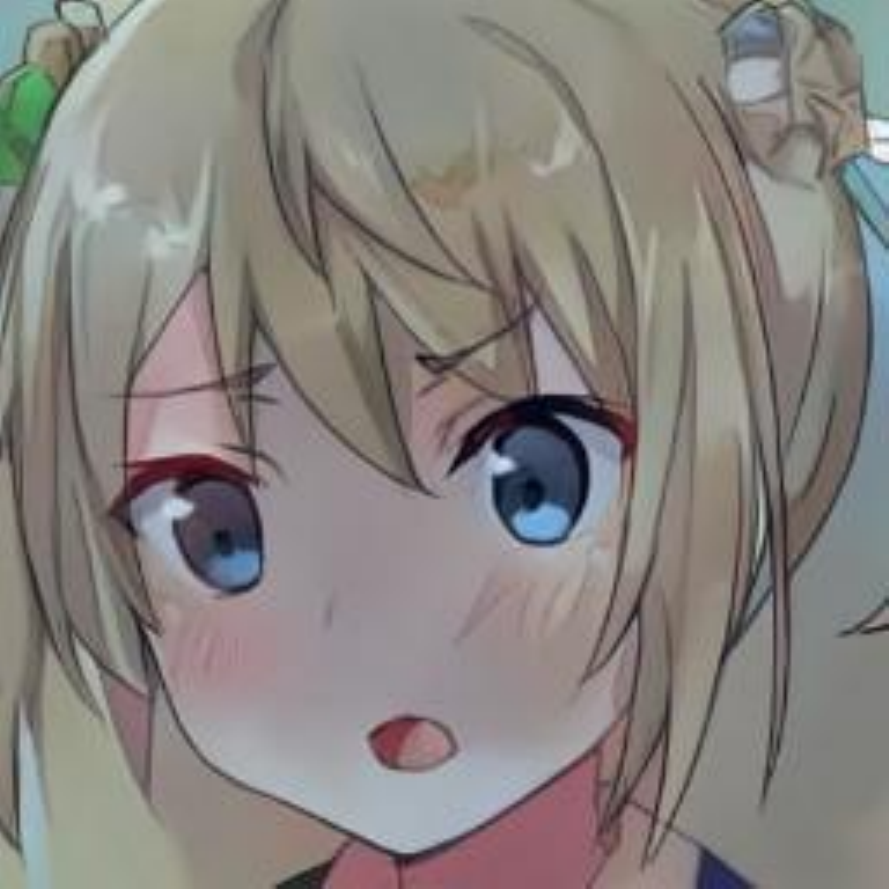}\vspace{4pt}
        \includegraphics[width=\linewidth]{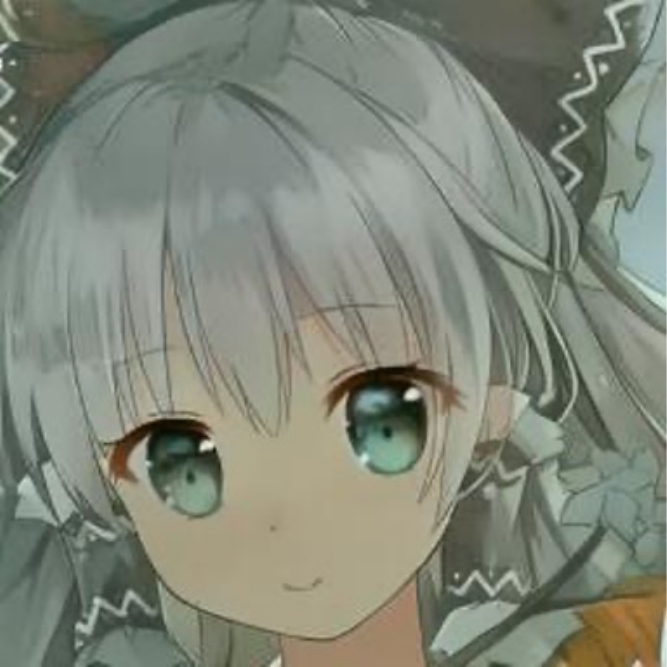}\vspace{4pt}
        \includegraphics[width=\linewidth]{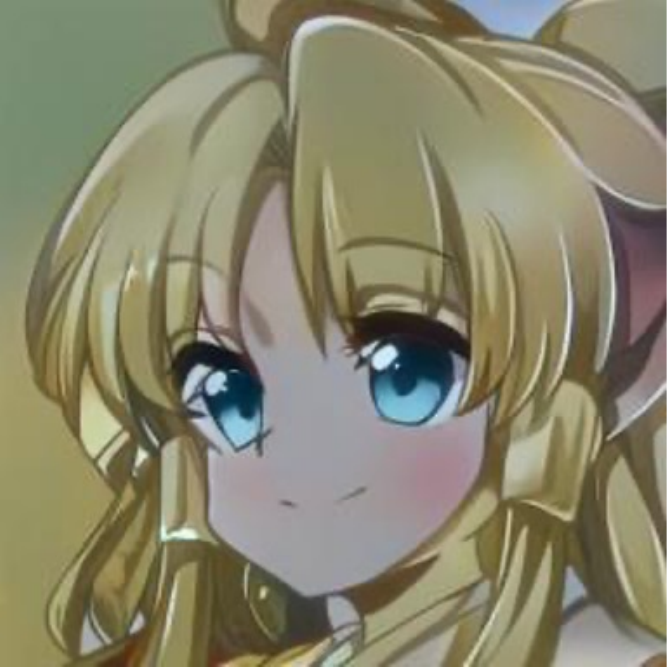}
    \end{minipage}
    }\hspace{-4pt}
    \subfigure[]{
    \begin{minipage}[b]{0.23\linewidth}
        \includegraphics[width=\linewidth]{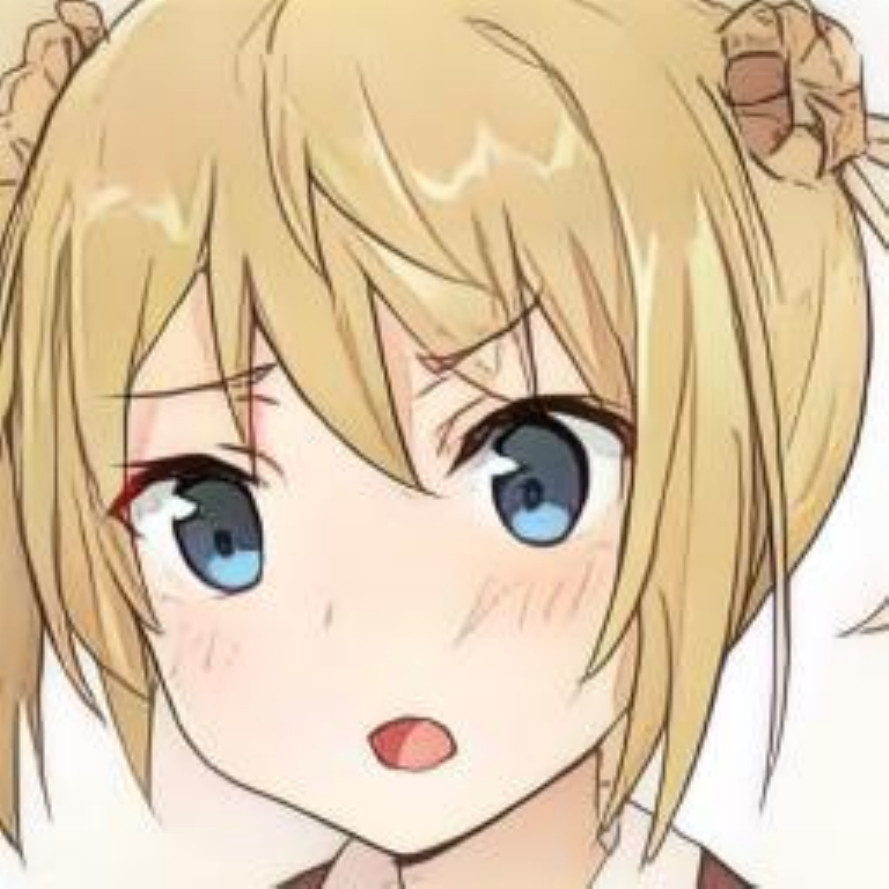}\vspace{4pt}
        \includegraphics[width=\linewidth]{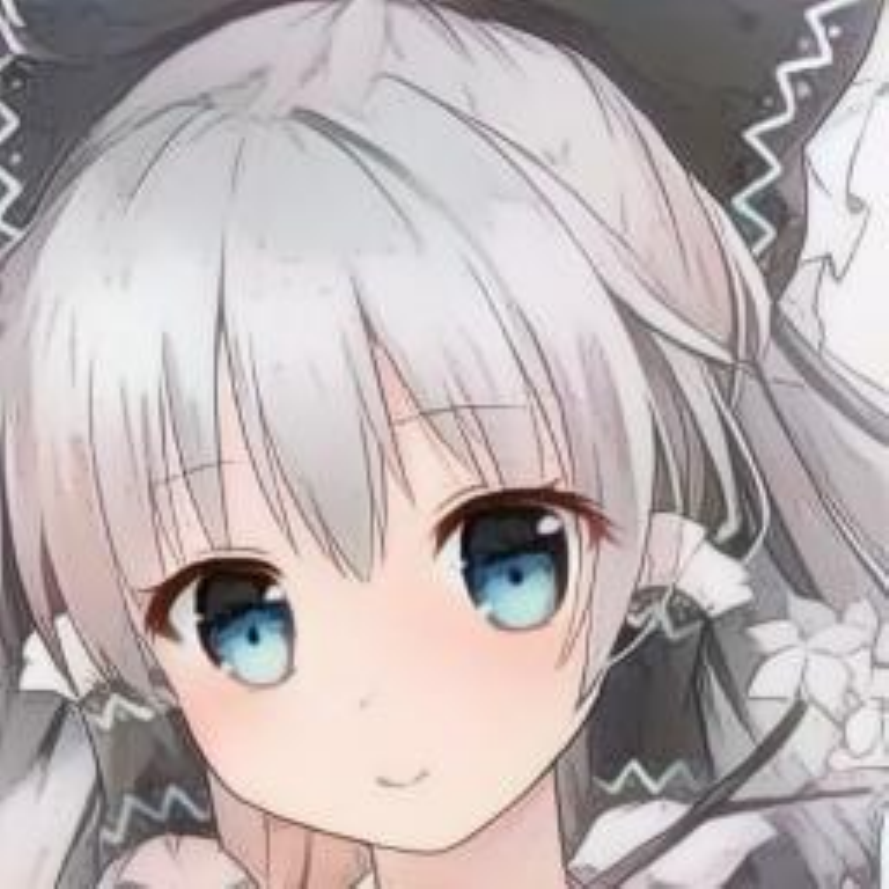}\vspace{4pt}
        \includegraphics[width=\linewidth]{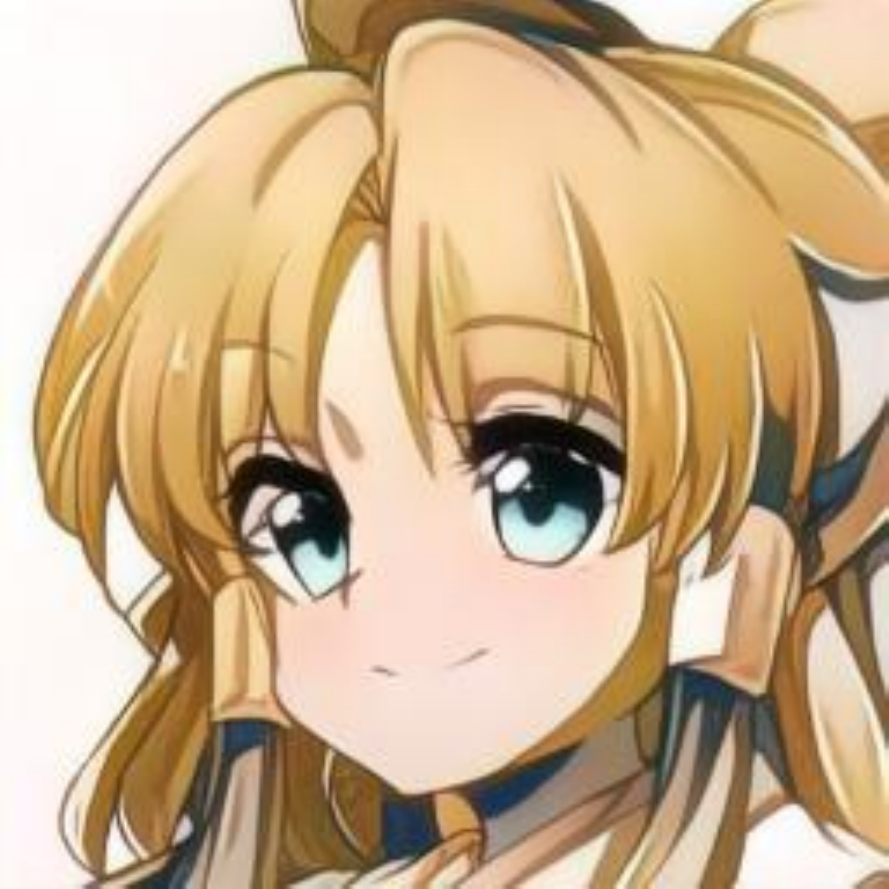}
    \end{minipage}
    }\hspace{-4pt}
    \subfigure[]{
    \begin{minipage}[b]{0.23\linewidth}
        \includegraphics[width=\linewidth]{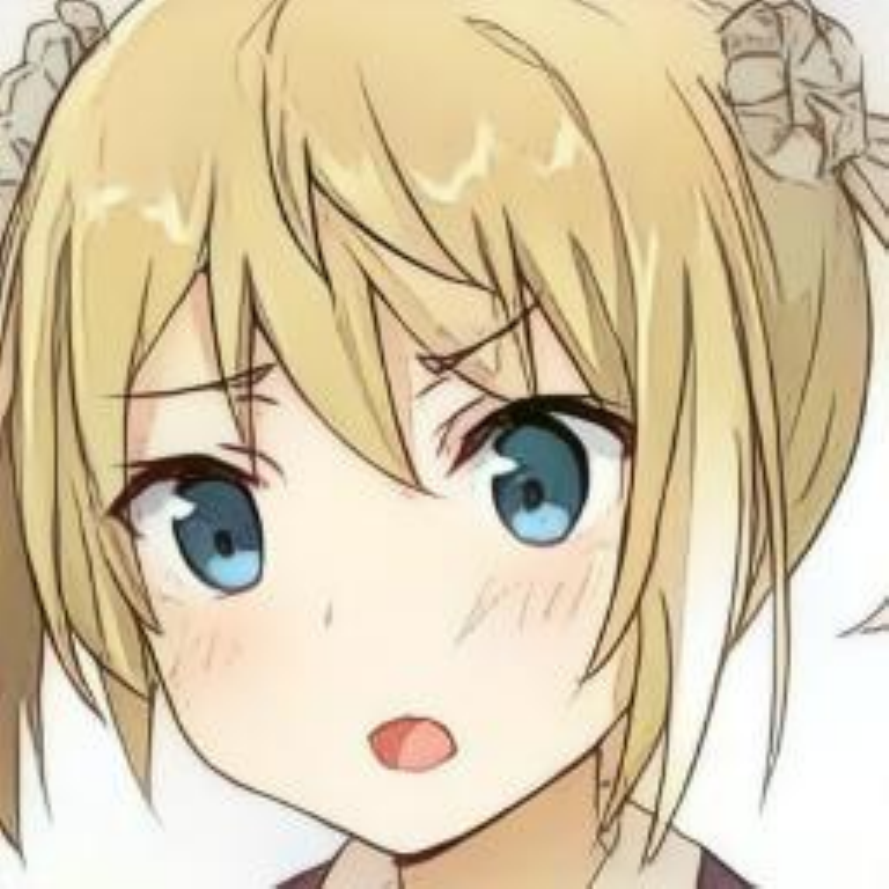}\vspace{4pt}
        \includegraphics[width=\linewidth]{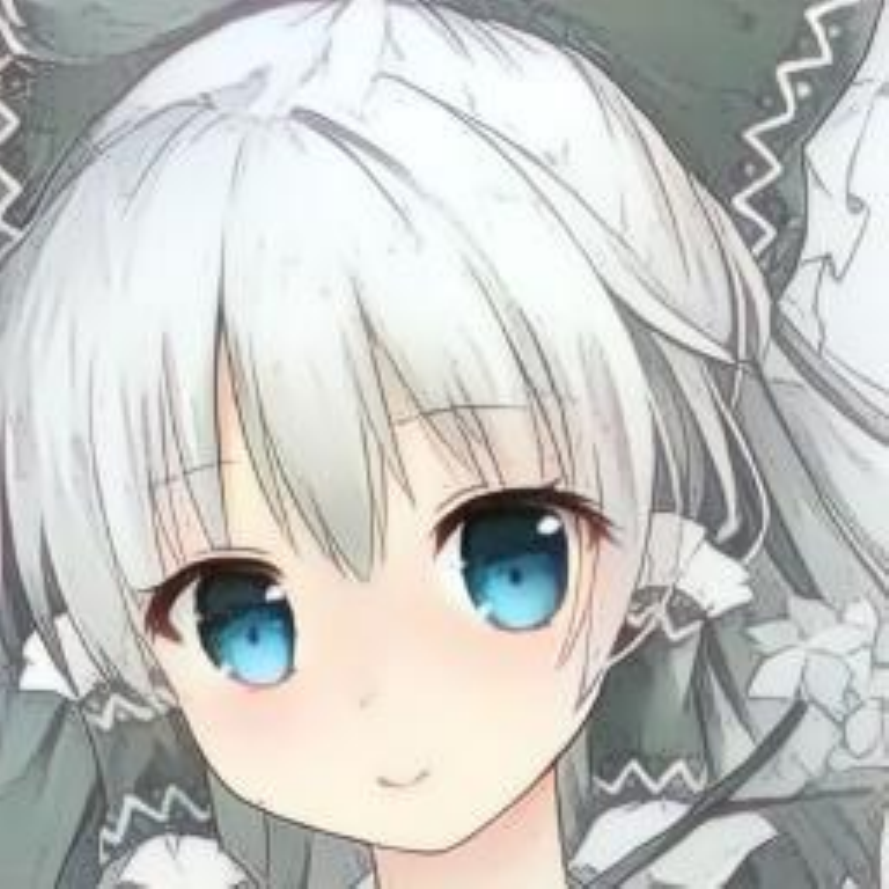}\vspace{4pt}
        \includegraphics[width=\linewidth]{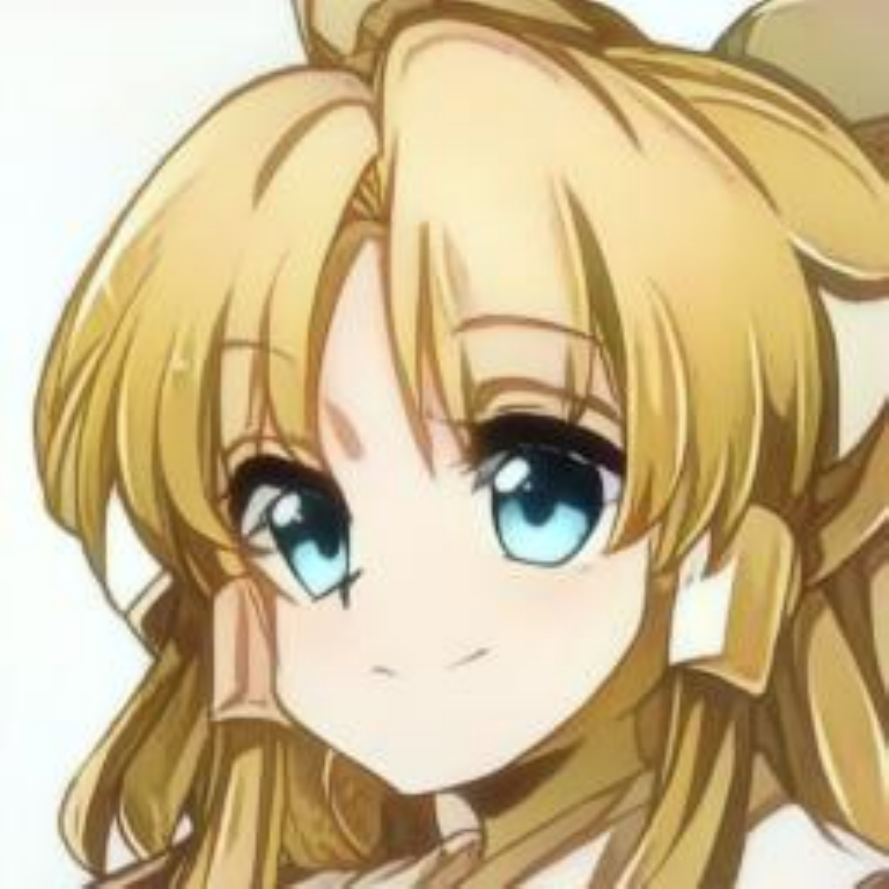}
    \end{minipage}
    }
    \caption{Reference-based line drawing colorization test for ablation study. (a) Reference images, (b) Results without fine-tuning, (c) Results with fine-tuning for 1 epoch, (d) Results with fine-tuning for 10 epochs.}
    \label{fig:ablation_colorization}
\end{figure}

We perform the image reconstruction test by distorting the ground truth image that serves as the reference image. Comparison results are shown in Fig.~\ref{fig:ablation_reconstruction}. The results without fine-tuning can reconstruct the overall structure of original image, but the color difference is obvious. After adding the image reconstruction loss to the fine tuning, the effect is significantly improved. The results of different fine-tuning schemes are not obvious in terms of visual differences

For reference-based line drawing colorization test, we show results in Fig.~\ref{fig:ablation_colorization}. Although the model without fine-tuning can distinguish regions that need different colors, there is still a color gap between the generated images and the reference images, and the result colored image looks dimmer. We think the model needs more training to generate results with accurate color information. However, training the diffusion model with classifier-free guidance is time-consuming, so we briefly fine-tune the model and get much better results.

We also compute PSNR, MS-SSIM and FID quantitative index for quantitative comparison. 
Results are shown in Table~\ref{tab:quantitative_ft}. After 10 epochs of fine-tuning, AnimeDiffusion continues to gain in FID score but little improvement in PSNR and MS-SSIM score. In fact, fine-tuning the model for 1 epoch is enough to have good colorization performance.
This also validates that our fine-tuning is mainly to fix color bias based on the pre-trained model with fundamental generation ability. Our designed hybrid training strategy can make model learn better colorization ability and save the training time cost.
\begin{table}[ht]
    \centering
    \caption{Quantitative Evaluation for Ablation Study Results}
    \label{tab:quantitative_ft}
    \begin{tabular}{lccc}
    \toprule
        AnimeDiffusion  & PSNR$\uparrow$ & MS-SSIM$\uparrow$ & FID$\downarrow$ \\
    \midrule
        Without Fine-tuning     & 12.4234 & 0.8079 & 55.1841 \\
        Fine-tuning (1 epoch)   & 25.4658 & 0.9596 & 44.1876 \\
        Fine-tuning (10 epochs) & \textbf{25.8992} & \textbf{0.9600} & \textbf{40.4392} \\
    \bottomrule
    \end{tabular}
\end{table}
\subsection{User Study}
\begin{figure}[ht]
\centering  
\includegraphics[width=\linewidth]{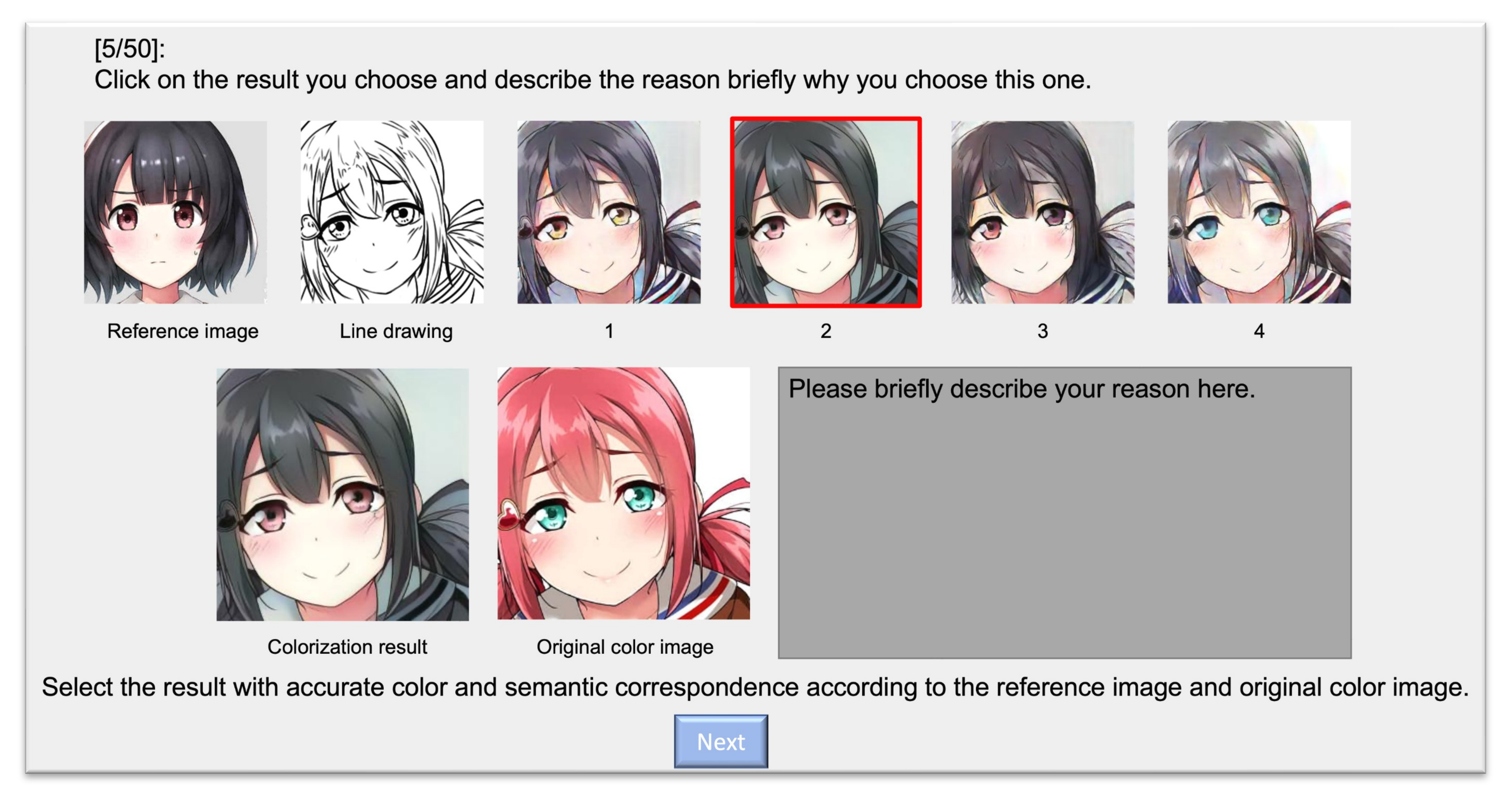}
\caption{A user interface of our user study.}
\label{fig:user_study}
\end{figure}

It is generally challenging to evaluate the visual quality of images, in particular for line drawing colorization. We randomly select 50 line drawings and 50 reference images to perform reference-based line drawing colorization using AnimeDiffusion compared with Lee et al.~\cite{lee2020reference}, Li et al.~\cite{li2022eliminating} and Cao et al.\cite{cao2022attention}. Then we conduct a user study to let participants compare colored results of these four methods, participants need subjectively evaluate the colored results according to the reference images and original color images of line drawings to choose the best one from four choices, along with a brief description of why they think this result is the visually best. A user interface of our user study is shown in Fig.~\ref{fig:user_study}.
20 participants take part in the user study and the percentage of each method chosen as the best is shown in Table~\ref{tab:user_study}. It is indicated that AnimeDiffusion has absolute advantages in human visual evaluation. 
\begin{table}[ht]
    \centering
    \caption{User Study Result}
    \label{tab:user_study}
    \begin{tabular}{lc}
    \toprule
        Methods & Percentage of chosen as best\\
    \midrule
        Lee et al.\cite{lee2020reference} & 13.0\% \\
        Li  et al.\cite{li2022eliminating} &17.4\% \\ 
        Cao et al.\cite{cao2022attention} & 21.0\% \\
        AnimeDiffusion  & \textbf{48.6\%} \\
    \bottomrule
    \end{tabular}
\end{table}

\section{Application}
\subsection{Famous Anime Character Recolorization}
\begin{figure}[htb]
    \centering
    \subfigure{
    \begin{minipage}[c]{0.23\linewidth}
    \stackunder[4pt]{\includegraphics[width=\linewidth]{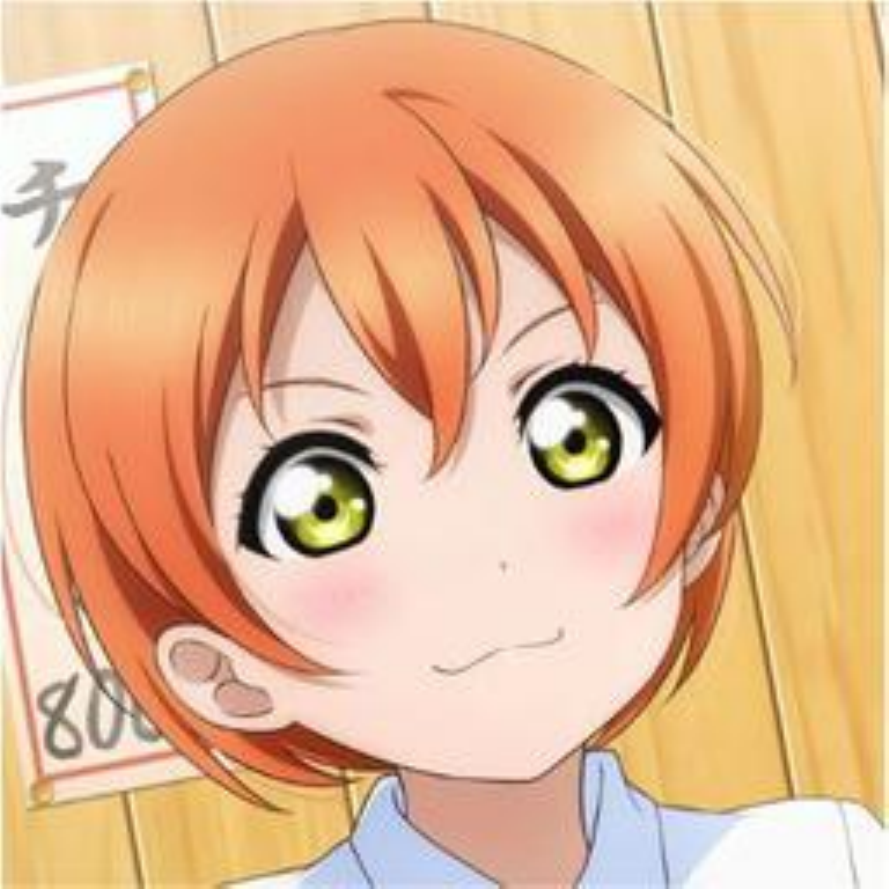}}{Anime 1}
    \end{minipage}
    \begin{minipage}[c]{0.71\linewidth}
    \includegraphics[width=0.325\linewidth]{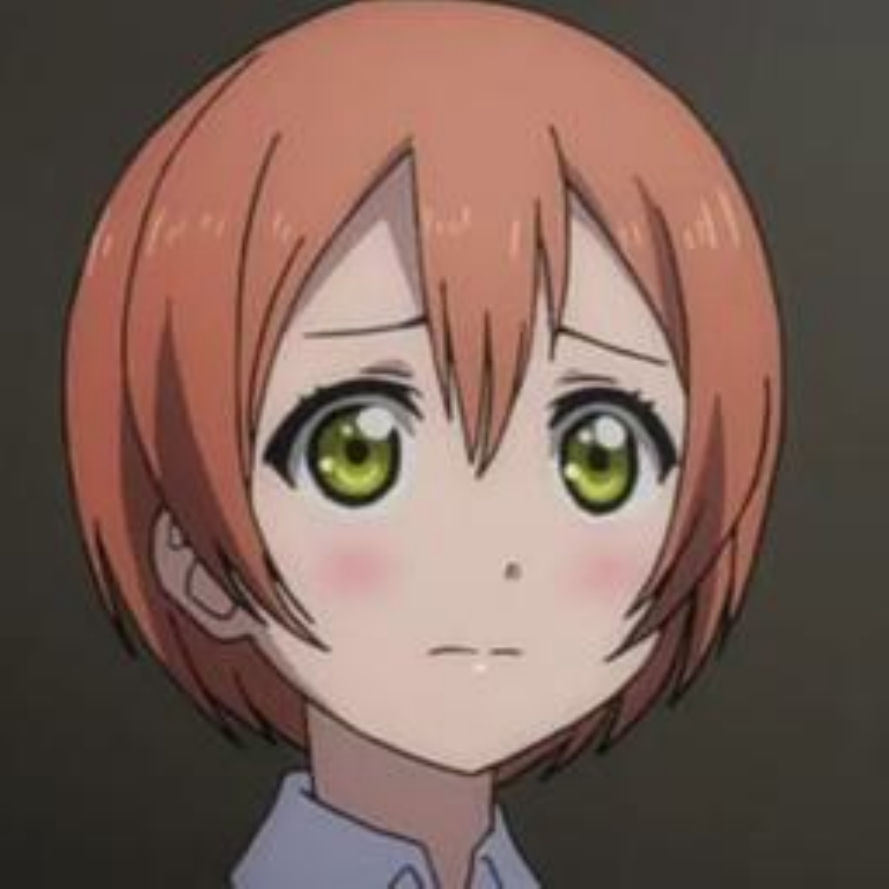}
    \includegraphics[width=0.325\linewidth]{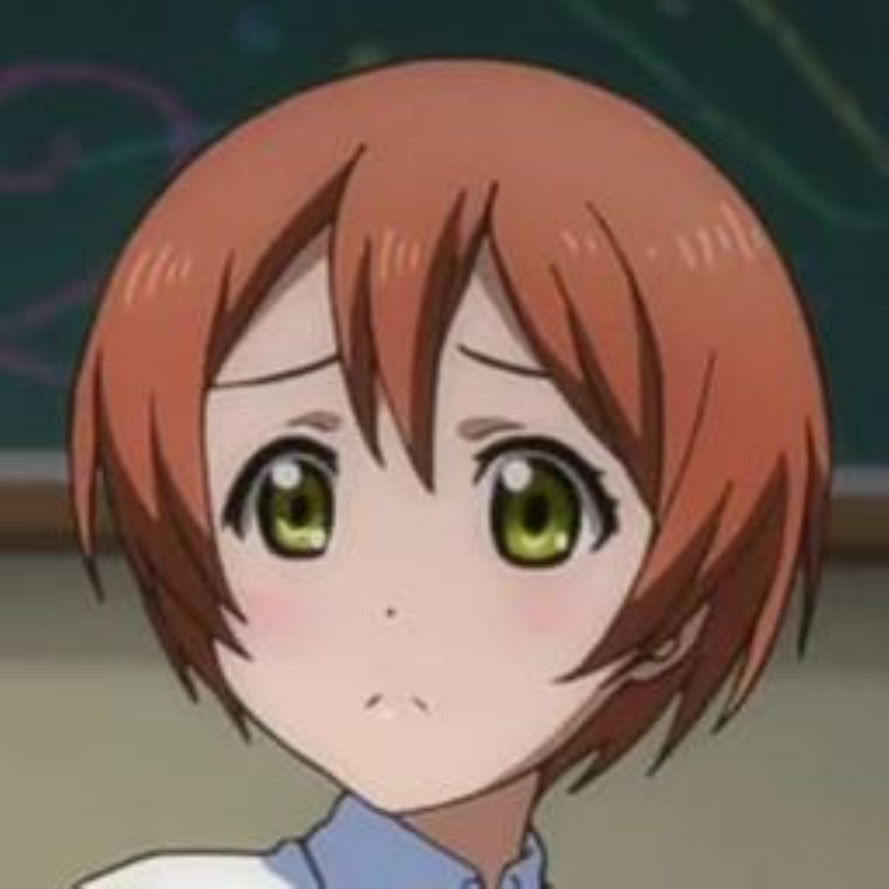}
    \includegraphics[width=0.325\linewidth]{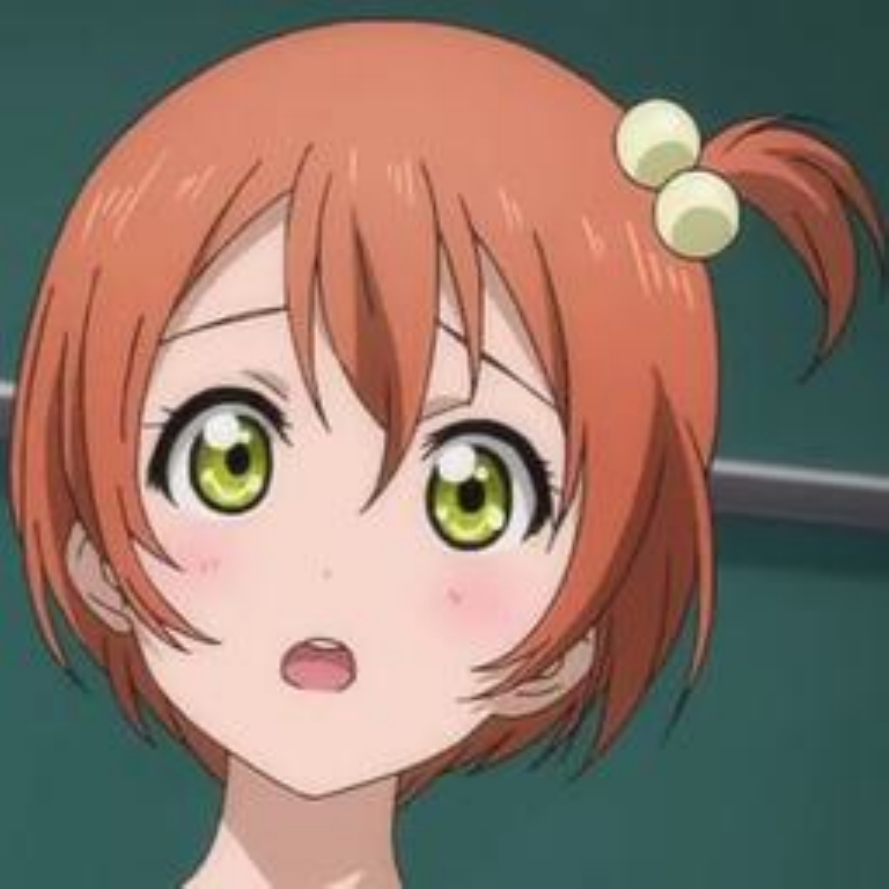}\vspace{4pt}
    \includegraphics[width=0.325\linewidth]{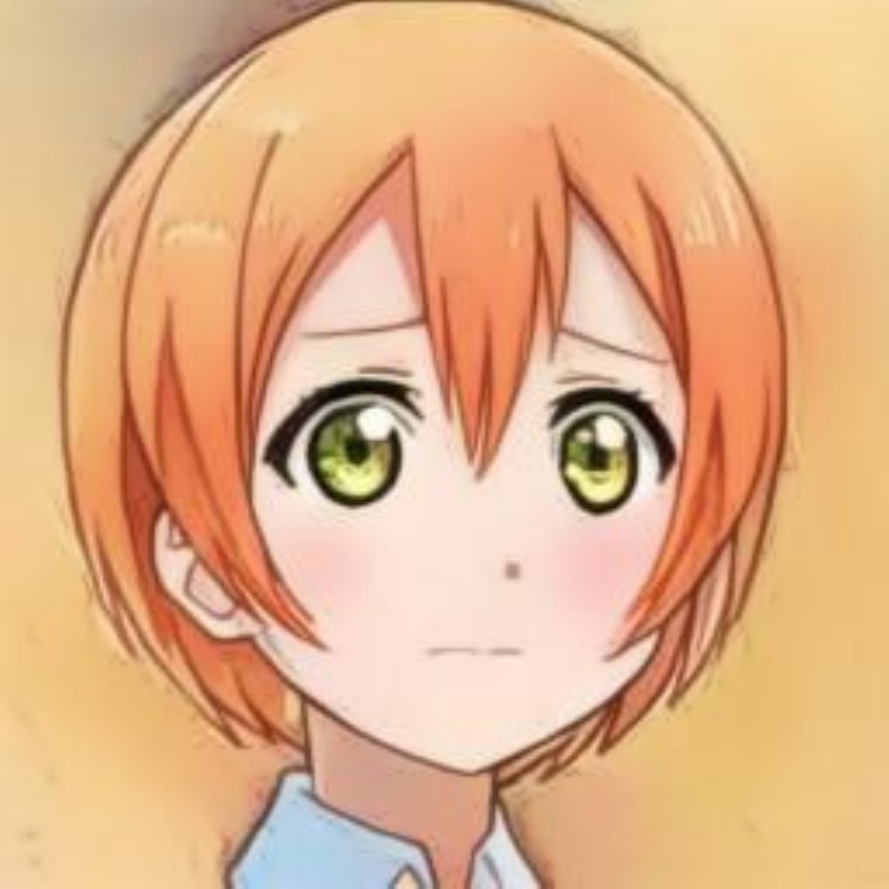}
    \includegraphics[width=0.325\linewidth]{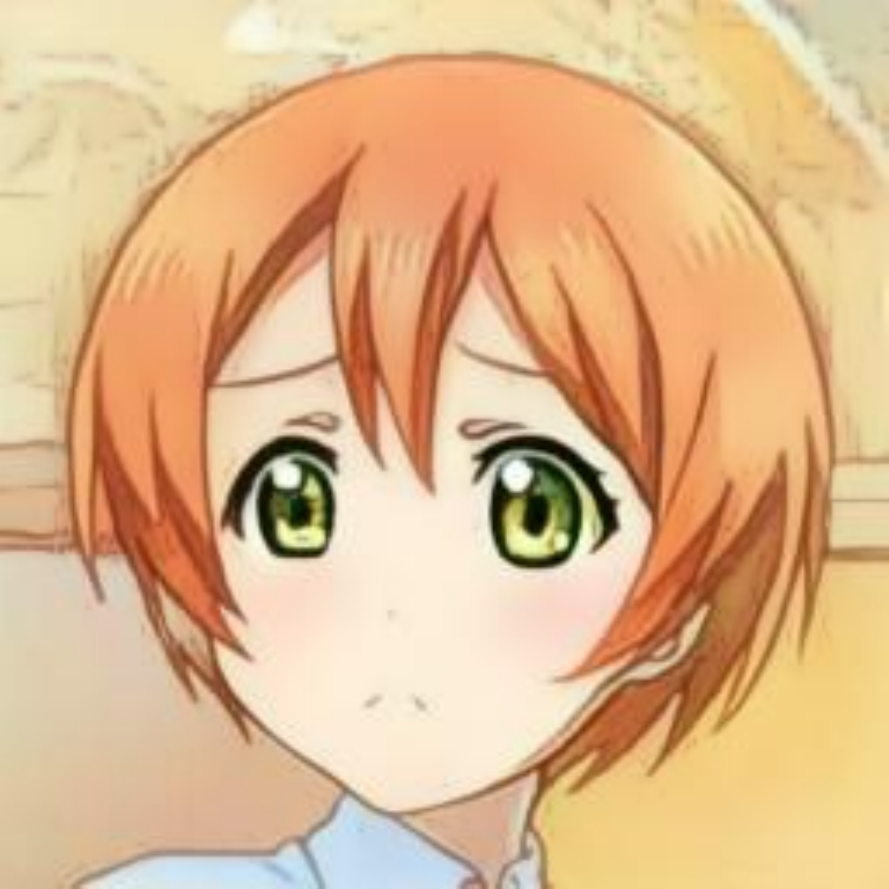}
    \includegraphics[width=0.325\linewidth]{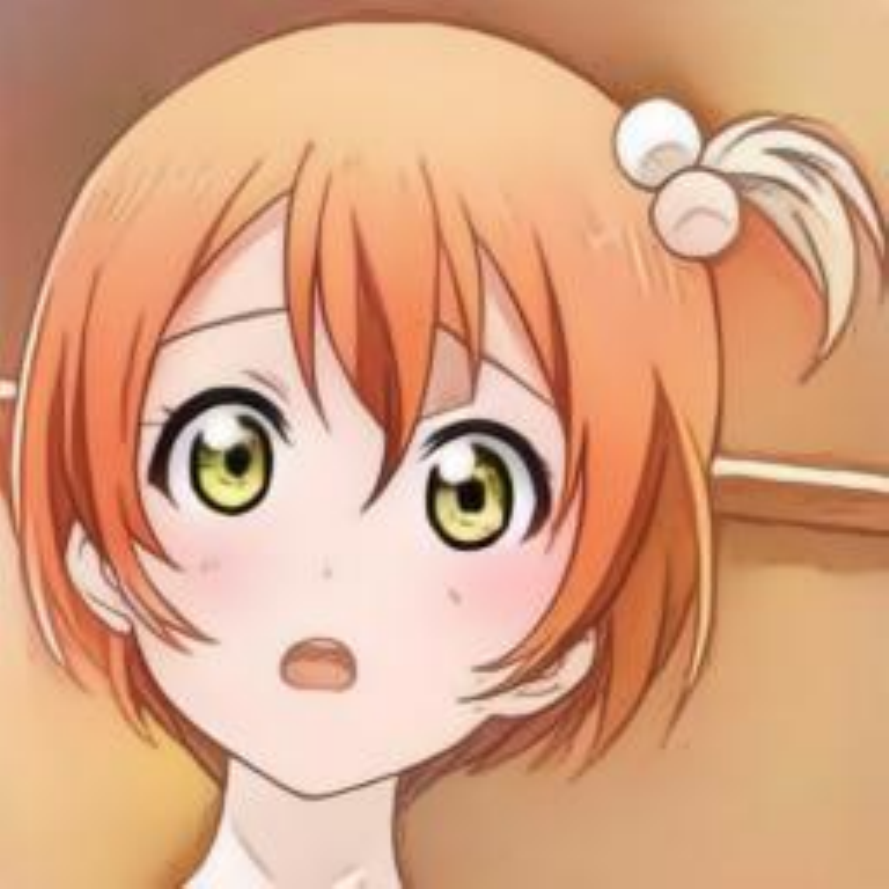}
    \end{minipage}
    }\vspace{2pt}
    \subfigure{
    \begin{minipage}[c]{0.23\linewidth}
    \stackunder[4pt]{\includegraphics[width=\linewidth]{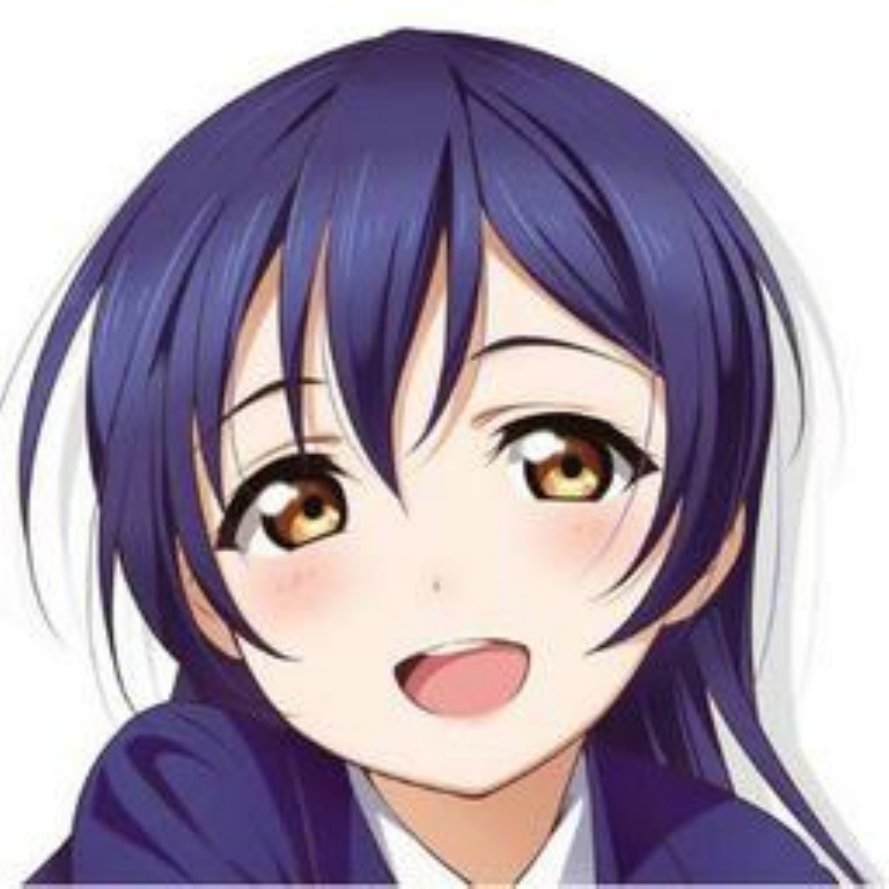}}{Anime 2}
    \end{minipage}
    \begin{minipage}[c]{0.71\linewidth}
    \includegraphics[width=0.325\linewidth]{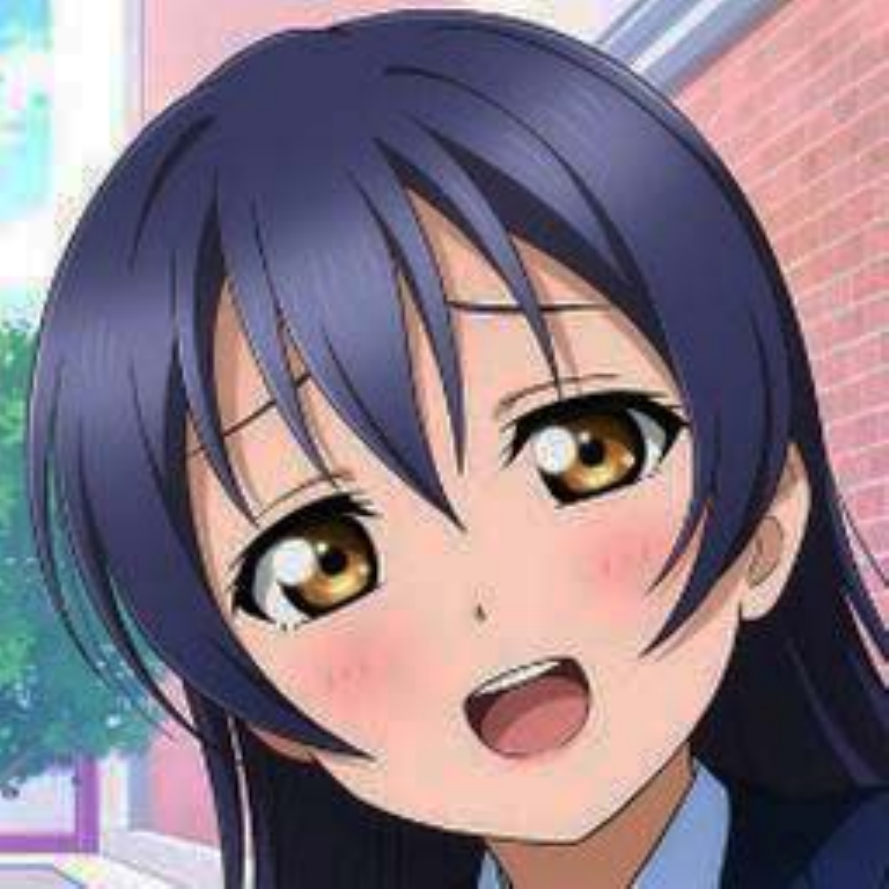}
    \includegraphics[width=0.325\linewidth]{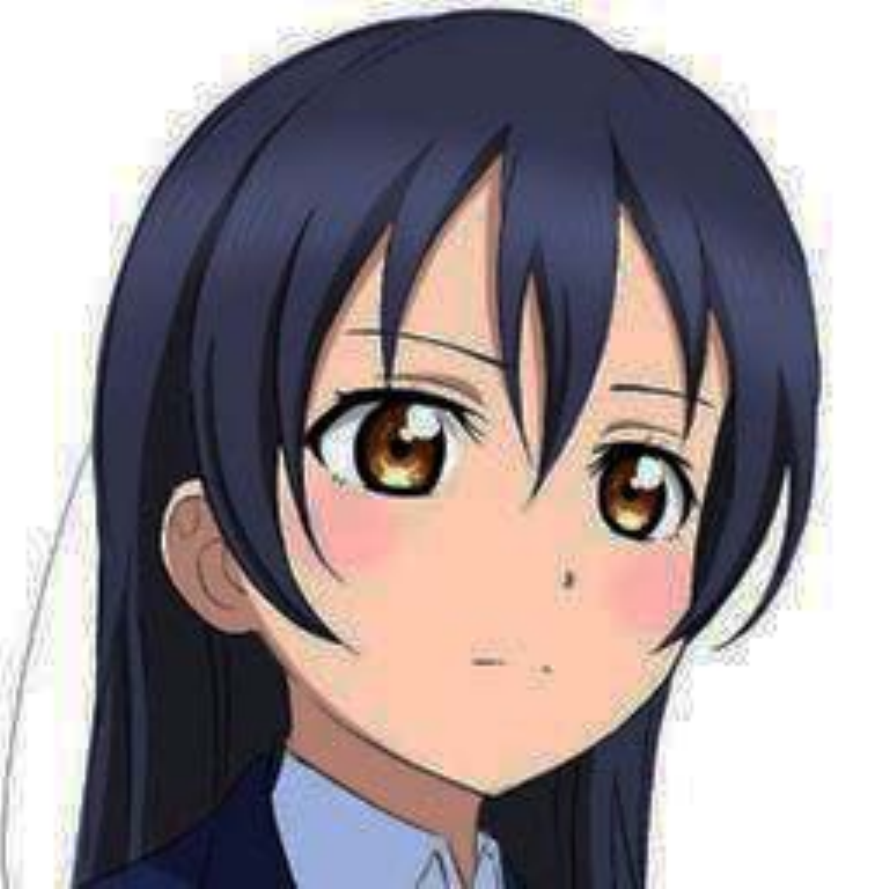}
    \includegraphics[width=0.325\linewidth]{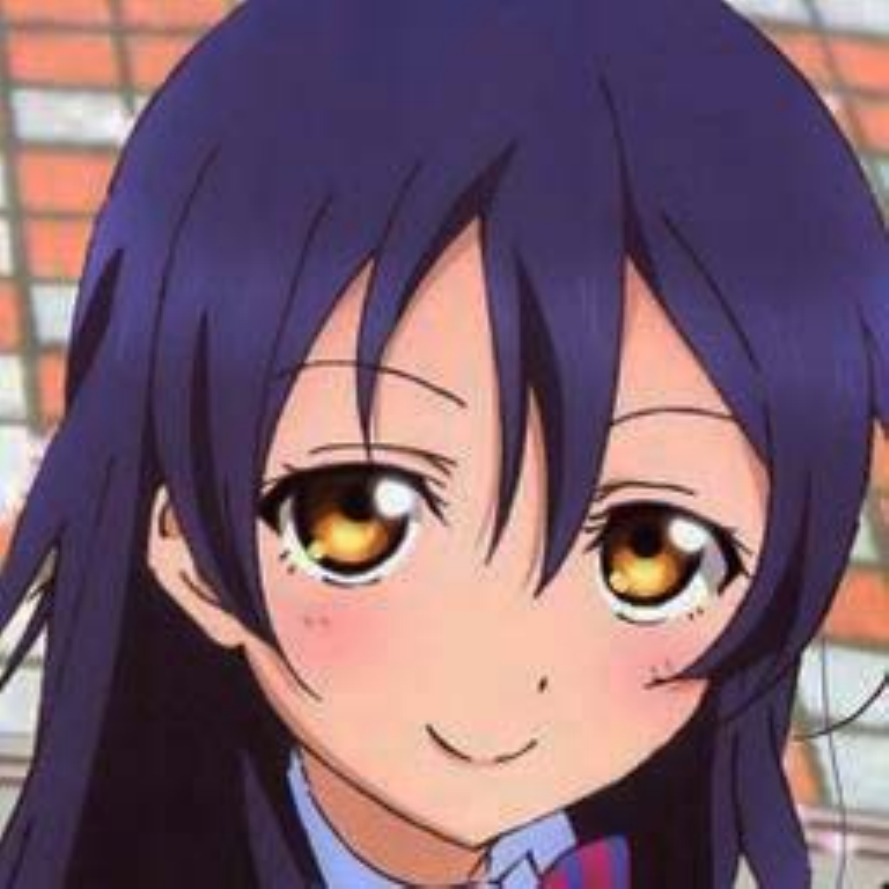}\vspace{4pt}
    \includegraphics[width=0.325\linewidth]{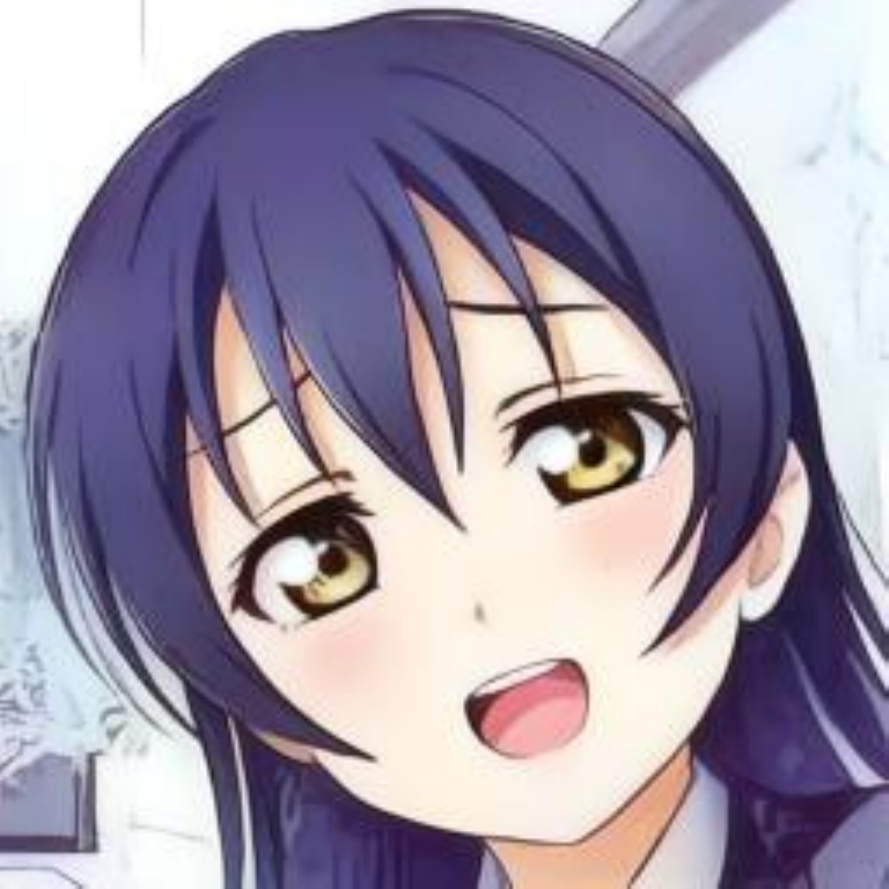}
    \includegraphics[width=0.325\linewidth]{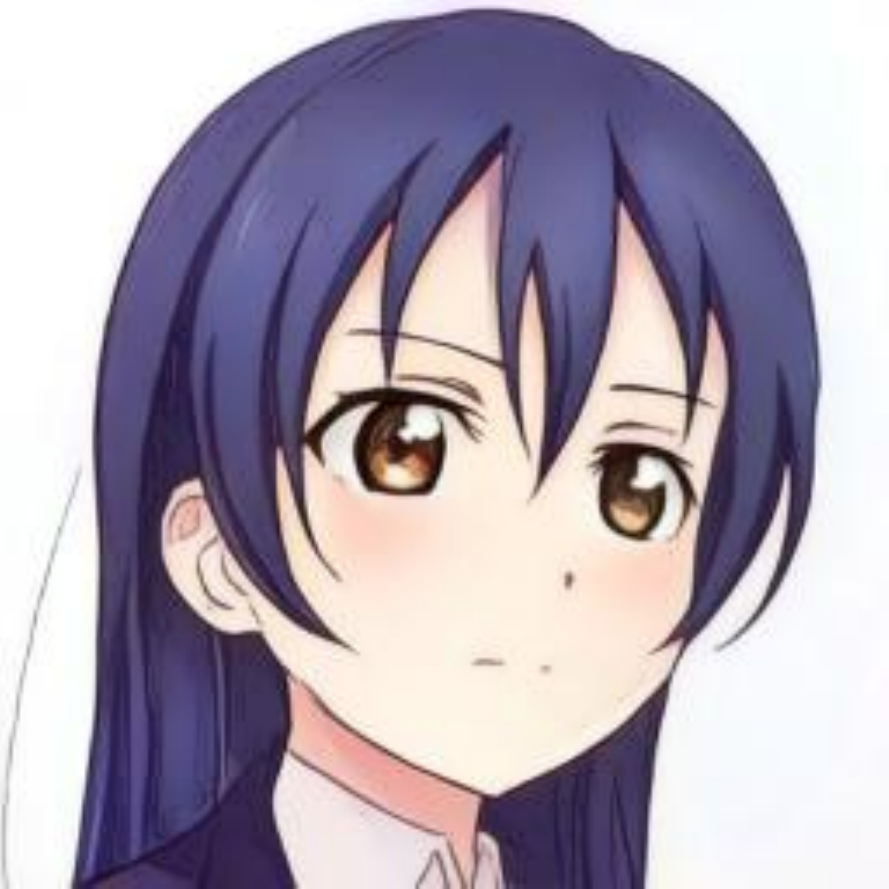}
    \includegraphics[width=0.325\linewidth]{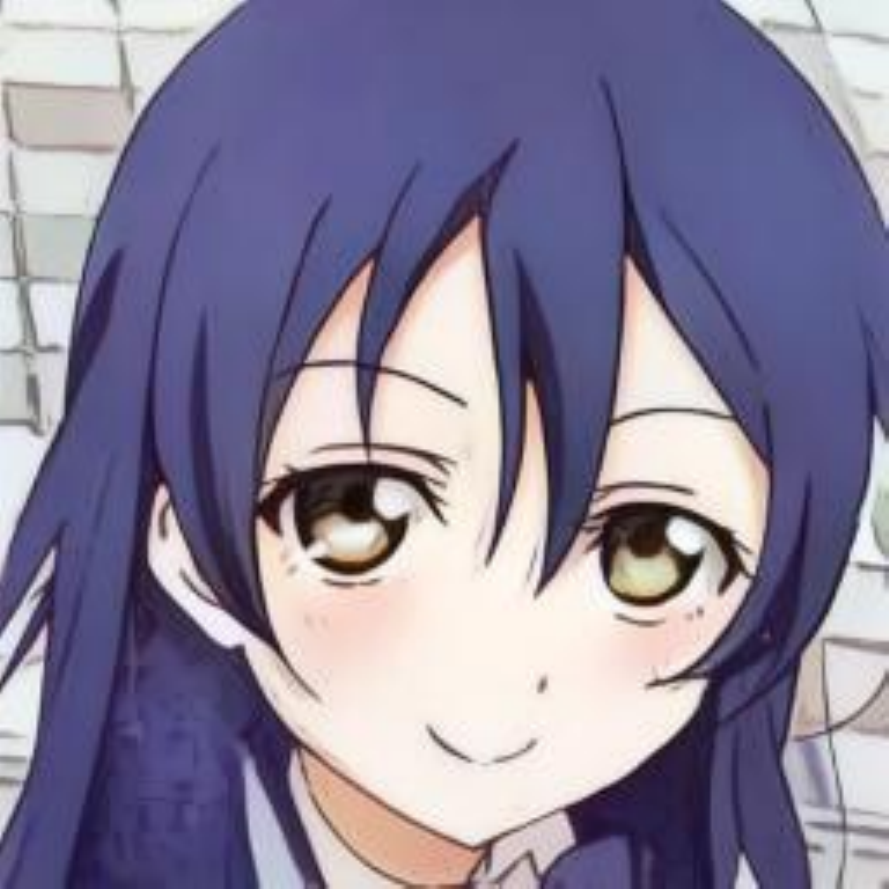}
    \end{minipage}
    }
    \caption{Illustration of famous anime character recolorization. Recolorization results have a uniform color style. The two anime characters are Hoshizora Rin and Sonoda Umi of LoveLive.}
    \label{fig:recolorization}
\end{figure}

Although we aim to train AnimeDiffusion to perform single line drawing colorization, however, in animation creation industry, reference-based colorization method can also be used to recolorize a series of images or even consecutive video frames of the same anime character.
Sometimes during the creation process, the same character appears different colors in different images, and using the recolorization technique, it is convenient to unify the same character's colors based on one reference image.
We apply AnimeDiffusion to perform recolorization task. In Fig.~\ref{fig:recolorization}, for each anime character, the top row is original colored image and the bottom row is our recolorized results.
In fact, we first convert original colored images to line drawings using XDoG extractor, then they are combined with the reference image together and are fed into AnimeDiffusion to generate colored results.
As is shown in Fig.~\ref{fig:recolorization}, according to one reference image, the other images of the same character can be recolorized with the same color style and accurate semantic information. 

\subsection{Original Anime Character Colorization}
\begin{figure}[htb]
    \centering
    \subfigure{
    \begin{minipage}[c]{0.23\linewidth}
    \stackunder[4pt]{\includegraphics[width=\linewidth]{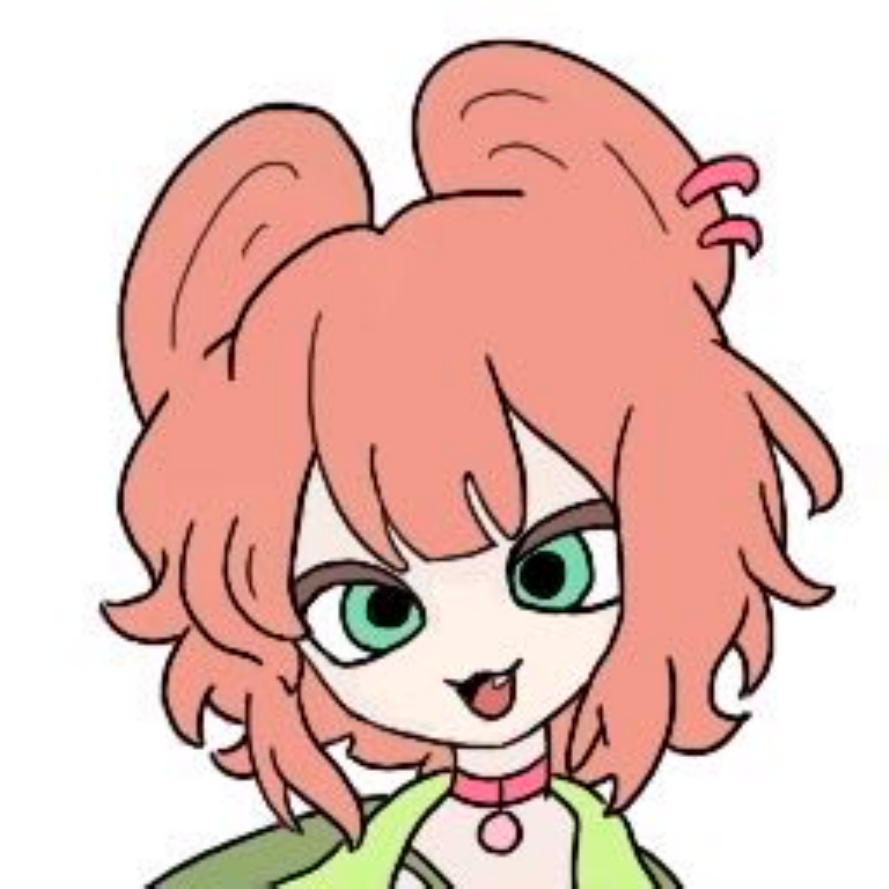}}{Reference}
    \end{minipage}
    \begin{minipage}[c]{0.71\linewidth}
    \includegraphics[width=0.325\linewidth]{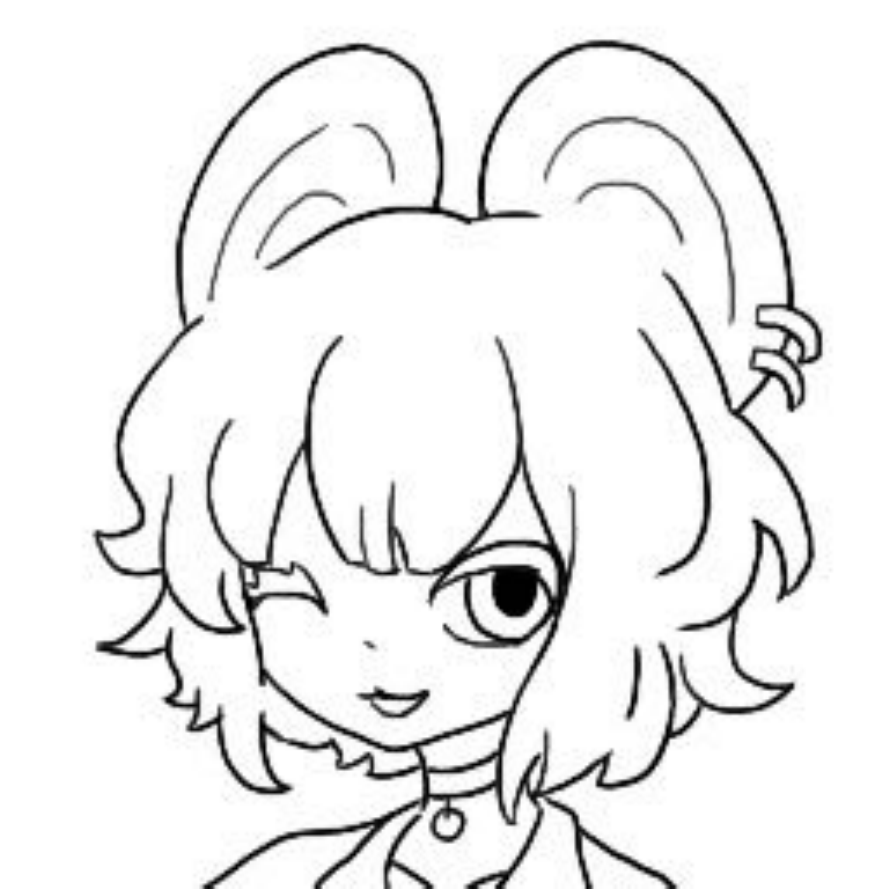}
    \includegraphics[width=0.325\linewidth]{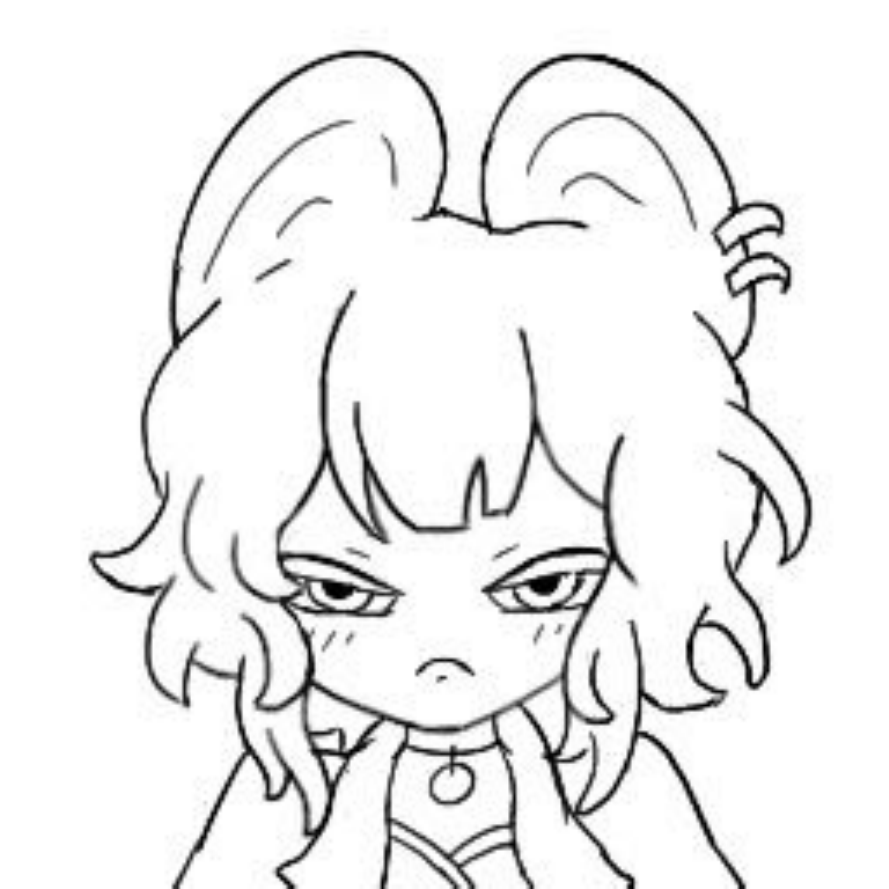}
    \includegraphics[width=0.325\linewidth]{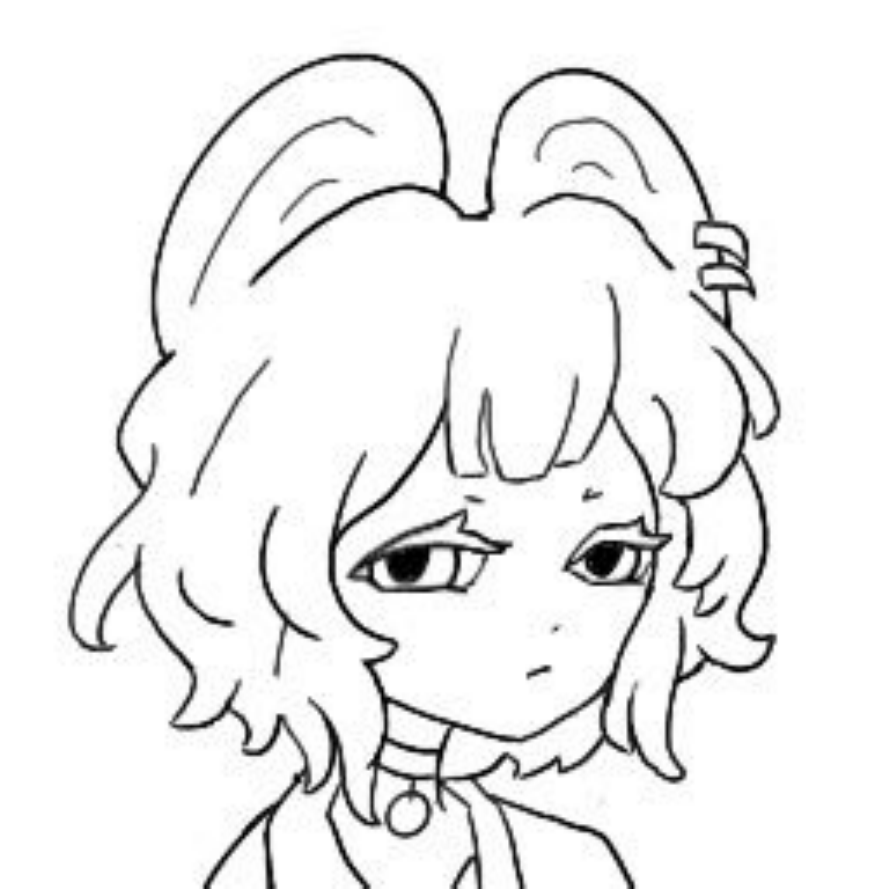}\vspace{4pt}
    \includegraphics[width=0.325\linewidth]{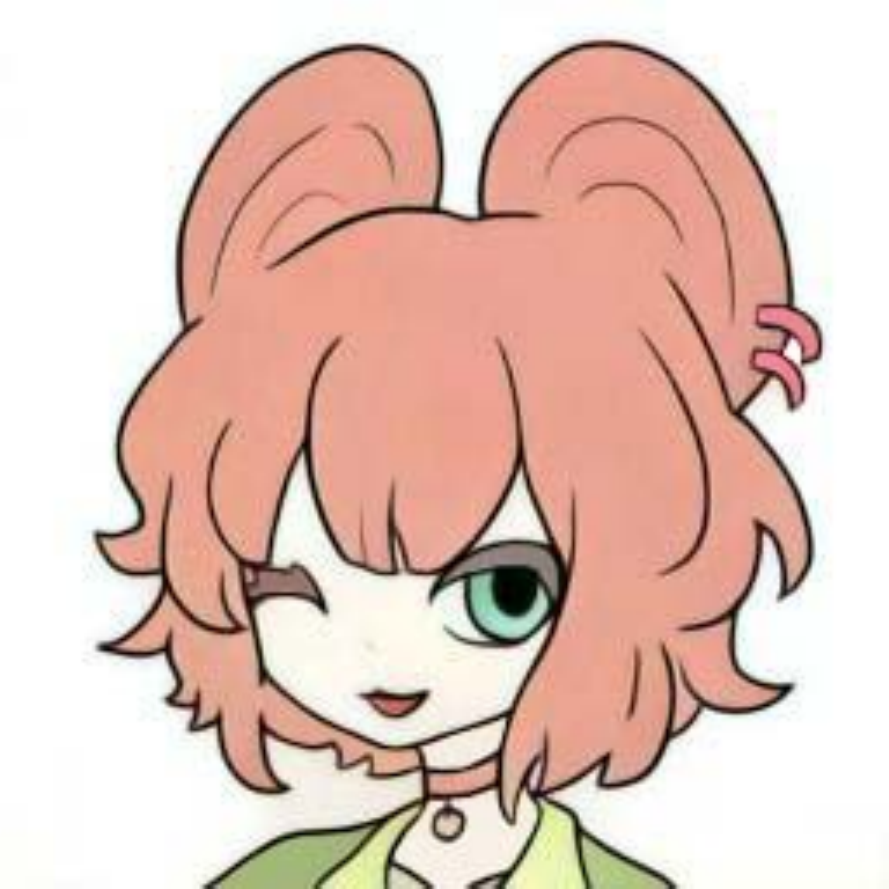}
    \includegraphics[width=0.325\linewidth]{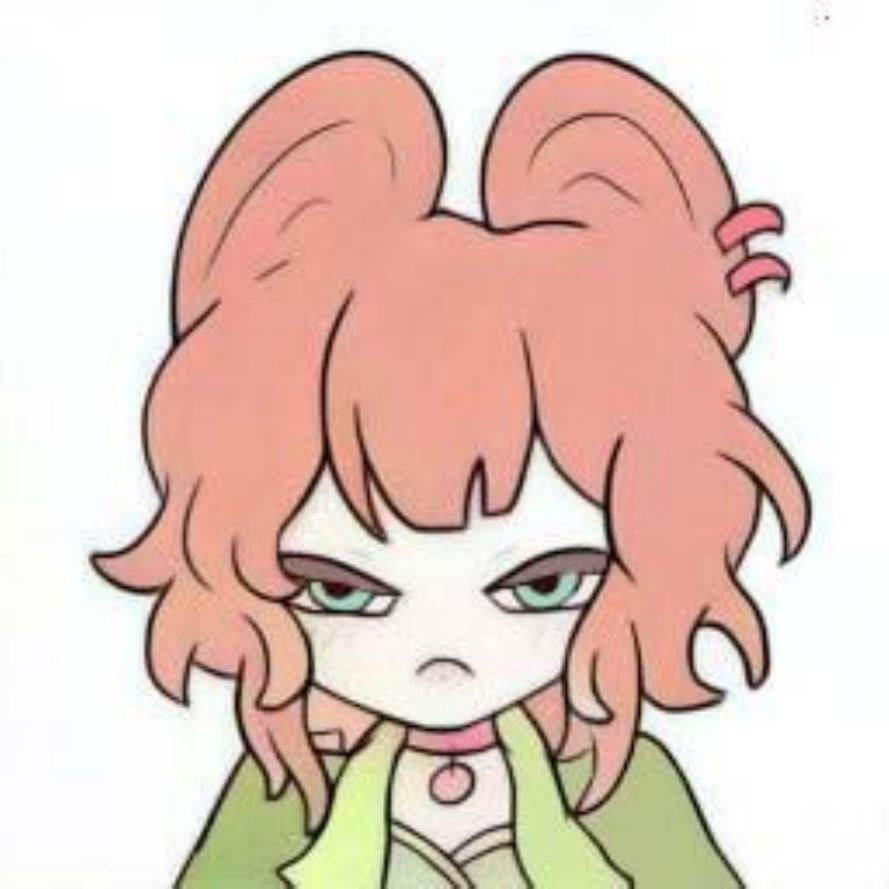}
    \includegraphics[width=0.325\linewidth]{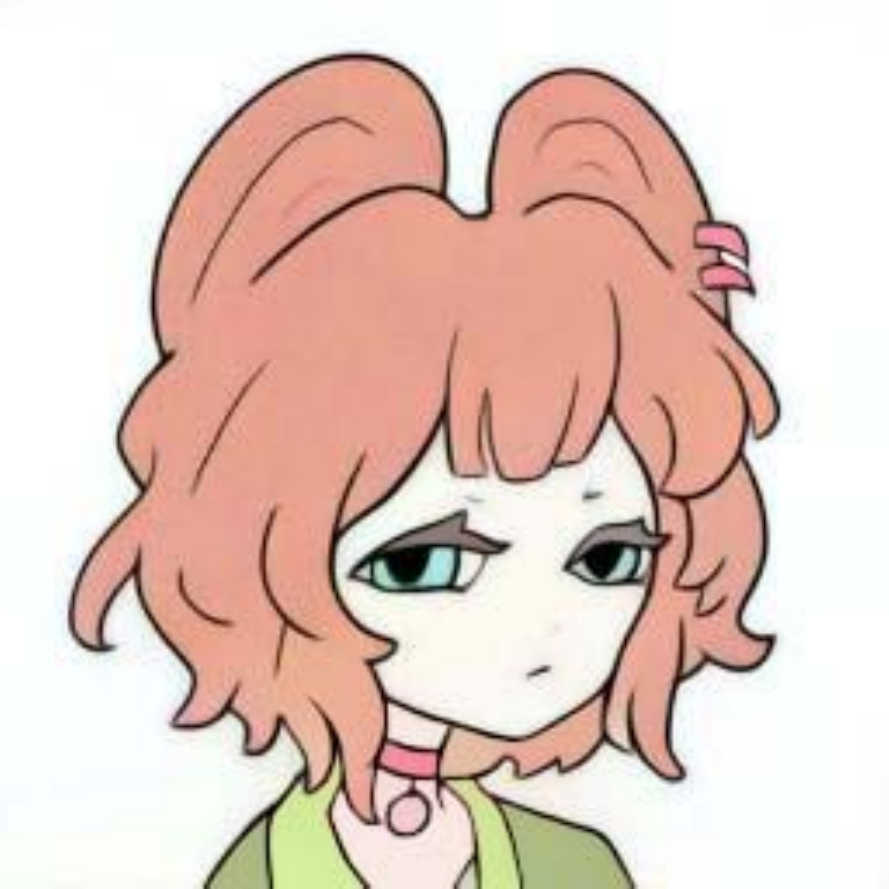}
    \end{minipage}
    }
    \caption{Illustration of original anime character colorization. The anime character with an exaggerated hairstyle is Little Yuyuan, which is created by Ms. Yuwen Wang.}
    \label{fig:original}
\end{figure}

We collaborate with professional artists and use our AnimeDiffusion to assist their work for original line drawings colorization. We invite professional artist to drawing line drawings attached with one reference color image. Since AnimeDiffusion pre-trained on our collected dataset has learned the semantic information of anime character face, it can generalize well to other hand-drawn characters, as is demontrated in Fig.~\ref{fig:original}. With our developed user interface, artist can easily do batch colorization of the same character according to one reference image. This greatly saves the artists' creation time and helps them to complete the creation more efficiently.

\subsection{Fashion Illustration Sketch Colorization}
\begin{figure}[htb]
    \centering
    \subfigure[]{
    \begin{minipage}[b]{0.31\linewidth}
        \includegraphics[width=\linewidth]{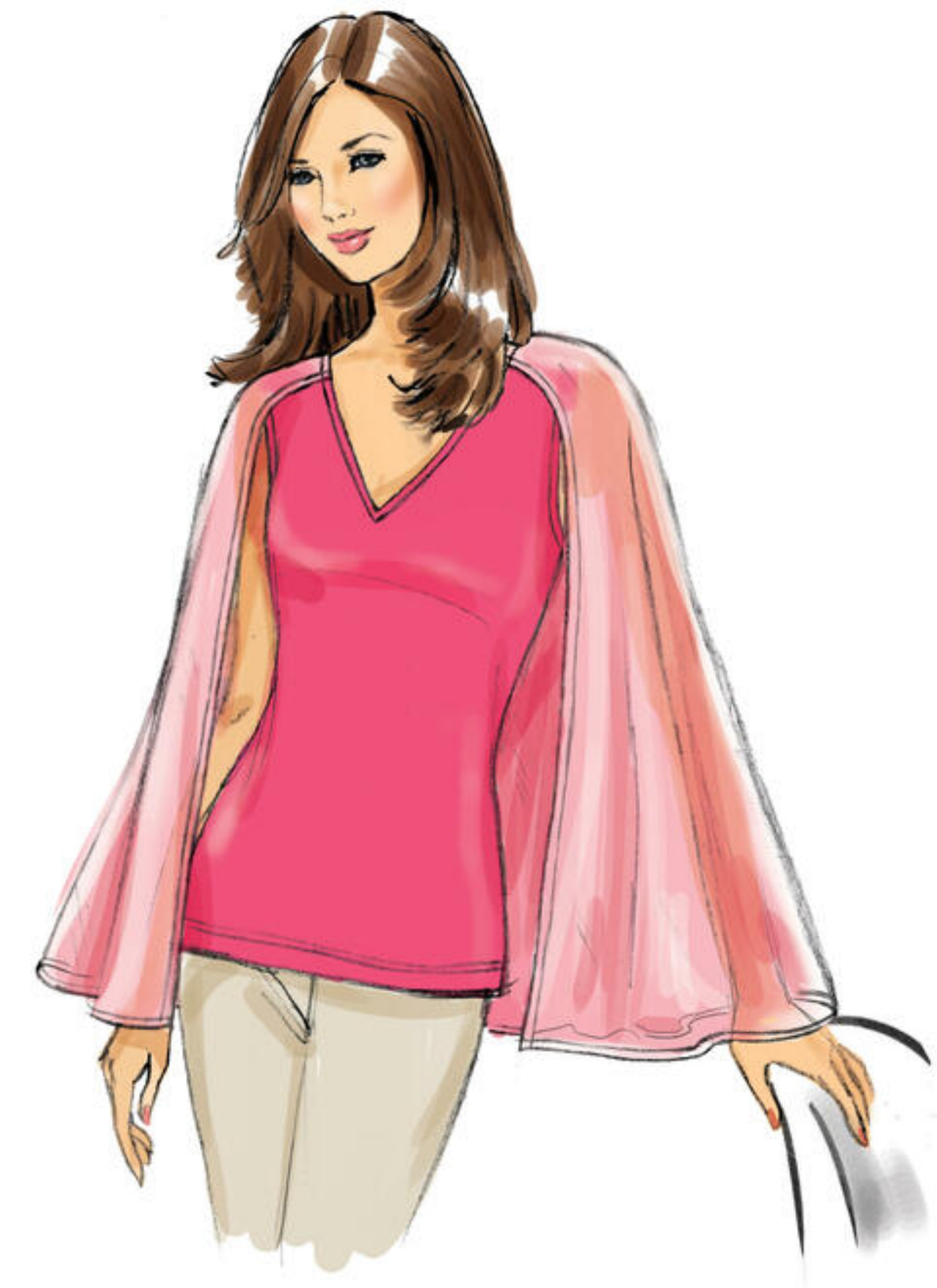}\vspace{4pt}
        \includegraphics[width=\linewidth]{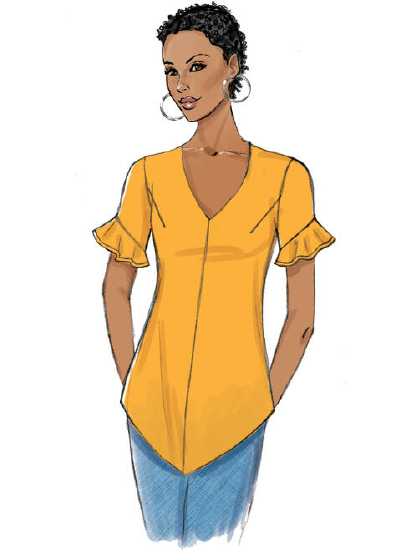}\vspace{4pt}
        \includegraphics[width=\linewidth]{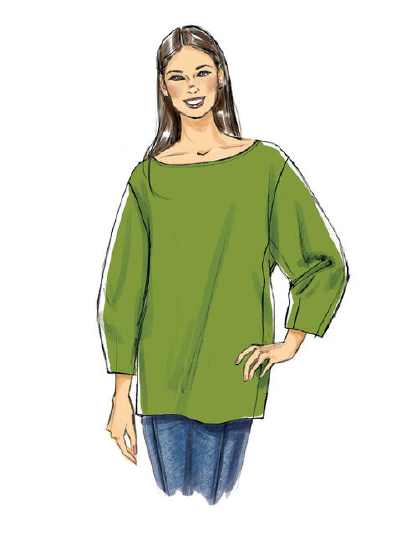}
    \end{minipage}
    }\hspace{-3pt}
    \subfigure[]{
    \begin{minipage}[b]{0.31\linewidth}
        \includegraphics[width=\linewidth]{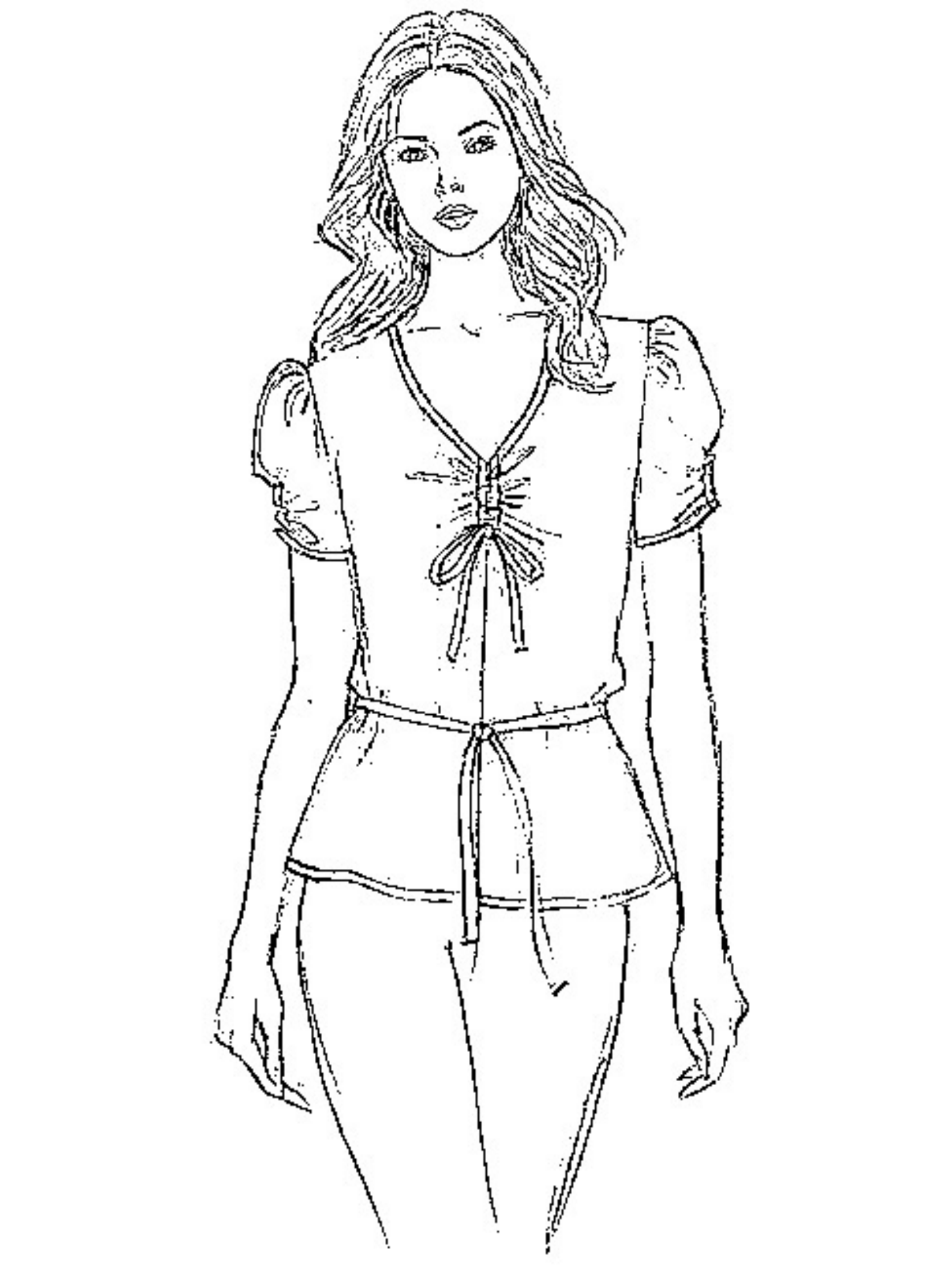}\vspace{4pt}
        \includegraphics[width=\linewidth]{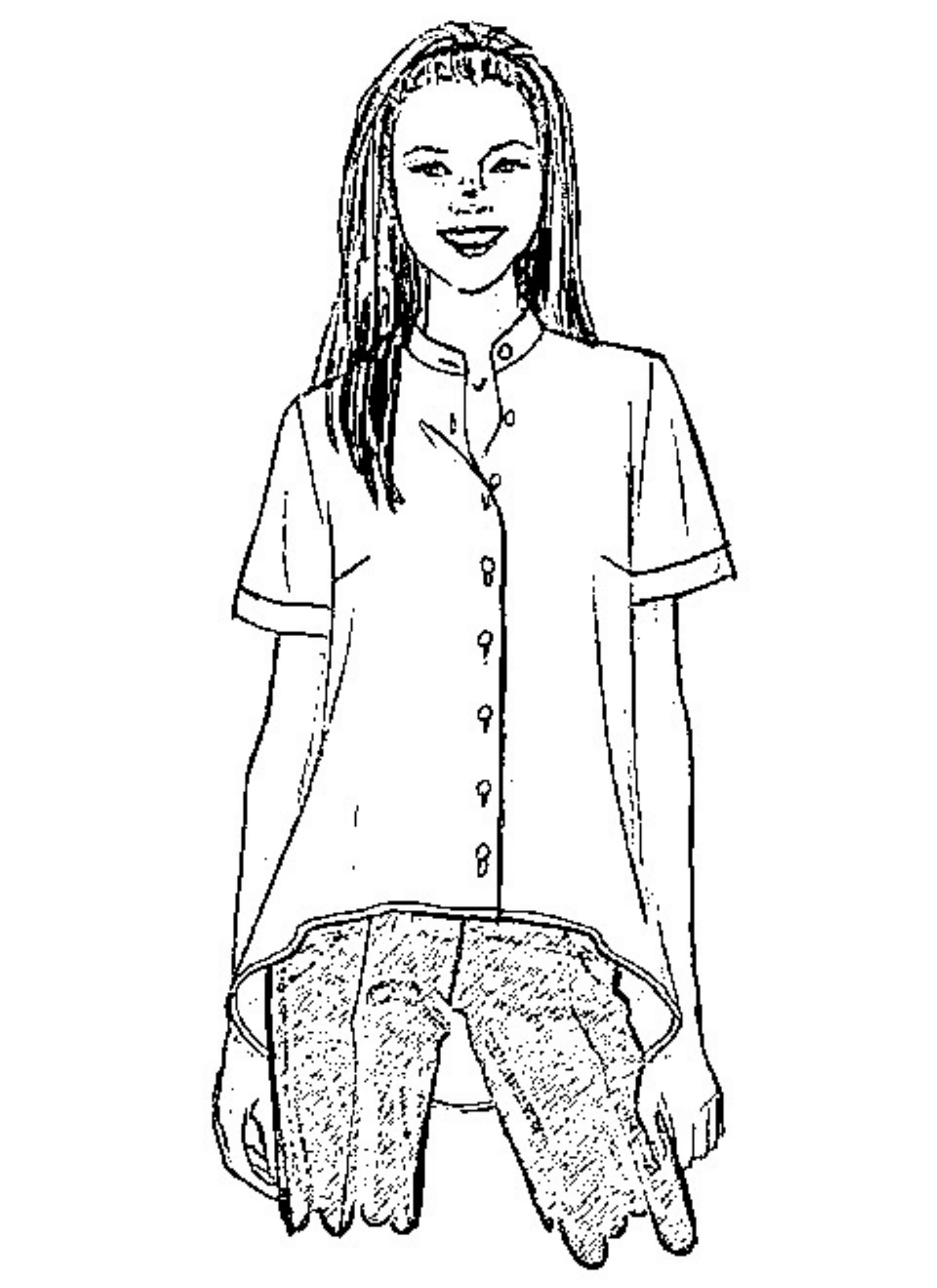}\vspace{4pt}
        \includegraphics[width=\linewidth]{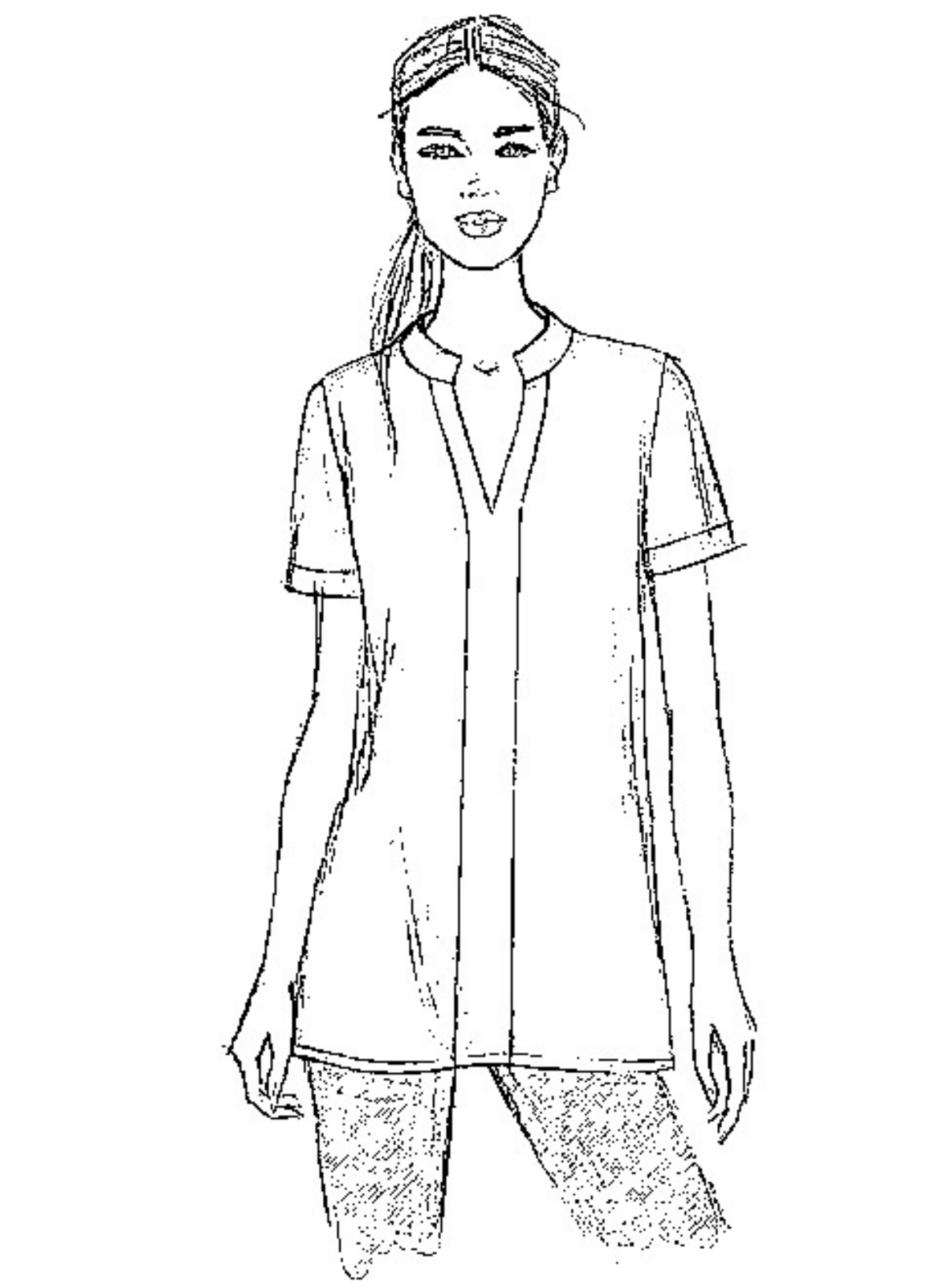}
    \end{minipage}
    }\hspace{-3pt}
    \subfigure[]{
    \begin{minipage}[b]{0.31\linewidth}
        \includegraphics[width=\linewidth]{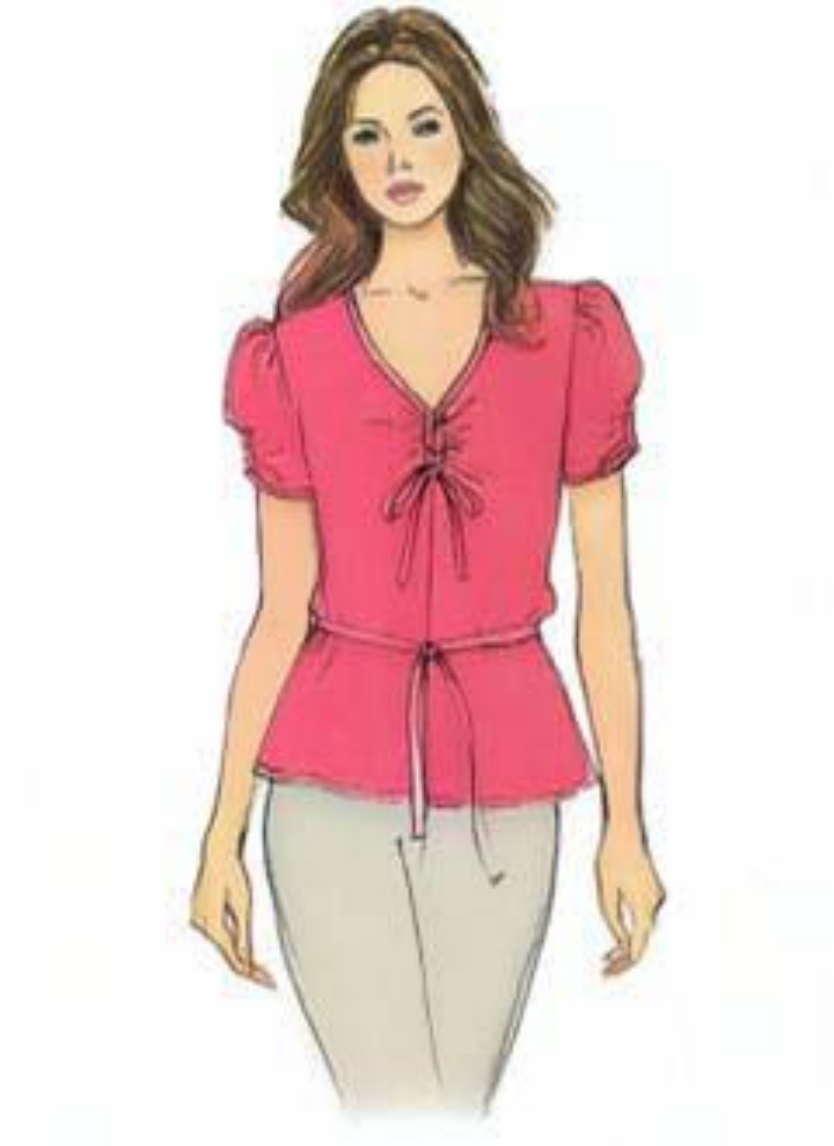}\vspace{4pt}
        \includegraphics[width=\linewidth]{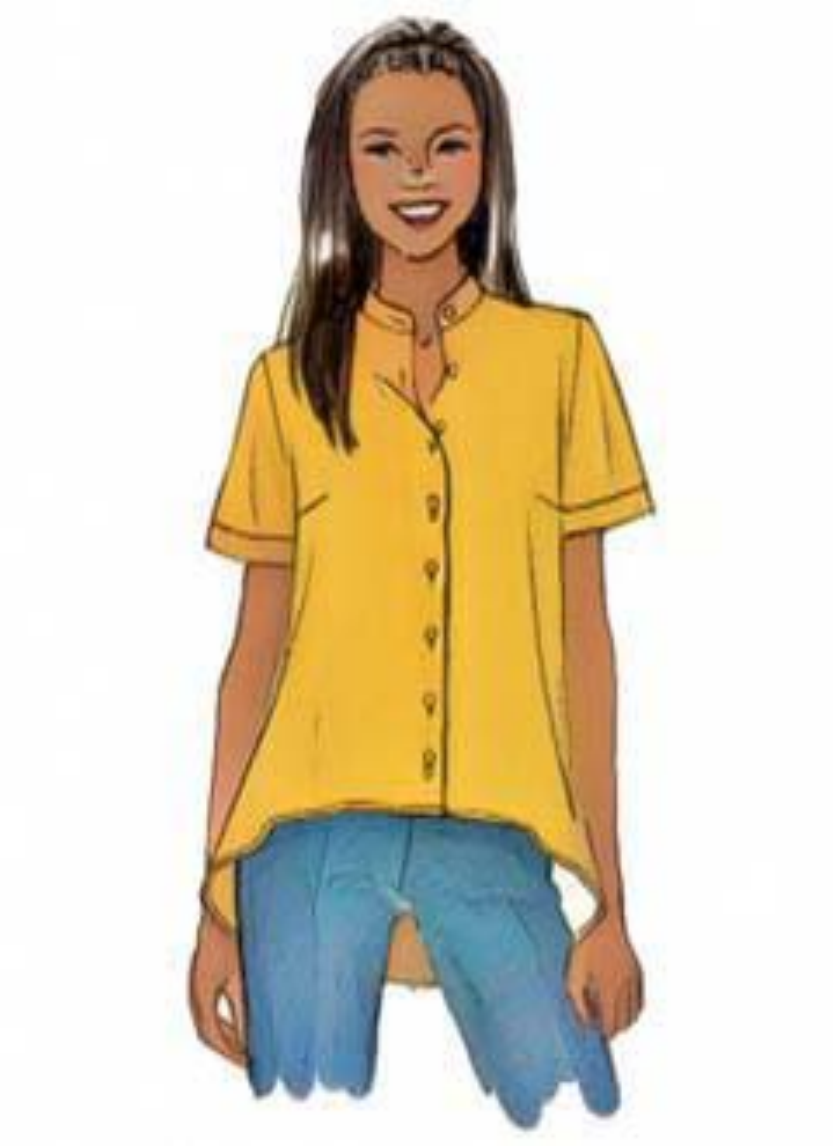}\vspace{4pt}
        \includegraphics[width=\linewidth]{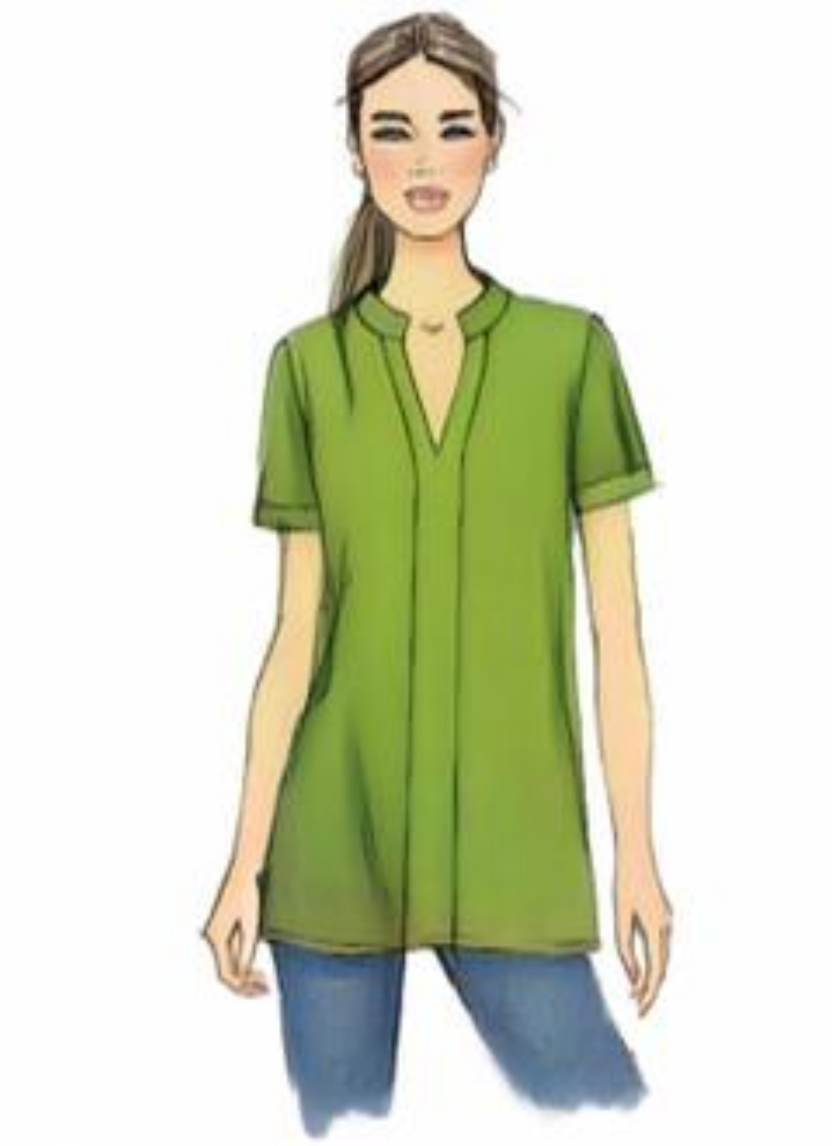}
    \end{minipage}
    }
    \caption{Illustration of fashion illustration sketch colorization. (a) Reference images, (b) Sketches, (c) Colorization results} 
    \label{fig:fashion}
\end{figure}

We also extend AnimeDiffusion to colorize fashion illustration sketches. Since fashion illustration is the same as animation, it is also the first to outline the line draft, and then fill in the color. We regard fashion illustation as another type of animation. As is shown in Fig.~\ref{fig:fashion}, given one color illustration and one sketch, AnimeDiffusion can generate colorization results which extend the range of accurate semantic correspondence to half of the body. We can not only keep the accurate color of the face, but also have good control over the clothing and torso, even the color of skin can be accurately distinguished.
Fashion designers can use our AnimeDiffusion user interface to easily colorize hand-drawn fashion sketches, which are used for the follow-up process of garment pattern making.

\begin{figure}[htb]
    \centering
    \subfigure[]{
    \begin{minipage}[b]{0.31\linewidth}
        \includegraphics[width=\linewidth]{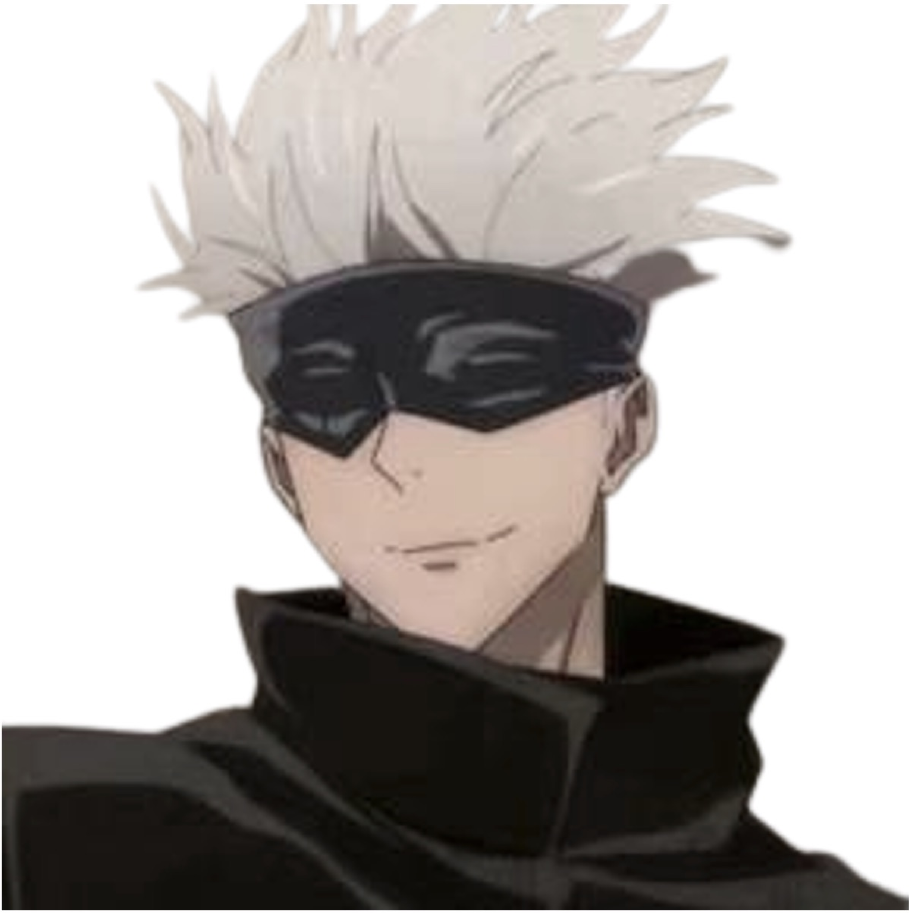}
    \end{minipage}
    }\hspace{-3pt}
    \subfigure[]{
    \begin{minipage}[b]{0.31\linewidth}
        \includegraphics[width=\linewidth]{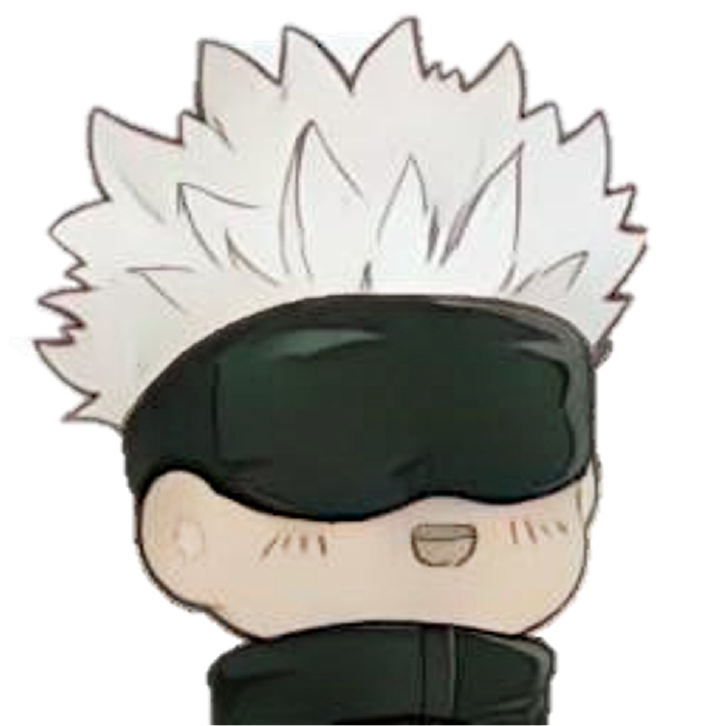}
    \end{minipage}
    }\hspace{-3pt}
    \subfigure[]{
    \begin{minipage}[b]{0.31\linewidth}
        \includegraphics[width=\linewidth]{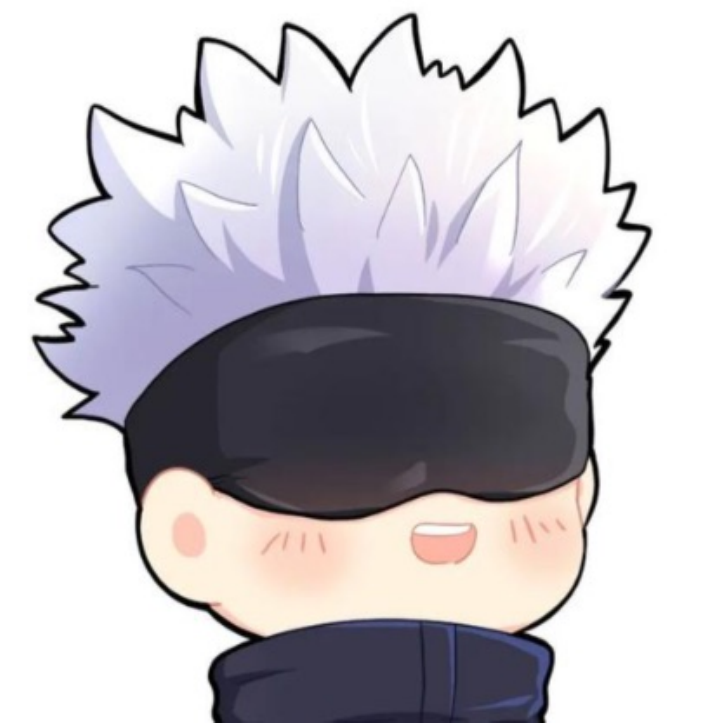}
    \end{minipage}
    }
    \caption{Limitation of our approach. (a) Reference images, (b) Colorization results, (c) Ground truth. Due to the large stylistic differences between the reference image and the line drawing, the color of the mouth is not available in the reference image, while the teeth are not correctly identified in our model in the line drawing, and the color of the teeth is not accurately reflected in our colorization results.} 
    \label{fig:limitation}
\end{figure}

\section{Conclusion And Future Work}
In this paper, we propose AnimeDiffusion, the first diffusion model tailored for anime face line drawing colorization. 
In order to train AnimeDiffusion we build a benchmark dataset for research purpose and also fill the gap of no available high resolution anime face dataset to evaluate line drawing colorization algorithms. 
To handle the high computation consumption problem of diffusion models, we design a novel hybrid training strategy which separates the image denoising task and image reconstruction task. 
Through extensive experiments and a user study, AnimeDiffusion has demonstrated better performance both qualitatively and quantitatively, outperforming other state-of-the-art GANs-based methods, with higher image quality and semantic color information. 
To the best of our knowledge, AnimeDiffusion is the first learning based work can accurately colorize anime face line drawing with heterochromatic pupils according to reference color image, without other special module for processing eyes or pupils in anime face. 

\meng{However, there is a limitation in out method. 
Our model uses paired training data in training, and there is some style correlation between the reference image and the line drawings. 
For special style line drawings, such as Chibi cartoons as shown in Fig \ref{fig:limitation}(c), if the corresponding semantic information does not exist in the reference image, the colorization result of the line drawing may appear to be inconsistent with the real image.}
In the future work, We will work on multi-modal input line drawing colorization such as combining text information and reference image together to make the interactive way of colorization more rich. 
This will greatly reduce the manual tasks of animators and improve the creation efficiency \meng{and colorization effect} of the animation creation industry.


%



\ifCLASSOPTIONcompsoc
  \section*{Acknowledgments}
  We thank Mr. Henry Tian, Ms. Mandy Wong, Ms. Rachel Liu and Mr. Leonard Chen for their help with anime knowledge and image samples selection. We thank Ms. Xiao Meng and Ms. Yuwen Wang for helping us create wonderful hand-painted line drawings.
\else
  \section*{Acknowledgment}
\fi

\ifCLASSOPTIONcaptionsoff
  \newpage
\fi



%
  

\bibliographystyle{IEEEtran}
\bibliography{reference}

\end{document}